\documentclass[11pt]{article}
\usepackage[utf8]{inputenc}
\usepackage{lmodern}
\usepackage[T1]{fontenc}
\usepackage[margin=1in]{geometry}
\usepackage{amsmath, amssymb}
\usepackage[colorlinks=true, linkcolor=black, citecolor=blue, urlcolor=black]{hyperref}
\usepackage[numbers,sort&compress]{natbib}
\usepackage{booktabs}
\usepackage{multirow}
\usepackage{graphicx}
\usepackage{amssymb}
\usepackage{longtable}
\usepackage{array}
\usepackage{authblk}
\usepackage{tabularx}
\usepackage{xcolor}
\usepackage{placeins}

\usepackage{tikz}
\usetikzlibrary{positioning, calc, backgrounds, fit, shapes.geometric, arrows.meta, decorations.pathreplacing}

\definecolor{colFoundation}{HTML}{7F8C8D}
\definecolor{colRL}{HTML}{E74C3C}
\definecolor{colDriving}{HTML}{3498DB}
\definecolor{colRobotics}{HTML}{E67E22}
\definecolor{colVideo}{HTML}{9B59B6}
\definecolor{colMedical}{HTML}{27AE60}
\definecolor{colEducation}{HTML}{F1C40F}
\definecolor{colArchitecture}{HTML}{1ABC9C}
\definecolor{yearcolor}{HTML}{2C3E50}

\title{World Models: A Comprehensive Survey of Architectures, Methodologies, Reasoning Paradigms, and Applications}

\author[1]{Arif Hassan Zidan}
\author[2]{Yi Pan}
\author[2]{Hanqi Jiang}
\author[9]{Ruiyu Yan}
\author[2]{Wei Ruan}
\author[2]{Zihao Wu}
\author[2]{Lifeng Chen}
\author[2]{Weihang You}
\author[2]{Xinliang Li}
\author[3]{Bowen Chen}
\author[2]{Huawen Hu} 
\author[10]{Peilong Wang}
\author[12]{Sizhuang Liu}
\author[5]{Jing Zhang}
\author[2]{Siyuan Li}

\author[2]{Zhengliang Liu}
\author[6]{Yu Bao}
\author[3]{Lin Zhao}
\author[7]{Lichao Sun}
\author[5]{Dajiang Zhu}
\author[4]{Xiang Li}
\author[8]{Jinglei Lv}
\author[4]{Quanzheng Li}
\author[11]{Wei Liu}
\author[2]{Tianming Liu\textsuperscript{*}}
\author[1]{Wei Zhang\textsuperscript{*}}

\affil[1]{School of Computer and Cyber Sciences, Augusta University, Augusta, GA, USA}
\affil[2]{School of Computing, University of Georgia, Athens, GA, USA}
\affil[3]{Department of Biomedical Engineering, New Jersey Institute of Technology, Newark, NJ, USA}
\affil[4]{Department of Radiology, Massachusetts General Hospital, Harvard Medical School, Boston, MA, USA}
\affil[5]{Department of Computer Science and Engineering,  University of Texas at Arlington, Arlington, TX, USA}
\affil[6]{Department of Graduate Psychology, James Madison University, Harrisonburg, VA, USA}
\affil[7]{Computer Science and Engineering, Lehigh University, Bethlehem, PA, USA}
\affil[8]{School of Biomedical Engineering, The University of Sydney, Sydney, Australia}
\affil[9]{Tandon School of Engineering, New York University, Brooklyn, NY, USA}
\affil[10]{Department of Radiation Oncology, City of Hope National Medical Center, Duarte, CA, USA}
\affil[11]{Department of Mayo Clinic Comprehensive Cancer Center, Mayo Clinics, Phoenix, AZ, USA}
\affil[12]{Savannah River Ecology Laboratory (SREL), University of Georgia, Aiken, SC, USA}

\date{}

\renewcommand{\hat}{\widehat}

\def\a\cos{\mathrm{arc\cos}}

\begin{document}

\maketitle

\vspace{-0.6em}
\noindent\textsuperscript{*}\textbf{Corresponding author(s).} E-mail(s):
\href{mailto:tliu@uga.edu}{tliu@uga.edu};
\href{mailto:wzhang2@augusta.edu}{wzhang2@augusta.edu}

\begin{abstract}
World models, internal simulators that learn the structure and dynamics of an environment, have emerged as a central paradigm in the pursuit of artificial general intelligence, enabling agents to predict, plan, and reason within learned representations. Despite rapid progress across reinforcement learning, robotics, autonomous driving, and video generation, the field still lacks a unified framework capable of integrating its diverse architectural choices, training methodologies, reasoning mechanisms, and application settings. This survey addresses that gap by introducing a comprehensive multi-axis taxonomy organized along four complementary dimensions: (i)~architecture, encompassing representation format, dynamics formulation, input modality, learning paradigm, and downstream application; (ii)~methodological family, including state-space and recurrent approaches, transformer-based models, diffusion-based generators, physics-informed networks, and language-augmented multimodal systems; (iii)~reasoning strategy, covering imagination-based planning, latent policy learning, counterfactual reasoning, and planning under uncertainty; and (iv)~application domain, spanning robotics, autonomous driving, video prediction, multimodal agents, reinforcement learning, scientific modeling, medical imaging, educational measurement, and business and finance. Tracing the historical development of the field from early cognitive-science foundations to milestone systems such as PlaNet, the Dreamer family, MuZero, Sora, Cosmos, and Genie, we examine how these dimensions interact and highlight the recent convergence of chain-of-thought reasoning with world-model imagination. Across these axes, we review evaluation protocols and benchmarks, identify persistent challenges—including compounding prediction errors, sim-to-real transfer, and fragmented evaluation practices—and outline future directions toward unified multimodal world models, foundation-scale interactive simulators, and safe deployment in safety-critical domains. By synthesizing these developments within a single cross-disciplinary framework, this survey provides a structured roadmap for advancing world-model research toward more general, robust, and capable autonomous systems.
\end{abstract}

\tableofcontents
\newpage
\section{Introduction}

The pursuit of artificial general intelligence (AGI) has long motivated researchers to develop intelligent systems that not only recognize meaningful patterns from multimodal data but also acquire a coherent and causal understanding of the environments in which they operate. Central to this ambition is the concept of the World Model (WM)—an internal simulator that captures environmental dynamics and enables both forward and counterfactual rollouts for perception, prediction, and decision-making~\cite{li2025embodied}. The intellectual foundations of this idea extend far beyond contemporary machine learning. In cognitive science, it has long been recognized that humans interpret the external world by abstracting it into simplified elements and relational structures. This perspective is articulated in Johnson-Laird’s theory of mental models~\cite{johnson1983mental} and resonates with early developments in artificial intelligence, such as Minsky’s frame representations proposed in the 1970s~\cite{minsky1974framework}. In brief, these perspectives highlight the longstanding interdisciplinary interest in constructing internal representations that enable reasoning about complex environments.
 
While these early frameworks were largely symbolic, the advent of deep learning ushered in a new era for operationalizing the world model concept. In reinforcement learning (RL), Ha and Schmidhuber~\cite{ha2018world} revitalized this idea by demonstrating that generative neural networks can learn compact spatial and temporal representations of an environment in an unsupervised manner. Notably, their work showed that agents could even be trained entirely within internally generated simulations derived from these learned representations. More recently, LeCun~\cite{lecun2022path} identified the world model as a central architectural component for autonomous intelligence, proposing that it should infer missing information about the current state of the world and predict plausible future states resulting from imagined sequences of actions.

Since these seminal contributions, the field has expanded rapidly in both scope and ambition. In model-based reinforcement learning (MBRL), the Dreamer series~\cite{hafner2019dream, hafner2020mastering, hafner2025mastering} demonstrated that agents can learn complex behaviors entirely through latent imagination, scaling from simple control tasks to diverse domains using a unified algorithmic framework. In parallel, DeepMind’s MuZero~\cite{schrittwieser2020mastering} achieved superhuman performance in several challenging domains without access to explicit environment rules by learning an implicit model that predicts only planning-relevant quantities.
 
Beyond MBRL, OpenAI's Sora~\cite{liu2024sora} introduced large-scale video generation as a form of world simulation, sparking debate about whether such models constitute genuine world models~\cite{ding2025understanding}. Foundation model approaches from Meta (V-JEPA 2~\cite{assran2025vjepa2}), DeepMind (Genie~\cite{bruce2024genie}), and NVIDIA (Cosmos~\cite{nvidia2025cosmos}) have further demonstrated that large-scale self-supervised pretraining can yield actionable world simulators for robotic planning, interactive environment generation, and physical AI, respectively. These efforts align with LeCun's broader vision~\cite{lecun2022path} of a modular cognitive architecture centered on a configurable predictive world model trained via his proposed Joint-Embedding Predictive Architecture (JEPA), offering a theoretically grounded alternative to purely generative approaches. Collectively, these developments have transformed world models from a niche MBRL topic into a central pillar of the broader pursuit of AGI.

Within this expanding landscape, a particularly promising development is the integration of chain-of-thought (CoT) reasoning~\cite{wei2022chain} with world models. Traditional CoT represents reasoning as explicit sequences of natural language tokens, a process that can be computationally expensive and limited by the relatively low information density of discrete textual representations~\cite{chen2025latentcotsurvey}. Recent research has begun to shift this reasoning process into latent spaces, where world models provide the underlying substrate for multi-step deliberation. For instance, Coconut~\cite{hao2024coconut} introduces continuous thought representations that enable breadth-first reasoning directly in latent space. Similarly, LCDrive~\cite{tan2025lcdrive} integrates CoT-style reasoning with action planning by interleaving action-proposal tokens with latent world model predictions, allowing agents to simulate counterfactual futures before committing to a trajectory. Building on this direction, FutureX~\cite{xiang2025futurex} proposes an auto-think mechanism that dynamically activates a latent world model only when scene complexity warrants deliberative reasoning. Collectively, these approaches suggest that world models may function not merely as predictive simulators but also as reasoning engines in their own right—potentially replacing linguistic chains of thought with grounded, spatiotemporal chains of imagination (CoI).

Notably, the growing prominence of world models is also driven by increasing recognition of the fundamental limitations of large language models (LLMs) built on transformer architectures~\cite{vaswani2023attentionneed, lin2022survey}. Despite the remarkable success of models such as GPT-4~\cite{openai2023gpt4} and reasoning-augmented systems like o1~\cite{zhong2025evaluation} in tasks involving language understanding and code generation, these systems operate primarily within the discrete and relatively low-dimensional space of text tokens. As a result, they lack a grounded understanding of the continuous and high-dimensional physical world~\cite{lecun2022path}. Meanwhile, LLMs generally lack persistent world-state representations, have limited capacity for causal reasoning, and struggle with long-horizon planning—capabilities that are routinely exhibited by biological organisms. This gap reflects Moravec’s paradox~\cite{moravec1988mind}: while high-level cognitive tasks such as language processing and chess playing appear tractable for machines, sensorimotor competencies that have been refined through billions of years of biological evolution remain far more difficult to replicate.

In contrast, world models aim to address these limitations by learning to predict the consequences of actions within physical or simulated environments, thereby constructing internal representations of object dynamics and temporal evolution. By enabling agents to simulate possible futures and evaluate alternative action sequences, world models provide a foundation for planning, reasoning, and adaptive decision-making in complex environments. The growing institutional investment in this direction—exemplified by the establishment of Advanced Machine Intelligence (AMI)~\cite{lecun_ami_2025}, DeepMind’s continued development of the Genie family of models, and NVIDIA’s Cosmos platform—signals an emerging consensus that the next frontier of artificial intelligence lies in building systems capable of modeling and interacting with the world itself.

Yet, as research activity accelerates and world models attract attention from increasingly diverse communities, a fundamental question remains: what exactly constitutes a world model? In model-based reinforcement learning (MBRL), the concept is often defined narrowly as a learned transition function $\hat{T}(s_{t+1} \mid s_t, a_t)$ coupled with a reward predictor $\hat{R}(r_t \mid s_t, a_t)$~\cite{sutton2018reinforcement}. In the broader AI literature, however, world models are increasingly viewed as general-purpose simulators capable of supporting counterfactual reasoning, causal inference, and hierarchical planning~\cite{lecun2022path, li2025embodied}. This diversity of perspectives has led to a fragmented research landscape, in which communities working on latent dynamics models, generative video prediction, object-centric representations, and language-grounded planning often develop their approaches in relative isolation.

Importantly, various recent surveys have begun to organize this rapidly evolving field. For example, Ding et al.~\cite{ding2025understanding} examine world models from the perspective of understanding versus prediction, while Li et al.~\cite{li2025embodied} propose a three-axis taxonomy for embodied AI. Additional domain-specific surveys have explored applications in autonomous driving~\cite{guan2024world}, robotic manipulation~\cite{li2025steprobot}, and 3D/4D scene modeling~\cite{worldbench2025survey}. Nevertheless, existing reviews typically focus on specific methodological perspectives or application domains and do not simultaneously address the full spectrum of architectural paradigms, methodological families, reasoning mechanisms, and application contexts within a unified framework. The world models are opening up an underexplored domains, including medical imaging and educational measurement—further highlighting the need for a comprehensive, cross-disciplinary survey.

To address this gap, this survey provides a structured and comprehensive review organized along several complementary axes:

\begin{enumerate}

\item \textbf{Architecture} (Section~\ref{sec:architecture}): key architectural design choices, including representation type, dynamics modeling approach, input modality, learning paradigm, and downstream use cases.

\item \textbf{Methodological Families} (Section~\ref{sec:methodology}): major modeling paradigms such as state-space and recurrent latent models, variational and generative approaches, transformer-based architectures, object-centric and compositional models, physics-informed models, and language-augmented multimodal systems.

\item \textbf{Reasoning Strategies} (Section~\ref{sec:reasoning}): mechanisms for decision-making and planning, including imagination-based planning, policy learning within learned models, counterfactual reasoning, long-horizon and hierarchical planning, and planning under uncertainty.

\item \textbf{Application Domains} (Section~\ref{sec:applications}): representative application areas, including robotics, autonomous driving, video prediction and scene understanding, multimodal agents, reinforcement learning and games, scientific modeling, medical imaging and video analysis, and educational measurement.

\end{enumerate}

\begin{figure}[htbp]
    \centering
    \includegraphics[width=0.9\linewidth]{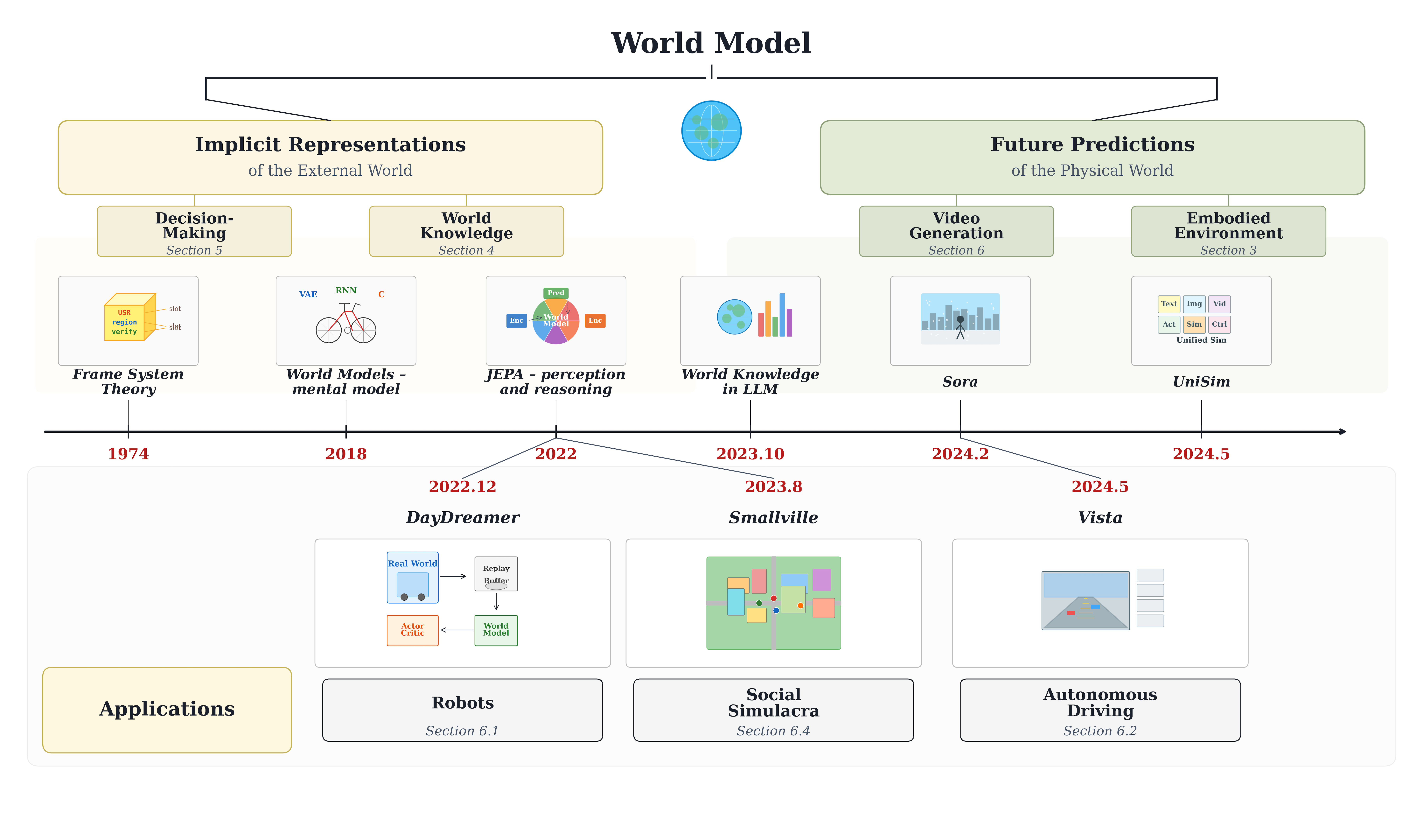}
    \caption{Overview of the world model landscape, organized into a conceptual taxonomy of implicit representations and future predictions, a historical timeline of milestone contributions from 1974 to 2024, and representative application-domain deployments spanning robotics, social simulation, and autonomous driving.}
    \label{fig:timeline}
\end{figure}

Figure~\ref{fig:timeline} summarizes this landscape visually, organized in three layers. \textit{Top:} The conceptual taxonomy divides world models into two complementary perspectives—implicit representations of the external world, encompassing decision-making (Section~5) and world knowledge learning (Section~4), and future predictions of the physical world, covering video generation (Section~6) and embodied environments (Section~3). \textit{Middle:} A historical timeline of milestone contributions, from Minsky's frame system theory (1974) through Ha and Schmidhuber's neural world models (2018), LeCun's JEPA (2022), world knowledge in LLMs (2023), to recent large-scale simulators including Sora and UniSim (2024). \textit{Bottom:} Representative application-domain deployments—DayDreamer for robotics (Section~6.1), Smallville for social simulacra (Section~6.4), and Vista for autonomous driving (Section~6.2)—illustrating the breadth of deployment contexts that world models now support. Additionally,  we further examine evaluation protocols and benchmarks (Section~\ref{sec:evaluation}), identify key challenges facing current approaches (Section~\ref{sec:challenges}), and outline promising directions for future research (Section~\ref{sec:future}). By providing a unified, multi-axis perspective, this survey aims to serve both as a comprehensive reference for researchers entering the field and as a roadmap for advancing world models toward robust, general-purpose autonomous intelligence.

\section{Background and Conceptual Foundations of World Model}

World models are state-of-the-art intelligent models that enable an intelligent agent to form compact representations of its environment and predict how that environment may evolve over time. In artificial intelligence, the modern idea is closely associated with model-based reinforcement learning and generative latent-dynamics modeling, where an agent learns not only how to act but also how the world changes in response to actions.

Conceptually, world models are rooted in a broader predictive view of intelligence. Rather than reacting only to current inputs, intelligent systems benefit from anticipating future states, estimating consequences of actions, and using internal simulation to guide behavior. This idea has strong connections to earlier traditions in cognitive science, neuroscience, and reinforcement learning, including predictive processing, predictive coding, and predictive representations. The following sections present the vital background, conceptual architecture, and foundations of World Model. 

\subsection{Definition and Basic Concepts}

In artificial intelligence, a world model is an internal predictive model that captures how an environment evolves over time and how that evolution depends on an agent’s actions. An early formulation by Schmidhuber described a model-building control system in which the controller was equipped with an additional module, the \textit{world model}, trained to predict future inputs from prior input–action pairs~\cite{schmidhuber1991}. In contemporary machine learning, the term has broadened while retaining this core idea: a world model is a learned representation of environment dynamics that supports prediction, simulation, and decision-making. Ha and Schmidhuber’s neural world-model framework made this formulation especially influential in modern deep learning by demonstrating that an agent can learn compressed spatial and temporal representations of an environment and then use them to support downstream control~\cite{ha2018world}.

To make this definition precise, let $o_t$ denote the observation at time $t$, $a_t$ the action, $r_t$ the reward, and $s_t$ a latent state intended to summarize the information necessary for predicting the future. A world model can then be written as a parameterized predictive system with parameters $\theta$ that approximates the environment dynamics

To formalize this definition, let $o_t$ denote the observation at time $t$, $a_t$ the action, $r_t$ the reward, and $s_t$ a latent state, intended to summarize the information necessary for predicting the future. A world model can then be expressed as a parameterized predictive system with parameters $\theta$ that approximates the environment dynamics:
\begin{equation}
p_\theta(s_{t+1}, o_{t+1}, r_t \mid s_t, a_t).
\end{equation}
In fully observed settings, one may identify $s_t$ with the true environment state. In partially observed settings, however, the model must infer a latent belief-like state from the history $h_t = (o_{\leq t}, a_{< t})$, for example through an encoder of the form

In fully observed settings, $s_t$ may coincide with the true environment state. In partially observed settings, however, the model must infer a latent, belief-like state from the interaction history $h_t = (o_{\leq t}, a_{< t})$, for example through an encoder of the form
\begin{equation}
s_t \sim q_\theta(s_t \mid h_t).
\end{equation}
This notation highlights an essential point: a world model need not reproduce the external world in full detail; rather, it must represent those aspects of experience that are sufficient for useful prediction and control.

The objective of a world model is therefore not merely to reconstruct observations, but to learn state representations that make future outcomes predictable. In simple settings, this may involve forecasting the next observation directly. In more realistic settings, particularly under partial observability, the model must maintain a latent state that summarizes past observations and actions sufficiently well to predict future observations, rewards, and other task-relevant signals. This perspective helps explain why later work shifted from raw observation prediction to latent dynamics modeling. PlaNet, for instance, learns environment dynamics from images in a compact latent space and explicitly combines deterministic and stochastic transition components, reflecting the insight that useful world models must track both persistent structure and uncertainty over multiple possible futures~\cite{hafner2019planet}. Dreamer extends this line of work by treating learned latent dynamics not only as a predictive model but also as a substrate for behavior learning through imagined rollouts~\cite{hafner2020dreamer}.

Most world models therefore contain several recurring components. First, they include a perception or representation module that compresses high-dimensional sensory inputs into a tractable state representation, such as a latent vector or token sequence. Second, they include an action-conditioned dynamics model that predicts how this latent state evolves over time. Third, many systems incorporate task-level heads, such as reward or continuation predictors, because control requires estimating not only what the world will look like, but also whether imagined trajectories are desirable or terminal. Ha and Schmidhuber instantiated this decomposition using a variational autoencoder for visual compression and a recurrent dynamics model for temporal prediction~\cite{ha2018worldmodels}. PlaNet and Dreamer refined the same blueprint into latent state-space models suitable for planning and policy optimization~\cite{hafner2019planet,hafner2020dreamer}, while Genie scaled the idea into a generative interactive environment built from a spatiotemporal video tokenizer, an autoregressive dynamics model, and a learned latent action interface~\cite{bruce2024genie}.

From a functional perspective, the central promise of world models is imagination. Once a model can simulate likely future trajectories under candidate actions, an agent can use those imagined trajectories to evaluate plans, improve a policy, or train partly or entirely within a learned environment rather than the real one. This is why world models are so closely connected to model-based reinforcement learning. Ha and Schmidhuber showed that policies could be trained inside a model-generated environment and then transferred back to the actual task~\cite{ha2018worldmodels}. Dreamer advanced this idea by learning behaviors purely through latent imagination, propagating value gradients through imagined trajectories in compact latent space~\cite{hafner2020dreamer}. Genie broadens the same concept beyond narrow task simulators by introducing action-controllable virtual worlds learned from unlabeled internet video, which the authors describe as a “foundation world model”~\cite{bruce2024genie}. Across these variants, the unifying principle is that the model serves as an internal sandbox for counterfactual interaction.

Formally, if a policy $\pi(a_t \mid s_t)$ acts inside the learned dynamics, the model can generate an imagined rollout
\begin{equation}
s_t, a_t, s_{t+1}, a_{t+1}, \dots, s_{t+H}
\end{equation}
for horizon $H$, with actions sampled from $\pi$ and transitions sampled from $p_\theta(s_{t+1} \mid s_t, a_t)$. Planning or policy learning can then optimize an expected return inside the model,
\begin{equation}
J(\pi)
=
\mathbb{E}_{p_\theta,\pi}\!\left[\sum_{k=0}^{H-1}\gamma^k r_{t+k}\right],
\end{equation}
where $\gamma \in [0,1)$ is the discount factor. This equation captures the operational role of a world model: it turns future interaction into a differentiable or at least simulatable object that can be searched, optimized, or evaluated before executing behavior in the real environment.

Several conceptual distinctions are useful when defining world models. One is the distinction between observation-space and latent-space models. Observation-space models attempt to predict future pixels, frames, or sensor readings directly, whereas latent-space models predict compressed hidden states that are typically more computationally efficient and more useful for planning. A second distinction is between deterministic and stochastic world models. Deterministic models are often simpler, but they may blur over genuinely uncertain futures; stochastic models are better suited to capturing ambiguity and multimodality. A third distinction concerns task-specific versus general-purpose world models. Earlier systems were usually trained on a single environment for planning or control, whereas newer systems such as Genie seek to learn broadly reusable, action-controllable generative environments from large-scale, weakly supervised data. These distinctions matter because they shape what the model can represent, how it is trained, and how its predictions can be used.

Meanwhile, a world model is not equivalent to a perfect simulator or a complete ontology of reality. Its value depends on whether it captures the aspects of the environment that matter for prediction and control. Ha and Schmidhuber explicitly noted that an unsupervised visual model may reproduce visually detailed but task-irrelevant structure while failing to capture features critical for successful behavior~\cite{ha2018worldmodels}. PlaNet likewise framed learned dynamics as a long-standing challenge because errors compound across multi-step prediction horizons, especially in image-based domains~\cite{hafner2019planet}. For this reason, the practical quality of a world model is typically judged not only by reconstruction fidelity, but also by whether it yields stable imagined rollouts, supports useful planning, improves sample efficiency, and generalizes beyond the exact trajectories observed during training. In this sense, the core idea of a world model is predictive abstraction: the model should compress experience into a form that is simple enough to simulate yet rich enough to support effective action.

A world model can also be described more compactly as a learned function 
$f_\theta$ that approximates the transition dynamics of an environment:
\begin{equation}
    \hat{s}_{t+1}, \hat{r}_{t+1} = f_\theta(s_t, a_t)
\end{equation}
where $s_t$ denotes the state (or observation) at time $t$, $a_t$ the action taken, $\hat{s}_{t+1}$ the predicted next state, and $\hat{r}_{t+1}$ the predicted reward. In practice, world models often operate in a learned latent space $z_t = \text{enc}(o_t)$ rather than directly over raw observations $o_t$, enabling compact representations and tractable long-horizon prediction.

Moreover, three properties distinguish a world model from a generic predictive model:
\begin{enumerate}
    \item \textbf{Action-conditioning}: The model predicts how the environment evolves \emph{in response to} specific actions, enabling counterfactual reasoning (``What would happen if I turned left instead of right?'').
    \item \textbf{Multi-step rollout}: The model can be applied autoregressively to generate trajectories of arbitrary length, supporting planning and simulation.
    \item \textbf{Utility for decision-making}: The model's predictions are used downstream---for policy optimization, planning, data augmentation, or safety verification---rather than serving as an end in themselves.
\end{enumerate}

\subsection{Key components of a world model}

Most modern world models operate within a partially observable Markov decision process (POMDP) framework $(\mathcal{S}, \mathcal{A}, T, R, \Omega, O, \gamma)$, where $\mathcal{S}$ is the state space, $\mathcal{A}$ the action space, $T$ the transition function, $R$ the reward function, $\Omega$ the observation space, $O$ the emission function, and $\gamma \in [0,1)$ a discount factor. A world model approximates $T$ and optionally $R$ and $O$. We will summarize multiple key components of World Model in the following four functional modules:

\textbf{Encoder.} The encoder maps raw, high-dimensional observations $o_t$ (images, point clouds, sensor readings) into a compact latent representation:
\begin{equation}
    z_t = q_\phi(z_t \mid o_{\leq t}, a_{<t})
    \label{eq:encoder}
\end{equation}
Notably, this compression step is essential, as raw observations are often too high-dimensional to support tractable multi-step prediction. The encoder can be either deterministic, such as a convolutional neural network, or stochastic, such as the posterior network in a variational autoencoder (VAE). For example, Ha and Schmidhuber~\cite{ha2018world} used a VAE-based encoder to compress $64\times 64$ image frames into 32-dimensional latent vectors. Similarly, the Dreamer family of models~\cite{hafner2020dreamer, hafner2021dreamerv2} employs a posterior encoder conditioned on observation history to infer both deterministic and stochastic components of the latent state.

\textbf{Dynamics model (transition predictor).} The dynamics model predicts the next latent state given the current state and action:
\begin{equation}
    \hat{z}_{t+1} = p_\theta(\hat{z}_{t+1} \mid z_t, a_t)
    \label{eq:dynamics}
\end{equation}
Importantly, the dynamics model constitutes the core of a world model. In RNN-based architectures~\cite{hafner2019planet}, it is typically implemented as a Recurrent State-Space Model (RSSM), which factorizes the latent state into a deterministic recurrent component $h_t$ that preserves long-range temporal dependencies, and a stochastic component $h_t$ that captures uncertainty in the environment. In transformer-based architectures~\cite{micheli2023iris, zhang2023storm}, the dynamics model is commonly formulated as an autoregressive transformer that predicts subsequent token(s) in a discrete latent sequence. In diffusion-based approaches~\cite{alonso2024diamond}, future states are generated through an iterative denoising process.

Besides, at a high level, current world models can be understood as a system composed of three tightly coupled components: a visual model, a memorial model, and a control model. Together, these components enable an agent to perceive its environment, retain and organize past information, and select actions on the basis of predicted future outcomes. This decomposition is especially useful because it clarifies how world models transform raw sensory inputs into structured internal representations that support reasoning, planning, and decision-making.

The visual model is responsible for perception and representation learning. Its primary role is to transform high-dimensional sensory observations, such as images, video frames, or other raw inputs, into a compact and informative latent representation. In many modern world models, this component is implemented using convolutional neural networks, variational autoencoders, vision transformers, or tokenizers that compress observations into latent vectors or discrete tokens. The importance of the visual model lies in its ability to filter out irrelevant perceptual details while preserving the features that are most critical for downstream prediction and control. Without such compression, directly modeling future trajectories in raw observation space would often be computationally prohibitive and statistically inefficient.

The memorial model serves as the temporal and predictive core of the architecture. Its function is to maintain a representation of past experience and to model how the latent state of the environment evolves over time. This component is called “memorial” because it provides the system with memory: it integrates current observations with historical context, allowing the agent to infer hidden structure, track temporal dependencies, and represent uncertainty about future states. In recurrent world models, the memorial model is often implemented using recurrent neural networks or recurrent state-space models, which combine deterministic memory states with stochastic latent variables. In more recent architectures, transformers and diffusion-based sequence models have also been used to capture long-range temporal dependencies and generate future latent trajectories. The memorial model is what allows a world model to go beyond static perception and function as a predictive simulator of environment dynamics.

The control model is the decision-making component. Given the latent state produced by the visual model and updated through the memorial model, the control model determines which action should be taken in order to maximize expected reward, achieve a specified goal, or satisfy a task constraint. In reinforcement learning settings, this component may take the form of a policy network, a value function, or a planning module that evaluates imagined future rollouts generated by the world model. More broadly, the control model translates predictive knowledge into purposeful behavior. Its effectiveness depends not only on the quality of the learned policy, but also on the fidelity of the perceptual and temporal representations provided by the other two components.

These three modules are not independent; rather, they operate as an integrated system. The visual model encodes the current observation into a latent state, the memorial model updates this latent state in light of prior context and predicts future states, and the control model uses these representations to evaluate alternatives and select actions. Their interaction enables the central capability of a world model: the ability to imagine possible futures before acting in the real environment. In this sense, the visual model answers the question of what is being observed, the memorial model addresses how the world changes over time, and the control model determines what should be done next.

This tripartite view also provides a useful conceptual framework for comparing different world model architectures. Some systems emphasize stronger visual encoding through powerful tokenization or representation learning, whereas others focus on more expressive memorial mechanisms for long-horizon prediction. Still others devote greater modeling capacity to the control component, especially in tasks that require sophisticated planning or policy optimization. Despite these differences, the visual–memorial–control decomposition captures a common structural logic underlying many world-model-based systems across reinforcement learning, robotics, autonomous systems, and scientific applications.

\textbf{Reward predictor.} The reward predictor estimates the scalar reward from the current latent state:
\begin{equation}
    \hat{r}_t = p_\psi(r_t \mid z_t)
    \label{eq:reward}
\end{equation}
Accurate reward prediction is essential for model-based RL, as policy optimization in imagination depends on the quality of predicted returns. MuZero~\cite{schrittwieser2020muzero} demonstrated that a world model whose dynamics operate entirely in a learned abstract space---predicting reward, value, and policy without ever reconstructing observations---suffices for superhuman performance in Go, chess, shogi, and Atari.

\textbf{Decoder (optional).} The decoder reconstructs observations from latent states:
\begin{equation}
    \hat{o}_t = p_\xi(o_t \mid z_t)
    \label{eq:decoder}
\end{equation}
Decoders serve two roles: (1) providing a reconstruction loss signal for training the encoder and dynamics model, and (2) enabling visualization of imagined trajectories. However, the decoder is architecturally optional. MuZero dispenses with it entirely, and JEPA-based models~\cite{lecun2022path, assran2023ijepa} predict in representation space rather than pixel space, sidestepping the computational burden and the blurriness associated with pixel-level reconstruction.

Beyond these four modules, several world models incorporate additional components. \emph{Continuation predictors} estimate episode termination probabilities. \emph{Discount predictors} model time-varying discount factors. DreamerV3~\cite{hafner2023dreamerv3} uses symlog-transformed predictions and categorical value representations to achieve cross-domain generality.

\subsection{Fundamental distinctions between World models and model-free RL}
A foundational distinction in reinforcement learning lies in whether the agent learns an explicit model of environmental dynamics or learns behavior directly from reward-driven interaction. This divide separates world-model-based approaches from model-free reinforcement learning (RL) and has important consequences for planning, sample efficiency, transfer, uncertainty handling, and interpretability \cite{moerland2023mbrl,ha2018worldmodels}.

At the most basic level, the two paradigms differ in \emph{what is being learned}. World-model methods learn predictive structure---for example, transition, observation, and reward dynamics in observation space or in a latent state space---so that future trajectories can be imagined or evaluated internally \cite{ha2018worldmodels,hafner2019planet,hafner2020dreamer}. By contrast, model-free RL typically learns a policy, value function, or both directly from interaction data, without requiring an explicit predictive model of the environment \cite{mnih2015dqn,schulman2017ppo,haarnoja2018sac}. In this sense, world models emphasize learning how the environment evolves, whereas model-free methods emphasize learning which action maximizes return.

This difference naturally induces a second distinction: \emph{planning versus direct policy execution}. Learned world models can be rolled forward to support online planning or latent imagination. PlaNet performs online planning in latent space, PETS uses learned dynamics for model-predictive control, Dreamer improves behavior by imagining trajectories in a learned latent model, and TD-MPC combines a latent dynamics model with trajectory optimization at decision time \cite{hafner2019planet,chua2018pets,hafner2020dreamer,hansen2022tdmpc}. In contrast, canonical model-free methods such as DQN, PPO, and SAC typically act through a direct forward pass of a learned policy or value-based decision rule rather than explicit search through hypothetical futures \cite{mnih2015dqn,schulman2017ppo,haarnoja2018sac}.

A third distinction concerns \emph{sample efficiency}. A recurrent motivation for world-model approaches is that a learned dynamics model allows the agent to reuse real experience more effectively through prediction, imagination, or planning. PILCO is a classic example of extreme data efficiency in model-based control, while PETS, MBPO, and Dreamer show that learned models can substantially improve performance per real environment step in modern continuous-control and visual-control settings \cite{deisenroth2011pilco,chua2018pets,janner2019mbpo,hafner2020dreamer}. Model-free methods, by construction, do not exploit an explicit learned simulator in this way; instead, they improve policies or value functions from real or replayed transitions alone \cite{mnih2015dqn,schulman2017ppo,haarnoja2018sac}.

However, the advantages of world models come with a characteristic liability: \emph{model bias}. If the learned dynamics are inaccurate, long imagined rollouts can drift away from the true environment and induce the policy to exploit modeling errors. PILCO explicitly frames model bias as a central problem and addresses it with probabilistic dynamics and uncertainty-aware planning, while MBPO shows that short branched rollouts can reduce the harmful effects of model exploitation in practice \cite{deisenroth2011pilco,janner2019mbpo}. Model-free RL avoids this particular failure mode because it does not rely on explicit multi-step predictions of environment dynamics, although it sacrifices some of the structural leverage available to model-based systems \cite{mnih2015dqn,haarnoja2018sac}.

The two paradigms also differ in \emph{representation learning}. In modern world-model architectures, latent states are trained not only to support action selection but also to summarize the hidden dynamics of the environment over time. World Models, PlaNet, and Dreamer all rely on compact latent representations that support prediction and imagination, rather than purely reactive control \cite{ha2018worldmodels,hafner2019planet,hafner2020dreamer}. By contrast, in standard model-free RL, learned representations are typically optimized only insofar as they improve policy or value estimation for the current task \cite{mnih2015dqn,schulman2017ppo,haarnoja2018sac}. This difference often makes world-model representations more naturally reusable for downstream planning or adaptation.

This distinction becomes especially important for \emph{generalization and transfer}. DARLA showed that disentangled representations can improve zero-shot transfer in RL, while Schema Networks demonstrated that a generative, causal model of environment dynamics can enable stronger transfer and combinatorial generalization than reactive baselines on structured tasks \cite{higgins2017darla,kansky2017schema}. More broadly, a learned world model can in principle be paired with new rewards, goals, or planners without discarding all previously acquired knowledge of environment dynamics. Model-free policies, in contrast, are typically more tightly coupled to the reward structure under which they were trained \cite{mnih2015dqn,schulman2017ppo}.

Another key difference is support for \emph{counterfactual and hypothetical reasoning}. Because world models specify how the world would evolve under alternative actions, they naturally support “what if” analysis. Woulda, Coulda, Shoulda formalized this idea by using structural causal models for counterfactual policy search from logged experience, and Schema Networks likewise emphasized generative causal structure for reasoning about unseen situations \cite{buesing2019counterfactual,kansky2017schema}. Standard model-free RL does not natively provide an explicit simulator for evaluating alternate futures; any such reasoning must be added externally or approximated indirectly through value estimation \cite{mnih2015dqn,schulman2017ppo}.

World models can also offer greater \emph{transparency of internal predictions}. In World Models and Dreamer, researchers can inspect reconstructions, latent rollouts, or imagined trajectories, while Schema Networks exposes an explicitly structured generative model of object interactions and consequences \cite{ha2018worldmodels,hafner2020dreamer,kansky2017schema}. Model-free policies such as DQN or SAC, by contrast, usually encode their knowledge more implicitly in policy and value parameters, which can make post hoc interpretation harder \cite{mnih2015dqn,haarnoja2018sac}.

A further distinction concerns \emph{uncertainty}. PETS uses probabilistic ensembles to capture uncertainty in learned dynamics, and PILCO models uncertainty directly through Gaussian process dynamics; more generally, approximate Bayesian techniques such as Monte Carlo dropout provide a practical route for predictive uncertainty estimation in deep models \cite{chua2018pets,deisenroth2011pilco,gal2016dropout}. Model-free RL can also reason about uncertainty, but typically through uncertainty in value estimation rather than explicit uncertainty over future world trajectories; examples include Bootstrapped DQN for deep exploration and distributional RL for learning return distributions \cite{osband2016bootstrappeddqn,bellemare2017distributional}. Thus, uncertainty in world models is often tied more directly to forecasting and planning.

Despite these contrasts, the boundary between world models and model-free RL is not absolute. Many strong modern agents are hybrid. Dreamer combines a learned world model with actor-critic learning in latent imagination, MBPO uses a learned model to supply synthetic data to an off-policy learner, TD-MPC combines latent dynamics with value learning for control, and SPR shows how predictive latent objectives can substantially improve otherwise model-free agents \cite{hafner2020dreamer,janner2019mbpo,hansen2022tdmpc,schwarzer2021spr}. The modern landscape is therefore better viewed as a continuum: world-model approaches place prediction and internal simulation at the center of control, whereas model-free methods place direct return optimization at the center.

In summary, the fundamental distinction is that world models learn an internal predictive account of the environment and use it for imagination, planning, or reasoning, whereas model-free RL learns to act effectively without requiring explicit environmental simulation. This difference propagates into downstream properties including planning ability, sample efficiency, vulnerability to model bias, transfer potential, counterfactual reasoning, and uncertainty handling \cite{moerland2023mbrl,ha2018worldmodels,hafner2020dreamer,mnih2015dqn}.

\subsection{Role of latent spaces in world models}

A defining design choice in modern world models is whether future predictions are made in the \emph{observation space} (e.g., raw pixels) or in a learned \emph{latent space}. Most successful world models operate in latent space, and this design choice has major implications for predictive accuracy, computational efficiency, and downstream task performance.

\textbf{Motivation for latent prediction.} Raw observations in real-world are high-dimensional and contain substantial task-irrelevant information. A $64\times 64$ RGB image has 12,288 dimensions; a $256\times 256$ image has 196,608. Predicting every pixel of every future frame is computationally expensive and forces the model to allocate capacity to visually complex but decision-irrelevant details (e.g., exact textures, lighting variations). Latent space prediction compresses observations into a compact representation $z_t$ that retains decision-relevant information while discarding perceptual noise, reducing the dimensionality by orders of magnitude (e.g., from 12,288 to 32--256 dimensions in typical implementations).

\textbf{Motivation for latent prediction.} Real-world observations are typically high-dimensional and contain substantial information that is irrelevant to decision-making. For example, a $64\times 64$ RGB image contains 12,288 dimensions, whereas a $256\times 256$ RGB image contains 196,608. Predicting every pixel of future frames is therefore computationally expensive and forces the model to devote representational capacity to visually complex but task-irrelevant details, such as textures or lighting variations. By contrast, latent-space prediction compresses observations into a compact representation, $z_t$ that preserves decision-relevant information while filtering out perceptual noise, often reducing dimensionality by several orders of magnitude (e.g., from 12,288 dimensions to 32--256 in typical implementations).

\textbf{Deterministic vs.\ stochastic latent spaces.} Early world models used deterministic encoders, but stochastic environments demand stochastic latent representations that capture aleatoric uncertainty. Ha and Schmidhuber~\cite{ha2018world} used a VAE with a Gaussian latent space, where the stochasticity of the latent code captures the inherent unpredictability of the environment. The RSSM~\cite{hafner2019planet} introduced a hybrid design: a deterministic recurrent state $h_t$ that preserves temporal memory, combined with a stochastic component $z_t$ sampled from a learned prior or posterior distribution. This dual structure has been adopted by the entire Dreamer line and many subsequent models. DreamerV2~\cite{hafner2021dreamerv2} further demonstrated that \emph{discrete} categorical latent variables (32 categorical distributions with 32 classes each, yielding a $32^{32}$ representational capacity) outperform continuous Gaussian latents on Atari games, likely because discrete representations better capture the discrete nature of game state transitions.

\textbf{Deterministic versus stochastic latent spaces.} Early world models often relied on deterministic encoders, but stochastic environments require representations that can capture aleatoric uncertainty. Ha and Schmidhuber~\cite{ha2018world}, for example, used a VAE with a Gaussian latent space, in which the stochastic latent code reflects the inherent unpredictability of the environment. The RSSM~\cite{hafner2019planet} introduced a hybrid formulation consisting of a deterministic recurrent state, $h_t$ that preserves temporal memory, together with a stochastic component, $z_t$ sampled from a learned prior or posterior distribution. This dual-state design has since become standard in the Dreamer family and many subsequent world models. DreamerV2~\cite{hafner2021dreamerv2} further showed that \emph{discrete} categorical latent variables can outperform continuous Gaussian latents on Atari benchmarks, likely because discrete representations better align with the discrete transition structure of many game environments.

\textbf{Continuous versus discrete tokenization.} An alternative to continuous latent spaces is discrete tokenization, in which observations are mapped to a finite vocabulary of learned codes. IRIS~\cite{micheli2023iris}, for instance, uses a VQ-VAE~\cite{kingma2014vae} tokenizer to transform image frames into discrete tokens and then models dynamics through next-token prediction with a transformer. This formulation creates a close parallel to language modeling and allows training with categorical cross-entropy objectives. STORM~\cite{zhang2023storm} adopts a hybrid strategy by combining stochastic continuous latents with transformer-based dynamics, thereby occupying an intermediate position between continuous and fully discrete approaches.

\textbf{Prediction in representation space (JEPA).} LeCun~\cite{lecun2022path} proposed the Joint-Embedding Predictive Architecture (JEPA) as a principled alternative to both pixel-space prediction and reconstruction-based latent modeling. In JEPA, the model predicts the representation of a future observation directly in embedding space rather than reconstructing the observation itself, and no decoder is required. This formulation avoids the bottleneck of pixel reconstruction, which often penalizes errors in task-irrelevant details, as well as the mode-averaging effects associated with pixel-wise losses. I-JEPA~\cite{assran2023ijepa} validated this approach for image representation learning, while V-JEPA~\cite{bardes2024vjepa} and V-JEPA 2~\cite{meta2025vjepa2} extended it to video, demonstrating strong performance in large-scale video understanding and zero-shot robotic planning.

\textbf{Latent-space structure and downstream performance.} The structure of the latent space directly influences the quality of imagined rollouts and, consequently, the effectiveness of policies trained on imagined trajectories. If the latent space fails to encode decision-relevant factors, such as object positions, velocities, or contact dynamics, then even an accurate dynamics model will generate uninformative predictions. Conversely, an overly detailed latent space that attempts to preserve every perceptual feature may waste capacity on irrelevant variation. The success of MuZero~\cite{schrittwieser2020muzero}, which learns a latent representation optimized solely for reward and value prediction without any reconstruction objective, illustrates that task-aligned latent spaces can outperform reconstruction-based alternatives. DreamerV3~\cite{hafner2023dreamerv3} further highlights the importance of carefully structured latent representations by using symlog transformations and categorical value distributions to maintain calibration across domains with highly variable reward scales.

Despite the success of latent world models, several open challenges remain, including: (1) \emph{latent-space collapse}, where distinct observations are mapped to identical codes and critical information is lost; (2) \emph{representation drift}, where the latent space shifts within training and undermines the consistency of imagined data; (3) \emph{disentanglement}, namely, learning latent dimensions that correspond to interpretable physical factors; and (4) \emph{scalability}, that is, designing latent spaces that remain both compact and expressive as environmental complexity increases.

\section{Categorization of World Models by Architecture}
\label{sec:architecture}
World model architectures can be analyzed along several complementary axes, each reflecting a key design dimension of how the model encodes observations, represents dynamics, handles uncertainty, and supports downstream decision-making. In brief, these axes provide a structured framework for comparing existing approaches and understanding the trade-offs among different architectural choices. World model architectures can be categorized along several complementary axes described below:

\subsection{Classification by representation} 
The choice of how to represent environment states is a fundamental design decision that determines what information the world model retains, how tractable multi-step prediction is, and what downstream tasks the model can support. We identify six principal representation families.

\subsubsection{Observation-space (pixel-level) representations.}
The most direct approach predicts future observations in raw observation space---typically RGB pixels or LiDAR point clouds. Early video prediction models~\cite{babaeizadeh2018sv2p, denton2018svg} operated in pixel space, as do recent diffusion-based world models such as DIAMOND~\cite{alonso2024diamond} and GameNGen~\cite{valevski2024gamengen}. The advantage is that no information is discarded: every visual detail is available for downstream use. The disadvantages are high dimensionality (a $256\times 256$ RGB frame has 196,608 dimensions), allocation of model capacity to decision-irrelevant details (textures, lighting), and the computational cost of multi-step prediction. Pixel-space models are most appropriate when visual fidelity is itself the objective, as in video generation or game simulation.

\subsubsection{Continuous latent representations.}
Most successful world models compress observations into continuous latent vectors via a learned encoder, then predict dynamics in this compact space. Ha and Schmidhuber~\cite{ha2018world} used a VAE to compress $64\times 64$ frames into 32-dimensional Gaussian latent codes. The RSSM~\cite{hafner2019planet} introduced a hybrid continuous latent state comprising a deterministic recurrent component $h_t$ (preserving temporal memory through GRU updates) and a stochastic component $z_t$ (sampled from a learned Gaussian distribution to capture environmental uncertainty):
\begin{equation}
    h_t = \text{GRU}(h_{t-1}, z_{t-1}, a_{t-1}), \quad z_t \sim \mathcal{N}(\mu_\theta(h_t), \sigma_\theta(h_t))
\end{equation}
This deterministic-stochastic split has been adopted across the entire Dreamer line~\cite{hafner2020dreamer, hafner2023dreamerv3} and many subsequent models (STORM~\cite{zhang2023storm}). Continuous latent spaces offer principled uncertainty quantification through the stochastic component and are well-suited for continuous control domains where smooth interpolation between states is meaningful. Their limitation is the tendency toward blurry reconstructions when a pixel-wise decoder is used, due to the Gaussian likelihood assumption.

\subsubsection{Discrete token representations.}
An alternative is to quantize observations into a finite vocabulary of discrete codes, typically via a VQ-VAE~\cite{kingma2014vae} tokenizer. IRIS~\cite{micheli2023iris} converts each image frame into a sequence of discrete tokens, then treats dynamics as next-token prediction---drawing a direct parallel with language modeling. DreamerV2~\cite{hafner2021dreamerv2} demonstrated that discrete categorical latent variables (32 categorical distributions with 32 classes each) outperform continuous Gaussian latents on Atari games, likely because discrete representations better capture the discrete nature of game state transitions. GAIA-1~\cite{hu2023gaia1} extends this approach to driving, tokenizing video frames and predicting future tokens autoregressively with a 9-billion-parameter transformer. Discrete representations enable the use of powerful autoregressive transformer architectures with categorical cross-entropy training, but sacrifice fine-grained spatial detail due to the quantization bottleneck.

\subsubsection{Joint-embedding (representation-space) predictions.}
LeCun~\cite{lecun2022path} proposed the Joint-Embedding Predictive Architecture (JEPA) as a principled alternative that avoids both pixel-space reconstruction and explicit tokenization. In JEPA, a predictor network maps the embedding of the current observation to the embedding of the next observation, with the target embeddings produced by an exponential moving average (EMA) encoder:
\begin{equation}
    \hat{z}_{t+1} = \text{predictor}_\theta(z_t), \quad z^{\text{target}}_{t+1} = \text{enc}_{\bar{\theta}}(o_{t+1})
\end{equation}
No decoder is used; the loss operates entirely in representation space. This avoids the pixel-reconstruction bottleneck (which penalizes models for failing to predict decision-irrelevant details) and the mode-averaging inherent in pixel-wise losses. I-JEPA~\cite{assran2023ijepa} validated this for images. V-JEPA~\cite{bardes2024vjepa} extended it to video, and V-JEPA 2~\cite{meta2025vjepa2}---pretrained on over one million hours of internet video---achieved state-of-the-art video understanding and enabled zero-shot robot planning. MuZero~\cite{schrittwieser2020muzero} can also be viewed as operating in a task-aligned representation space, since its latent dynamics are optimized solely for reward and value prediction without any reconstruction objective.

\subsubsection{Structured and object-centric representations.}
Rather than treating the world state as a monolithic vector or token sequence, object-centric models decompose it into a set of \emph{slots}, each representing a distinct entity with its own properties:
\begin{equation}
    s_t = \{\text{slot}^1_t, \text{slot}^2_t, \ldots, \text{slot}^N_t\}
\end{equation}
Kipf et al.~\cite{kipf2020cswm} introduced the Contrastive Structured World Model (C-SWM), which learns object-centric representations via contrastive learning and models dynamics through a graph neural network over object slots. RoboDreamer~\cite{zhou2024robodreamer} decomposes language instructions into primitive components and uses compositional diffusion models conditioned on each. DreMa~\cite{barcellona2025drema} integrates Gaussian Splatting with physics simulators for object-level scene manipulation. Object-centric representations support combinatorial generalization to novel object configurations and are more interpretable, but they scale poorly with the number of objects ($O(N^2)$ for pairwise interaction modeling) and assume that the environment can be cleanly decomposed into discrete entities.

\subsubsection{3D and occupancy-based representations.}
For domains with rich spatial structure, world models can represent the environment as 3D occupancy grids, voxels, or point clouds. OccWorld~\cite{wang2024occworld} uses a GPT-like model to autoregressively predict future 3D occupancy tokens for autonomous driving, enabling spatially consistent scene forecasting. Copilot4D~\cite{zhang2024copilot4d} learns to forecast LiDAR point clouds via discrete diffusion, achieving over 65\% reduction in Chamfer distance for 1-second prediction. Kong et al.~\cite{kong2025_3d4d} surveyed 3D and 4D world modeling approaches, establishing taxonomies across video-based, occupancy-based, and LiDAR-based generation. These representations are particularly valuable for autonomous driving and robotics, where 3D spatial reasoning is essential for safe planning, but they incur substantial memory and compute costs that scale cubically with spatial resolution.

\subsection{Classification by dynamics} 
\label{sec:dynamics}
 
A fundamental dimension for understanding world models lies in how they represent and learn \emph{dynamics}, i.e., the temporal evolution of latent states. 
Rather than focusing on architectural components, this perspective categorizes models based on the form of the transition mechanism, typically expressed as \( p(s_{t+1} \mid s_t, a_t) \). 
This viewpoint is particularly relevant for domains such as healthcare, where dynamics correspond to disease progression, treatment response, and longitudinal physiological changes.

In the following, we categorize world models according to their formulation of dynamics, ranging from explicit parametric transitions to implicit generative processes and adaptive memory-based mechanisms.

\subsubsection{Deterministic Dynamics}

Deterministic dynamics models assume that the next state is a single-valued function of the current state, i.e., \( s_{t+1} = f_\theta(s_t, a_t) \). 
This formulation simplifies learning and enables efficient multi-step rollout in latent space. Early world models such as World Models~\cite{ha2018worldmodels} and PlaNet~\cite{hafner2019planet} adopt this paradigm, using recurrent networks to propagate latent states. 
Subsequent works such as Dreamer~\cite{hafner2020dreamer} further demonstrate that deterministic latent dynamics can support long-horizon imagination and policy optimization.

However, deterministic formulations inherently struggle to capture multi-modal futures and uncertainty, often leading to averaged or overconfident predictions.  This limitation is particularly problematic in medical applications, where disease trajectories exhibit significant inter-patient variability.

\subsubsection{Stochastic Dynamics}

To address the limitations of deterministic transitions, stochastic dynamics models introduce latent variables to represent uncertainty, modeling transitions as distributions rather than point estimates.

In practice, this is often implemented through variational state-space models, where latent stochastic variables capture unobserved factors influencing temporal evolution.  Modern latent world models, including extensions of PlaNet and Dreamer, incorporate stochastic components to improve expressivity while retaining stable temporal structure.

Stochastic dynamics are especially important in medical settings, where uncertainty arises from both measurement noise and intrinsic biological variability.  Such models enable the representation of multiple plausible disease trajectories and support probabilistic reasoning in downstream tasks.

\subsubsection{Implicit Generative Dynamics}

An alternative formulation models dynamics implicitly through generative processes, without explicitly parameterizing transition functions. 
Diffusion-based models represent a prominent example, where temporal evolution is captured through iterative denoising processes over entire trajectories.

This paradigm has recently been extended to medical world modeling. 
MRI Contrast Enhancement Kinetics World Model (CEKWorld)~\cite{kong2026cekworld} formulates contrast-agent dynamics as a continuous-time generative process conditioned on non-contrast MRI. 
By leveraging latent diffusion, it synthesizes temporally dense sequences from sparsely sampled clinical data. 
To address the challenges of low temporal resolution, the model introduces spatiotemporal consistency constraints in latent space, mitigating content distortion and temporal discontinuity observed in sparse training regimes~\cite{kong2026cekworld}.

Implicit generative dynamics provide strong flexibility and expressivity, making them well-suited for modeling complex biological processes. 
However, their lack of explicit transition structure can limit interpretability and complicate integration with decision-making frameworks.

\subsubsection{Representation-space Predictive Dynamics}

Recent work explores modeling dynamics directly in representation space, focusing on predicting future latent embeddings rather than reconstructing observations.  This paradigm is closely related to joint-embedding predictive architectures (JEPA), where the objective is to learn invariant and predictive representations.

LeWorldModel~\cite{leworldmodel2024} adopts this approach by learning end-to-end predictive embeddings from pixel inputs, improving stability and scalability by avoiding explicit reconstruction.  Similarly, Brain-JEPA~\cite{brainjepa2024} extends this paradigm to neural data, modeling brain dynamics from fMRI signals using spatiotemporal masking and gradient-based target positioning.

In this formulation, dynamics are defined as transformations in representation space, capturing essential temporal structure while discarding irrelevant details.  This abstraction is particularly advantageous for high-dimensional data such as brain imaging, where direct pixel-level modeling is both inefficient and unnecessary.

\subsubsection{Memory-Augmented Dynamics}

A more recent direction extends dynamics modeling with external memory, allowing transitions to be conditioned on retrieved past experiences. 
In this formulation, dynamics are no longer purely parametric but depend on both learned representations and an external memory module.

CLARITY~\cite{clarity2025} exemplifies this paradigm by modeling context-aware disease trajectories in latent space. 
By incorporating patient-specific information and historical cases, the model dynamically adapts its predictions, enabling personalized forecasting of disease progression and treatment outcomes.

More broadly, medical world model frameworks~\cite{medicalwm2024} emphasize the importance of incorporating domain knowledge and patient-specific context into dynamics modeling. 
Such approaches are particularly effective under distribution shift, where purely parametric models may fail to generalize across populations or institutions.

Memory-augmented dynamics thus provide a mechanism for bridging global knowledge and local adaptation, a property that is essential for real-world deployment in healthcare settings.

\subsubsection{Discussion and Open Challenges}

The classification above highlights a spectrum of dynamics formulations, ranging from explicit deterministic transitions to implicit generative processes and adaptive memory-based systems. 
These approaches reflect different trade-offs between efficiency, expressivity, interpretability, and adaptability.

Despite significant progress, several open challenges remain. 
First, accurately modeling long-term dynamics under sparse and irregular observations remains difficult, particularly in medical domains where data acquisition is constrained. 
Second, balancing expressivity and interpretability is an ongoing challenge, as highly flexible generative models often lack clear physical or clinical meaning. 
Third, robustness under distribution shift continues to be a major concern, motivating the integration of memory and test-time adaptation mechanisms.

Finally, a promising direction lies in hybrid models that combine stochastic dynamics, implicit generative modeling, and memory augmentation. 
Such systems may offer a unified framework capable of capturing complex temporal processes while remaining adaptable to real-world variability.
\subsection{Classification by Modality}
\label{sec:modality}

The modality of input data constitutes a fundamental axis along which world models diverge in design, capability, and applicability. Since world models must internalize the dynamics of the environments they represent, the sensory channels through which environmental information is perceived directly shape both the learned representations and the downstream utility of the model. We organize existing world models into five broad modality categories: \textit{visual-only}, \textit{language-only}, \textit{3D geometric}, \textit{proprioceptive and tactile}, and \textit{multimodal fusion}. Table~\ref{tab:modality} summarizes representative works along this taxonomy.


\begin{table*}[ht]
\centering
\caption{Representative world models classified by input modality.}
\label{tab:modality}
\setlength{\tabcolsep}{5pt}
\renewcommand{\arraystretch}{1.15}
\small
\begin{tabular}{@{} l c c c c c c @{}}
\toprule
\textbf{Model}
  & \textbf{Visual} & \textbf{Language} & \textbf{LiDAR/3D} & \textbf{Proprioceptive} & \textbf{Tactile} & \textbf{Audio} \\
\midrule
%
\multicolumn{7}{@{}l}{\textit{\textbf{Visual-only}}} \\[2pt]
Dreamer (v1--v3)~\cite{hafner2019dream,hafner2020mastering,hafner2025mastering}
  & \checkmark & & & & & \\
IRIS~\cite{micheli2023transformers}
  & \checkmark & & & & & \\
DIAMOND~\cite{alonso2024diffusion}
  & \checkmark & & & & & \\
GameNGen~\cite{valevski2024diffusion}
  & \checkmark & & & & & \\
Sora~\cite{liu2024sora}
  & \checkmark & & & & & \\
Genie / Genie\,2~\cite{bruce2024genie}
  & \checkmark & & & & & \\
\midrule
%
\multicolumn{7}{@{}l}{\textit{\textbf{Language-only}}} \\[2pt]
RAP~\cite{hao2023reasoning}
  & & \checkmark & & & & \\
WebDreamer~\cite{gu2024your}
  & & \checkmark & & & & \\
WorldLLM~\cite{levy2025worldllm}
  & & \checkmark & & & & \\
\midrule
%
\multicolumn{7}{@{}l}{\textit{\textbf{3D Geometric (LiDAR / Point Cloud / Occupancy)}}} \\[2pt]
LiDARGen~\cite{zyrianov2022learning}
  & & & \checkmark & & & \\
LidarDM~\cite{zyrianov2024lidardm}
  & & & \checkmark & & & \\
OccWorld~\cite{zheng2023occworld}
  & & & \checkmark & & & \\
\midrule
%
\multicolumn{7}{@{}l}{\textit{\textbf{Proprioceptive and Tactile}}} \\[2pt]
DayDreamer~\cite{wu2023daydreamer}
  & \checkmark & & & \checkmark & & \\
MLA~\cite{mla2025multisensory}
  & \checkmark & \checkmark & \checkmark & & \checkmark & \\
Tactile-VLA~\cite{tactile_vla2025}
  & \checkmark & \checkmark & & & \checkmark & \\
\midrule
%
\multicolumn{7}{@{}l}{\textit{\textbf{Multimodal Fusion}}} \\[2pt]
GAIA-1 / GAIA-2~\cite{hu2023gaia,chitta2025gaia2}
  & \checkmark & \checkmark & & & & \\
Cosmos~\cite{nvidia2025cosmos}
  & \checkmark & \checkmark & & & & \\
V-JEPA\,2~\cite{assran2025vjepa2}
  & \checkmark & & & \checkmark & & \\
UniSim~\cite{yang2023unisim}
  & \checkmark & & \checkmark & & & \\
VLWM~\cite{chen2025vlwm}
  & \checkmark & \checkmark & & & & \\
\bottomrule
\end{tabular}
\end{table*}

\subsubsection{Visual-Only World Models}

The most extensively studied category of world models operates on visual observations, typically in the form of RGB images or video frames. This design choice is natural: vision provides a high-bandwidth, information-rich channel that captures spatial layout, object appearance, and temporal dynamics simultaneously. Pioneering work in model-based reinforcement learning established this paradigm. Ha and Schmidhuber~\cite{ha2018world} trained a variational autoencoder (VAE) and a recurrent neural network (RNN) on image observations to learn compact latent representations of visual environments, enabling agents to act within internally generated ``dreams.'' The Dreamer series~\cite{hafner2019dream, hafner2020mastering, hafner2025mastering} extended this approach through the Recurrent State-Space Model (RSSM), learning visual dynamics in a compact latent space and training policies entirely through imagined rollouts. More recently, DreamerV3~\cite{hafner2025mastering} demonstrated that a single, fixed configuration can master over 150 diverse tasks spanning continuous control, Atari games, and even Minecraft, establishing visual world models as general-purpose decision-making substrates.

An important architectural shift has been the adoption of transformer-based and diffusion-based dynamics models operating on visual tokens. IRIS~\cite{micheli2023transformers} discretizes visual observations using a VQ-VAE and models dynamics with an autoregressive transformer, achieving competitive performance on the Atari 100k benchmark. DIAMOND~\cite{alonso2024diffusion} replaces the autoregressive transition model with a diffusion process, generating the next observation conditioned on past frames and actions. This diffusion-based paradigm has been pushed further by GameNGen~\cite{valevski2024diffusion}, which demonstrated real-time interactive simulation of the classic game DOOM at over 20 frames per second using a fine-tuned Stable Diffusion model, achieving visual fidelity nearly indistinguishable from the actual game by human raters.

At the foundation model scale, video generation has been reframed as a form of world simulation. OpenAI's Sora~\cite{liu2024sora} demonstrated that scaling video diffusion transformers trained on diverse internet video can produce temporally coherent visual sequences that appear to respect physical constraints such as object permanence and rigid-body dynamics, although the extent to which such models constitute genuine world models remains debated~\cite{ding2025understanding}. DeepMind's Genie~\cite{bruce2024genie} took a complementary approach, learning a latent action space from unlabeled video in a self-supervised manner, enabling the generation of interactive and controllable 2D environments from single image prompts.

Despite their success, visual-only world models face inherent limitations. Raw pixel reconstruction is computationally expensive and may force models to allocate representational capacity to task-irrelevant visual details such as textures and lighting variations. Moreover, purely visual representations lack the semantic abstraction needed for high-level reasoning and struggle to capture aspects of the world state that are not directly observable in the image, such as object masses, material properties, or off-screen entities.

\subsubsection{Language-Only World Models}

A distinct but increasingly influential line of research employs natural language as the primary representational medium for world modeling. Rather than predicting future pixel values, language-only world models predict future \emph{states} expressed as textual descriptions, leveraging the semantic abstraction and compositional structure inherent in natural language. This paradigm capitalizes on the extensive world knowledge already encoded in large language models (LLMs) through pretraining on internet-scale text corpora.

Hao et al.~\cite{hao2023reasoning} proposed the Reasoning via Planning (RAP) framework, which repurposes an LLM to serve simultaneously as a world model---predicting state transitions---and as a reasoning agent, combined with Monte Carlo Tree Search for strategic exploration of the reasoning space. Notably, RAP on LLaMA-33B surpassed chain-of-thought prompting on GPT-4 in plan generation tasks, demonstrating that LLMs can serve as effective approximate world models when coupled with principled planning algorithms. In a similar spirit, WebDreamer~\cite{gu2024your} treats LLMs as world models for web-based environments, simulating the outcomes of candidate browser actions in natural language before committing to one, achieving substantial improvements over reactive baselines on benchmarks such as VisualWebArena.

Language-based world models offer several attractive properties. Natural language provides a semantically rich and human-interpretable interface, enabling straightforward inspection of predicted world states and facilitating human-in-the-loop correction. Furthermore, language-based predictions are computationally far less expensive to generate than pixel-level reconstructions, making them particularly suitable for high-level task planning over long horizons. Recent work on WorldLLM~\cite{levy2025worldllm} induces explicit natural-language hypotheses about transition regularities via Bayesian inference over agent experience, yielding an interpretable and low-dimensional world model that guides both forward prediction and curiosity-driven exploration.

However, language-only world models suffer from significant limitations. Textual descriptions are inherently lossy: they cannot capture fine-grained spatial relationships, continuous dynamics, or precise metric information. LLM-based world models also exhibit compounding prediction errors over long horizons, with accuracy degrading substantially as rollout length increases~\cite{gu2024your}. Additionally, these models inherit the well-documented limitations of LLMs, including hallucination, difficulty with precise numerical reasoning, and fragile performance on tasks requiring grounded physical understanding.

\subsubsection{3D Geometric World Models}

A third category of world models operates on three-dimensional geometric representations, including LiDAR point clouds, occupancy grids, and depth maps. These representations provide explicit 3D structural information that is complementary to the appearance-centric nature of RGB imagery and is especially critical for applications in autonomous driving and outdoor robotics where precise spatial reasoning is required.

LiDAR-based world modeling has advanced rapidly in recent years. LiDARGen~\cite{zyrianov2022learning} formulated point cloud generation as a denoising diffusion process in an equirectangular projection, producing physically plausible LiDAR scans that can be conditionally sampled for tasks such as point cloud densification without retraining. Subsequent diffusion-based approaches, including R2DM~\cite{nakashima2024lidar} and RangeLDM~\cite{hu2024rangeldm}, improved fidelity and scalability by operating in latent range-view spaces. LidarDM~\cite{zyrianov2024lidardm} further demonstrated compositional scene generation, synthesizing LiDAR data within realistic driving environments. Beyond generation, occupancy-based world models such as OccWorld~\cite{zheng2023occworld} predict future 3D occupancy states of a scene, providing a structured volumetric representation that unifies spatial prediction across both static and dynamic elements.

3D geometric world models are particularly valuable for autonomous driving, where a comprehensive survey of driving-specific approaches can be found in~\cite{guan2024world, worldbench2025survey}. These models can faithfully represent the geometry of roads, buildings, and vehicles, capturing information that is absent from monocular camera images. However, they typically lack semantic richness---a LiDAR point cloud encodes \emph{where} objects are but not necessarily \emph{what} they are or how they will behave---motivating the integration of 3D geometric data with other modalities.

\subsubsection{Proprioceptive and Tactile World Models}

Embodied agents, particularly robots, interact with their environments through sensory channels that extend well beyond vision. Proprioceptive signals---such as joint positions, velocities, and torques---provide direct information about the agent's own body state, while tactile sensors convey contact forces and surface properties during physical interaction. Incorporating these modalities into world models is essential for contact-rich manipulation tasks where visual observation alone is insufficient.

DayDreamer~\cite{wu2023daydreamer} demonstrated that Dreamer-style world models can be applied to physical robots by learning dynamics from a combination of visual observations and proprioceptive inputs, enabling a quadruped robot to learn locomotion gaits entirely from real-world interaction without simulation. Several recent works have further expanded the sensory repertoire. The Multisensory Language-Action (MLA) model~\cite{mla2025multisensory} proposes an encoder-free multimodal alignment scheme that repurposes the LLM backbone itself as a perception module, directly interpreting 2D images, 3D point clouds, and tactile tokens through positional correspondence. Similarly, Tactile-VLA~\cite{tactile_vla2025} and BiTLA~\cite{bitla2025} integrate tactile feedback with vision-language-action architectures, demonstrating improvements on contact-rich manipulation benchmarks.

The inclusion of proprioceptive and tactile modalities addresses a fundamental blind spot of vision-only approaches: the physics of contact. A robot grasping a fragile object must sense grip force in real time; a hand manipulating a deformable object must feel how the material yields. These signals are inherently low-dimensional compared to vision but carry information that is impossible to obtain from cameras alone. The challenge lies in the heterogeneity of these sensor modalities---tactile readings, joint encoders, and force-torque sensors each operate at different frequencies, dimensionalities, and noise characteristics---requiring careful architectural design for effective fusion.

\subsubsection{Multimodal Fusion World Models}

Perhaps the most ambitious and rapidly growing category encompasses world models that jointly process multiple input modalities, seeking to construct richer and more robust internal representations of the world by leveraging the complementary strengths of different sensory channels. The intuition is straightforward: the physical world is inherently multimodal, and an agent that can simultaneously process visual appearance, linguistic context, geometric structure, and physical feedback should develop a more complete understanding of environmental dynamics than one relying on any single channel.

In autonomous driving, GAIA-1~\cite{hu2023gaia} demonstrated a multimodal world model that conditions video generation on text descriptions and structured control inputs, enabling controllable synthesis of driving scenarios. Its successor, GAIA-2~\cite{chitta2025gaia2}, further advanced controllability through multi-view latent diffusion with fine-grained scene-level conditioning. UniSim~\cite{yang2023unisim} takes a different approach, constructing neural feature grids from recorded driving logs to simultaneously simulate both camera and LiDAR data in a closed-loop fashion, enabling the first closed-loop evaluation of autonomy systems on counterfactual safety-critical scenarios.

At the foundation model scale, NVIDIA's Cosmos~\cite{nvidia2025cosmos} provides a platform for training world foundation models that ingest video and text to produce generative simulators for physical AI applications. Meta's V-JEPA 2~\cite{assran2025vjepa2}, a 1.2 billion-parameter model trained via self-supervised video prediction using the Joint-Embedding Predictive Architecture (JEPA), achieves state-of-the-art performance on visual understanding benchmarks while simultaneously enabling zero-shot robot planning by integrating visual predictions with proprioceptive inputs. This model exemplifies LeCun's~\cite{lecun2022path} vision of a non-generative world model that predicts in abstract representation space rather than reconstructing raw pixels, avoiding the computational burden and representational waste of pixel-level generation.

A particularly notable recent development is the Vision-Language World Model (VLWM)~\cite{chen2025vlwm}, which perceives the environment through visual observations but predicts world evolution using language-based abstraction. By compressing video inputs into hierarchical text descriptions (termed ``Tree of Captions'') and training a language model to predict action-state trajectories, VLWM achieves strong performance on high-level task planning benchmarks while maintaining full interpretability of the predicted world states. This approach bridges the visual and language modalities in a complementary fashion: vision handles low-level perception while language carries the representational burden of dynamics prediction.

The key architectural challenge in multimodal world models lies in the fusion mechanism. Early approaches typically employ modality-specific encoders followed by concatenation or cross-attention in a shared latent space~\cite{mai2024efficient}. More recent designs explore tighter integration strategies, such as the Global Workspace approach~\cite{devillers2025multimodal}, which draws inspiration from cognitive neuroscience's Global Workspace Theory to learn a shared bottleneck representation across modalities, allowing each sensory channel to inform the world model while maintaining modality-specific processing pathways.

\subsubsection{Discussion and Open Challenges}

The modality landscape of world models reveals several important trends and unresolved tensions. First, there is a clear trajectory toward \textit{modality convergence}: state-of-the-art systems increasingly combine multiple sensory inputs, mirroring the multimodal nature of biological perception. Second, the choice of \textit{prediction target modality} is decoupling from the input modality---models may ingest video but predict in language space (VLWM) or ingest images but predict in abstract latent space (V-JEPA 2), suggesting that the optimal representation for world dynamics may not coincide with any raw sensory modality.

Several open challenges remain. \textit{Scalable multimodal alignment} across more than two or three modalities remains technically difficult, as the number of pairwise alignment objectives grows quadratically. \textit{Temporal synchronization} poses practical challenges when fusing modalities that operate at vastly different sampling rates (e.g., 30~Hz cameras, 1~kHz tactile sensors, and 10~Hz LiDAR). \textit{Missing modality robustness}---the ability of a world model to degrade gracefully when one or more sensory channels are unavailable---has received little attention but is critical for real-world deployment. Finally, the question of \textit{optimal prediction modality} remains open: it is not yet clear whether world models should predict in pixel space, language space, latent embedding space, or some hybrid thereof, as each choice entails distinct trade-offs among fidelity, interpretability, computational cost, and downstream task utility.

\subsection{Classification by learning paradigms}
\label{sec:learning_paradigms}

The learning paradigm—\emph{how} a world model acquires knowledge of environmental dynamics constitutes a fundamental classification axis that is orthogonal to representation type (Section~3.1), dynamics architecture (Section~3.2), and input modality (Section~3.3). The same architectural backbone, such as the RSSM underlying the Dreamer family, can be trained through online reinforcement learning~\cite{hafner2019dream}, offline batch learning, or large-scale self-supervised pretraining, yielding models with markedly different data requirements, generalization capabilities, and deployment characteristics. This section organizes existing world models into six broad learning-paradigm categories—self-supervised and unsupervised learning, online model-based reinforcement learning, offline and batch learning, foundation model pretraining and adaptation, supervised and imitation learning, and hybrid multi-stage paradigms—and analyzes the trade-offs each entails. Table~\ref{tab:learning_paradigms} provides a summary of representative models classified along this axis.

\subsubsection{Self-Supervised and Unsupervised Learning}
\label{sec:ssl}

The dominant learning paradigm for modern world models is self-supervised learning (SSL), in which models learn compact spatiotemporal representations from raw, unlabeled observations by optimizing pretext objectives that exploit the inherent structure of sensory data. This paradigm eliminates the need for reward signals or ground-truth labels, enabling training on large-scale datasets collected without task-specific annotation.

\paragraph{Reconstruction-based self-supervision.}
The earliest and most widely adopted SSL strategy for world models is learning through observation reconstruction. Ha and Schmidhuber~\cite{ha2018world} pioneered this approach by training a VAE in an unsupervised manner to compress image observations into compact latent vectors, paired with an RNN that predicted future latent distributions. The entire world model was learned without any task reward, relying solely on the reconstruction objective of the VAE and the predictive loss of the RNN. The Dreamer series~\cite{hafner2019dream, hafner2020mastering, hafner2025mastering} extended this paradigm through the RSSM, where the world model is trained to reconstruct observations, predict rewards, and predict episode terminations from latent states. A critical finding from DreamerV3~\cite{hafner2025mastering} is that the world model's representations are predominantly shaped by its \emph{task-agnostic reconstruction objective} rather than by task-specific reward gradients, suggesting that unsupervised dynamics learning provides the primary representational substrate upon which task-specific signals merely fine-tune behavior.

Reconstruction-based self-supervision also underlies video prediction approaches. Stochastic video prediction models such as SV2P~\cite{babaeizadeh2018sv2p} and SVG~\cite{denton2018svg} learn to predict future video frames by introducing variational latent variables to capture the inherent uncertainty of future outcomes. More recently, diffusion-based world models—including DIAMOND~\cite{alonso2024diamond} and GameNGen~\cite{valevski2024gamengen}—reformulate next-observation prediction as a denoising process, where the model learns to reverse a gradual noise-corruption procedure. The denoising score-matching objective is itself a form of self-supervision: no labels are required beyond the observations themselves.

While reconstruction-based approaches have proven remarkably effective, they carry a fundamental limitation: pixel-level reconstruction forces models to allocate representational capacity to task-irrelevant visual details such as textures, lighting variations, and background clutter, potentially at the expense of capturing decision-relevant dynamics.

\paragraph{Contrastive and non-contrastive representation learning.}
An alternative SSL strategy avoids pixel-level reconstruction entirely, instead learning structured representations through contrastive or non-contrastive objectives. Contrastive Structured World Models (C-SWM)~\cite{kipf2020cswm} learn object-centric state representations by contrasting temporally adjacent state-transition pairs against randomly sampled negatives, yielding physically meaningful latent spaces that support accurate multi-step prediction without any reconstruction decoder. Self-Predictive Representations (SPR)~\cite{schwarzer2020spr} encourage task-relevant latent consistency by training representations to predict their own future latent states under a learned transition model, achieving substantial improvements in data efficiency on the Atari 100k benchmark.

Non-contrastive methods, which avoid the need for explicit negative samples, have also gained traction. Approaches inspired by BYOL~\cite{grill2020bootstrap} and VICReg~\cite{bardes2021vicreg} learn representations by enforcing agreement between different augmented views of the same observation while using regularization mechanisms—such as exponential moving average targets, variance-invariance-covariance constraints, or stop-gradient operations—to prevent representational collapse. DINO-WM~\cite{zhou2025dinowm} builds world dynamics models on compact DINOv2 patch embeddings rather than raw pixels, predicting future patch features from offline behavioral trajectories and enabling zero-shot planning across diverse environments without reward models or expert demonstrations. This demonstrates that pre-trained visual representations encoding rich semantic structure can serve as powerful inductive biases for dynamics learning.

\paragraph{Joint-Embedding Predictive Architectures (JEPA).}
Perhaps the most theoretically principled SSL paradigm for world modeling is the Joint-Embedding Predictive Architecture (JEPA) proposed by LeCun~\cite{lecun2022path}. Unlike generative approaches that predict raw observations, JEPA learns to predict future \emph{representations} in a learned latent space, deliberately discarding unpredictable and task-irrelevant perceptual details. This design choice reflects a fundamental insight: the optimal representation for world dynamics may not coincide with any raw sensory modality, and forcing pixel-level prediction wastes model capacity on information that is irrelevant for downstream decision-making.

I-JEPA~\cite{assran2023ijepa} instantiated this principle for static images, predicting the representations of masked image regions from visible context entirely in latent space, and achieving competitive performance with substantially higher computational efficiency than both masked autoencoder and contrastive baselines. V-JEPA~\cite{bardes2024vjepa} extended this to the temporal domain by predicting masked spatiotemporal regions in video, learning motion-sensitive representations without pixel reconstruction. V-JEPA 2~\cite{assran2025vjepa2}, pretrained on over one million hours of internet video, demonstrated that scaling JEPA pretraining yields video representations with state-of-the-art understanding and prediction capabilities—achieving 77.3\% top-1 accuracy on Something-Something v2 for motion understanding and 39.7 recall-at-5 on Epic-Kitchens-100 for action anticipation. Critically, V-JEPA 2 can be post-trained into an action-conditioned world model (V-JEPA 2-AC) using fewer than 62 hours of unlabeled robot video, enabling zero-shot robotic planning on physical Franka arms without any task-specific training or reward. The JEPA paradigm thus represents a shift from ``generative'' to ``predictive'' self-supervised learning, where the model learns what aspects of the future are predictable and relevant rather than attempting to reconstruct every detail.

\subsubsection{Online Model-Based Reinforcement Learning}
\label{sec:online_mbrl}

In online model-based reinforcement learning (MBRL), the world model is learned concurrently with the agent's policy through iterative interaction with the environment. The agent alternates between collecting real experience, updating its internal model of the environment, and using that model to improve its behavior—either through synthetic data generation or direct planning.

\paragraph{Interleaved model learning and data augmentation.}
The conceptual roots of this paradigm trace back to Sutton's Dyna architecture~\cite{sutton1990dyna}, which interleaved real environment interaction with model-based updates by using the learned dynamics to generate synthetic transitions for value function training. In modern MBRL, this strategy has been refined to balance the benefits of model-generated data against the risks of compounding model errors. MBPO~\cite{janner2019mbpo} limits model rollouts to short horizons initiated from real states sampled from a replay buffer, providing a theoretical monotonic improvement guarantee while substantially improving data efficiency. STEVE~\cite{buckman2018sample} further explored how uncertainty-aware model rollouts can be incorporated into value estimation, using ensembles to quantify epistemic uncertainty and avoid exploitation of poorly modeled regions.

\paragraph{Imagination-based policy learning.}
A more ambitious use of online world models places imagined latent trajectories at the center of policy optimization. PlaNet~\cite{hafner2019planet} demonstrated that compact latent dynamics models can support planning directly from image observations, achieving 50\(\times\) data efficiency improvements over model-free methods using the Cross-Entropy Method for online trajectory optimization. The Dreamer family~\cite{hafner2019dream, hafner2020mastering, hafner2025mastering} extended this by combining learned latent dynamics with actor-critic learning entirely within imagined rollouts: after fitting a world model from real experience, the agent rolls forward imagined latent trajectories and uses them to train value and policy networks via backpropagation through the differentiable dynamics. Because the transition dynamics are implemented as differentiable neural networks, gradients of long-term value estimates can be propagated through imagined trajectories, effectively transforming imagination into a differentiable computation graph. DayDreamer~\cite{wu2023daydreamer} validated this paradigm on physical robots, demonstrating that a quadruped could learn locomotion gaits entirely from real-world interaction without any simulation.

\paragraph{Search-based online reasoning.}
An alternative online paradigm performs explicit planning at inference time through search procedures. MuZero~\cite{schrittwieser2020mastering} learns a latent dynamics model sufficient for predicting only planning-relevant quantities—rewards, values, and policy logits—and combines it with Monte Carlo Tree Search to achieve superhuman performance across Go, chess, shogi, and Atari without access to environment rules. This approach embodies the principle of \emph{value equivalence}~\cite{grimm2020value}: the latent space is shaped to preserve decision-relevant structure rather than to faithfully reconstruct observations. EfficientZero~\cite{ye2021efficientzero} improved data efficiency by incorporating self-supervised consistency objectives, while TD-MPC~\cite{hansen2022tdmpc} integrated learned latent dynamics with trajectory optimization for continuous control. Unlike amortized approaches that encode planning into a parametric policy, search-based methods explicitly recompute optimal actions for each state, trading increased computational cost at inference time for improved flexibility and robustness to out-of-distribution states.

\subsubsection{Offline and Batch Learning}
\label{sec:offline}

In safety-critical domains such as healthcare, autonomous driving, and industrial control, online exploration may be prohibitively expensive, dangerous, or ethically impermissible. Offline world model learning addresses this constraint by training entirely from pre-collected, static datasets without any further environment interaction.

The core challenge in this setting is \emph{distribution shift}: when the learned world model is used to simulate trajectories beyond the support of the training data, prediction errors can compound rapidly, potentially leading to ``model exploitation''—where the policy discovers unrealistically high-reward trajectories that exist only as artifacts of model inaccuracies. MOPO~\cite{yu2020mopo} addresses this by learning an ensemble of dynamics models and penalizing the policy's return estimates by the ensemble's disagreement, providing a conservative lower bound on the true expected return. The Diffusion World Model (DWM)~\cite{ding2025dwm} departs from the standard one-step autoregressive paradigm by predicting multi-step future states and rewards concurrently via diffusion, achieving a 44\% performance gain over one-step models on D4RL benchmarks. More recently, RLVR-World~\cite{chen2025rlvr} proposed using reinforcement learning with verifiable rewards as a post-training paradigm for world models, directly optimizing transition prediction metrics rather than maximum likelihood, demonstrating substantial gains across text games, web navigation, and robotic manipulation.

A critical insight from the offline learning literature is the tension between \emph{predictive accuracy} and \emph{decision utility}: a world model that achieves low reconstruction error does not necessarily provide the most useful representation for policy optimization~\cite{lambert2020objective}. Models may preserve visually rich but task-irrelevant details while failing to represent reward-critical structure, helping explain why policy performance does not always correlate with prediction quality. This observation has motivated growing interest in \emph{decision-oriented} world models that are explicitly optimized for downstream control utility.

\subsubsection{Foundation Model Paradigm: Large-Scale Pretraining and Adaptation}
\label{sec:foundation}

Inspired by the transformative success of the ``pretrain then fine-tune'' paradigm in natural language processing and computer vision, an emerging class of world models is trained at industrial scale on massive, heterogeneous datasets before being adapted to specific downstream tasks through lightweight fine-tuning or zero-shot transfer.

\paragraph{World foundation models.}
At the largest scale, World Foundation Models (WFMs) are pretrained on diverse, internet-scale data to learn general-purpose dynamics priors. OpenAI's Sora~\cite{liu2024sora} demonstrated that training a diffusion transformer on large-scale internet video yields temporally coherent visual sequences that appear to respect certain physical constraints, though subsequent analysis revealed these properties to be inconsistent~\cite{ding2025understanding, kang2025howfar}. NVIDIA's Cosmos~\cite{nvidia2025cosmos} released open-weight WFMs at 7B and 14B parameters trained on over 20 million hours of real-world video, with its successor Cosmos-Predict2.5~\cite{nvidia2025cosmos25} transitioning to flow-matching with RL-based post-training on 200 million clips. DeepMind's Genie series~\cite{bruce2024genie, parkerholder2024genie2, deepmind2026genie3} progressively scaled interactive world model generation from 2D environments to 3D navigable worlds generated from text prompts at 720p and 24 fps. The Large World Model (LWM)~\cite{liu2024lwm} pursued an orthogonal scaling axis by extending context length to one million tokens via RingAttention, enabling joint modeling of long video and text sequences. These efforts represent the convergence of video generation and world simulation at industrial scale, though whether scale alone can yield genuine physical understanding—as opposed to sophisticated pattern matching—remains actively debated.

\paragraph{Pretrain-then-fine-tune for world models.}
Below the scale of foundation models, the pretrain-then-fine-tune paradigm has been adopted to improve the data efficiency of task-specific world model training. WPT~\cite{xu2025wpt} pretrains a generalist world model on reward-free, non-expert offline data from diverse tasks, then fine-tunes it online for specific downstream tasks, demonstrating clear improvements in sample efficiency—particularly on hard exploration tasks where learning from scratch fails entirely. MOTO~\cite{rafailov2023moto} explores offline pretraining followed by online fine-tuning for model-based robot learning, while Vid2Act~\cite{pan2024model} uses a mixture world model pretrained on multi-task offline trajectories and adaptively transfers dynamics knowledge to novel tasks through domain-selective distillation.

\paragraph{Cross-domain transfer and adaptation.}
A key promise of pretrained world models is their ability to transfer knowledge across domains, embodiments, and environments. SimDist~\cite{levy2026simulation} pretrains world model components in simulation and finetunes only the dynamics model for real-world deployment, exploiting the modular structure of world models to directly target the sim-to-real dynamics gap. V-JEPA 2-AC~\cite{assran2025vjepa2} achieves zero-shot robotic planning on physical robots using a world model post-trained on fewer than 62 hours of unlabeled robot video—without collecting any data from the target environment or performing any task-specific training. The Open X-Embodiment initiative~\cite{openxembodiment2024} demonstrated that policies trained on a mixture of over one million trajectories from 22 robot embodiments achieve 50\% higher success rates than single-domain policies, establishing that cross-embodiment transfer is viable at scale.

\subsubsection{Supervised and Imitation Learning}
\label{sec:supervised}

While self-supervised and reinforcement learning paradigms dominate the world model literature, supervised and imitation learning remain important in settings where ground-truth labels, expert demonstrations, or structured prior knowledge are available.

\paragraph{Learning from demonstrations.}
In robotic manipulation, world models are frequently trained on large-scale teleoperated demonstrations. RT-1~\cite{brohan2023rt1} trained a scalable transformer on over 130,000 real-robot episodes covering more than 700 tasks, achieving 97\% success rate on trained tasks. While RT-1 and its successors (RT-2~\cite{zitkovich2023rt2}, OpenVLA~\cite{kim2024openvla}) are primarily framed as Vision-Language-Action policies rather than explicit world models, their training paradigm—supervised next-action prediction from observation-action pairs—is structurally analogous to world model training, and the implicit dynamics knowledge they acquire through large-scale demonstration learning increasingly overlaps with the capabilities of explicit world models.

\paragraph{Language and structured supervision.}
Language provides a rich form of weak supervision for world model learning. GAIA-1~\cite{hu2023gaia1} jointly conditions video generation on text descriptions and structured control inputs, where the text serves as a supervisory signal specifying scene composition and ego-vehicle behavior. WorldLLM~\cite{levy2025worldllm} induces explicit natural-language hypotheses about transition regularities via Bayesian inference over agent experience, yielding an interpretable world model guided by linguistic structure. Language-conditioned latent dynamics models such as Dynalang~\cite{lin2024dynalang} treat language understanding itself as emergent from the dynamics learning objective: the model learns to associate diverse language types—descriptions, instructions, corrections—with their environmental consequences through supervised prediction.

\paragraph{Physics-supervised learning.}
A distinct form of supervision arises when world models are trained on data generated by physics simulators with known ground-truth dynamics. Physics-informed model-based RL~\cite{ramesh2023pimbrl} incorporates known physics equations as soft or hard constraints within the world model training loop. Machine learning interatomic potentials such as SchNet~\cite{schutt2017schnet} and NequIP~\cite{batzner2022nequip} are trained on quantum-mechanical calculations to approximate potential energy surfaces, functioning as supervised world models of molecular dynamics. These approaches trade generality for physical accuracy and are discussed further in Section~4.5.

\subsubsection{Hybrid and Multi-Stage Learning Paradigms}
\label{sec:hybrid}

In practice, the most effective world model systems increasingly combine multiple learning paradigms in staged or interleaved training pipelines, leveraging the complementary strengths of each approach.

\paragraph{Self-supervised pretraining followed by RL fine-tuning.}
The most prominent hybrid paradigm combines large-scale self-supervised pretraining with task-specific reinforcement learning. V-JEPA 2~\cite{assran2025vjepa2} exemplifies this two-stage approach: a 1.2-billion-parameter video encoder is first pretrained via self-supervised masked prediction on over one million hours of internet video, then post-trained as an action-conditioned world model (V-JEPA 2-AC) using a small amount of robot interaction data, with planning performed via the Cross-Entropy Method at deployment. Cosmos-Predict2.5~\cite{nvidia2025cosmos25} follows a similar trajectory: self-supervised pretraining via flow-matching is followed by RL-based post-training to improve physical plausibility. The finding from DreamerV3~\cite{hafner2025mastering} that world model representations are predominantly shaped by unsupervised reconstruction objectives further motivates this paradigm, suggesting that task-agnostic pretraining on large unlabeled datasets could substantially reduce the online interaction budget required for downstream task learning.

\paragraph{Simulation pretraining with real-world adaptation.}
A complementary hybrid approach leverages the abundant data available in simulation to pretrain world models that are subsequently adapted to the real world. DayDreamer~\cite{wu2023daydreamer} demonstrated that world models can bridge this gap by learning directly on physical robots, while SimDist~\cite{levy2026simulation} takes a modular approach—pretraining the full world model stack in simulation and finetuning only the dynamics component using a small number of real-world trajectories. Domain randomization and curriculum learning strategies further support this paradigm by training world models across distributions of simulated environments to improve robustness to real-world variation.

\paragraph{Multi-task and multi-modal joint training.}
The Unified World Model (UWM)~\cite{he2025uwm} couples video prediction and action generation through a shared diffusion process during pretraining on large-scale robotic datasets, enabling a single model to serve simultaneously as a simulator and a policy. Motus~\cite{motus2024} introduces a Mixture-of-Transformer architecture integrating three specialized experts—for understanding, video generation, and action prediction—with flexible mode-switching within a single framework, achieving a 45\% absolute improvement over prior vision-language-action baselines by pretraining on unlabeled video with optical flow as an embodiment-agnostic action proxy.

\subsubsection{Discussion and Open Challenges}
\label{sec:lp_discussion}

\begin{table*}[htbp]
\centering
\caption{Representative world models classified by learning paradigm.}
\label{tab:learning_paradigms}
\setlength{\tabcolsep}{5pt}
\renewcommand{\arraystretch}{1.2}
\small
\begin{tabularx}{\textwidth}{@{} >{\raggedright\arraybackslash}p{4.2cm} >{\raggedright\arraybackslash}X c c c c @{}}
\toprule
\textbf{Learning Paradigm} & \textbf{Representative Models} & \textbf{Data Req.} & \textbf{Labels} & \textbf{Generalization} & \textbf{Compute} \\
\midrule
\multicolumn{6}{@{}l}{\textit{\textbf{Self-Supervised / Unsupervised}}} \\[1pt]
~~Reconstruction-based & Dreamer v1--v3~\cite{hafner2019dream,hafner2020mastering,hafner2025mastering}, DIAMOND~\cite{alonso2024diamond}, GameNGen~\cite{valevski2024gamengen} & Med--High & None & Moderate & Moderate \\
~~Contrastive / Non-contrastive & C-SWM~\cite{kipf2020cswm}, SPR~\cite{schwarzer2020spr}, DINO-WM~\cite{zhou2025dinowm} & Med & None & Moderate & Low--Med \\
~~JEPA & I-JEPA~\cite{assran2023ijepa}, V-JEPA~\cite{bardes2024vjepa}, V-JEPA 2~\cite{assran2025vjepa2} & Very High & None & High & High \\
\midrule
\multicolumn{6}{@{}l}{\textit{\textbf{Online Model-Based RL}}} \\[1pt]
~~Data augmentation (Dyna-style) & Dyna-Q~\cite{sutton1990dyna}, MBPO~\cite{janner2019mbpo}, STEVE~\cite{buckman2018sample} & Low & Reward & In-domain & Moderate \\
~~Imagination-based & PlaNet~\cite{hafner2019planet}, Dreamer~\cite{hafner2019dream}, DayDreamer~\cite{wu2023daydreamer} & Low--Med & Reward & In-domain & Moderate \\
~~Search-based & MuZero~\cite{schrittwieser2020mastering}, EfficientZero~\cite{ye2021efficientzero}, TD-MPC~\cite{hansen2022tdmpc} & Low--Med & Reward & In-domain & Med--High \\
\midrule
\multicolumn{6}{@{}l}{\textit{\textbf{Offline / Batch Learning}}} \\[1pt]
& MOPO~\cite{yu2020mopo}, DWM~\cite{ding2025dwm}, RLVR-World~\cite{chen2025rlvr} & Medium & Reward & Data-limited & Low--Med \\
\midrule
\multicolumn{6}{@{}l}{\textit{\textbf{Foundation Model Pretraining}}} \\[1pt]
& Sora~\cite{liu2024sora}, Cosmos~\cite{nvidia2025cosmos}, Genie~\cite{bruce2024genie}, LWM~\cite{liu2024lwm} & Very High & None/Weak & Cross-domain & Very High \\
\midrule
\multicolumn{6}{@{}l}{\textit{\textbf{Supervised / Imitation}}} \\[1pt]
& RT-1~\cite{brohan2023rt1}, GAIA-1~\cite{hu2023gaia}, Dynalang~\cite{lin2024dynalang}, WorldLLM~\cite{levy2025worldllm} & Med--High & Required & In-domain & Moderate \\
\midrule
\multicolumn{6}{@{}l}{\textit{\textbf{Hybrid / Multi-Stage}}} \\[1pt]
& V-JEPA 2-AC~\cite{assran2025vjepa2}, Cosmos-Predict2.5~\cite{nvidia2025cosmos25}, UWM~\cite{he2025uwm}, Motus~\cite{motus2024} & Flexible & Partial & High & High \\
\bottomrule
\end{tabularx}
\end{table*}

Table~\ref{tab:learning_paradigms} presents representative world models classified by learning paradigms, where Data req.'' indicates the typical scale of training data, Labels'' specifies whether task-specific supervision is required, and ``Generalization'' characterizes the breadth of transfer capability.

The six learning paradigms surveyed above reveal several important cross-cutting trends and unresolved tensions.

First, there is a clear trajectory toward \textit{paradigm convergence}: the most capable world models increasingly combine self-supervised pretraining with task-specific reinforcement learning or supervised fine-tuning, mirroring the ``pretrain then adapt'' recipe that has proven transformative in NLP and computer vision. The finding that DreamerV3's representations are dominated by unsupervised objectives~\cite{hafner2025mastering} provides theoretical motivation for this trajectory, suggesting that the bulk of world knowledge can be acquired from unlabeled data.

Second, the question of \textit{objective alignment} remains largely unresolved. Self-supervised objectives such as pixel reconstruction and denoising are not necessarily aligned with downstream decision-making requirements~\cite{lambert2020objective}. Decision-oriented approaches—exemplified by MuZero's value-equivalence principle~\cite{grimm2020value} and JEPA's selective prediction in representation space~\cite{lecun2022path}—offer principled alternatives, but a unified framework that systematically balances predictive accuracy with decision utility is still lacking.

Third, the emergence of \textit{scaling laws for world models} represents a critical open question. Foundation models such as Cosmos and Sora demonstrate that larger models trained on more data yield improved generation quality, but whether this improvement translates to genuine physical understanding—rather than more sophisticated pattern matching—remains contested. Controlled experiments by Kang et al.~\cite{kang2025howfar} suggest that video generation models exhibit case-based generalization rather than abstract physical reasoning, highlighting the need for evaluation protocols that distinguish statistical regularity from causal understanding.

Fourth, \textit{data efficiency and data scale} present a fundamental tension. JEPA and foundation model paradigms achieve strong cross-domain generalization but require millions of hours of video data for pretraining. Online MBRL methods achieve remarkable data efficiency—PlaNet demonstrated 50\(\times\) improvements over model-free methods—but generalize poorly beyond their training distribution. Bridging this gap through sample-efficient transfer from pretrained representations to novel environments is among the most important open problems.

Fifth, \textit{continual and lifelong learning} of world models has received comparatively little attention. Deployed world models must adapt to evolving environments, new objects, and shifting dynamics without catastrophically forgetting previously acquired knowledge. Current systems are almost universally trained in a single phase and frozen at deployment, a limitation that fundamentally constrains their real-world applicability.

Finally, the interplay between learning paradigm and \textit{causal understanding} warrants deeper investigation. Most self-supervised world models learn correlational rather than causal models of dynamics, as demonstrated by the failure modes identified in LLM-based world simulators~\cite{wang2024bytesized32} and video generation models~\cite{kang2025howfar}. Whether architectural innovations, training objectives, or data curation strategies can enable world models to acquire genuine causal understanding through any learning paradigm remains one of the field's most fundamental open questions.

\subsection{Classification by downstream use}

While prior subsections have organized world models by their architectural backbone and learning paradigm, an equally revealing lens is their \emph{downstream application}. The same foundational idea-learning an internal model of environment dynamics and using it to anticipate future states-manifests very differently depending on whether the end goal is mastering a board game, navigating an urban intersection, folding laundry, diagnosing tumors, generating cinematic video, or reasoning over natural language. Examining world models through the lens of downstream use exposes not only domain-specific design constraints (observation modality, action space, safety requirements, real-time latency budgets) but also recurring architectural motifs that transcend individual fields. In what follows, we survey six major application families, highlight representative systems in each, and close with a cross-cutting discussion of open challenges.

\subsubsection{Reinforcement Learning and Planning}
\label{sec:downstream:rl}

Reinforcement learning (RL) and game-playing remain the original and most mature proving ground for world models. PlaNet~\cite{hafner2019planet} demonstrated that a recurrent state-space model trained entirely from image observations could support effective planning via the cross-entropy method, eliminating the need for a model-free policy altogether. The Dreamer line of work extended this idea to actor-critic learning entirely within imagined trajectories: Dreamer~\cite{hafner2020dreamer} introduced latent imagination for continuous-action tasks, DreamerV2~\cite{hafner2021dreamerv2} replaced Gaussian latents with discrete representations and matched model-free performance on the Atari benchmark, and DreamerV3~\cite{hafner2023dreamerv3} introduced a suite of robustness techniques-symlog predictions, layer normalization, and free-bits KL balancing-that allowed a single set of hyperparameters to master domains ranging from Minecraft to continuous locomotion.

In parallel, the MuZero family~\cite{schrittwieser2020muzero} demonstrated that a learned dynamics model, combined with Monte Carlo tree search (MCTS), could achieve superhuman performance in Go, chess, shogi, and Atari without access to ground-truth game rules. EfficientZero~\cite{ye2021efficientzero} augmented MuZero with self-supervised temporal consistency losses and achieved human-level Atari performance with only two hours of real-time experience, dramatically improving sample efficiency.  For continuous control, TD-MPC~\cite{hansen2022tdmpc} proposed learning an implicit latent dynamics model jointly optimized with temporal-difference objectives and local trajectory optimization,  and its successor TD-MPC2~\cite{hansen2023td} scaled this paradigm to a single 317M-parameter world model capable of solving 80 diverse continuous-control tasks,  establishing a new benchmark for generalist model-based control. A persistent limitation across RL world models is compounding prediction error over long horizons, which degrades both planning and policy optimization quality.

\subsubsection{Autonomous Driving}
\label{sec:downstream:driving}

Autonomous driving presents world models with a distinctive set of challenges: multi-agent interaction, long-tail safety scenarios, multi-sensor fusion, and the need to generate physically plausible future states at both the geometric and photometric level. GAIA-1~\cite{hu2023gaia} was among the first large-scale generative world models for driving, using a video diffusion architecture conditioned on text, action, and map inputs to produce realistic multi-second driving rollouts. Its successor, GAIA-2~\cite{russell2025gaia}, extended the framework to controllable multi-view generation with improved geometric consistency. DriveDreamer~\cite{wang2023drivedreamer} adopted a two-stage approach-first learning structured traffic dynamics, then generating video conditioned on predicted layouts-and DriveDreamer-2~\cite{zhao2024drivedreamer2} further incorporated large language models for diverse scenario description and generation. WoVoGen~\cite{lu2024wovogen} advanced multi-camera consistency through world-volume-aware diffusion, explicitly lifting generation into 3D to enforce cross-view coherence.

Beyond video-centric formulations, several works have explored geometric world models for driving. OccWorld~\cite{zheng2024occworld} learns a 3D occupancy-based world model that jointly predicts future scene occupancy and ego-vehicle motion, enabling downstream planning in a structured volumetric representation.  Vista~\cite{gao2024vista} demonstrated that a single autoregressive video world model can generalize across diverse driving conditions while offering versatile controllability over viewpoint, dynamics, and weather.  UniSim~\cite{yang2023unisim} pushed the boundary further by learning a universal real-world simulator that can render novel viewpoints and simulate the visual consequences of arbitrary agent actions, bridging the gap between world modeling and closed-loop simulation.  A key open question in driving world models is the extent to which photorealistic generation translates into improved downstream planning performance versus merely serving as a data-augmentation tool.

\subsubsection{Robotics and Embodied AI}
\label{sec:downstream:robotics}

Transferring world models from simulation to physical hardware introduces challenges of partial observability, noisy actuators, and the prohibitive cost of real-world data collection. DayDreamer~\cite{wu2023daydreamer} provided the first demonstration that the Dreamer framework could learn locomotion and manipulation behaviors entirely on physical robots, achieving quadruped walking in one hour and robotic arm pick-and-place in ten minutes of real interaction-without any simulation pre-training.  RoboDreamer~\cite{zhou2024robodreamer} extended world models to compositional multi-step manipulation by learning compositional world dynamics that decompose complex instructions into sequenced sub-goals, enabling robot imagination over long task horizons.

A complementary line of research focuses on learning general-purpose visual representations that implicitly encode world dynamics. V-JEPA~2~\cite{assran2025vjepa2} pre-trained a joint-embedding predictive architecture on over one million hours of internet video, then post-trained an action-conditioned predictor (V-JEPA~2-AC) with fewer than 62 hours of unlabeled robot trajectories. The resulting model enabled zero-shot prehensile manipulation on Franka arms in new lab environments via model-predictive control-without any task-specific reward or additional data collection.  While not world models per se, the Robotics Transformer series-RT-1~\cite{brohan2023rt1} and RT-2~\cite{zitkovich2023rt2}-demonstrated that large-scale transformer policies trained on diverse robot data can exhibit emergent generalization,  and RT-2 in particular showed that vision-language pre-training allows robot policies to inherit semantic reasoning from web-scale data. The interplay between explicit world-model planning (DayDreamer, RoboDreamer) and implicit world knowledge encoded in large foundation models (V-JEPA~2, RT-2) remains an active frontier, with hybrid architectures increasingly favored.

\subsubsection{Healthcare and Medical Imaging}
\label{sec:downstream:healthcare}

Healthcare represents a nascent yet high-impact frontier for world models, where the ability to simulate disease progression or physiological dynamics can directly inform clinical decision-making. CEKWorld~\cite{kong2026cekworld} introduced the first world model for MRI contrast enhancement kinetics, learning to simulate the spatiotemporal dynamics of contrast agent distribution from a single non-contrast scan.  By modeling pharmacokinetic evolution through spatiotemporal consistency learning, CEKWorld eliminates the need for exogenous contrast injection, reducing patient risk and cost while enabling continuous temporal-resolution imaging.  CLARITY~\cite{ding2025clarity} proposed a latent-space world model for disease trajectory forecasting, conditioning predictions on treatment decisions to support treatment planning in oncology.  Similarly, MeWM~\cite{yang2025mewm} demonstrated generative simulation of tumor evolution conditioned on clinical interventions, improving optimal treatment protocol selection by 13\% in F1 score. 

Beyond treatment planning, self-supervised world models have entered neuroscience. Brain-JEPA~\cite{dong2024brain} applied a JEPA framework with gradient positioning and spatiotemporal masking to fMRI data, learning brain dynamics representations that transfer to diverse downstream tasks.  A recent survey by Qazi et al.~\cite{qazi2025beyond} provides the first systematic review of world models in healthcare, organizing the field into four capability levels: temporal prediction, action-conditioned prediction, counterfactual rollouts, and planning or control.  The survey highlights that most medical world models remain at level one or two, with counterfactual reasoning and closed-loop planning representing largely unrealized opportunities.  Critical challenges include the extreme scarcity of longitudinal paired data, the need for rigorous uncertainty quantification in clinical deployment, and regulatory barriers to using generative models in diagnostic pipelines.

\subsubsection{Video Generation and Creative Simulation}
\label{sec:downstream:video}

The convergence of world modeling and video generation has produced some of the most publicly visible systems in AI. OpenAI's Sora~\cite{brooks2024video} framed video generation explicitly as \emph{world simulation}, using a diffusion-transformer architecture to produce photorealistic videos with emergent physical behaviors such as object persistence and fluid dynamics,  though it notably struggles with fine-grained physical consistency. Genie~\cite{bruce2024genie} from DeepMind approached interactive world modeling by learning a latent action space entirely from unlabeled internet video, enabling playable 2D environments generated from a single image.  Genie~2~\cite{parkerholder2024genie2} scaled this concept to photorealistic 3D environments with persistent geometry and object interactions,  positioning foundation world models as a potential replacement for hand-authored game engines.

In the domain of interactive simulation, GameNGen~\cite{valevski2024diffusion} demonstrated that a diffusion model conditioned on action sequences could simulate the DOOM game engine in real time at over 20 frames per second,  with visual quality nearly indistinguishable from the original engine according to human evaluators.  DIAMOND~\cite{alonso2024diamond} showed that diffusion-based world models preserve visual details critical for downstream RL performance in Atari, outperforming prior discrete-token world models.  NVIDIA's Cosmos~\cite{nvidia2025cosmos} introduced a world foundation model platform explicitly designed for physical AI applications, providing pre-trained video generation backbones that downstream systems can fine-tune for robotics, driving, or creative tasks. A fundamental tension in this space is whether video-generation world models truly learn physical laws or merely produce visually plausible hallucinations-a distinction with profound implications for safety-critical downstream uses.

\subsubsection{Language Reasoning and Decision-Making}
\label{sec:downstream:language}

An emerging research direction reinterprets large language models (LLMs) themselves as world models, leveraging their latent knowledge for planning and reasoning. RAP~\cite{hao2023rap} formalized this perspective by treating an LLM as a world model within a Monte Carlo tree search framework: the LLM simulates the consequences of candidate reasoning steps, a reward function scores partial solutions, and MCTS guides search toward high-quality completions.  This approach achieved substantial improvements on mathematical reasoning, plan generation, and logical inference benchmarks.  WebDreamer~\cite{gu2024your} applied a similar model-based planning paradigm to web agents, using an LLM to simulate the outcomes of candidate browser actions (clicking, typing, navigating) and selecting actions via tree search over imagined web page states, outperforming reactive baselines on WebArena benchmarks. 

Dynalang~\cite{lin2024dynalang} explored a different integration point, using language as an additional observation modality within a DreamerV3-based world model. By grounding language tokens in the same latent dynamics as visual observations, Dynalang enables agents to benefit from language hints, instructions, and contextual descriptions during both world model learning and policy optimization.  WorldLLM~\cite{levy2025worldllm} introduced curiosity-driven theory-making, where an LLM iteratively builds and refines explicit world theories to improve its predictive accuracy and downstream decision-making.  These language-centric world models highlight a fundamental question: whether the implicit world knowledge captured during language pre-training is sufficient for reliable planning, or whether explicit dynamics learning remains necessary for robust decision-making in complex environments.

\subsubsection{Discussion and Open Challenges}
\label{sec:downstream:discussion}

Several cross-cutting themes emerge from this survey of downstream applications. First, there is a clear progression from \emph{task-specific} world models (e.g., MuZero for board games, OccWorld for driving) toward \emph{foundation world models} (e.g., Cosmos, Genie~2, V-JEPA~2) that aim to serve as general-purpose backbones across domains. Whether such generalist models can match specialist performance in safety-critical settings remains unresolved. Second, the evaluation gap is widening: video-generation world models are often assessed on perceptual quality metrics (FID, FVD), whereas RL world models are measured on task reward, and medical world models on clinical endpoints-making cross-domain comparison and progress tracking difficult. Third, the question of whether world models learn genuine \emph{causal} structure or merely correlational shortcuts is increasingly urgent, particularly as these models enter healthcare and autonomous driving where counterfactual reasoning is essential. Finally, the computational cost of training and deploying large-scale world models remains a bottleneck, with real-time inference posing particular challenges for embodied applications. Addressing these challenges will require not only architectural advances but also new evaluation frameworks, safety guarantees, and interdisciplinary collaboration.

\section{Categorization of World Models by Methodological Families}
\label{sec:methodology}
This section focuses on the main \emph{methodological families} used to build and train the World Models. Each family reflects different assumptions about how environmental dynamics should be modeled, which inductive biases are most effective, and how learned representations should relate to underlying physical processes. We begin with state-space and recurrent latent models (Section~\ref{sec:methodology}.1), which have long been central to model-based reinforcement learning, and then examine transformer-based world models (Section~\ref{sec:transformer_world_models}), which replace recurrence with self-attention and tokenized representations. We next consider diffusion-based world models (Section~\ref{sec:diffusionworldmodel}), followed by physics-informed and structured approaches (Section~\ref{sec:Physicsworldmodel}) that embed domain knowledge directly into the architecture. We conclude with language-augmented and multi-modal world models, which combine natural language with diverse sensory inputs to support richer grounding and broader generalization.

\subsection{State-space and recurrent latent world models}
Structured State Space Models (SSMs) such as S4~\cite{gu2022s4} and Mamba~\cite{gu2023mamba} offer $O(n)$ sequence processing with strong long-range dependency modeling, presenting an alternative to both recurrent and attention-based architectures. S4 introduced structured parameterizations that efficiently capture long-range dependencies through the HiPPO framework. Mamba extended this with input-dependent (selective) state transitions, achieving transformer-competitive performance on language and other modalities at 5$\times$ higher throughput. Recent work has begun to explore SSMs explicitly as world-model backbones: Po et al.~\cite{po2025longcontext} proposed long-context state-space video world models (ICCV 2025), demonstrating that linear RNNs can capture long-range temporal dependencies without increasing per-frame generation cost. Their linear complexity and strong temporal modeling make SSMs promising candidates for long-horizon environment simulation.

Whereas, recurrent neural networks, particularly LSTMs~\cite{hochreiter1997lstm} and GRUs~\cite{cho2014gru}, were the first deep architectures used for world modeling. Ha and Schmidhuber~\cite{ha2018world} combined a VAE (the ``vision'' model V) with a Mixture Density Network-RNN (MDN-RNN, the ``memory'' model M) to learn a compact generative model of RL environments. The VAE compressed each image frame into a latent vector $z_t$, while the MDN-RNN predicted the distribution of the next latent $p(z_{t+1} | z_t, a_t)$, capturing stochastic dynamics. A minimal linear controller (only 867 parameters) trained via CMA-ES could solve Car Racing and VizDoom tasks by planning entirely in the model's imagination.

Hafner et al.~\cite{hafner2019planet} introduced the Recurrent State-Space Model (RSSM) in PlaNet, which splits the latent state into a deterministic recurrent component $h_t$ and a stochastic component $z_t$. This dual-path design was critical: the deterministic path preserves long-range memory, while the stochastic path captures environmental uncertainty. PlaNet used the Cross-Entropy Method (CEM) for online planning and demonstrated 50$\times$ data efficiency improvements over model-free methods on continuous control tasks from pixels.

The Dreamer series~\cite{hafner2020dreamer, hafner2021dreamerv2, hafner2023dreamerv3} extended the RSSM with actor-critic learning in imagined trajectories. DreamerV1 learned value functions and policies by backpropagating analytic gradients through the world model's predictions. DreamerV2 introduced discrete categorical latent representations, enabling mastery of Atari games. DreamerV3 achieved robustness across 150+ tasks with a fixed set of hyperparameters through normalization and transformation techniques, and was the first algorithm to collect diamonds in Minecraft from scratch without human demonstrations.


GAN-based world models exploit adversarial training for sharper outputs. GameGAN~\cite{kim2020gamegan} learned to simulate Pac-Man by ingesting gameplay screenshots and keyboard actions, using a dynamics engine, memory module, and rendering engine---all implemented as neural networks trained end-to-end. While producing visually crisp frames, GAN-based approaches suffer from training instability and mode collapse, limiting their adoption for complex environments.

\subsection{Transformer-based world models}
\label{sec:transformer_world_models}

The transformer architecture~\cite{vaswani2017attention}, originally developed for natural language processing, has emerged as a dominant paradigm for world modeling owing to its capacity for long-range dependency modeling, parallelizable training, and favorable scaling properties. This section surveys the evolution of transformer-based world models, from early sequence-modeling formulations of reinforcement learning to large-scale interactive environment generators.

\textbf{Foundational formulations.}
The application of transformers to sequential decision-making was catalyzed by two concurrent works. The first one is \emph{Decision Transformer}~\cite{chen2021decision} recast offline reinforcement learning as conditional sequence modeling: a causally masked GPT architecture autoregressively generates actions conditioned on desired returns-to-go, past states, and actions, entirely bypassing temporal-difference learning and value function estimation.  The Second one is \emph{Trajectory Transformer}~\cite{janner2021trajectory} adopted a complementary model-based perspective, discretizing continuous state-action-reward trajectories into tokens and modeling their joint distribution with a transformer, thereby enabling beam-search-based planning over long horizons.  Together, these works established that the sequence prediction capabilities of transformers could subsume both policy learning and dynamics modeling within a unified framework. 

Concurrent developments in video generation provided key architectural ingredients. For exapmle, VideoGPT~\cite{yan2021videogpt} demonstrated that a VQ-VAE tokenizer paired with an autoregressive transformer could generate temporally coherent video sequences, foreshadowing the \emph{tokenize-then-predict} paradigm that would become standard in visual world models.  Scaling these ideas to multi-task settings, Gato~\cite{reed2022gato} serialized data from over 600 diverse tasks-spanning Atari games, robotic manipulation, image captioning, and dialogue-into a single token sequence consumed by a 1.2B-parameter decoder-only transformer, demonstrating the feasibility of a universal generalist agent.  The Multi-Game Decision Transformer~\cite{lee2022multigame} investigated scaling properties by training a single transformer policy across multiple Atari games simultaneously,  while Algorithm Distillation~\cite{laskin2023algorithm} showed that entire RL learning histories could be distilled into a causal transformer, enabling \emph{in-context} reinforcement learning-improvement across episodes without any parameter updates. 

\textbf{RL agents trained in transformer world models.}
A pivotal advance came with IRIS~\cite{micheli2023iris}, which demonstrated that transformers are sample-efficient world models for imagination-based reinforcement learning.  IRIS employs a discrete autoencoder to tokenize visual observations into a sequence of codes  $z_t = (z_t^1, \ldots, z_t^K)$ and trains an autoregressive transformer to predict future token sequences conditioned on past observations and actions: 
\begin{equation}
    p(z_{t+1}, r_t, \gamma_t \mid z_{\leq t}, a_{\leq t}) = \prod_{k=1}^{K} p\!\left(z_{t+1}^k \mid z_{t+1}^{<k}, z_{\leq t}, a_{\leq t}\right) \cdot p(r_t, \gamma_t \mid z_{\leq t}, a_{\leq t}),
\end{equation}
where $r_t$ and $\gamma_t$ denote the predicted reward and termination signal, respectively. On the Atari 100k benchmark, IRIS achieved a mean human-normalized score of 1.046, surpassing human performance on 10 of 26 games and establishing a new state of the art for methods without lookahead search. 

Several works rapidly built upon this paradigm.  TransDreamer~\cite{chen2022transdreamer} proposed the Transformer State-Space Model (TSSM), directly replacing the GRU-based Recurrent State-Space Model (RSSM) in the Dreamer framework~\cite{hafner2023dreamerv3} with a transformer backbone while retaining stochastic latent variables, and demonstrated improved performance on tasks requiring long-range memory.  TWM~\cite{robine2023twm} applied a Transformer-XL architecture over compact latent representations of states, actions, and rewards, efficiently capturing long-term dependencies through segment-level recurrence.  STORM~\cite{zhang2023storm} combined a GPT-like transformer with a categorical VAE encoder, introducing stochastic latent variables that improved robustness over IRIS's deterministic tokenization;  it achieved 126.7\% mean human performance on Atari 100k with only 4.3 hours of training on a single GPU. 

More recent efforts have targeted efficiency and expressiveness. $\Delta$-IRIS~\cite{micheli2024deltairis} introduced a context-aware tokenizer that encodes stochastic frame-to-frame deltas rather than full observations, dramatically reducing the token count per timestep and achieving state-of-the-art performance on Crafter while training an order of magnitude faster than IRIS.  DART~\cite{agarwal2024dart} paired a transformer decoder for world modeling with a transformer encoder for policy learning, incorporating explicit memory tokens to handle partial observability.  Dedieu et al.~\cite{dedieu2025improving} proposed three complementary improvements-Dyna-style warmup training, nearest-neighbor patch tokenization, and block teacher forcing-that together enabled a transformer world model to surpass both DreamerV3 and human-level performance on Craftax-classic for the first time, reaching 69.7\% reward versus 53.2\% for DreamerV3. 

\textbf{Transformer models as real-world and game simulators.}
Beyond closed-loop RL training, transformers have been scaled to serve as general-purpose interactive simulators. Genie~\cite{bruce2024genie}, an 11-billion-parameter model from Google DeepMind, represents the first generative interactive environment trained from unlabeled Internet video.  Its architecture comprises three components: a spatiotemporal video tokenizer (ST-ViViT), a latent action model that infers discrete action codes from video without ground-truth action labels, and an ST-Transformer dynamics model that autoregressively predicts future latent frames conditioned on inferred actions.  Genie enables controllable generation of diverse 2D platformer environments from text, sketch, or image prompts.  Its successor, Genie~2~\cite{parkerholder2024genie2}, extends this paradigm to rich 3D environments using an autoregressive latent diffusion architecture, generating consistent interactive worlds for up to one minute from a single image prompt  with emergent object interactions, NPC behavior, and persistent world memory. 

GAIA-1~\cite{hu2023gaia1}, a 9-billion-parameter model from Wayve, applies the transformer world model paradigm to autonomous driving simulation.  Its 6.5B-parameter autoregressive transformer predicts future discrete image tokens conditioned on video, text, and action inputs, while a 2.6B-parameter video diffusion decoder renders predictions back to pixel space,  exhibiting emergent understanding of 3D geometry and traffic dynamics.  UniSim~\cite{yang2024unisim} learns an interactive real-world simulator by orchestrating diverse training data-internet images, robotics trajectories, and navigation sequences-to simulate visual outcomes of both high-level language instructions and low-level motor commands;  policies trained entirely within UniSim transfer zero-shot to real robotic manipulation.  Oasis~\cite{oasis2024} demonstrated real-time open-world generation using a Diffusion Transformer (DiT) backbone, producing Minecraft-like interactive environments at 20~FPS.  Pandora~\cite{xiang2024pandora} coupled a pretrained large language model with a video diffusion model for world simulation controlled by free-form natural language.  OpenAI's Sora~\cite{brooks2024sora} further established the paradigm of large-scale video generation models as implicit world simulators, demonstrating emergent physical reasoning in a diffusion transformer trained on internet-scale video. 

\textbf{Domain-specific adaptations.}
Transformer-based world models have seen extensive adoption in \emph{autonomous driving}. UniAD~\cite{hu2023uniad}, a unified detection, tracking, mapping, motion forecasting, occupancy prediction, and planning within a single transformer framework where all perception modules serve the downstream planning objective through unified query interfaces.  OccWorld~\cite{zheng2024occworld} formulates the driving world model in 3D semantic occupancy space: a VQ-VAE scene tokenizer produces discrete volumetric tokens, and a GPT-like spatial-temporal transformer autoregressively forecasts future occupancy grids and ego trajectories.  Copilot4D~\cite{zhang2024copilot4d} applies discrete diffusion on VQ-VAE-tokenized LiDAR point clouds using a spatial-temporal transformer,  reducing prior state-of-the-art Chamfer distance by over 65\% for one-second predictions on nuScenes.  MILE~\cite{hu2022mile} jointly learns a world model and driving policy through model-based imitation learning, executing complex maneuvers from plans generated entirely in imagination.  ViDAR~\cite{yang2024vidar} introduced visual point cloud forecasting as a scalable self-supervised pretraining objective that jointly learns semantics, 3D structure, and temporal dynamics,  while Vista~\cite{gao2024vista} developed a generalizable driving world model with high-fidelity, action-conditioned, long-horizon prediction trained on large-scale web-sourced driving video. 

In \emph{robotics}, the Robotics Transformer series has demonstrated the efficacy of transformer-based models for real-world manipulation at scale. RT-1~\cite{brohan2023rt1} trained a 35-million-parameter transformer on over 130,000 real robot episodes spanning 700+ tasks, establishing robust multi-task generalization.  RT-2~\cite{brohan2023rt2} co-fine-tuned large vision-language models (PaLI-X 55B and PaLM-E 12B) on both web-scale data and robotic trajectories, representing robot actions as text tokens.  This enabled significant transfer of web knowledge to robotic control-improving success rates on novel objects from 32\% to 62\%-and emergent semantic reasoning capabilities including chain-of-thought planning for multi-step manipulation tasks. 

\textbf{Comparative analysis and key challenges.}
Transformer-based world models offer several structural advantages over recurrent and diffusion-based alternatives. Unlike GRU-based models such as the RSSM in the Dreamer line~\cite{hafner2023dreamerv3}, transformers attend directly to any past timestep, circumventing the information bottleneck of fixed-size hidden states and enabling superior long-range dependency modeling.  Compared to diffusion-based world models such as DIAMOND~\cite{alonso2024diamond},  autoregressive transformers generate predictions in a single forward pass rather than requiring iterative denoising steps,  yielding substantial inference-time efficiency-a critical advantage for online planning within imagination. 

However, transformers introduce distinct challenges. The \emph{quadratic computational complexity} $\mathcal{O}(T^2)$ with respect to context length $T$ remains a fundamental bottleneck, particularly for high-resolution visual world models where each observation may comprise hundreds of tokens.  Mitigation strategies have included delta encoding to reduce per-frame token counts~\cite{micheli2024deltairis},  Transformer-XL segment-level recurrence~\cite{robine2023twm},  and fused single-token-per-timestep representations~\cite{zhang2023storm}.  The \emph{discrete tokenization bottleneck} poses another limitation: VQ-VAE-based approaches necessarily discard fine-grained visual information during quantization, a weakness that diffusion-based world models~\cite{alonso2024diamond} explicitly address by operating in continuous pixel space.  Furthermore, \emph{autoregressive error accumulation} over extended prediction horizons can degrade imagined trajectory quality, a problem partially mitigated by stochastic latent variables in STORM~\cite{zhang2023storm}  and TransDreamer~\cite{chen2022transdreamer}.  As transformer world models scale from game benchmarks toward open-ended real-world simulation, addressing these challenges while preserving the architecture's strengths in expressiveness, parallelism, and scalability constitutes an active and rapidly evolving research frontier.

\subsection{Diffusion-Based World Models}
\label{sec:diffusionworldmodel}

Diffusion models~\cite{ho2020ddpm} have emerged as a powerful paradigm for world modeling, offering a compelling alternative to discrete-token and recurrent approaches. Rooted in denoising score matching, these models learn to reverse a gradual noise-corruption process, generating high-fidelity samples through iterative denoising. When adapted for world modeling, diffusion-based approaches inherit several attractive properties: capturing complex, multi-modal distributions over future states, preserving fine-grained visual details that discrete tokenization may discard, and naturally representing environmental stochasticity through the sampling process itself.

\textbf{Foundational formulations.}
The core idea is to recast image generation as a conditional next-observation prediction problem. Given past observations $o_{<t}$ and an action $a_t$, the diffusion world model learns $p(o_{t+1} \mid o_{<t}, a_t)$ by training a denoising network $\epsilon_\theta$ to remove noise from a corrupted target observation. The training objective follows the standard denoising score matching loss:
\begin{equation}
\mathcal{L}(\theta) = \mathbb{E}_{o_{t+1}, \epsilon, k} \left[ \left\| \epsilon - \epsilon_\theta\!\left(o_{t+1}^{(k)},\; o_{<t},\; a_t,\; k\right) \right\|^2 \right],
\end{equation}
where $o_{t+1}^{(k)}$ denotes the observation corrupted at noise level $k$ and $\epsilon \sim \mathcal{N}(0, I)$. At inference time, predictions are generated by starting from pure noise and applying the learned denoising process for $K$ steps. A key architectural distinction is whether the diffusion process operates in pixel space or a learned latent space. Pixel-space approaches such as DIAMOND~\cite{alonso2024diamond} operate directly on raw observations, preserving all visual details at higher computational cost. Latent diffusion approaches~\cite{rombach2022ldm} first compress observations through a pretrained VAE, then apply diffusion in the lower-dimensional latent space. This strategy, adopted by GameNGen~\cite{valevski2024gamengen}, Sora~\cite{openai2024sora}, NVIDIA Cosmos~\cite{nvidia2025cosmos}, and many autonomous driving world models, significantly reduces computational costs while maintaining high perceptual quality.

An important early contribution bridging diffusion models and sequential decision-making was Diffuser~\cite{janner2022diffuser}, which directly modeled the joint distribution over state-action trajectories $p(\tau) = p(s_1, a_1, \ldots, s_T, a_T)$ using a temporal U-Net. Planning was accomplished by sampling from this trajectory distribution, with classifier-guided sampling and inpainting techniques steering trajectories toward high-reward regions or goal constraints. Decision Diffuser~\cite{ajay2023dd} extended this framework with classifier-free guidance, conditioning trajectory generation directly on target returns, constraints, or skills without a separately trained reward function. A further development, the Diffusion World Model (DWM)~\cite{ding2025dwm}, departs from the one-step autoregressive paradigm by predicting multi-step future states and rewards concurrently in a single forward pass, achieving a 44\% performance gain over one-step dynamics models on D4RL benchmarks.

\textbf{RL agents trained in diffusion world models.}
A defining application of diffusion-based world models is their use as imagination environments for RL agent training, analogous to the Dreamer series~\cite{hafner2019dreamerv1,hafner2021dreamerv2,hafner2025dreamerv3} but with the world model replaced by a diffusion process. DIAMOND~\cite{alonso2024diamond} (NeurIPS 2024 Spotlight) represents the most rigorous exploration of this paradigm, replacing discrete-token world models (e.g., IRIS~\cite{micheli2023iris}) with a pixel-space diffusion model conditioned on a short history of past frames and actions. Several design choices proved critical. First, DIAMOND adopted the EDM framework~\cite{karras2022edm} rather than DDPM~\cite{ho2020ddpm}, which becomes unstable under autoregressive generation with few denoising steps due to compounding errors. Second, EDM enabled stable long-horizon rollouts with as few as three denoising steps per frame. Third, for transitions with multiple plausible outcomes (e.g., opponent movement in Boxing), more denoising steps enabled better mode selection, whereas single-step denoising averaged across modes, producing blurry predictions. On Atari 100k, DIAMOND achieved a mean human-normalized score of 1.46, a new state of the art for agents trained entirely within a world model. Qualitative analysis revealed that improved performance in certain games was directly attributable to better modeling of critical visual details (e.g., small ball positions, score digits) that discrete-token approaches failed to preserve. DIAMOND was further demonstrated as an interactive neural game engine on 87 hours of Counter-Strike: Global Offensive gameplay, scaling from 4.4M to 381M parameters via a two-stage low-resolution dynamics prediction and diffusion-based upsampling pipeline. A subsequent extension, StateSpaceDiffuser~\cite{statespacediffuser2025}, addresses DIAMOND's limited temporal context by integrating a state-space model (Mamba), enabling long-context diffusion world modeling with linear per-frame complexity.

In robotics, diffusion world models have recently been adopted to refine pre-trained manipulation policies. World4RL~\cite{jiang2025world4rl} uses a diffusion-based world model pre-trained on multi-task robotic datasets as a high-fidelity simulator for end-to-end RL fine-tuning, entirely avoiding real-world interaction. Similarly, DiWA~\cite{chandra2025diwa} leverages a latent world model to adapt diffusion-based robotic skills offline, achieving effective fine-tuning on the CALVIN benchmark with orders of magnitude fewer physical interactions than model-free baselines. These works demonstrate that the train-in-imagination paradigm, initially validated on Atari, is now maturing toward real-world robotic deployment.

\textbf{Diffusion models as real-world and game simulators.}
Diffusion-based world models have also been deployed as standalone interactive simulators. GameNGen~\cite{valevski2024gamengen} demonstrated that a diffusion model built upon Stable Diffusion v1.4 could simulate classic DOOM at over 20 FPS on a single TPU, trained in two phases: an RL agent first played DOOM and its sessions were recorded, then a diffusion model learned next-frame prediction conditioned on past frames and actions. A key contribution was conditioning noise augmentation, which stabilized autoregressive generation over multi-minute sessions (next-frame PSNR 29.4, near-chance human distinguishability). GameGen-X~\cite{che2025gamegenx} extended this to interactive open-world environments with controllable actions and dynamic scenes.

UniSim~\cite{yang2023unisim} proposed a universal real-world simulator trained on diverse data spanning internet images, robotics, navigation, and human activity. It predicts future frames conditioned on multi-modal actions---from language instructions to low-level motor controls---unified via T5 embeddings with discretized control signals. UniSim enabled training vision-language planners and RL policies that transferred zero-shot to real robots, demonstrating diffusion world models' potential for bridging the sim-to-real gap.

OpenAI's Sora~\cite{openai2024sora} introduced large-scale video generation as world simulation via a DiT~\cite{peebles2023dit} architecture on spacetime latent patches, claiming emergent world-simulator properties. However, subsequent work revealed these to be inconsistent~\cite{ding2025understanding}, and Kang et al.~\cite{kang2025howfar} showed through controlled experiments that such models exhibit case-based generalization---mimicking nearest training examples rather than abstracting physical rules---fueling debate about whether visual-only training can yield true world models.

The scaling of diffusion world models has entered the foundation model era. NVIDIA Cosmos~\cite{nvidia2025cosmos} released open-weight WFMs in diffusion (7B/14B) and autoregressive variants, trained on 20 million hours of real-world video. Its successor Cosmos-Predict2.5~\cite{nvidia2025cosmos25} transitioned to flow-matching with RL-based post-training on 200 million clips. Together with CogVideoX~\cite{yang2025cogvideox}, which scales diffusion transformers via expert adaptive normalization, these efforts mark the convergence of video generation and world simulation at industrial scale.

\textbf{Domain-specific adaptations.}
Autonomous driving has been particularly fertile ground for diffusion-based world models, driven by the need for photorealistic simulation, action-controllable prediction, and scalable data augmentation. GAIA-1~\cite{hu2023gaia1} pioneered this direction with a 6.5B-parameter transformer world model paired with a 2.6B-parameter video diffusion decoder for controllable driving scenario generation conditioned on video, text, and action inputs. DriveDreamer~\cite{wang2023drivedreamer} introduced a two-stage pipeline: the first stage learns structured traffic constraints via diffusion, while the second enables future state prediction conditioned on past trajectories. DriveDreamer-2~\cite{zhao2024drivedreamer2} leveraged LLMs to convert user queries into agent trajectories for corner-case generation, and DriveDreamer4D~\cite{zhao2025drivedreamer4d} extended this line to 4D driving scene representation by combining video generation with 4D Gaussian Splatting. GenAD~\cite{zheng2024genad} developed the generalized predictive model framework with action controllability and trajectory conditioning. Vista~\cite{gao2024vista} pushed resolution to $576 \times 1024$ pixels while supporting multiple action control modes (steering angle/speed, trajectory commands, goal points) through dynamic conditioning priors. GAIA-2~\cite{wayve2025gaia2} scaled to multi-view generation with agent-level semantic control, trained on approximately 25 million video sequences across three countries. OccSora~\cite{wang2025occsora} proposed 4D occupancy generation as an alternative representation, generating geometry-aware voxel predictions that offer richer 3D structural information than image-based approaches. A common pattern across these models is latent diffusion with structured conditioning: BEV representations, 3D bounding boxes, HD map layouts, or ego-vehicle trajectories inject geometric and semantic structure into the diffusion process. Discrete diffusion has also been explored: Copilot4D~\cite{zhang2024copilot4d} discretized LiDAR observations via VQ-VAE and employed discrete diffusion for unsupervised future point-cloud prediction, demonstrating that diffusion-based world modeling extends beyond the image domain.

In robotic manipulation and control, Diffusion Policy~\cite{chi2023diffusionpolicy} adapted trajectory diffusion to visuomotor control, generating action sequences conditioned on visual observations using a temporal U-Net or Transformer denoiser with receding-horizon execution. While not a world model per se, it demonstrates the broader utility of diffusion for action generation in embodied settings. Several works have since combined diffusion-based action generation with explicit world models: RoboDreamer~\cite{robodreamer2024} used language-guided imagination for long-horizon robotic planning, DreMa~\cite{drema2024} integrated diffusion-based visual generation with model-based RL for dexterous manipulation, and IRASim~\cite{irasim2025} proposed a fine-grained interaction-aware world model for robot manipulation. A particularly promising direction is the unification of video prediction and action generation: the Unified World Model (UWM)~\cite{he2025uwm} couples video and action diffusion during pre-training on large-scale robotic datasets, enabling a single model to serve simultaneously as simulator and policy, while TesserAct~\cite{tesseract2025} extends world modeling to 4D embodied representations for physically grounded robotic planning.

\textbf{Comparative analysis and key challenges.}
Compared to alternative generative paradigms, diffusion models offer several distinctive advantages for world modeling. First, they consistently produce sharper predictions than VAE-based approaches (which suffer from Gaussian-likelihood blurriness) and discrete-token models (which may lose fine-grained details during quantization); DIAMOND's Atari results demonstrate that this visual superiority translates directly to improved downstream RL performance. Second, the stochastic sampling process naturally represents environmental uncertainty, with multiple denoising trajectories yielding diverse plausible futures. Third, diffusion models support rich conditioning mechanisms (classifier-free guidance, cross-attention, inpainting) that incorporate diverse information sources without fundamental architectural changes.

However, several key challenges remain. Inference speed is a fundamental bottleneck: iterative denoising is inherently slower than single-pass generation, and while few-step sampling, distillation, and consistency models~\cite{song2023consistency} can reduce the gap, real-time generation remains challenging for high-resolution observations; flow-matching formulations as adopted by Cosmos-Predict2.5~\cite{nvidia2025cosmos25} offer a promising direction. Autoregressive error accumulation persists: noise augmentation~\cite{valevski2024gamengen} and EDM-based formulations~\cite{alonso2024diamond} provide effective but incomplete mitigation, while StateSpaceDiffuser~\cite{statespacediffuser2025} and the multi-step parallel prediction of DWM~\cite{ding2025dwm} represent two recent complementary approaches. Computational cost at scale remains prohibitive for many research groups, with Cosmos training on 200 million video clips at up to 14B parameters. Finally, as the debate around Sora illustrates and the controlled experiments of Kang et al.~\cite{kang2025howfar} confirm, diffusion models trained purely on visual data may learn statistical regularities that mimic physical behavior without acquiring genuine physical understanding; ensuring that predictions respect conservation laws, object permanence, and causal relationships remains an open challenge connecting to the physics-informed approaches discussed in Section~\ref{sec:Physicsworldmodel}.

\subsection{Physics-informed and structured world model}
\label{sec:Physicsworldmodel}

The preceding sections have examined world models grounded in recurrent latent dynamics, variational and adversarial generation, transformer-based sequence prediction, and diffusion-based denoising. A common thread across these approaches is their reliance on general-purpose function approximation: the network architecture imposes minimal assumptions about the structure of the environment, and physical plausibility must emerge entirely from data. While this flexibility enables broad applicability, it comes at a cost. Models that lack structural priors tend to violate conservation laws over long rollouts, struggle to generalize across object counts or physical parameters, and require large datasets to learn regularities that are already well-characterized by classical mechanics. This section surveys an alternative family of world models that embed physical laws, relational structure, or compositional object representations directly into the architecture or training objective. We organize the discussion into five sub-families: physics-conserving neural dynamics models, graph neural network-based dynamics simulators, object-centric world models, equivariant and symmetry-preserving models, and neuro-symbolic and compositional approaches.

\textbf{Physics-conserving neural dynamics.}
Classical mechanics provides two equivalent formalisms for describing conservative systems. In the Hamiltonian formulation, the state of a system is described by generalized coordinates $\mathbf{q}$ and conjugate momenta $\mathbf{p}$, and the dynamics are governed by Hamilton's equations:
\begin{equation}
\frac{d\mathbf{q}}{dt} = \frac{\partial \mathcal{H}}{\partial \mathbf{p}}, \qquad \frac{d\mathbf{p}}{dt} = -\frac{\partial \mathcal{H}}{\partial \mathbf{q}},
\label{eq:hamilton}
\end{equation}
where $\mathcal{H}(\mathbf{q}, \mathbf{p})$ denotes the Hamiltonian, which typically corresponds to the total energy of the system. \textbf{Hamiltonian Neural Networks} (HNNs)~\cite{greydanus2019hnn} were the first to exploit this structure by parameterizing $\mathcal{H}$ with a multilayer perceptron and computing the time derivatives of $\mathbf{q}$ and $\mathbf{p}$ via automatic differentiation through Eq.~\eqref{eq:hamilton}. Because the architecture enforces Hamilton's equations exactly, the resulting dynamics conserve energy by construction, produce perfectly time-reversible trajectories, and generalize to longer horizons than standard neural network baselines that directly regress state derivatives.

An alternative formulation is provided by Lagrangian mechanics, where the dynamics follow from the Euler--Lagrange equation:
\begin{equation}
\frac{d}{dt}\frac{\partial \mathcal{L}}{\partial \dot{\mathbf{q}}} - \frac{\partial \mathcal{L}}{\partial \mathbf{q}} = 0,
\label{eq:euler_lagrange}
\end{equation}
with the Lagrangian $\mathcal{L}(\mathbf{q}, \dot{\mathbf{q}}) = T - V$ defined as the difference between kinetic energy $T$ and potential energy $V$. \textbf{Lagrangian Neural Networks} (LNNs)~\cite{cranmer2020lnn} parameterize the Lagrangian directly with a neural network and derive the equations of motion through Eq.~\eqref{eq:euler_lagrange}. A key advantage over HNNs is that the Lagrangian formulation does not require canonical coordinates, making it applicable to systems where the transformation to phase-space variables is nontrivial. \textbf{Deep Lagrangian Networks} (DeLaN)~\cite{lutter2019delan} pursued a complementary approach by decomposing the Lagrangian into separate neural networks for kinetic and potential energy, parameterizing the mass matrix $\mathbf{M}(\mathbf{q})$ as a positive-definite neural network output to ensure physical consistency, and demonstrating strong results on robot dynamics identification.

Subsequent work extended these foundational ideas along several axes. \textbf{Symplectic ODE-Net} (SymODEN)~\cite{zhong2020symoden} combined the Hamiltonian formulation with neural ordinary differential equations~\cite{chen2018neuralode}, learning dynamics from observed state trajectories while enforcing symplectic structure and incorporating control inputs. \textbf{Constrained Hamiltonian and Lagrangian Neural Networks} (CHNN/CLNN)~\cite{finzi2020chnn} showed that embedding systems into Cartesian coordinates and enforcing holonomic constraints explicitly via Lagrange multipliers dramatically simplifies the learning problem, yielding approximately 100$\times$ improvements in accuracy and data efficiency over prior HNN and LNN methods. \textbf{Hamiltonian Generative Networks} (HGNs)~\cite{toth2020hgn} bridged the gap between physics-conserving dynamics and high-dimensional observations by combining a VAE-like inference network with a Hamiltonian dynamics module, enabling the consistent learning of Hamiltonian dynamics directly from image sequences without restrictive domain assumptions.

Underpinning many of these models is the \textbf{Neural ODE} framework~\cite{chen2018neuralode}, which parameterizes continuous-time dynamics $d\mathbf{h}/dt = f_\theta(\mathbf{h}(t), t)$ with a neural network and computes forward trajectories using black-box ODE solvers. Neural ODEs provide memory-efficient training through the adjoint method, constant memory cost regardless of integration depth, and adaptive computation that allocates more solver steps to regions of complex dynamics. When combined with Hamiltonian or Lagrangian structure, Neural ODEs ensure that the learned dynamics respect both the continuous nature of physical time and the conservation laws encoded in the energy function.

A complementary line of work draws on Koopman operator theory, which seeks a (possibly infinite-dimensional) linear operator that governs the evolution of observables of a nonlinear dynamical system. Li et al.~\cite{li2020koopman} proposed learning \textbf{compositional Koopman operators} by integrating Koopman theory with graph neural networks, enabling linear prediction in a learned embedding space while preserving the compositional structure of multi-body systems. This approach achieves efficient long-horizon prediction through simple matrix exponentiation in the lifted space, at the cost of requiring the learned embedding to faithfully linearize the dynamics.

\textbf{Graph neural network-based dynamics simulators.}
While physics-conserving models encode conservation laws into the dynamics, a parallel line of research encodes relational structure through graph neural networks (GNNs). The central insight is that many physical systems---collections of particles, rigid bodies, deformable meshes---are naturally described as graphs, where nodes represent entities and edges represent interactions. \textbf{Interaction Networks} (INs)~\cite{battaglia2016in} introduced the first general-purpose learnable physics engine based on this principle. An IN decomposes a physical scene into objects and pairwise relations, computes interaction effects through learned neural network modules, and aggregates these effects to predict per-object state updates. The original work demonstrated accurate prediction of gravitational dynamics, rigid-body collisions, and mass-spring systems, establishing the template for all subsequent GNN-based simulators.

Battaglia et al.~\cite{battaglia2018relational} subsequently provided a unified theoretical framework for relational inductive biases in deep learning, formalizing the graph network as a general computational primitive that subsumes INs, message-passing neural networks, and related architectures. This framework clarified how different choices of graph structure, message functions, and aggregation operators encode different physical assumptions, and motivated a series of increasingly powerful simulators.

\textbf{Graph Network-based Simulators} (GNS)~\cite{sanchez2020gns} scaled the IN paradigm to complex, multi-material physical domains. By representing the simulation state as a particle graph and applying learned message-passing updates, GNS achieved accurate simulation of fluids, rigid solids, and deformable materials within a single framework, generalizing to particle counts and domain configurations not seen during training. \textbf{MeshGraphNets}~\cite{pfaff2021meshgraphnets} extended this approach to mesh-based simulation domains such as computational fluid dynamics and structural mechanics, operating on adaptive triangular or tetrahedral meshes rather than particle graphs. By processing the mesh as a graph with learned message-passing, MeshGraphNets achieved one to two orders of magnitude speedup over traditional numerical solvers while maintaining physical accuracy, and received the ICLR 2021 Outstanding Paper Award.

\textbf{Neural Relational Inference} (NRI)~\cite{kipf2018nri} addressed the complementary problem of simultaneously inferring interaction structure and dynamics from observational data. Using a variational autoencoder framework where the encoder infers a latent interaction graph and the decoder predicts dynamics conditioned on this graph, NRI learns to discover which entities interact and how, without supervision on the graph structure. \textbf{DPI-Net}~\cite{li2019dpinet} extended particle-based GNN simulators to manipulation tasks involving rigid bodies, deformable objects, and fluids, demonstrating that learned particle dynamics can support downstream planning for robotic control. Li et al.~\cite{li2020visualgrounding} further showed that such learned physical models can be grounded in visual observations, enabling prediction of physical dynamics directly from images through a perception-to-simulation pipeline.

\textbf{Object-centric world models.}
A closely related but distinct research direction decomposes visual scenes into discrete object representations, enabling compositional reasoning about dynamics. The key enabler for this line of work is \textbf{Slot Attention}~\cite{locatello2020slotattention}, an iterative competitive attention mechanism that decomposes perceptual inputs into a fixed set of exchangeable representation vectors called ``slots,'' each corresponding to an object or entity in the scene. Slot Attention requires no supervision on object identity or segmentation, instead relying on reconstruction objectives to discover object structure. Its success has established slot-based representations as the de facto standard for unsupervised object discovery in visual world models.

\textbf{SAVi} (Slot Attention for Video)~\cite{kipf2022savi} extended Slot Attention to the temporal domain, conditioning slot initialization on the first frame to enable consistent object tracking across video sequences. \textbf{Contrastive Structured World Models} (C-SWM)~\cite{kipf2020cswm} took a different approach, learning object-centric state representations through contrastive learning rather than reconstruction. C-SWM encodes each object as a vector in a structured latent space and models transitions via a GNN operating on these object representations, demonstrating that contrastive objectives can yield physically meaningful latent spaces that support accurate multi-step prediction in environments with multiple interacting objects.

\textbf{SlotFormer}~\cite{wu2023slotformer} combined slot-based object representations with Transformer-based dynamics prediction, autoregressively predicting future slot states and decoding them into video frames. By operating on object-level tokens rather than pixel patches, SlotFormer achieves strong performance on unsupervised visual dynamics simulation while maintaining compositional generalization to novel object configurations. More recent work has further integrated object-centric representations with model-based reinforcement learning. \textbf{Object-Centric Dreamer}~\cite{ocrssm2023} combines Slot Attention with the RSSM architecture via a GNN-based per-object dynamics model, enabling model-based RL agents to reason about individual objects. \textbf{SOLD} (Slot Object-Centric Latent Dynamics)~\cite{mosbach2025sold} extends this to relational manipulation learning from pixels, using slot representations with learned relational dynamics for robotic control tasks.

Several additional contributions have expanded the scope of object-centric world models such as \textbf{OP3}~\cite{veerapaneni2020op3} introduced entity abstraction for model-based planning, learning to decompose scenes into interactable entities. \textbf{GENESIS}~\cite{engelcke2020genesis} proposed a generative model that jointly performs scene decomposition and generation, producing spatially disentangled object representations. \textbf{DreamWeaver}~\cite{dreamweaver2025} demonstrated compositional world model learning directly from pixels at scale, while recent work on \textbf{STICA}~\cite{stica2025} and \textbf{FIOC-WM}~\cite{fiocwm2025} has explored causality-aware and factored interactive object-centric world models, respectively, incorporating causal reasoning into the object-centric dynamics.

\textbf{Equivariant and symmetry-preserving models.}
Physical laws are characterized by symmetries: the laws of motion are invariant to translations, rotations, and reflections of the coordinate system. Standard neural networks do not respect these symmetries, requiring them to learn invariances from data augmentation or large datasets. Equivariant architectures address this by building symmetry directly into the network structure. \textbf{E(n) Equivariant Graph Neural Networks} (EGNNs)~\cite{satorras2021egnn} introduced a GNN architecture that is equivariant to rotations, translations, and reflections in $n$-dimensional Euclidean space, achieving up to 1000$\times$ improvements in data efficiency on molecular dynamics tasks compared to non-equivariant baselines. EGNNs operate by updating both node features and coordinate positions through message-passing, with the coordinate updates constructed to preserve equivariance by design.

However, strict equivariance can be counterproductive when external fields break the symmetry of the system. \textbf{Subequivariant Graph Neural Networks} (SGNNs)~\cite{han2022sgnn} formalized this observation by introducing the concept of subequivariance---a relaxation of full $E(3)$ equivariance to the subgroup of symmetries that are actually preserved by the physical system. For example, a system under gravity is invariant to rotations about the vertical axis and horizontal translations but not to arbitrary 3D rotations. SGNNs demonstrated superior performance over both non-equivariant and fully equivariant models on dynamics prediction tasks involving gravity and other external fields.

\textbf{SEGNO}~\cite{segno2024} (Second-order Equivariant Graph Neural ODE) further advanced this line by combining equivariant GNNs with neural ODE integration, enforcing second-order continuity in the learned dynamics. By operating on both positions and velocities and preserving equivariance through the ODE integration, SEGNO achieves more accurate long-horizon prediction for systems governed by second-order differential equations. Recent work on \textbf{flow-equivariant world models}~\cite{flowequivariant2025} has begun to explore how equivariance constraints can be incorporated into broader world model architectures beyond particle systems, extending these principles to latent dynamics models for reinforcement learning.

\textbf{Neuro-symbolic and compositional approaches.}
The approaches discussed thus far embed physical structure implicitly through architectural choices. An alternative strategy is to combine neural learning with explicit symbolic reasoning, seeking to recover interpretable physical laws from data. Cranmer et al.~\cite{cranmer2020symbolic} demonstrated \textbf{symbolic distillation from graph neural networks}, training a GNN-based simulator and then extracting symbolic expressions for the learned interaction functions using symbolic regression. This approach successfully recovered known force laws (e.g., Newtonian gravity, Coulomb's law) and Hamiltonians from simulation data, and even discovered a novel analytical formula relevant to dark matter dynamics in astrophysics, demonstrating that neural network-based world models can serve as stepping stones toward interpretable scientific discovery.

\textbf{Cosmos}~\cite{agarwala2024cosmos} (not to be confused with NVIDIA's Cosmos platform) proposed neurosymbolic grounding for compositional world models, learning to ground visual observations in symbolic object descriptions and compose dynamics rules from these symbols. This approach achieves strong compositional generalization---the ability to predict dynamics for combinations of objects and interactions not seen during training---by factoring the world model into perception (neural) and dynamics (symbolic) components. \textbf{Physics-informed model-based reinforcement learning} (PiMBRL)~\cite{ramesh2023pimbrl} took a more direct approach, incorporating known physics equations as soft or hard constraints within the world model training loop for model-based RL, demonstrating improved sample efficiency and physical plausibility on control tasks where partial physics knowledge is available.

The neuro-symbolic direction also intersects with the growing literature on \textbf{semantic world models}, which ground dynamics prediction in natural language or symbolic scene descriptions rather than raw pixels. While these approaches are discussed more fully in Section~3.6, they share with physics-informed methods the core insight that structured, interpretable representations can improve generalization and data efficiency relative to monolithic neural dynamics models.

\textbf{Comparative analysis and open challenges.}
The five sub-families surveyed in this section share a common thesis: encoding known structural properties of the physical world into world model architectures yields improvements in data efficiency, generalization, and long-horizon stability relative to unstructured approaches. Physics-conserving models (HNN, LNN, and their extensions) excel at eliminating energy drift in long rollouts but require knowledge of the appropriate coordinate system and energy decomposition. GNN-based simulators (IN, GNS, MeshGraphNets) provide a natural framework for relational reasoning over multi-body systems and achieve remarkable speedups over numerical simulation, but their accuracy degrades for contact-rich dynamics involving friction and collision. Object-centric models (Slot Attention, C-SWM, SlotFormer) enable compositional reasoning and generalization to novel object configurations, but unsupervised object discovery remains fragile in visually complex real-world scenes. Equivariant architectures (EGNN, SGNN, SEGNO) offer dramatic data efficiency gains by encoding symmetries, but require careful analysis of which symmetries actually hold in the target domain. Neuro-symbolic approaches achieve the strongest compositional generalization and interpretability but struggle to scale to high-dimensional, continuous-valued observations.

Several important open challenges cut across all sub-families. First, no existing architecture seamlessly unifies physics conservation, object-centricity, and equivariance within a single framework; combining these complementary inductive biases remains an active research frontier. Second, contact-rich dynamics---involving friction, collision, and deformation---remain difficult for all approaches, as these phenomena introduce discontinuities that violate the smoothness assumptions underlying many physics-informed models. Third, while these methods demonstrate strong results on synthetic benchmarks such as Physion~\cite{bear2022physion} and controlled simulation environments, bridging the gap to real-world deployment with noisy, partial observations remains a significant challenge. Fourth, integrating the structural inductive biases surveyed here with the representational power of foundation-scale models (Section~3.4) represents a promising but largely unexplored direction. Finally, whether physical behavior should be explicitly encoded through architectural constraints or can emerge from sufficient data and scale---the tension between physics-informed and physics-emergent approaches---remains a fundamental open question in the field.

\subsection{Language-augmented and multi-modal world models}

The world model architectures surveyed in the preceding sections---state-space and recurrent latent models, variational and generative approaches, transformer-based systems, diffusion models, and physics-informed designs---operate predominantly within visual or low-dimensional state spaces, learning dynamics from pixels, point clouds, or structured physical representations. While these paradigms have proven remarkably effective for environment simulation and policy learning, they share a fundamental limitation: they lack the ability to incorporate the rich semantic and compositional structure that natural language provides. Human cognition, by contrast, routinely integrates linguistic knowledge with perceptual experience to form predictive models of the world---we can imagine what will happen when we ``push the red cup off the table'' precisely because language supplies abstract relational and causal structure that complements raw sensory prediction~\cite{lecun2022path}. This observation has motivated a rapidly growing body of work that augments world models with language and multimodal inputs, fundamentally altering how these systems represent states, predict transitions, and generalize across tasks. Crucially, the integration of language is not merely an interface convenience; it changes the model architecture, the learning objective, and the nature of the predicted quantities. This section surveys six methodological families that embody different strategies for this integration, focusing throughout on \emph{how} language and multimodality reshape world model design rather than on specific downstream applications, which are discussed in Section~\ref{sec:applications}.

\paragraph{Language-conditioned latent dynamics models.}
The most direct approach to incorporating language into world models is to augment existing latent dynamics architectures---particularly the RSSM framework underlying the Dreamer family~\cite{hafner2023dreamerv3}---with language as an additional conditioning signal. \textbf{Dynalang}~\cite{lin2024dynalang} introduced the foundational insight that language understanding can be unified with future prediction as a single self-supervised objective. Built upon the DreamerV3 RSSM, Dynalang extends the model to accept both image observations and text tokens as input, learning to predict future text and image representations jointly. The key methodological contribution is that language grounding emerges naturally from the dynamics learning objective: the model learns to associate diverse language types---descriptions, instructions, corrections---with their environmental consequences, without requiring explicit language-grounding supervision. This approach demonstrates that the transition function $p_\theta(z_{t+1} \mid z_t, a_t, l_t)$, where $l_t$ denotes a language embedding, can serve as a universal grounding mechanism.

Subsequent work has explored variations of this language-conditioned dynamics paradigm. \textbf{LIMT}~\cite{aljalbout2025limt} leverages pre-trained language models such as Sentence-BERT to extract semantically meaningful task embeddings that condition both the dynamics model and the actor-critic networks, achieving approximately 30\% higher success rates than model-free baselines under equivalent sample budgets and enabling zero-shot generalization to novel task descriptions. \textbf{LUMOS}~\cite{nematollahi2025lumos} extends the paradigm to real-world robotic manipulation by learning a language-conditioned visuomotor policy within the latent space of an offline world model, demonstrating that language-steerable skills learned from fewer than 1\% hindsight language annotations can transfer zero-shot to physical robots on the CALVIN benchmark. \textbf{LED-WM}~\cite{ledwm2025} takes a complementary approach by conditioning the world model on \emph{environmental descriptions} rather than task instructions, using an attention mechanism to explicitly ground textual descriptions of dynamics rules to entities in the observation space, thereby improving policy generalization to environments with novel dynamics. Collectively, these methods establish that language conditioning at the latent dynamics level improves sample efficiency, enables compositional task specification, and facilitates transfer across environments---all without fundamentally departing from the RSSM architectural paradigm.

\paragraph{Large language models as world models.}
A conceptually distinct approach repurposes large language models themselves as world models, leveraging their pre-trained knowledge to simulate state transitions and predict action consequences in text space. The \textbf{Reasoning via Planning (RAP)} framework~\cite{hao2023rap} exemplifies this paradigm by casting the LLM in a dual role: as both a world model that predicts state transitions and a reasoning agent that proposes actions. RAP integrates Monte Carlo Tree Search (MCTS) with the LLM's predictions, using the world model component to evaluate state transitions while task-specific rewards guide the search. This formulation enabled LLaMA-33B to surpass chain-of-thought prompting on GPT-4 by 33\% in plan generation tasks, establishing that the structured exploration afforded by treating an LLM as an explicit world model can substantially outperform sequential reasoning.

The viability of LLMs as world models has been investigated along multiple axes. Xie et al.~\cite{xie2025precondition} demonstrated that LLMs can be fine-tuned to learn PDDL-style precondition and effect knowledge---determining action applicability based on world state and predicting resulting states after action execution---from synthetic data, with human-participant studies confirming alignment with human understanding of world dynamics. In a complementary direction, \textbf{GLAM}~\cite{carta2023glam} showed that LLMs can achieve functional grounding through online reinforcement learning in text-based environments, using PPO to progressively update an LLM policy for improved sample efficiency and generalization.

However, systematic evaluation has revealed fundamental limitations of current LLMs as world simulators. Wang et al.~\cite{wang2024bytesized32} introduced the ByteSized32-State-Prediction benchmark comprising 76,369 state transitions and found that even GPT-4 achieves only 59.9\% accuracy on non-trivial transitions, with cumulative accuracy dropping below 1\% after ten sequential steps. The identified failure modes---environment-driven transitions, arithmetic reasoning, common-sense physics, and scientific knowledge---suggest that LLMs learn correlational rather than causal models of world dynamics. More recent work by Gkountouras et al.~\cite{gkountouras2025causal} addresses this limitation directly by integrating causal representation learning with LLMs, constructing a framework in which a causal world model with variables linked to natural language expressions serves as a simulator that the LLM can query. This causally-aware approach outperforms pure LLM-based reasoners, particularly for longer planning horizons, pointing toward a productive synthesis between statistical language models and structured causal reasoning.

\paragraph{Text-conditioned video generation as world simulation.}
A third methodological family transforms video generation architectures into world simulators through language conditioning, raising fundamental questions about whether text-to-video models can constitute genuine world models. \textbf{GAIA-1}~\cite{hu2023gaia1} pioneered this direction with a 9-billion-parameter architecture that casts world modeling as unsupervised sequence modeling over jointly tokenized video, text, and action inputs. The architecture pairs a 6.5-billion-parameter autoregressive transformer with a 2.6-billion-parameter video diffusion decoder, demonstrating emergent properties including contextual awareness and geometry understanding in driving scenarios. Text conditioning enables fine-grained control over ego-vehicle behavior and scene composition, establishing the viability of language as a control interface for large-scale generative world models.

\textbf{UniSim}~\cite{yang2023unisim} generalized this approach by training a video diffusion model on diverse data spanning internet images, robotics, navigation, and human activity, unifying multiple action modalities---from language instructions to low-level motor controls---through T5 language model embeddings concatenated with discretized control signals. The resulting model supports temporally extended actions in various modalities and enables training vision-language planners and reinforcement learning policies that transfer zero-shot to real robots. \textbf{Pandora}~\cite{xiang2024pandora} introduced a hybrid autoregressive-diffusion architecture that uses free-form text as the action space and video as the state space, integrating a pre-trained 7-billion-parameter LLM with a pre-trained video generation model through lightweight fine-tuning. This design enables real-time, language-controllable world simulation across diverse domains while generating videos of unlimited duration through autoregressive chaining.

At industrial scale, NVIDIA's \textbf{Cosmos}~\cite{nvidia2025cosmos} released open-weight World Foundation Models at 7B and 14B parameters trained on over 20 million hours of real-world video, integrating text, image, and video conditioning within unified diffusion and autoregressive variants. The subsequent Cosmos-Predict2.5~\cite{nvidia2025cosmos25} transitioned to flow-matching with reinforcement learning-based post-training on 200 million clips, while the companion Cosmos-Reason model provides richer text grounding through chain-of-thought reasoning about physical dynamics. The \textbf{Large World Model (LWM)}~\cite{liu2024lwm} pursued an orthogonal scaling axis by extending context length to one million tokens via RingAttention, enabling joint modeling of long video and text sequences in a unified autoregressive framework. DeepMind's \textbf{Genie}~\cite{bruce2024genie} and its successors demonstrated that interactive world models can be built from unlabeled video alone by learning latent action models, with Genie 3 accepting text prompts to generate navigable 3D environments at 24 frames per second. Despite these impressive capabilities, the debate over whether text-conditioned video generation constitutes genuine world modeling remains active: controlled experiments by Kang et al.~\cite{kang2025howfar} have shown that such models exhibit case-based generalization---mimicking nearest training examples rather than abstracting physical rules---suggesting that visual fidelity alone does not guarantee causal understanding.

\paragraph{Vision-language joint embedding approaches.}
In contrast to generative approaches that predict future observations in pixel space, joint embedding predictive architectures learn to predict future \emph{representations} rather than raw sensory data. LeCun's \textbf{JEPA} framework~\cite{lecun2022path} provides the theoretical foundation for this paradigm, arguing that generative pixel-space prediction wastes model capacity on perceptually irrelevant details and proposing instead that world models should operate by aligning representations of compatible inputs in a shared embedding space. This principle has been operationalized through a series of increasingly capable models.

\textbf{V-JEPA 2}~\cite{assran2025vjepa2}, pre-trained on over one million hours of video through self-supervised spatiotemporal representation learning, predicts masked spatio-temporal regions entirely in a learned latent space without any pixel reconstruction. The resulting representations achieve 77.3\% top-1 accuracy on Something-Something v2 and, critically for world modeling, can be post-trained for robotic action-conditioned planning on Franka robot arms using fewer than 62 hours of unlabeled robot video. \textbf{VL-JEPA}~\cite{fung2024vljepa} extends the joint embedding principle to the vision-language domain by predicting continuous text embeddings rather than autoregressively generating tokens, achieving competitive performance with 50\% fewer trainable parameters (1.6 billion) than classical vision-language models while natively supporting open-vocabulary classification, text-to-video retrieval, and visual question answering. This approach demonstrates that JEPA principles extend naturally to multimodal settings, providing an alternative to the cross-attention mechanisms used in generative approaches.

A related line of work investigates how pre-trained visual foundation models can serve as representation backbones for world modeling. \textbf{DINO-WM}~\cite{zhou2025dinowm} builds world dynamics models on compact DINOv2 patch embeddings rather than raw pixels, predicting future patch features from offline behavioral trajectories and achieving zero-shot planning across six environments including mazes, push manipulation, and multi-particle scenarios without requiring reward models or expert demonstrations. While DINO-WM is language-free, its success illustrates a broader methodological point: pre-trained visual representations that encode semantic structure---whether acquired through implicit language alignment during pre-training or through self-supervised objectives---provide a powerful inductive bias for dynamics learning, complementing the explicit language conditioning approaches described above.

\paragraph{Semantic world models.}
Perhaps the most radical departure from conventional world modeling is the proposal that world models need not predict future observations or latent states at all, but can instead reason directly in language and semantic space. \textbf{Semantic World Models (SWM)}~\cite{berg2025swm} reformulate world modeling as visual question answering about the future: given current observations and candidate actions, the model---built upon the PaliGemma 3B vision-language model---directly answers questions about the semantic effects of those actions without generating any pixel-level predictions. A new projection matrix conditions the VLM on actions, enabling both sampling-based planning and gradient-based policy improvement through the model's differentiable predictions. The results are striking: on the LangTable benchmark, SWM improves success rates from 14.4\% to 81.6\% through gradient-based optimization, dramatically outperforming pixel-based world models by reasoning in a space that is both lower-dimensional and more directly aligned with task-relevant information.

\textbf{VLWM}~\cite{chen2025vlwm} extends this paradigm by using natural language as the abstract world state representation for planning. Given visual observations, VLWM first infers overall goal achievements, then predicts a trajectory composed of interleaved actions and world state changes expressed as structured text. The model supports both System-1 reactive planning through a learned action policy and System-2 reflective planning through cost minimization over predicted textual trajectories, with the latter improving Elo scores by 27\% over the former. These semantic approaches suggest a provocative thesis: for tasks where the relevant state can be expressed linguistically, bypassing pixel-space prediction entirely may yield both more efficient and more capable world models. The open question is whether this advantage holds for domains requiring fine-grained spatial reasoning, where semantic abstraction may discard information critical for precise manipulation.

\paragraph{Unified multimodal architectures.}
The approaches described above augment existing architectures with language or replace visual prediction with linguistic reasoning. A final methodological family pursues a more ambitious goal: building end-to-end architectures that natively integrate vision, language, and action within a single world model framework. \textbf{WorldGPT}~\cite{ge2024worldgpt} constructs a generalist world model upon a multimodal large language model backbone, acquiring understanding of world dynamics through analyzing millions of videos and incorporating a cognitive architecture combining memory offloading, knowledge retrieval, and context reflection to enhance performance on long-term tasks. \textbf{Motus}~\cite{motus2024} introduces a Mixture-of-Transformer (MoT) architecture integrating three specialized experts---for understanding, video generation, and action prediction---with a UniDiffuser-style scheduler that enables flexible switching between world model, vision-language-action model, inverse dynamics model, and video generation modes within a single architecture. The use of optical flow as an embodiment-agnostic proxy for actions, compressed into low-dimensional latent codes, enables large-scale action pre-training from unlabeled video, achieving a 45\% absolute improvement over prior vision-language-action baselines on the RoboTwin 2.0 benchmark.

Language compositionality has also been leveraged to improve world model generalization. \textbf{RoboDreamer}~\cite{robodreamer2024} factorizes video generation through the natural compositionality of language, parsing instructions into lower-level primitives that condition separate models to generate videos compositionally, enabling generalization to unseen combinations of objects and actions. \textbf{SuSIE}~\cite{black2024susie} takes a lightweight approach by using a language-conditioned image editing diffusion model---fine-tuned from InstructPix2Pix on video data---as a world model for subgoal prediction, alternating between high-level image editing and low-level goal-reaching to achieve state-of-the-art performance on the zero-shot CALVIN benchmark. These unified architectures illustrate a broader trend toward treating vision, language, and action as co-equal modalities within a single predictive framework, rather than treating language as an auxiliary input to a primarily visual system.

\paragraph{Comparative analysis and open challenges.}
The six methodological families surveyed above represent fundamentally different answers to the question of how language should participate in world modeling. Language-conditioned latent dynamics models preserve the well-understood RSSM framework while gaining compositional task specification and improved generalization. LLM-based world models offer rich prior knowledge and flexible text-based reasoning but currently lack the causal fidelity required for reliable multi-step simulation. Text-conditioned video generation achieves photorealistic world simulation at scale but faces unresolved questions about whether visual generation implies genuine physical understanding. Joint embedding approaches avoid the computational burden and information-theoretic inefficiency of pixel-space prediction, operating instead in compact representation spaces that may be more suitable for decision-making. Semantic world models bypass visual prediction entirely, reasoning in language space with dramatic efficiency gains for tasks where semantic abstraction suffices. Unified multimodal architectures seek to subsume all modalities within a single framework, though the architectural complexity required to support flexible mode-switching remains a practical challenge.

Several cross-cutting open problems emerge from this analysis. First, the question of \emph{causal grounding} remains largely unresolved: most language-conditioned world models learn correlational rather than causal relationships between language and dynamics, and establishing whether a model has truly grounded language to physical consequences---rather than exploiting surface-level statistical regularities---requires new evaluation protocols that go beyond task success metrics~\cite{gkountouras2025causal}. Second, \emph{compositional generalization} is frequently claimed but rarely demonstrated rigorously; while language provides compositional structure, it remains unclear whether current models achieve systematic generalization to novel combinations of concepts or merely interpolate within training distributions. Third, \emph{long-horizon consistency} poses a fundamental challenge across all families: maintaining coherent world state predictions over extended text-conditioned rollouts is difficult whether the model operates in pixel space, latent space, or language space. Fourth, the optimal \emph{level of abstraction} for world model prediction---pixels, latent representations, or semantic descriptions---likely depends on the downstream task, but principled methods for selecting or adapting this abstraction level are lacking. Finally, the \emph{computational cost} of large-scale language-augmented world models, from Cosmos at 14 billion parameters to LWM at one million token context, presents significant barriers to accessibility and reproducibility, motivating research into efficient architectures that preserve the benefits of multimodal integration at reduced scale.

\section{Categorization of World Models by Reasoning Strategy}
\label{sec:reasoning}
A world model is only as useful as the reasoning it enables. Whereas the previous sections examined how world models are structured (Section~\ref{sec:architecture}) and trained (Section~\ref{sec:methodology}), this section focuses on how their predictions are used for decision-making. We organize these reasoning strategies into four categories. Imagination-based planning (Section~\ref{sec: Imagination-based planning}) uses the model to simulate candidate action sequences and choose promising trajectories through search or optimization. Policy learning with a world model trains a reactive policy within imagined rollouts, reducing reliance on online environment interaction. Counterfactual reasoning (Section~\ref{sec:counterfactual}) asks what would have happened under an alternative action, supporting causal analysis and credit assignment. Finally, planning under uncertainty (Section~\ref{sec:planning_uncertainty}) addresses imperfect predictions and partial observability, emphasizing robust decision-making under model uncertainty.

\subsection{Imagination-based planning}
\label{sec: Imagination-based planning}

A fundamental challenge in intelligent systems is how to reason about future outcomes without relying exclusively on costly interaction with the physical world. Humans routinely address this problem through mental simulation, imagining possible consequences before acting~\cite{hamrick2019analogues}. Inspired by this capability, modern world models increasingly incorporate imagination-based planning, in which agents evaluate hypothetical futures using internal models of environmental dynamics~\cite{moerland2023model}.
\begin{figure}[htbp]
    \centering
    \includegraphics[width=0.8\linewidth]{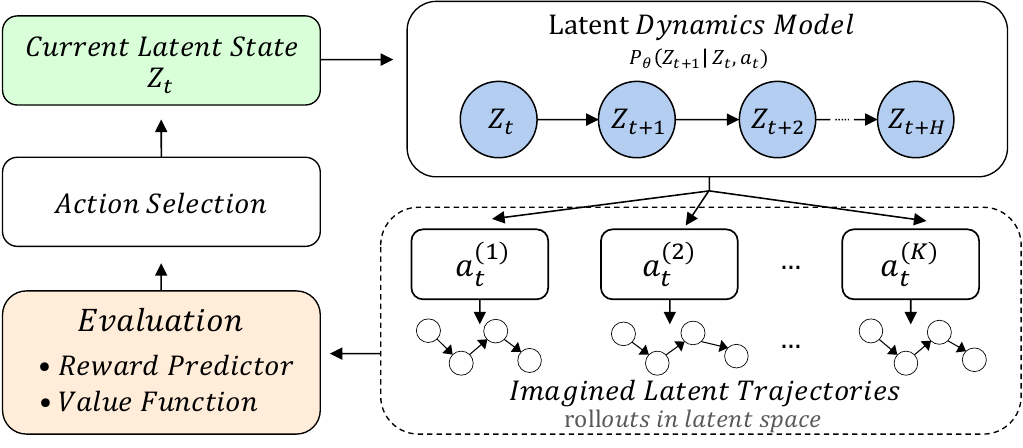}
    \caption{Imagination-based planning in latent world models. 
Starting from the current latent state $Z_t$, a learned dynamics model is used to simulate future states in latent space. 
Multiple candidate actions are evaluated through imagined rollouts, which are scored using a reward predictor and a value function. 
The selected action is then executed and the process repeats. 
For clarity, the figure illustrates a simplified branching over actions; in practice, planning is performed over action sequences and rollouts are conditioned on actions at each step.}
    \label{fig:imagination}

\end{figure}
In world models, imagination can be formalized as trajectory optimization in learned latent dynamics. Rather than interacting with the environment to evaluate actions, an agent internally constructs and evaluates possible future trajectories, effectively decoupling decision-making from physical experience. This perspective reframes planning as a computational problem: optimizing actions over simulated latent trajectories under a learned transition model.

Formally, let $z_t$ denote the latent representation of the system at time $t$. Given a candidate sequence of actions $a_{t:t+H}$, the model simulates future latent states over a planning horizon $H$~\cite{hafner2019planet}:
\begin{equation}
Z_{t+k} \sim p_{\theta}(Z_{t+k} \mid Z_{t+k-1}, a_{t+k-1}), \quad k = 1, \dots, H,
\label{eq:latent_rollout}
\end{equation}
where $p_{\theta}$ is a learned probabilistic transition model. These simulated states are evaluated through task-relevant predictors:
\begin{equation}
r_{t+k} \sim p_{\phi}(r \mid Z_{t+k}), \quad v_{t+k} = \mathbb{E}[V(Z_{t+k})],
\label{eq:reward_value_pred}
\end{equation}
allowing the agent to estimate long-term consequences entirely within latent space~\cite{hafner2019dreamerv1}. Crucially, operating in latent space abstracts away high-dimensional perceptual detail while preserving decision-relevant structure, enabling planning to scale beyond raw observation space.

Despite differences in implementation, existing approaches can be unified under a single principle: decision-making is achieved by optimizing over imagined trajectories. The primary distinction lies in when this optimization is performed. One class of methods amortizes imagination into a parametric policy during training, while another performs explicit optimization at decision time through online reasoning. This distinction reflects a fundamental trade-off between amortized computation and on-demand deliberation in intelligent systems~\cite{moerland2023model}.

\subsubsection{Imagination during learning: background planning.}
In the first paradigm, imagination is primarily used to improve policies during training by generating synthetic experience. Instead of relying solely on real environment interaction, the world model produces large numbers of hypothetical trajectories, which are treated as additional training data for updating policy and value functions~\cite{sutton1990dyna}. 

Crucially, because the transition dynamics are implemented as differentiable neural networks, gradients of long-term value estimates can be propagated through imagined trajectories via backpropagation through time. This effectively turns imagination into a differentiable computation graph, enabling end-to-end optimization of long-horizon behavior and alleviating the need for explicit multi-step credit assignment from real trajectories.

Representative methods include \textbf{PlaNet}~\cite{hafner2019planet} and the \textbf{Dreamer family}~\cite{hafner2019dreamerv1, hafner2023dreamerv3}, which learn latent dynamics models and optimize policies entirely within imagined rollouts. These approaches demonstrate that a substantial portion of policy learning can be driven by internally generated experience, leading to significant improvements in sample efficiency compared to model-free reinforcement learning.

From a broader perspective, this paradigm can be interpreted as amortizing the cost of planning into the training phase: the agent learns a parametric policy that implicitly encodes the outcome of many imagined trajectories, thereby enabling fast decision-making at inference time without explicit search.

\subsubsection{Imagination at decision time: forward search.}
In the second paradigm, imagination is performed at inference time to guide action selection through online reasoning. Rather than relying solely on a pre-trained policy, the agent expands possible futures from the current state and evaluates their long-term consequences under a learned dynamics model, effectively performing planning on demand~\cite{hansen2022tdmpc}. 

This process is typically implemented via search procedures such as \textbf{Monte Carlo Tree Search (MCTS)}, where candidate action sequences are iteratively explored and refined based on predicted rewards and value estimates. Representative approaches include \textbf{MuZero}~\cite{schrittwieser2020mastering}, \textbf{EfficientZero}~\cite{ye2021efficientzero}, and \textbf{TD-MPC}~\cite{hansen2022tdmpc}, which integrate learned latent dynamics with search or trajectory optimization to achieve strong performance across a range of decision-making tasks.

A central concept in this setting is value equivalence~\cite{grimm2020value}, which shifts the objective of representation learning from reconstructing observations to preserving decision-relevant quantities such as rewards, values, and policies. As a result, the latent space is shaped to support accurate planning rather than faithful reconstruction, aligning model learning with downstream control objectives.

Unlike amortized approaches, this paradigm explicitly recomputes optimal actions for each state, trading increased computational cost for improved flexibility and robustness. In particular, online search enables the agent to adapt to out-of-distribution states and recover from model or policy errors by re-evaluating alternative futures at decision time.

From a broader perspective, this paradigm can be viewed as allocating computational resources to real-time deliberation rather than pre-computation: planning is performed explicitly when needed, allowing the agent to dynamically refine its decisions based on the current context and predicted outcomes.

\subsubsection{Cross-domain advantages of latent imagination.}
The advantages of imagination-based planning stem from a single underlying property: the decoupling of decision evaluation from physical interaction. By shifting computation from the external environment to an internal latent simulation, physical experience is effectively transformed into an infinitely reusable computational resource.

First, this decoupling unlocks substantial gains in sample efficiency through data amplification. Because internal rollouts are generated without the temporal and physical bottlenecks of real-world interaction, a single observed trajectory can be expanded into a multitude of imagined futures. This computational parallelization enables significantly faster policy improvement compared to purely model-free approaches. Second, imagination enables safe and risk-free exploration. Since hypothetical trajectories are evaluated entirely within the latent space, agents can safely simulate high-risk, rare, or irreversible outcomes. This eliminates physical consequences during the trial-and-error process, which is particularly crucial for safety-critical domains such as robotics and autonomous systems~\cite{moerland2023model}. Third, imagination provides a natural computational substrate for counterfactual and causal reasoning. By fixing the initial latent state and intervening on candidate action sequences, agents can evaluate alternative futures under identical environmental conditions. This elevates the model beyond simple predictive modeling, allowing for causal analysis of decision outcomes through explicit "what-if" queries.

Ultimately, these advantages are not restricted to classical control settings, but extend broadly to sequential decision-making under uncertainty. From a broader perspective, latent imagination serves as a general-purpose mechanism for transforming observational data into structured predictions over possible futures, thereby seamlessly unifying prediction, planning, and reasoning within a single computational framework.

\subsubsection{Open challenges: compounding errors and objective mismatch.}
Despite its advantages, imagination-based planning remains fundamentally constrained by imperfections in learned world models. A primary limitation arises from compounding errors, where small inaccuracies in one-step predictions accumulate over successive rollout steps, leading to a progressive divergence between imagined and real trajectories. This phenomenon can be understood as a form of distribution shift: as rollouts extend further into the future, the model is increasingly queried on states that lie outside the support of its training data, resulting in internally consistent yet physically implausible imagined futures.

A second challenge lies in objective mismatch~\cite{lambert2020objective}. Many world models are trained using generative or reconstruction-based objectives, such as predicting observations or short-term signals, which are not necessarily aligned with the requirements of downstream decision-making. Consequently, improvements in predictive accuracy do not always translate into better control performance, revealing a fundamental misalignment between representation learning and decision objectives. Recent approaches such as \textbf{SPR}~\cite{schwarzer2020spr} partially address this issue by encouraging task-relevant latent consistency, but a general solution remains elusive.

Importantly, these two challenges are deeply intertwined: compounding errors are exacerbated when the learned representation fails to preserve decision-relevant structure, while objective mismatch can amplify long-horizon prediction errors by prioritizing perceptual fidelity over controllability. From a broader perspective, this suggests that imagination-based planning is not merely a problem of improving predictive accuracy, but a joint problem of representation learning, dynamics modeling, and decision alignment.

Addressing these limitations requires advances along multiple axes, including improving long-horizon consistency, learning task-aligned latent representations, and developing objectives that explicitly couple model learning with decision-making. More fundamentally, it calls for a shift from viewing world models as passive predictors toward treating them as decision-oriented abstractions optimized for planning and control.

\subsection{Policy learning with a world model}
Reinforcement learning (RL) has achieved impressive results in a wide range of sequential decision-making problems. However, its heavy reliance on large-scale environment interaction remains a major limitation, especially in domains where data collection is expensive, slow, or unsafe~\cite{gu2024review,zheng2025deepresearcher}. This challenge has led to growing interest in model-based reinforcement learning, where an agent learns an internal model of the environment and uses it to support decision-making. Within this broader paradigm, world models have emerged as an important framework for improving sample efficiency and enabling policy learning through internal simulation rather than direct interaction alone~\cite{dong2026learning,li2025robotic}.

A world model generally consists of two closely related components: a representation model and a dynamics model~\cite{ding2025understanding,zuo2025gaussianworld}. The representation model maps high-dimensional observations into a compact latent state, while the dynamics model predicts how this latent state evolves under the effect of actions. In many formulations, the model may also predict rewards and termination signals, so that imagined trajectories can be used not only for forecasting future states but also for evaluating candidate decisions. The role of this design is not limited to dimensionality reduction. More importantly, it defines a predictive latent space in which policy learning or planning can be carried out more efficiently than in the original observation space.

Once such a model is learned, the main question becomes how it should be used for control. In other words, the key issue is not only whether the model can represent the environment accurately, but also how imagined trajectories contribute to policy optimization. Existing methods differ mainly in this respect. Broadly speaking, policy learning in world model-based RL can be organized into three major approaches: model-generated data augmentation, latent imagination for actor-critic learning, and gradient-based optimization through differentiable dynamics~\cite{xu2026specialist, Zidan2026}.

The first approach uses the learned model to generate synthetic transitions that supplement real experience. This idea can be traced back to early work such as Dyna-Q~\cite{hwang2014model}, where real interaction and model-based updates are interleaved during learning. In modern model-based RL, this strategy often appears in a more controlled form: the agent performs short imagined rollouts from states sampled from a replay buffer and then uses the resulting transitions for value or policy updates. Methods such as MBPO~\cite{yu2020mopo} are representative of this line of work. Rather than relying on long-horizon simulations, MBPO limits model rollouts to short segments in order to reduce the effect of compounding errors while still improving data efficiency. Related methods such as STEVE~\cite{buckman2018sample} further explore how uncertainty-aware model rollouts can be incorporated into target estimation. The strength of this family of methods lies in its relative conservatism: the model is used to expand training data, but policy learning is not placed entirely at the mercy of long imagined trajectories.

A second line of work performs policy learning directly in latent imagination. In this setting, the world model is not merely a source of additional synthetic transitions; instead, imagined latent trajectories become the main medium through which policy evaluation and improvement are carried out. PlaNet~\cite{hafner2019learning} is an important early example of this direction, showing that compact latent dynamics can support planning directly from image observations. The Dreamer family~\cite{okada2021dreaming,deng2022dreamerpro} extends this idea further by combining learned latent dynamics with actor-critic learning. After fitting a latent world model from real experience, Dreamer rolls forward imagined latent trajectories and uses them to train value and policy networks entirely in latent space. This approach offers two advantages. First, policy optimization becomes computationally efficient because it avoids repeatedly reconstructing high-dimensional observations. Second, the representation and control components become more tightly connected, since the latent state must preserve information that is useful not only for prediction, but also for reward estimation and long-horizon decision-making. For this reason, latent imagination methods place particularly strong demands on representation quality.

A third approach takes advantage of differentiability in the learned dynamics and performs policy optimization through analytic gradients. In model-free RL, policy gradients are typically estimated through likelihood-ratio methods or actor-critic approximations, both of which may suffer from high variance. By contrast, if the learned transition model is differentiable, one can backpropagate reward-related signals through imagined trajectories and directly update policy parameters. This general idea is reflected in work on Stochastic Value Gradients (SVG)~\cite{heess2015learning}, which showed that a learned stochastic dynamics model can be used to propagate gradients through trajectories for policy learning. Such methods are especially attractive in continuous control, where actions and latent transitions are naturally amenable to end-to-end differentiation. At the same time, they are highly sensitive to model fidelity. If the learned dynamics become inaccurate in regions visited by the policy, then the resulting gradients may guide the policy toward behaviors that are effective only inside the model rather than in the true environment.

Although these three approaches are closely related, they make different assumptions about how the model should support control. Data augmentation methods use the model mainly as a provider of additional experience and therefore tend to be more conservative. Latent imagination methods rely more directly on imagined trajectories during policy optimization and thus require stronger latent representations. Gradient-based methods go even further by allowing the internal structure of the model itself to shape policy updates, which can be highly efficient but also more vulnerable to model errors. As a result, the value of a world model cannot be judged independently of the policy learning mechanism built upon it.

An important point, however, is that not all world models are designed around observation reconstruction. Much of the literature on latent imagination, including PlaNet and Dreamer, learns latent states that remain closely tied to reconstructive or predictive modeling of the environment. By contrast, decision-oriented models such as MuZero~\cite{he2023model} take a different route. MuZero does not aim to reconstruct raw observations; instead, it learns a latent dynamics model sufficient for predicting task-relevant quantities such as reward, value, and policy-related transitions. This distinction is conceptually important for a survey of policy learning. It suggests that, from the perspective of control, a useful world model need not reproduce every detail of the external world. In some cases, it may be more effective to model only the aspects of dynamics that matter for decision-making.

Despite the promise of these methods, policy learning with world models faces several persistent challenges. One of the most important is compounding error. Even if one-step prediction error appears small, small inaccuracies may accumulate over multiple imagined steps and distort long-horizon return estimates. This becomes especially problematic when the policy is trained on imagined trajectories, because the policy may adapt to flaws in the learned model rather than to the true environment. In extreme cases, the agent discovers trajectories that achieve unrealistically high reward under the model but fail when executed in the real environment. This phenomenon is often described as model exploitation and remains a central concern in world model-based RL.

A related challenge is the mismatch between the objective used to train the world model and the requirements of downstream control. A model that performs well under reconstruction or prediction metrics does not necessarily provide the most useful representation for policy learning. For example, it may preserve visually rich but task-irrelevant details while failing to represent reward-critical structure. This problem helps explain why policy learning performance does not always correlate with prediction quality in a straightforward way. It also motivates the growing interest in task-aware latent representations and decision-oriented world modeling, as illustrated by the contrast between reconstructive latent models and approaches such as MuZero.

To mitigate these problems, several practical strategies have been proposed. One common approach is to use short-horizon rollouts so that model-generated errors do not have time to accumulate excessively~\cite{lu2026contextual}. Another is uncertainty-aware modeling, often implemented through ensembles or probabilistic dynamics models, which helps identify when imagined trajectories are less trustworthy~\cite{lou2024uncertainty}. Multi-step consistency objectives, such as latent overshooting, have also been introduced to improve long-horizon prediction quality. In addition, some work explores discrete or otherwise structured latent spaces in order to reduce the tendency of policies to exploit unrealistic details of continuous latent representations. Although these methods differ in form, they are all motivated by the same core requirement: policy learning must remain robust even when the learned model is imperfect.

The relevance of this problem becomes even clearer in the setting of Offline RL. Because the agent can no longer interact with the environment to correct its mistakes, it must rely entirely on a fixed dataset and whatever inductive support is provided by the learned model. In this context, world models can offer an important mechanism for policy evaluation, imagined rollout, or constrained policy improvement. At the same time, offline settings also make model bias more dangerous, since erroneous imagined trajectories cannot be corrected through further exploration. For this reason, the relationship between world models and offline RL has become an active area of recent research.

Recent work has also explored more expressive sequence models for world modeling, especially transformer-based architectures~\cite{burchi2025learning}. Compared with recurrent dynamics models, transformers are often better at capturing long-range temporal dependencies and complex sequential structure, which may improve latent prediction and imagined rollout quality. This trend has broadened the design space of world model-based RL, particularly in domains that require longer memory or richer sequence understanding. More generally, the field is moving toward increasingly reusable predictive models that can support policy learning across multiple tasks instead of being trained from scratch for a single environment. From the perspective of policy learning, this shift is significant because it points toward a future in which control policies are learned on top of more general latent dynamics models rather than task-specific simulators alone.

In summary, world models provide a promising framework for reinforcement learning by shifting part of the learning process from direct environment interaction to internal simulation. Their role in policy learning is not limited to improving sample efficiency. They also introduce new mechanisms for control, including model-generated updates, latent imagination, and gradient-based optimization through differentiable dynamics. At the same time, the success of these methods depends critically on whether the learned model captures the aspects of the environment that are relevant for decision-making, and whether policy optimization can remain robust to model bias. For this reason, policy learning in world model-based RL is best understood not simply as an extension of model-free RL, but as a distinct framework in which representation learning, dynamics modeling, and control are deeply intertwined.

\subsection{Counterfactual reasoning}
\label{sec:counterfactual}

Counterfactual reasoning represents a fundamental shift in the role of world models, transforming them from predictive simulators into causal reasoning systems. While conventional world models estimate future trajectories conditioned on observed states and actions~\cite{ha2018world, hafner2019planet}, counterfactual reasoning addresses a qualitatively different question: given an observed trajectory, what would have occurred under an alternative decision at some past time step? This formulation aligns with the highest level of Pearl’s ladder of causation, moving beyond association and intervention toward reasoning about alternative histories~\cite{pearl2018book}.

In the context of world models, counterfactual reasoning can be formalized as latent-conditioned re-simulation under intervention. Unlike standard forward rollouts, which propagate dynamics from arbitrary states, counterfactual queries must remain anchored to the specific latent conditions that generated the observed trajectory. As a result, counterfactual reasoning requires not only forward prediction, but also reconstruction of the latent causal factors underlying past observations.

\subsubsection{The abduction--action--prediction pipeline.}The computation of counterfactuals follows the classical abduction--action--prediction paradigm in causal inference~\cite{pearl2009causality}, which can be formalized as:\begin{equation}z_t \sim q_\phi(z_t \mid o_{\le t}, a_{<t}), \quad a_t \leftarrow do(a_t'), \quad z_{t+1}' = f_\theta(z_t, a_t').\end{equation}

The first term corresponds to abduction, where the model infers not only the latent state but also the underlying exogenous noise (unobserved environmental randomness) that generated the specific observed trajectory. This step is inherently underdetermined: multiple latent configurations may explain the same observations, introducing ambiguity into the inferred causal state. In world models, this ambiguity is particularly pronounced, as the representation must simultaneously support accurate forward prediction and encode the precise causal context required for valid interventions.The second term represents the intervention, replacing the factual action with a counterfactual alternative while preserving the inferred latent context. Unlike standard policy evaluation, this operation explicitly enforces a $do$-operator semantics, ensuring that only the specific causal mechanism associated with the action is modified, while all other environmental factors remain rigidly fixed.The third term performs prediction, rolling the system forward under the modified dynamics~\cite{hafner2019dreamerv1}. Notably, this rollout must remain strictly consistent with both the inferred latent state and the intervened action, placing strong requirements on the stability and causal fidelity of the learned transition model.

Crucially, both factual and counterfactual trajectories share the exact same inferred latent state at time $t$, differing only in the intervened action. Counterfactual reasoning is therefore not a naive re-simulation from arbitrary states, but a causally conditioned rollout that strictly isolates the effect of decision changes under identical latent circumstances.

Ultimately, the abduction--action--prediction pipeline reveals that counterfactual reasoning is fundamentally a problem of disentangling inference, intervention, and prediction within a shared latent space. The reliability of counterfactual conclusions therefore depends not only on predictive accuracy, but critically on the identifiability and causal structure of the learned representation.

\subsubsection{Why counterfactuals matter: isolating decision effects.}
The core value of counterfactual reasoning lies in its unique capacity to isolate the causal contribution of decisions. While standard predictive models can forecast outcomes conditioned on observed trajectories, they inherently fail to attribute responsibility to specific actions. Counterfactual reasoning overcomes this by enabling the direct comparison of alternative decisions under strictly identical latent conditions, thereby definitively disentangling the true effects of an action from the underlying environmental dynamics.

In the context of offline reinforcement learning—where online interaction is typically limited or unsafe~\cite{moerland2023model}—counterfactual world models empower agents to reinterpret historical trajectories. By estimating how different actions would have altered long-term returns~\cite{buesing2019woulda}, this approach effectively transforms static observational datasets into a rich substrate for causal evaluation, facilitating robust policy improvement without requiring further environment interaction.

Beyond policy optimization, this capability is paramount in safety-critical domains such as autonomous driving and medical decision-making~\cite{chen2025medicalwm}. In these environments, determining whether an adverse outcome could have been prevented is just as critical as predicting its occurrence. Here, counterfactual reasoning underpins accountability, risk analysis, and decision auditing, fundamentally elevating the role of world models from mere prediction to causal explanation.

Ultimately, counterfactual reasoning serves as a critical complement to imagination-based planning by introducing a causal interpretation layer over simulated trajectories. While imagination proactively explores potential futures, counterfactuals retroactively explain realized outcomes by identifying how alternative interventions would have reshaped them. Together, they establish a dual perspective on decision-making, seamlessly unifying forward-looking prediction with backward-looking causal attribution within a cohesive world modeling framework.

\subsubsection{Toward causally structured world models.}
Reliable counterfactual reasoning imposes stringent requirements on representation structure. A latent state that is strictly sufficient for forward prediction is not necessarily suitable for counterfactual analysis unless it admits a modular causal decomposition. In conventional latent-variable models with entangled representations, intervening on a single action can inadvertently induce widespread and unstructured changes across the latent space. This violates a core tenet of causal inference: interventions should remain strictly localized, affecting only their specific downstream mechanisms.

This limitation exposes a fundamental gap between predictive sufficiency and causal adequacy. Representations optimized purely for reconstructing observations or minimizing prediction error frequently fail to preserve the modular structure necessary for valid interventions~\cite{dehaan2019causal, arjovsky2019irm}. Consequently, true counterfactual reasoning demands more than just an accurate dynamics model; it necessitates a representation space inherently designed to support the localized and interpretable manipulation of isolated causal factors.

To bridge this gap, recent approaches increasingly integrate structural causal models (SCMs) into the design of world models, actively enforcing representations that respect causal modularity~\cite{schoelkopf2021toward}. A central objective in this pursuit is causal disentanglement—specifically, the separation of variables directly influenced by agent actions from those governing the broader, autonomous environment~\cite{seitzer2021causal, thomas2017independently}. This directly operationalizes the independent causal mechanisms (ICM) principle~\cite{peters2017elements}, which posits that the generative process of an environment should be factorized into independent, composable components that do not mutually interfere upon local intervention.

Under such causally structured representations, counterfactual interventions manifest as precise, localized modifications of specific mechanisms, rather than arbitrary and unpredictable perturbations of a monolithic latent space. Ultimately, this paradigm shift dictates that future world models must evolve beyond serving merely as compact predictive encodings. Instead, they must function as causally structured representations, where the very geometry of the latent space faithfully reflects the true underlying factorization of the environment's dynamics.

\subsubsection{Fundamental limits: non-identifiability and counterfactual validity.}
Despite its profound conceptual appeal, counterfactual reasoning in world models faces formidable foundational challenges. From a theoretical perspective, the core bottleneck lies in non identifiability the well-documented phenomenon where multiple distinct latent dynamics models can perfectly explain the observed factual data while yielding entirely divergent counterfactual predictions~\cite{locatello2019challenging}. Consequently, achieving mere observational accuracy is fundamentally insufficient to guarantee counterfactual validity.

From a practical standpoint, counterfactual queries inherently risk pushing models into out-of-distribution (OOD) regimes. This vulnerability is particularly acute when the proposed counterfactual action deviates significantly from the data-collecting behavioral policy~\cite{janner2019mbpo}. In such extrapolation scenarios, model rollouts may remain internally coherent yet causally fallacious—a critical failure mode termed counterfactual hallucination~\cite{lambert2020objective}. More fundamentally, this exposes a severe vulnerability under distribution shift: predictive accuracy and counterfactual correctness can easily become uncoupled, laying bare the inherent limitations of relying solely on standard generative training objectives.

Addressing these formidable bottlenecks requires advancing far beyond merely scaling up predictive models. It necessitates the integration of stronger causal structural priors, the development of uncertainty-aware inference mechanisms capable of flagging OOD interventions, and the creation of rigorous evaluation protocols explicitly designed to assess causal reliability rather than just reconstruction loss. Ultimately, the realization of trustworthy counterfactual world models demands representations that do more than passively reproduce observed histories; they must reliably and consistently support grounded reasoning about alternative realities.

\subsection{Planning under uncertainty}
\label{sec:planning_uncertainty}

A world model is only as useful as the decisions it enables. Because learned dynamics models are inevitably imperfect---trained on finite data and approximating complex physical processes---every plan derived from such a model is subject to uncertainty. Two distinct sources contribute. \emph{Aleatoric} (environmental) uncertainty arises from inherently stochastic transitions: the same action applied in the same state may produce different outcomes due to sensor noise, unobserved variables, or genuine randomness in the environment. \emph{Epistemic} (model) uncertainty stems from the learner's incomplete knowledge: regions of the state--action space that are poorly covered by training data yield unreliable predictions. Effective planning must account for both.

\textbf{Stochastic dynamics and distributional rollouts.}
The most direct approach encodes aleatoric uncertainty into the world model itself. The RSSM~\cite{hafner2019planet} maintains a stochastic latent variable $z_t$ sampled from a learned prior at each step, so that multi-step rollouts naturally produce a \emph{distribution} over future trajectories rather than a single deterministic path. DreamerV2~\cite{hafner2021dreamerv2} replaced Gaussian latents with discrete categorical distributions, improving the model's ability to represent multi-modal outcomes (e.g., a pedestrian turning left or right). DreamerV3~\cite{hafner2023dreamerv3} extended this with symlog-transformed predictions and categorical value heads, ensuring that the actor--critic trained on imagined rollouts remains calibrated across domains with vastly different reward magnitudes. In these systems, the policy is optimized with respect to the \emph{expected} return under the model's stochastic dynamics, implicitly averaging over aleatoric uncertainty.

\textbf{Ensemble-based epistemic uncertainty.}
Epistemic uncertainty is harder to capture because a single neural network tends to be overconfident on out-of-distribution inputs. Ensembles of world models offer a practical remedy. PETS~\cite{chua2018pets} trains an ensemble of probabilistic neural networks, each predicting a Gaussian distribution over next states, and plans using trajectory sampling across ensemble members. The \emph{disagreement} among ensemble predictions serves as a proxy for epistemic uncertainty: high disagreement signals that the model is extrapolating beyond its training data. Plan2Explore~\cite{sekar2020plan2explore} exploited this signal directly, using ensemble disagreement as an intrinsic reward to drive exploration toward states where the world model is most uncertain---a strategy that learns task-agnostic world models which can be rapidly adapted to new objectives. MBPO~\cite{janner2019mbpo} used an ensemble to determine a trust horizon: the policy trains on imagined rollouts only up to the horizon beyond which model predictions become unreliable, then falls back to real data. This adaptive truncation prevents the policy from exploiting inaccurate long-horizon predictions.

\textbf{Pessimistic and conservative planning.}
In offline settings---where the agent must learn entirely from a fixed dataset without further environment interaction---optimistic exploitation of model inaccuracies is particularly dangerous, because the agent cannot collect corrective data. MOPO~\cite{yu2020mopo} penalizes the reward of imagined transitions by a term proportional to the ensemble's predictive uncertainty, encouraging the policy to avoid state--action regions where the model is unreliable:
\begin{equation}
    \tilde{r}(s, a) = \hat{r}(s, a) - \lambda \cdot u(s, a)
\end{equation}
where $u(s, a)$ quantifies uncertainty (typically the maximum standard deviation across ensemble members) and $\lambda$ controls the degree of conservatism. MOReL~\cite{kidambi2020morel} takes a more aggressive approach, terminating imagined rollouts entirely when the model detects that the agent has left the support of the offline data. RAMBO-RL~\cite{rigter2022rambo} formulates the problem as a two-player game between the policy and an adversarial dynamics model, training the model to minimize policy performance within a constrained set of plausible dynamics. This adversarial framing provides worst-case robustness without requiring explicit uncertainty quantification.

\textbf{Safe planning under learned dynamics.}
Safety-critical applications impose hard constraints that the agent must satisfy even when the world model is wrong. Garc\'{\i}a and Fern\'{a}ndez~\cite{garcia2015safe} surveyed the landscape of safe RL, distinguishing between methods that modify the optimization criterion (risk-sensitive objectives) and those that constrain the policy's admissible set. In the world-model setting, As et al.~\cite{as2022constrained} proposed Bayesian world models for constrained policy optimization: a Gaussian process dynamics model provides calibrated uncertainty bounds, and the policy is optimized to maximize return while satisfying probabilistic safety constraints derived from these bounds. In autonomous driving, world models are used to simulate rare but dangerous scenarios (collisions, sudden braking, adverse weather) that are underrepresented in training data, enabling stress-testing of planning systems against tail events that would be impractical or unethical to reproduce in the real world~\cite{gao2024vista, hu2023gaia1}. In medical decision-making, the consequences of model error can be severe; MeWM~\cite{yang2025mewm} incorporates an inverse dynamics model that evaluates the predicted outcomes of treatment plans, providing a feedback loop that mitigates the risk of acting on inaccurate simulations.

\textbf{Disentangling controllable and uncontrollable dynamics.}
Not all uncertainty is equal from a planning perspective. Some variation in future states is a direct consequence of the agent's actions (controllable), while other variation is driven by exogenous factors beyond the agent's influence (uncontrollable)---the behavior of other vehicles, stochastic weather, or unseen patient physiology. Iso-Dream~\cite{pan2022iso} explicitly separates controllable and noncontrollable visual dynamics within the world model's latent space, enabling the planner to focus on aspects of the future it can influence while treating the rest as noise to be robust against. This decomposition is particularly relevant for multi-agent environments such as urban driving, where the ego agent's plan must be robust to the unpredictable actions of surrounding actors.

Despite the progress outlined above, several challenges remain. First, calibration---ensuring that model uncertainty estimates are neither overconfident nor excessively conservative---remains difficult for deep neural networks, and miscalibrated uncertainty directly translates into suboptimal or unsafe plans. Second, the computational cost of ensemble-based methods scales linearly with ensemble size, making them expensive for large-scale world models with billions of parameters. Third, there is no unified framework that jointly addresses aleatoric uncertainty, epistemic uncertainty, and hard safety constraints within a single planning algorithm; current methods typically handle one or two of these concerns in isolation. Finally, the interaction between long-horizon compounding error and uncertainty quantification is poorly understood: as rollout horizons increase, the distinction between model error and genuine stochasticity blurs, and current uncertainty estimators struggle to decompose the two.

\section{Categorization of World Models by Application Domains}
\label{sec:applications}
This section surveys nine major application areas, moving roughly from the most established to the most emerging. Robotics (Section~\ref{sec:robotics}) and autonomous driving (Section~\ref{sec:autodriving}) remain core settings for world model research, where latent dynamics models have shown strong gains in sample efficiency and sim-to-real transfer. Video prediction and scene understanding extend these models to forecasting visual futures, while multimodal and language-grounded agents use foundation models as implicit world simulators for embodied decision-making. Reinforcement learning and games (Section~\ref{sec:rl_games}) continue to serve as key testbeds for new architectures. We then examine newer applications in scientific and domain-specific modeling (Section~\ref{sec:scientific_modeling}), medical imaging and clinical decision-making (Section~\ref{sec:medical}), educational measurement (Section~\ref{sec:edu_measurement}), and business and finance (Section~\ref{sec:fin}), where world models are being adapted beyond the physical environments in which they first emerged.

\subsection{Robotics}
\label{sec:robotics}
Robotics is one of the most demanding application domains for world models because prediction quality is ultimately judged by physical execution rather than simulator return alone. Unlike agents acting in relatively clean discrete environments, robotic systems must cope with continuous kinematics, contact discontinuities, partial observability caused by occlusion and sensor noise, and the potentially irreversible consequences of a bad prediction. In this setting, world models are attractive for at least three reasons: they can reduce the cost of real-world data collection by supporting synthetic rollouts in latent or pixel space, enable counterfactual evaluation of candidate action sequences before physical commitment, and learn transferable representations of object and scene dynamics that generalize across tasks and embodiments. Recent surveys frame this landscape from complementary perspectives. Li et al.~\cite{li2025steprobot} review robotic manipulation through the joint lenses of perception, prediction, and control, while Li et al.~\cite{li2025embodied} propose a broader taxonomy based on functionality, temporal modeling, and spatial representation. Consistent with those reviews, this subsection keeps robotics as the application focus while organizing the discussion using several recurring technical families: latent dynamics models, self-supervised predictive architectures, generative world simulators, and structured or object-centric models. Table~\ref{tab:robotics_overview} summarizes representative examples.

\begin{table}[htbp]
\centering
\caption{Representative world models in robotics, organized by paradigm. Real Robot indicates evaluation on physical hardware beyond simulation.}
\label{tab:robotics_overview}
\setlength{\tabcolsep}{4pt}
\footnotesize
\begin{tabular}{@{}l c l l c@{}}
\toprule
\textbf{Model} & \textbf{Year} & \textbf{Paradigm} & \textbf{Key Contribution} & \textbf{Real} \\
\midrule
DayDreamer~\cite{wu2023daydreamer}       & 2022 & Latent dynamics   & RSSM on 4 physical platforms          & \checkmark \\
I-JEPA~\cite{assran2023ijepa}            & 2023 & Predictive/JEPA   & Masked image representation            & \\
UniSim~\cite{yang2023unisim}             & 2023 & Generative sim.   & Universal generative simulator         & \checkmark \\
TD-MPC2~\cite{hansentd}                  & 2023 & Latent dynamics   & Latent MPC + value learning            & \\
V-JEPA~\cite{bardes2023v}                & 2024 & Predictive/JEPA   & Masked video representation            & \\
DreMa~\cite{barcellona2024dream}         & 2024 & Structured / 3D   & Compositional digital twin             & \\
RoboDreamer~\cite{zhou2024robodreamer}   & 2024 & Latent dynamics   & Compositional language goals            & \\
DINO-WM~\cite{zhou2024dino}              & 2025 & Predictive/JEPA   & Frozen DINOv2 + zero-shot MPC          & \\
RWM~\cite{li2025robotic}                 & 2025 & Latent dynamics   & Dual-autoregressive offline MBRL       & \\
V-JEPA 2~\cite{assran2025vjepa2}         & 2025 & Predictive/JEPA   & 1.2B param.\ action-conditioned MPC    & \\
COSMOS~\cite{nvidia2025cosmos}            & 2025 & Generative sim.   & 720p video diffusion, 20M hrs          & \\
UWM~\cite{zhu2025unified}                & 2025 & Generative sim.   & Unified video + action diffusion       & \\
Cosmos Policy~\cite{nvidia2026cosmospolicy} & 2026 & Generative sim. & Video model adapted into policy        & \checkmark \\
PointWorld~\cite{huang2026pointworld}     & 2026 & Structured / 3D   & 3D point-flow world modeling           & \checkmark \\
\bottomrule
\end{tabular}
\end{table}

\paragraph{Latent Imagination for Robot Learning.}
The most direct application of world model principles to physical robotics is the use of latent dynamics models: learned simulators that compress sensory observations into compact state representations and roll out imagined futures for policy improvement. This paradigm, rooted in the Recurrent State-Space Model (RSSM) used by the Dreamer family~\cite{hafner2019dream,hafner2020mastering}, factorizes the latent state into a deterministic recurrent component $h_t = f_\theta(h_{t-1}, z_{t-1}, a_{t-1})$ and a stochastic component $z_t \sim q_\phi(z_t \mid h_t, x_t)$. In robotics, that factorization is attractive because it offers a tractable way to model both predictable motion and harder-to-predict contact events while training policies largely through imagined rollouts rather than repeated physical trials. For example, DayDreamer~\cite{wu2023daydreamer} demonstrated latent imagination on real robots—wheeled, quadruped, arm, and dexterous platforms—showing that it remains effective despite noise, delays, and irreversibility, not just in simulators. RoboDreamer~\cite{zhou2024robodreamer} builds on this by factorizing language instructions into reusable primitive skills with latent goal embeddings, enabling novel task compositions at test time.

A second important direction is offline deployment, where policies must be trained from pre-collected datasets rather than through repeated online interaction. \textbf{RWM}~\cite{li2025robotic} models this setting with a dual-autoregressive design and a self-supervised consistency objective aimed at improving long-horizon rollouts under dataset bias and partial observability. Its uncertainty-aware extension, \textbf{RWM-U}~\cite{li2025uncertainty}, further propagates epistemic uncertainty during rollout and couples the world model to uncertainty-penalized policy optimization, making the method substantially more robust on offline real-robot settings than earlier offline MBRL baselines. A complementary design is represented by \textbf{TD-MPC2}~\cite{hansentd}, which couples a latent dynamics model to short-horizon planning and value learning. Rather than using the model only for imagination-based policy training, TD-MPC2 uses it directly at decision time, illustrating how latent world models in robotics now span both imagination-heavy RL and tightly integrated planning-and-control pipelines.

\paragraph{Self-Supervised Predictive Architectures.}

A parallel and increasingly influential line of work develops predictive models in representation space rather than by reconstructing pixels, following LeCun's Joint-Embedding Predictive Architecture (JEPA) framework~\cite{lecun2022path}. Strictly speaking, some members of this family originated as general visual representation learners rather than robot-specific world models. They are nevertheless relevant here because later robotic planning systems directly build on their predictive abstractions, and because they make explicit a key design claim for robotics: prediction in an abstract latent space may be a better use of model capacity than reconstructing every texture, shadow, and background detail in the scene.

Building on this principle, I-JEPA~\cite{assran2023ijepa} showed that predicting latent representations of masked image regions produces semantically meaningful features without pixel reconstruction, providing a foundation for action-conditioned planning. V-JEPA~\cite{bardes2023v} extended latent prediction to video, masking and predicting spatiotemporal regions to bridge passive visual pretraining with active control. DINO-WM~\cite{zhou2024dino} made this bridge explicit for robotics, combining a frozen DINOv2 encoder with an action-conditioned transition model, enabling offline training and planning entirely in latent space. Finally, V-JEPA 2~\cite{assran2025vjepa2} scaled these ideas to a 1.2B-parameter video model with action conditioning, demonstrating that large-scale latent video pretraining can support robotic manipulation planning without pixel-level generation. Collectively, these works highlight that effective robotic world models need not be photorealistic, as long as latent predictions preserve task-relevant structure.

\begin{table}[htbp]
\centering
\caption{Comparison of self-supervised predictive world model architectures (JEPA family) for robotics. Offline indicates the model can be trained on pre-collected trajectories without environment interaction. Zero-shot indicates test-time generalization without per-task fine-tuning.}
\label{tab:jepa_comparison}
\small
\begin{tabular}{lccccc}
\toprule
\textbf{Model} & \textbf{Input} & \textbf{Encoder} & \textbf{Action-Cond.} & \textbf{Offline} & \textbf{Zero-shot} \\
\midrule
I-JEPA~\cite{assran2023ijepa}    & Image  & ViT (learned) & \texttimes & \checkmark & \texttimes \\
V-JEPA~\cite{bardes2023v}    & Video  & ViT (learned) & \texttimes & \checkmark & \texttimes \\
DINO-WM~\cite{zhou2025dinowm}    & Image  & DINOv2 (frozen)& \checkmark & \checkmark & \checkmark \\
V-JEPA 2~\cite{assran2025vjepa2} & Video  & ViT-H (learned)& \checkmark & \checkmark & \checkmark \\
\bottomrule
\end{tabular}
\end{table}

\paragraph{Generative World Simulators for Robotics.}
A third paradigm treats the world model as a generative simulator that produces action-conditioned video rollouts for planning, policy learning, or synthetic data generation. The core appeal is straightforward: a visually faithful, action-conditioned simulator can partially bridge the gap between abundant unlabeled video and scarce robot demonstrations, while providing a controllable testbed for counterfactual evaluation prior to real deployment.

Early work in this direction, exemplified by UniSim~\cite{yang2023unisim}, demonstrated that a single generative model trained on heterogeneous sources---robot demonstrations, human videos, and simulated trajectories---can serve as more than a passive predictor. It can act as a substrate for training planners and policies that subsequently transfer to physical robots. This insight motivates a key architectural question: how should video generation and action modeling be coupled? Unified World Models (UWM)~\cite{zhu2025unified} address this by jointly diffusing over video and actions within one backbone, allowing action-free video to contribute to the same pretrained representation as action-labeled robot data---a practically important property given the relative scarcity of high-quality demonstrations.

Scale and generality push this line further. COSMOS~\cite{nvidia2025cosmos} pretrains on massive video corpora as a broad prior over physical dynamics, treating task-level adaptation as a lightweight downstream step. Its companion Cosmos-Transfer1~\cite{nvidia2025cosmostransfer} targets sim-to-real transfer specifically, translating simulated trajectories into more photorealistic videos while preserving scene structure and motion. Cosmos Policy~\cite{nvidia2026cosmospolicy} goes a step further by collapsing the classical separation between world model and policy: future observations, values, and actions become jointly learnable outputs of one pretrained video backbone, blurring the boundary between prediction and control.

A complementary axis of variation concerns what the simulator predicts. Rather than generating 2D video, PointWorld~\cite{huang2026pointworld} operates in 3D point-flow space, predicting scene motion directly from RGB-D observations and candidate actions. This is especially appealing for manipulation, where interactions are more naturally described by spatial displacement than by image synthesis, and where tying the model to geometry facilitates embodiment transfer. At a different level of abstraction, GenSim2~\cite{hua2024gensim2} and RoboGen~\cite{wang2023robogen} leverage large language models to automatically generate diverse task curricula, reward functions, and training environments, using generative rollouts for policy evaluation without real data collection. Together, these approaches illustrate that the value of a world simulator need not be confined to policy execution: it can also substantially reduce the human engineering effort required to construct generalizable training pipelines.

\paragraph{Structured and Object-Centric World Models.}
Manipulation problems typically depend on a small number of entities whose identities, poses, affordances, and pairwise relations determine success or failure. Representing the entire scene as a monolithic latent vector obscures precisely this relational structure, motivating a class of models that instead maintain explicit decomposition over objects and their interactions.

One influential route combines slot-based perceptual decomposition~\cite{locatello2020object} with graph-style relational reasoning over object tokens. The robotics motivation is practical: preserving object-level state and pairwise geometry yields better extrapolation in contact-rich settings and supports combinatorial generalization to unseen configurations---both failure modes that flat scene representations handle poorly.

DreMa~\cite{barcellona2024dream} frames this idea as a learnable digital twin, coupling explicit world structure to imagination and planning so that predicted states remain geometrically consistent with what the robot can physically execute. This matters especially in contact-rich manipulation, where small errors in relative geometry accumulate rapidly and tend to dominate failure modes. 3D-VLA~\cite{zhen20243dvla} pursues a complementary angle by grounding reasoning and action generation in point clouds rather than 2D observations, gaining a more direct handle on occlusion, viewpoint change, and spatial arrangement. Together, these works converge on a shared lesson: robotic world models tend to benefit from inductive biases about objects, geometry, and interaction structure rather than from undifferentiated scene-level latents alone.

\paragraph{World Models for Sim-to-Real Transfer.}
The sim-to-real gap remains one of the most practically significant obstacles in robot learning. Its sources are multiple and interacting: contact dynamics differ from reality in stiffness, friction, and damping; visual appearance diverges in lighting, material reflectance, and sensor noise; and subtle mismatches in interactive dynamics accumulate into large trajectory deviations at deployment. World models address this challenge from two complementary directions---generating realistic training data that narrows the domain gap at the input level, and learning representations that abstract away distributional variation irrelevant to task execution.

The more direct route is appearance transfer: a generative model translates simulated trajectories into visually realistic videos while preserving scene geometry and motion, as in COSMOS-Transfer1~\cite{nvidia2025cosmostransfer}. A related strategy expands task diversity rather than visual realism alone---GenSim2~\cite{hua2024gensim2} uses LLM-generated tasks, assets, and scenarios to increase coverage over the variation that most often breaks sim-trained policies at deployment.

The second direction operates at the representation level. Models such as V-JEPA~\cite{bardes2023v} and DINO-WM~\cite{zhou2024dino} predict in latent space rather than reconstructing pixels, naturally deemphasizing low-level texture and lighting while preserving geometric and semantic structure more relevant to control. This helps explain why latent predictive models often transfer better than their visual fidelity alone would suggest.

\paragraph{Benchmarks and Evaluation.}
Evaluation in this area spans standardized simulation benchmarks and increasingly common physical hardware trials, reflecting growing emphasis on real-world validation. A persistent challenge is the evaluation validity gap: high simulation performance often fails to predict real-robot success, motivating benchmarks specifically designed to measure sim-to-real correlation. 


On the simulation side, the DeepMind Control Suite~\cite{tunyasuvunakool2020dm_control} remains standard for continuous-control sample efficiency, while RLBench~\cite{james2020rlbench} provides 100 manipulation tasks with demonstrations for evaluating world-model-based planning and imitation learning, and Meta-World~\cite{yu2020metaworld} stresses multi-task transfer across 50 tasks. For real-robot evaluation, the Open X-Embodiment dataset~\cite{openxembodiment2024}---over one million trajectories across 22 embodiments---has become central for measuring representation and policy transfer across morphologies. SimplerEnv~\cite{li2024evaluating} occupies a distinct niche: designed explicitly to improve correspondence between simulation scores and real-robot outcomes, it is particularly useful for testing whether simulated progress is likely to survive deployment.


\subsection{Autonomous driving and control systems}
\label{sec:autodriving}
Autonomous driving represents one of the most mature and commercially motivated application domains for world models, driven by the need to ensure safe navigation in open-world environments that are inherently unpredictable, high-dimensional, and safety-critical. Unlike laboratory-scale robotic manipulation, autonomous driving operates at the intersection of real-time perception, multi-agent interaction, regulatory compliance, and physical dynamics—making predictive simulation and counterfactual reasoning not merely convenient but operationally essential. A survey dedicated to this intersection is provided by Guan et al. \cite{guan2024world}, who document the rapid proliferation of learned environment models across the full perception, prediction, and planning stack of modern autonomous driving pipelines. Collectively, the models surveyed in this section span three complementary capabilities: generating photorealistic driving scenarios for training data augmentation, predicting action-conditioned future observations for planning, and learning compact 4D scene representations that preserve geometric structure over long horizons.

\textbf{Scene generation and data augmentation.} A primary motivation for deploying world models in autonomous driving is the scarcity of safety-critical edge cases—unusual weather, atypical road configurations, or near-miss events—that are difficult to collect at scale from real-world fleet operations. World models address this by generating photorealistic, action-controllable driving scenarios that supplement real data for downstream perception and planning training. \textbf{GAIA-1} \cite{hu2023gaia1} pioneered large-scale generative world modeling for this domain with a 9-billion-parameter architecture that jointly tokenizes video, text, and action inputs, enabling controllable generation of diverse scenarios conditioned on ego-vehicle behavior and textual scene descriptions. Its successor \textbf{GAIA-2} \cite{wayve2025gaia2} scaled this paradigm to multi-camera, multi-agent generation with fine-grained semantic control over individual traffic participants, trained on approximately 25 million video sequences across three countries —demonstrating that generative world models can function as scalable data engines for rare scenario synthesis. \textbf{MagicDrive} \cite{gao2023magicdrive} extended this line to spatially consistent multi-view generation across six camera angles simultaneously, introducing explicit 3D geometric control through BEV map and bounding box conditioning that ensures cross-view consistency absent from purely image-level generation approaches.

\textbf{Action-conditioned video prediction.} Beyond data augmentation, world models serve as closed-loop simulators that predict future observations given planned ego-vehicle actions, enabling planners to evaluate action consequences without physical execution. \textbf{DriveDreamer} \cite{wang2023drivedreamer} established a two-stage paradigm: a first stage learning structured traffic constraints via diffusion from historical data, and a second enabling action-conditioned future frame prediction. \textbf{DriveDreamer-2} \cite{zhao2024drivedreamer2} advanced this by integrating large language model guidance for corner-case generation, converting user-specified natural language queries into agent trajectory specifications that condition the world model to synthesize challenging rare scenarios on demand. \textbf{DriveDreamer4D} \cite{zhao2025drivedreamer4d} further extended this line to explicit 4D scene representation by combining video generation with 4D Gaussian Splatting, producing spatially consistent multi-view predictions with geometric supervision that support novel viewpoint synthesis. \textbf{Vista} \cite{gao2024vista} pushed the resolution frontier to 576×1024 pixels while introducing flexible action control modes—encompassing steering angle and speed commands, waypoint trajectory conditioning, and goal-point specification—unified through a dynamic conditioning prior within a single architecture. \textbf{GenAD} \cite{zheng2024genad} proposed a generalized predictive model framework emphasizing action controllability and trajectory conditioning across diverse driving environments, establishing that a unified world model architecture can capture dynamics across multiple geographic regions and sensor configurations.

\textbf{4D occupancy and geometric world models.} Image-based world models represent the environment as sequences of 2D projections, discarding explicit 3D geometric structure. A parallel line of work embeds 4D scene representations—combining spatial 3D structure with temporal evolution—directly into the world model architecture. \textbf{OccSora} \cite{wang2025occsora} proposed generating future 4D occupancy voxel grids as an alternative world representation, producing geometry-aware predictions that explicitly encode volumetric scene structure and support reasoning about occluded regions invisible in camera images. \textbf{OccWorld} \cite{zheng2024occworld} introduced a joint world model that simultaneously predicts future ego motion and 3D occupancy evolution in a unified latent space, demonstrating that occupancy-based world modeling enables more accurate long-horizon trajectory planning than camera-only counterparts by preserving geometric structure across the prediction horizon. These occupancy-based approaches are complementary to image-based methods: while they sacrifice photorealistic fidelity, they offer structured 3D representations that are more directly amenable to geometric planning and safety verification.

\textbf{LiDAR and multimodal world modeling.} Production autonomous driving systems routinely fuse camera, LiDAR, and radar inputs; however, most generative world models operate primarily in the image domain. \textbf{Copilot4D} \cite{zhang2024copilot4d} demonstrated that world model principles extend naturally to LiDAR point-cloud modalities by discretizing 3D point clouds via VQ-VAE and applying discrete diffusion for unsupervised future point-cloud prediction, achieving accurate occupancy forecasting without any labeled supervision. \textbf{UniWorld} \cite{min2023uniworld} proposed a unified multimodal world model that learns joint representations over camera images and LiDAR inputs, enabling consistent future prediction across both sensor modalities through a shared latent dynamics model and demonstrating improved downstream 3D detection and motion forecasting through world-model-based pre-training.

\textbf{World models for end-to-end planning.} Beyond perception-level simulation, world models have been directly integrated into end-to-end driving planners. \textbf{LCDrive} \cite{tan2025lcdrive} integrates chain-of-thought style reasoning with latent world model predictions, interleaving action-proposal tokens with world model rollouts to evaluate candidate trajectories counterfactually before committing to an action. \textbf{FutureX} \cite{xiang2025futurex} introduces an auto-think mechanism that dynamically activates a latent world model only when scene complexity warrants deliberative reasoning, reducing computational overhead during routine driving while maintaining full predictive capacity in complex scenarios. Think Twice Before Driving \cite{jia2023think} formalizes this deliberative planning paradigm by decoupling state estimation from policy optimization through an explicit world model, showing that agents that simulate before acting achieve substantially lower collision rates on closed-loop evaluation.

\textbf{Benchmarks and evaluation.} The field has converged on a set of standard benchmarks that evaluate different aspects of world model performance in the autonomous driving context, summarized in Table 2. The nuScenes dataset \cite{caesar2020nuscenes} provides 1,000 driving scenes with synchronized multi-camera and LiDAR data, and is widely used for evaluating future prediction quality through video FID, Fréchet Video Distance (FVD), and trajectory-conditioned PSNR. The Waymo Open Dataset \cite{sun2020scalability} provides larger-scale coverage at higher sensor resolution, enabling evaluation of long-horizon scene prediction fidelity. \textbf{CARLA} \cite{dosovitskiy2017carla} an open-source urban driving simulator, serves as the primary closed-loop evaluation environment, where world model-based planners are assessed on route completion rate, infraction rate, and driving score across diverse weather and traffic scenarios. The recently introduced \textbf{nuPlan} \cite{karnchanachari2024towards} provides reactive multi-agent closed-loop simulation, enabling evaluation of world models under interactive traffic participant behavior rather than pre-recorded scenario replay—an increasingly important capability as the field transitions from open-loop to closed-loop evaluation regimes.

\begin{table}[h]
\centering
\caption{Representative benchmarks for autonomous driving world model evaluation.}
\label{tab:ad_benchmarks}
\setlength{\tabcolsep}{4pt}
\begin{tabular}{@{}lllll@{}}
\toprule
\textbf{Benchmark} & \textbf{Year} & \textbf{Modality} & \textbf{Scale} & \textbf{Key Metric} \\
\midrule
nuScenes           & 2020 & Camera + LiDAR & 1,000 scenes   & FVD, traj.\ PSNR \\
Waymo Open         & 2020 & Camera + LiDAR & 1,950 segments & Prediction ADE/FDE \\
CARLA              & 2017 & Simulation     & Unlimited       & Driving score, infraction rate \\
nuPlan             & 2024 & Camera + LiDAR & 1,500 hours    & Closed-loop reactive score \\
\bottomrule
\end{tabular}
\end{table}

\textbf{Challenges and future directions.} Autonomous driving has proven to be one of the most fertile application domains for world model research, with large-scale fleet data availability, clear evaluation metrics, and direct commercial impact jointly motivating rapid progress. The field is currently transitioning from open-loop video prediction—where models are evaluated on single-step or short-horizon frame quality—toward closed-loop planning evaluation, where the world model must support reliable action selection over extended horizons under interactive traffic dynamics. Several key challenges remain. First, long-horizon consistency continues to limit deployed systems: current world models struggle to maintain geometrically and semantically coherent predictions beyond a few seconds, with compounding errors degrading scene structure over extended horizons. Second, multi-agent modeling—accurately predicting the joint future behavior of all traffic participants conditioned on the ego-vehicle's planned actions—requires capturing complex inter-agent dependencies that current architectures model only approximately. Third, the sim-to-real gap persists: models trained on recorded fleet data may not faithfully represent the interactive dynamics that emerge in closed-loop deployment, where the ego-vehicle's actions influence the behavior of surrounding agents. Fourth, safety verification of world model-based planners demands formal guarantees that remain elusive for high-capacity learned models, limiting their adoption in certified safety-critical deployments. Finally, the computational cost of diffusion-based and transformer-based world models at driving-relevant resolutions poses deployment challenges, motivating research into efficient architectures that preserve predictive quality under strict real-time latency constraints.

\subsection{Video prediction and scene understanding}

Video prediction---forecasting future frames given past observations and optional conditioning signals---is one of the oldest and most natural formulations of world modeling. Scene understanding extends this beyond raw pixel forecasting to extracting structured representations of the depicted world: object identities, spatial relations, physical properties, and causal interactions. Together, these capabilities form a foundation for numerous downstream applications, from autonomous navigation to embodied agent training and creative content generation.

Unlike the methodological treatment of diffusion-based world models in Section~\ref{sec:diffusionworldmodel}, which examines \emph{how} generative architectures learn environment dynamics, this section surveys video prediction and scene understanding as an \emph{application domain}. The focus is on what tasks are being solved, how models are evaluated, and what fundamental limitations persist---spanning multiple methodological families rather than detailing any single architecture. We refer readers to Section~\ref{sec:diffusionworldmodel} for architectural specifics of the diffusion-based systems mentioned here. We also distinguish this discussion from domain-specific video prediction in autonomous driving (Section~6.2), which addresses sensor-specific and safety-critical requirements unique to that setting, and from multimodal agent systems (Section~6.4), which discuss some of the same models from the perspective of agent training environments. Table~\ref{tab:video_prediction_overview} summarizes representative models organized by paradigm.

\begin{table}[htbp]
\centering
\caption{Representative world models for video prediction and scene understanding, organized by paradigm.}
\label{tab:video_prediction_overview}
\footnotesize
\begin{tabular}{>{\raggedright\arraybackslash}p{2.8cm} c >{\raggedright\arraybackslash}p{2.8cm} >{\raggedright\arraybackslash}p{4.5cm} c}
\toprule
\textbf{Model} & \textbf{Year} & \textbf{Paradigm} & \textbf{Key Contribution} & \textbf{Interactive} \\
\midrule
SV2P~\cite{babaeizadeh2018sv2p}              & 2018 & Stochastic VAE       & Variational latent for multi-future prediction & \\
SVG~\cite{denton2018svg}                     & 2018 & Stochastic VAE       & Learned prior for stochastic video generation & \\
SAVP~\cite{lee2018savp}                      & 2018 & VAE + GAN            & Combined variational and adversarial training & \\
VideoGPT~\cite{yan2021videogpt}              & 2021 & Autoregressive       & VQ-VAE + GPT for video token prediction & \\
Sora~\cite{openai2024sora}                   & 2024 & Diffusion transformer & Minute-long high-fidelity video generation & \\
SVD~\cite{blattmann2023svd}                  & 2023 & Latent diffusion     & Scalable image-to-video diffusion model & \\
CogVideoX~\cite{yang2025cogvideox}           & 2024 & Diffusion transformer & Expert adaptive normalization for text-to-video & \\
Open-Sora~\cite{opensora2024}                & 2024 & Diffusion transformer & Open-source efficient video generation & \\
Genie~\cite{bruce2024genie}                  & 2024 & Latent action model  & Interactive 2D worlds from unlabeled video & \checkmark \\
Genie 2~\cite{parkerholder2024genie2}        & 2024 & Latent action model  & Persistent interactive 3D environments & \checkmark \\
DIAMOND~\cite{alonso2024diamond}             & 2024 & Pixel diffusion      & Pixel-space diffusion world model, Atari 100k SOTA & \checkmark \\
GameNGen~\cite{valevski2024gamengen}         & 2024 & Latent diffusion     & Neural DOOM engine at 20+ FPS & \checkmark \\
Oasis~\cite{oasis2024}                       & 2024 & Autoregressive       & Transformer-based interactive game simulator & \checkmark \\
Pandora~\cite{xiang2024pandora}              & 2024 & Hybrid               & Language-action conditioned video world model & \checkmark \\
Diffusion Forcing~\cite{chen2024diffusionforcing} & 2024 & Causal diffusion & Next-token meets full-sequence diffusion & \\
GaussianWorld~\cite{zuo2025gaussianworld}    & 2025 & 3D Gaussian          & Streaming 3D occupancy prediction & \\
\bottomrule
\end{tabular}
\end{table}

\paragraph{From Deterministic to Stochastic Video Prediction.}
Early deep learning approaches to video prediction employed deterministic convolutional architectures that directly regressed future frames from past observations. While conceptually straightforward, these models produced characteristically blurry predictions because they averaged over the inherently multi-modal distribution of possible futures---a pixel in an occluded region, for instance, could plausibly take many different values, and a mean-squared-error loss encourages the model to hedge by outputting the mean. Stochastic video prediction models addressed this fundamental limitation by introducing latent random variables that capture the uncertainty over future outcomes. \textbf{SV2P}~\cite{babaeizadeh2018sv2p} augmented an action-conditioned video prediction network with a time-varying latent variable sampled from a learned posterior, enabling the generation of multiple plausible futures from a single past sequence. \textbf{SVG}~\cite{denton2018svg} introduced a learned prior network that generates latent codes autoregressively, conditioning each step on the previous frame's representation to produce temporally coherent stochastic predictions. \textbf{SAVP}~\cite{lee2018savp} combined variational and adversarial training objectives, using a VAE for diversity and a GAN discriminator for sharpness, demonstrating that the two approaches are complementary rather than competing. Oprea et al.~\cite{oprea2020review} provided a comprehensive review documenting the evolution from deterministic to stochastic to adversarial approaches, establishing video prediction as a well-defined research area within computer vision.

These early stochastic models operated primarily on low-resolution, short-horizon sequences---typically predicting 10--30 frames at resolutions below $256 \times 256$. Their contributions were nevertheless foundational: they established the principle that video prediction is fundamentally a distribution estimation problem, introduced the variational and adversarial toolkits that subsequent work would scale, and identified the core tension between sample diversity and temporal consistency that continues to shape the field.

\paragraph{Autoregressive and Token-Based Video Prediction.}
A parallel line of work reformulated video prediction as a sequence modeling problem over discrete tokens, drawing on the success of autoregressive language models. \textbf{VideoGPT}~\cite{yan2021videogpt} pioneered this approach by training a VQ-VAE to discretize video frames into spatial-temporal token sequences, then modeling their joint distribution with a GPT-style transformer. This formulation naturally handles variable-length generation and benefits from the well-understood scaling properties of transformer architectures. \textbf{Phenaki}~\cite{villegas2022phenaki} extended autoregressive video generation to variable-length sequences conditioned on time-varying text prompts, enabling the generation of multi-minute videos that follow evolving narrative descriptions---a significant step toward open-domain video synthesis. \textbf{VideoPoet}~\cite{kondratyuk2024videopoet} further unified multiple video generation tasks (text-to-video, image-to-video, video-to-audio, stylization) within a single large language model framework, demonstrating that a unified token vocabulary can subsume diverse modalities.

The autoregressive paradigm was subsequently applied to interactive world simulation. \textbf{Oasis}~\cite{oasis2024} demonstrated that a transformer operating on discrete video tokens can serve as an interactive game simulator, generating real-time game frames conditioned on player actions---representing an autoregressive counterpart to the diffusion-based game engines discussed below. The key advantage of autoregressive approaches is their natural compatibility with scaling laws established in language modeling; the key limitation is that discrete tokenization may discard fine-grained visual details, and sequential token generation can be slow for high-resolution video.

\paragraph{Large-Scale Diffusion Models as World Simulators.}
The most significant recent development in video prediction has been the scaling of diffusion-based video generation models to internet-scale training data, producing systems whose outputs exhibit high visual fidelity and apparent physical consistency. This paradigm shift was crystallized by OpenAI's \textbf{Sora}~\cite{openai2024sora} (see Section~\ref{sec:diffusionworldmodel} for architectural details), whose technical report~\cite{brooks2024video} argued that the system exhibits emergent properties of a world simulator---including consistent 3D geometry, object permanence, and camera-aware scene generation---motivating systematic investigation into whether large-scale video generators constitute genuine world models. Concurrently, \textbf{Make-A-Video}~\cite{singer2023makeavideo} and \textbf{Imagen Video}~\cite{ho2022imagenvideo} demonstrated that pretrained text-to-image models could be extended to video through learned temporal dynamics, establishing cascaded diffusion architectures as a viable paradigm for high-definition video synthesis.

This debate has been informed by several concurrent large-scale systems. While some of the following were developed primarily as video generators rather than world models, they are included here because their scaling insights, architectural innovations, and emergent behaviors are directly informing the design of next-generation world simulators---and several are being actively studied for latent world-modeling properties. \textbf{Stable Video Diffusion} (SVD)~\cite{blattmann2023svd} demonstrated that carefully curated pretraining data enables high-quality image-to-video generation, establishing an influential open-weight baseline whose data curation methodology has been widely adopted. Google's \textbf{Lumiere}~\cite{bartal2024lumiere} introduced a space-time U-Net that generates entire video durations in a single pass, improving global temporal consistency---a property critical for world simulation. \textbf{W.A.L.T.}~\cite{gupta2024walt} proposed a unified transformer for both image and video generation using a shared latent space. Meta's \textbf{Movie Gen}~\cite{polyak2024moviegen} scaled to 30 billion parameters with synchronized audio generation, demonstrating the feasibility of multimodal world-scale generation. \textbf{CogVideoX}~\cite{yang2025cogvideox} introduced expert adaptive normalization for scalable text-to-video generation, while \textbf{Open-Sora}~\cite{opensora2024} democratized access to the DiT-based paradigm through an open-source implementation. NVIDIA's \textbf{Cosmos}~\cite{nvidia2025cosmos} and its successor \textbf{Cosmos-Predict2.5}~\cite{nvidia2025cosmos25}, pretrained on 20 million hours of video, explicitly position video generation as a foundation for physical AI and world simulation.

Despite their visual impressiveness, the status of these systems as world models remains contested. Ding et al.~\cite{ding2025understanding} examined the boundary between understanding and prediction in video generation, arguing that perceptual fidelity does not imply causal understanding. More pointedly, Kang et al.~\cite{kang2025howfar} conducted controlled experiments demonstrating that current video generation models exhibit \emph{case-based generalization}---reproducing visual patterns similar to training examples rather than abstracting underlying physical laws---and fail systematically on scenarios requiring genuine physical reasoning such as novel object interactions and out-of-distribution forces. These findings suggest that scaling visual-only training may be necessary but not sufficient for building world models that truly understand physical dynamics, a theme we revisit in the challenges discussion below.

\paragraph{Interactive World Models and Neural Game Engines.}
A stringent test of whether a video generation model constitutes a world model is its capacity for \emph{interactive} deployment: can it maintain a coherent, responsive environment conditioned on a continuous stream of user actions? The architectural details of the systems discussed below are covered in Section~\ref{sec:diffusionworldmodel}; here we focus on their application-level capabilities and what they reveal about the boundary between video generation and environment simulation. Section~6.4 further discusses how some of these systems serve as training environments for embodied agents.

DeepMind's \textbf{Genie} family~\cite{bruce2024genie, parkerholder2024genie2, deepmind2026genie3} progressively demonstrated interactive world generation from unlabeled video: Genie learned latent action models for 2D environments, Genie~2 scaled to persistent 3D worlds from single image prompts, and Genie~3 extended to text-prompted 3D worlds at 720p and 24 FPS. On the game simulation front, \textbf{DIAMOND}~\cite{alonso2024diamond} and \textbf{GameNGen}~\cite{valevski2024gamengen} showed that diffusion models can function as real-time game engines at interactive rates, while \textbf{GameGen-X}~\cite{che2025gamegenx} extended this to open-world environments with richer action spaces. \textbf{Pandora}~\cite{xiang2024pandora} bridged text-based interaction with visual world simulation by supporting both natural language actions and video state representations. \textbf{Diffusion Forcing}~\cite{chen2024diffusionforcing} provided a theoretical bridge between autoregressive and diffusion paradigms by applying variable noise levels across tokens, enabling causal generation with planning capabilities.

These interactive systems collectively suggest that the boundary between video generation and environment simulation is dissolving. However, they also reveal a persistent challenge: maintaining long-horizon consistency under interactive control, where the model must generate coherent responses to arbitrary action sequences rather than following a fixed trajectory, substantially amplifies the compounding error problem. Moreover, all of the systems discussed above operate on 2D image projections, discarding the 3D geometric structure that underpins physical reasoning about occlusion, depth, and spatial interaction---limitations that motivate the next direction.

\paragraph{3D and 4D Scene Understanding from Video.}
A complementary research direction extends video world models to explicit 3D and 4D (3D + time) scene representations, bridging video prediction with spatial scene understanding. Kong et al.~\cite{kong2025_3d4d} provided the first comprehensive survey dedicated to 3D and 4D world modeling, establishing taxonomies spanning three complementary approaches: \emph{video-based} (VideoGen) methods that generate future 2D frames and optionally lift them to 3D, \emph{occupancy-based} (OccGen) methods that directly predict volumetric scene structure, and \emph{LiDAR-based} (LiDARGen) methods that forecast future point clouds.

Within the occupancy prediction paradigm, \textbf{GaussianWorld}~\cite{zuo2025gaussianworld} introduced streaming 3D occupancy prediction using Gaussian splatting, representing scene elements as 3D Gaussians that evolve over time and enable efficient rendering from novel viewpoints. This approach preserves geometric structure that purely 2D video prediction discards, supporting downstream applications such as view synthesis, spatial planning, and collision detection.

A key challenge for temporal coherence in both 2D and 3D video world models is the compounding error inherent in autoregressive generation: small prediction errors accumulate over time, leading to degraded scene structure and visual artifacts in long sequences. Po et al.~\cite{po2025longcontext} addressed this with \emph{state-space video world models} for long-context generation, replacing attention-based temporal modeling with linear recurrent architectures (structured state-space models) that offer constant per-frame memory and computation costs. This enables generation over substantially longer horizons than transformer-based approaches, whose quadratic attention cost limits practical sequence lengths. The 3D and 4D world modeling direction is particularly promising for embodied AI applications, where agents must reason about spatial layout, occlusion, and object persistence in three dimensions rather than through 2D projections.

It is worth noting that ``scene understanding'' encompasses not only geometric reconstruction but also semantic and relational comprehension: recognizing objects, inferring their physical properties, understanding inter-object relations, and reasoning about causal structure from video. Object-centric approaches that decompose visual scenes into discrete entity representations---via slot-based mechanisms (Slot Attention, SAVi, SlotFormer) and relational dynamics via graph neural networks (C-SWM, Object-Centric Dreamer)---are surveyed in detail in Section~\ref{sec:Physicsworldmodel}. Those models provide compositional generalization and physical reasoning capabilities that complement the generation-focused approaches discussed here. A promising but largely unexplored frontier lies in integrating object-centric scene understanding with large-scale video generation models, which would combine the visual fidelity of diffusion-based generators with the structured reasoning afforded by explicit entity representations.

\paragraph{Physical Plausibility and the World Model Debate.}
The rapid advancement of large-scale video generation has precipitated a fundamental question: do these models \emph{understand} the physical world, or do they merely produce visually plausible simulations that mimic physical behavior without internalizing its underlying principles? This question has significant implications for whether video generation models can serve as reliable world models for downstream decision-making.

Several recent works have developed benchmarks and analyses specifically targeting this question. \textbf{PhysDreamer}~\cite{zhang2024physdreamer} explored physics-based interaction with 3D objects via video generation, demonstrating that coupling generative models with explicit physical simulation can improve the physical plausibility of generated interactions. \textbf{WorldSimBench}~\cite{qin2025worldsimbench} proposed a comprehensive evaluation framework that assesses video generation models specifically as world simulators, introducing metrics beyond visual quality to measure action controllability, physical consistency, and long-horizon coherence.

These investigations converge on a nuanced conclusion: current video generation models capture statistical regularities that correlate with physical behavior---objects generally fall downward, liquids flow, and rigid bodies collide---but they do not enforce physical laws as hard constraints. Consequently, they can produce physically implausible outputs for scenarios underrepresented in training data. Bridging this gap may require integrating explicit physical priors (as discussed in Section~\ref{sec:Physicsworldmodel}), grounding video generation in 3D scene representations, or developing hybrid architectures that combine the visual fidelity of diffusion models with the physical guarantees of simulation engines.

Despite remarkable progress, several fundamental challenges continue to limit the deployment of video prediction and scene understanding models as reliable world models.

\emph{Temporal consistency and long-horizon coherence.} Maintaining coherent object identities, physical properties, and geometric relationships across long generated sequences remains the single most significant limitation. Autoregressive models suffer from compounding prediction errors, while diffusion models struggle with global temporal planning across extended sequences. State-space architectures~\cite{po2025longcontext} and hierarchical generation strategies offer promising directions but have not yet fully resolved this challenge.

\emph{Physical understanding versus pattern matching.} As discussed above, current models capture visual correlations with physical behavior but do not enforce physical laws~\cite{kang2025howfar}. Closing this gap likely requires architectural innovations that integrate explicit physical reasoning---such as the physics-informed approaches reviewed in Section~\ref{sec:Physicsworldmodel}---with the visual generation capabilities of large-scale diffusion and autoregressive models.

\emph{Controllability and interactivity.} Enabling precise, fine-grained user control over generated content---from camera trajectories and object manipulations to semantic scene editing---remains challenging. Interactive world models (Genie, DIAMOND, GameNGen) have demonstrated real-time responsiveness within constrained domains, but extending this to open-world settings with diverse action spaces requires substantially richer conditioning mechanisms.

\emph{Evaluation methodology.} Standard perceptual metrics (FID, FVD, SSIM) capture visual quality but fail to assess physical plausibility, causal consistency, or interactive fidelity~\cite{luo2025beyondfvd}. The field urgently needs evaluation protocols that measure whether generated videos respect physical laws, maintain causal coherence, and support reliable downstream decision-making---properties that are essential if video world models are to serve as substrates for planning and control.

\emph{Computational cost and real-time deployment.} Diffusion-based models require multiple denoising steps per frame, and transformer-based models face quadratic scaling with sequence length, both limiting real-time applicability. Distillation, consistency models, and linear-complexity architectures represent active research directions, but achieving real-time generation at high resolution with full physical fidelity remains an open challenge.

\emph{From video generation to genuine world modeling.} Perhaps the most consequential open question is whether the video prediction paradigm---training on large corpora of passively observed video---can yield systems that truly model environment dynamics, or whether genuine world modeling requires active interaction, causal intervention, or explicit physical grounding. The resolution of this question will determine whether video generation models remain powerful content creation tools or evolve into foundational components of autonomous intelligent systems.

\subsection{Multimodal agents and language-grounded systems}
The convergence of large language models, vision-language models, and learned world models has given rise to a new generation of multimodal agents capable of perceiving, reasoning about, and acting in complex environments. Unlike the methodological discussion in Section~3.6, which examines how language and multimodal inputs reshape world model architectures and training paradigms, this section surveys the \emph{application systems} that deploy these capabilities across diverse domains---from robotic manipulation and open-world game playing to graphical user interface navigation and interactive world simulation. A unifying theme across these systems is the use of foundation models as implicit or explicit world models: large pretrained networks that encode rich knowledge about environment dynamics, physical affordances, and semantic relationships, thereby grounding agent behavior in multimodal understanding of the world~\cite{ahn2022saycan,driess2023palme}. The rapid maturation of this application domain since 2022 has been accompanied by the development of standardized benchmarks and open-source model releases that are accelerating progress toward generalist embodied agents.

\textbf{Language-grounded robotic manipulation and planning.} A foundational challenge in deploying language-conditioned robots is bridging the gap between the abstract semantic knowledge encoded in large language models and the physical affordances of a particular robot in a particular environment. \textbf{SayCan}~\cite{ahn2022saycan} addressed this by combining the semantic scoring of an LLM---which rates how \emph{useful} each primitive skill is for a given instruction---with a learned value function that rates how \emph{feasible} that skill is in the current state. The product of these two scores selects executable, contextually relevant actions, enabling a mobile manipulator to complete long-horizon kitchen tasks specified in natural language with a 74\% planning success rate across 101 real-world trials. \textbf{Inner Monologue}~\cite{huang2023innermonologue} extended this paradigm to closed-loop execution by incorporating environment feedback---success detection, scene descriptions, and human corrections---into an ongoing ``inner monologue'' that allows the LLM planner to retry and replan upon failure, substantially improving robustness over open-loop approaches. Building on this compositional philosophy, \textbf{Socratic Models}~\cite{zeng2022socratic} demonstrated that multiple pretrained foundation models (vision-language models, LLMs, audio models) can be composed through language as an intermediate representation without any additional training, establishing a zero-shot framework for embodied task completion.

A parallel line of work reframes robot policy generation as code synthesis. \textbf{Code as Policies}~\cite{liang2023codepolicies} showed that LLMs can generate executable Python programs that compose perception APIs, control primitives, and spatial reasoning functions, achieving zero-shot generalization to novel tasks across multiple robot embodiments without additional training. \textbf{VoxPoser}~\cite{huang2023voxposer} advanced this idea into three dimensions by using LLMs and VLMs to compose 3D affordance and constraint maps from RGB-D observations and language instructions; these volumetric value maps then serve as objective functions for motion planners, achieving a 76.7\% success rate on unseen manipulation instructions. In the domain of reward design, \textbf{Eureka}~\cite{ma2024eureka} demonstrated that GPT-4 can generate reward functions via evolutionary optimization of reward code, outperforming human expert reward designs on 83\% of 29 reinforcement learning environments spanning 10 robot morphologies---including the first successful demonstration of pen spinning with a simulated dexterous hand.

\textbf{Open-world game agents.} Open-world games---particularly Minecraft---have emerged as a primary testbed for multimodal agents that must perceive, plan, and act over long horizons in complex, procedurally generated environments. \textbf{MineDojo}~\cite{fan2022minedojo} established the foundational benchmark for this domain, providing 1,581 programmatic tasks with natural language goal specifications alongside an internet-scale knowledge base comprising 730K YouTube videos, 7K wiki pages, and 340K Reddit posts, and receiving the NeurIPS 2022 Outstanding Datasets and Benchmarks Award. Building on this infrastructure, \textbf{Voyager}~\cite{wang2023voyager} introduced the first LLM-powered lifelong learning agent, combining an automatic curriculum for open-ended exploration, an ever-growing skill library of executable code, and an iterative prompting mechanism with environment feedback and self-verification. Voyager discovered 3.3 times more unique items, traveled 2.3 times longer distances, and reached technology tree milestones 15.3 times faster than prior methods.

Subsequent systems addressed complementary challenges. \textbf{DEPS}~\cite{wang2023deps} (Describe, Explain, Plan, and Select) introduced interactive planning with LLMs that incorporates error explanation and plan refinement, becoming the first zero-shot agent to robustly accomplish over 70 distinct Minecraft tasks. \textbf{JARVIS-1}~\cite{wang2023jarvis1} combined multimodal language models with a multimodal memory system that stores and retrieves relevant past experiences, enabling completion of over 200 different tasks and achieving a 12.5\% completion rate on the notoriously difficult diamond pickaxe challenge---a five-fold improvement over prior records. \textbf{STEVE-1}~\cite{lifshitz2023steve1} took a more data-efficient approach by instruction-tuning the Video Pretraining (VPT) model for text and visual instruction following, producing a capable Minecraft agent trained for only \$60 of compute. Collectively, these systems demonstrate that combining foundation model reasoning with structured memory, code generation, and curriculum learning can yield agents with increasingly open-ended capabilities.

\textbf{Vision-Language-Action models for generalist robot control.} Vision-Language-Action (VLA) models represent an end-to-end paradigm in which a single model maps multimodal inputs---images and language instructions---directly to executable robot actions. This approach was foreshadowed by \textbf{Gato}~\cite{reed2022gato}, DeepMind's 1.2B-parameter generalist agent that performed 604 tasks spanning Atari games, image captioning, dialogue, and real robot block stacking within a single set of weights. \textbf{PaLM-E}~\cite{driess2023palme} scaled this vision substantially by injecting continuous embodied observations directly into the embedding space of the 562B-parameter PaLM language model, demonstrating positive transfer across internet-scale language, vision, and robotics domains.

The Robotics Transformer family established the VLA paradigm for real-world deployment. \textbf{RT-1}~\cite{brohan2023rt1} trained a scalable transformer on 130K real-robot episodes covering over 700 tasks collected from 13 robots over 17 months, achieving a 97\% success rate on trained tasks. \textbf{RT-2}~\cite{zitkovich2023rt2} was the first model to explicitly formalize the VLA concept, adapting large vision-language models to output robot actions represented as text tokens and exhibiting emergent capabilities such as chain-of-thought reasoning for multi-stage semantic manipulation. The \textbf{Open X-Embodiment} initiative~\cite{openxembodiment2024} assembled the largest open-source real robot dataset---over one million trajectories from 22 embodiments across 60 datasets contributed by 34 research laboratories worldwide---and showed that RT-1-X and RT-2-X models trained on this mixture achieved 50\% higher success rates than policies trained on single-domain data alone.

Recent open-source efforts have democratized access to VLA models. \textbf{Octo}~\cite{ghosh2024octo}, a 93M-parameter transformer-based diffusion policy pretrained on 800K episodes from Open X-Embodiment, achieved performance comparable to the 55B-parameter RT-2-X while being nearly 600 times smaller. \textbf{OpenVLA}~\cite{kim2024openvla}, a 7B-parameter open-source VLA trained on 970K real-world demonstrations, outperformed the closed RT-2-X by 16.5\% in absolute success rate across 29 tasks with seven times fewer parameters. \textbf{$\pi_0$}~\cite{black2024pi0}, developed by Physical Intelligence, introduced flow matching for action generation within a VLA framework built on the PaliGemma VLM backbone, enabling complex dexterous manipulation across seven robot platforms and 68 tasks. \textbf{VIMA}~\cite{jiang2023vima} demonstrated that diverse manipulation tasks can be expressed through multimodal prompts interleaving text and visual tokens, achieving 2.9 times higher success rates than alternatives in the most challenging generalization setting of its benchmark. These developments mark a clear trajectory from closed, proprietary systems toward open, reproducible generalist robot policies.

\textbf{GUI and web navigation agents.} Multimodal agents have also been deployed for interacting with digital interfaces, where the environment consists of graphical user interfaces rather than physical spaces. \textbf{CogAgent}~\cite{hong2024cogagent}, an 18B-parameter vision-language model specializing in GUI understanding, operates using only screenshots as input and outperformed LLM-based methods that consume extracted HTML on both the Mind2Web benchmark for PC navigation and the AITW benchmark for Android navigation, earning a CVPR 2024 Highlight distinction. \textbf{CRADLE}~\cite{tan2024cradle} proposed a modular framework for General Computer Control that restricts agents to interact with arbitrary software exclusively through screenshots and keyboard/mouse actions, becoming the first agent to follow main storylines in the complex open-world game Red Dead Redemption 2. \textbf{WebArena}~\cite{zhou2024webarena} and \textbf{VisualWebArena}~\cite{koh2024visualwebarena} established realistic benchmarks with self-hosted websites and visually grounded tasks, revealing that even the most capable multimodal models achieve only approximately 16\% task success---underscoring the substantial gap between current agent capabilities and human-level web navigation performance.

\textbf{Interactive world simulators as training environments.} A complementary application paradigm uses generative world models not as agents themselves but as interactive environments for training and evaluating embodied agents. \textbf{UniSim}~\cite{yang2023unisim} learned a universal simulator from diverse data sources and demonstrated that both vision-language planners and reinforcement learning policies trained purely within UniSim-generated simulations transferred successfully to physical robots, earning the ICLR 2024 Outstanding Paper Award. The \textbf{Genie} family from Google DeepMind has progressively expanded the scope of generative interactive environments: the original \textbf{Genie}~\cite{bruce2024genie} learned to generate interactive 2D environments from unlabeled internet video using a latent action model, \textbf{Genie 2}~\cite{parkerholder2024genie2} scaled this to persistent, action-controllable 3D worlds generated from a single image prompt, and \textbf{Genie 3}~\cite{deepmind2026genie3} extended generation to dynamic 3D worlds from text prompts at 720p resolution and 24 frames per second with real-time navigation. In a distinct but related direction, \textbf{RAP}~\cite{hao2023rap} formalized the repurposing of an LLM as both a world model and a reasoning agent, incorporating Monte Carlo Tree Search for strategic planning and demonstrating that RAP on LLaMA-33B surpassed chain-of-thought prompting on GPT-4 with a 33\% relative improvement. These results collectively suggest that learned world simulators---whether pixel-based or language-based---can serve as effective training substrates that reduce reliance on costly real-world data collection.

\textbf{Benchmarks and evaluation.} The evaluation of multimodal agent systems relies on a growing ecosystem of standardized benchmarks spanning diverse domains and capability levels. Table~\ref{tab:multimodal_benchmarks} summarizes the principal evaluation suites. In household robotics, \textbf{ALFRED}~\cite{shridhar2020alfred} provides 25K natural language instructions paired with 8K demonstrations in the AI2-THOR simulator, while \textbf{CALVIN}~\cite{mees2022calvin} evaluates language-conditioned policy learning across 34 tasks in four environments. For tabletop manipulation, \textbf{VIMA-Bench}~\cite{jiang2023vima} defines 17 task types with four-level generalization evaluation across over 600K trajectories, and \textbf{Language-Table}~\cite{lynch2023languagetable} offers 600K trajectories with 87K language instructions, achieving 93.5\% success on a real robot. The \textbf{Open X-Embodiment}~\cite{openxembodiment2024} dataset provides the largest cross-embodiment evaluation at over one million trajectories from 22 robot platforms. For digital agents, \textbf{WebArena}~\cite{zhou2024webarena} and \textbf{VisualWebArena}~\cite{koh2024visualwebarena} benchmark web navigation with 812 and 910 tasks respectively, while \textbf{AgentBench}~\cite{liu2024agentbench} evaluates LLMs as agents across eight distinct environments. A persistent finding across these benchmarks is that even state-of-the-art systems exhibit substantial performance gaps relative to human baselines, particularly on long-horizon tasks requiring sustained reasoning, recovery from errors, and compositional generalization.

\begin{table}[htbp]
\centering
\caption{Representative benchmarks for multimodal agent evaluation.}
\label{tab:multimodal_benchmarks}
\small
\begin{tabular}{llllp{3.2cm}}
\toprule
\textbf{Benchmark} & \textbf{Year} & \textbf{Domain} & \textbf{Scale} & \textbf{Key Metric} \\
\midrule
ALFRED & 2020 & Household (AI2-THOR) & 25K instructions & Task success rate \\
MineDojo & 2022 & Minecraft & 1,581 tasks & Task completion rate \\
CALVIN & 2022 & Language-cond.\ manip. & 34 tasks, 4 envs & Multi-step completion \\
VIMA-Bench & 2023 & Tabletop manipulation & 17 task types & 4-level generalization \\
Language-Table & 2023 & Tabletop manipulation & 600K trajectories & Real-robot success rate \\
Open X-Embodiment & 2024 & Cross-embodiment & 1M+ trajectories & Cross-embodiment transfer \\
WebArena & 2024 & Web navigation & 812 tasks & Task success rate \\
VisualWebArena & 2024 & Visual web navigation & 910 tasks & Functional evaluation \\
AgentBench & 2024 & Multi-domain agents & 8 environments & Multi-turn success rate \\
\bottomrule
\end{tabular}
\end{table}

\textbf{Challenges and future directions.} Despite rapid progress, several fundamental challenges remain. First, long-horizon consistency continues to limit deployed systems: current agents struggle to maintain coherent plans over hundreds of sequential decisions, with errors compounding as task horizons grow. Second, the question of whether foundation-model-based agents possess genuine causal understanding---as opposed to sophisticated pattern matching---remains unresolved, with implications for robustness and out-of-distribution generalization. Third, safety and reliability requirements for real-world deployment far exceed what current systems can guarantee, particularly in domains involving physical interaction with humans. Fourth, no single benchmark currently spans the full spectrum of multimodal agent capabilities, creating a need for unified evaluation frameworks that assess perception, reasoning, planning, and action in an integrated manner. Fifth, while VLA models have demonstrated impressive generalization, most still require massive training datasets; leveraging world models for data augmentation and sample-efficient learning represents a promising direction for reducing data requirements. Finally, nearly all current systems operate as single agents, leaving multi-agent coordination in shared physical or digital environments as a largely unexplored frontier. Addressing these challenges will require not only advances in individual model capabilities but also deeper integration of explicit world models with the implicit world knowledge encoded in foundation models.

\subsection{Reinforcement learning and Games}
\label{sec:rl_games}

Reinforcement learning (RL) formalizes sequential decision-making as finding a policy $\pi: \mathcal{S} \to \Delta(\mathcal{A})$ that maximizes the expected cumulative discounted return $J(\pi) = \mathbb{E}_\pi[\sum_{t=0}^{\infty} \gamma^t r_t]$ within an MDP $\mathcal{M} = (\mathcal{S}, \mathcal{A}, T, R, \gamma)$, where the transition dynamics $T(s' \mid s, a)$ and reward function $R(r \mid s, a)$ are both unknown~\cite{sutton2018reinforcement}. The central difficulty is that the agent must simultaneously discover what the environment does and optimize against it—a coupling that drives the prohibitive sample complexity of model-free methods, which require tens of millions of interactions before converging. World models address this at its root by learning approximate dynamics $\hat{T}$ and $\hat{R}$ from observed transitions, thereby decoupling environment discovery from policy optimization: once the model is learned, the agent can generate unlimited synthetic experience, plan multi-step action sequences, and assign credit for long-horizon returns—all without additional real environment steps. From the perspective of how the learned model is used, world model RL methods fall into three non-exclusive strategies~\cite{moerland2023model}: (i) \textbf{Dyna-style data augmentation}~\cite{sutton1991dyna}, which adds imagined transitions to the replay buffer to increase training data for model-free policy optimization; (ii) \textbf{imagination-based policy gradient}, which backpropagates policy gradients directly through differentiable multi-step rollouts in the world model; and (iii) \textbf{model-based planning}, which performs explicit look-ahead search (via Monte Carlo Tree Search, MCTS; model predictive control, MPC; or related algorithms) to select actions with superior long-term consequences at decision time. Game environments serve as the canonical testbed for this research because they provide well-defined ground-truth dynamics, reproducible evaluation, and a carefully calibrated spectrum of difficulty that isolates specific RL challenges: from low-dimensional continuous control that tests pure dynamics learning, through pixel-based Atari that demands simultaneous representation learning and planning, to board games and open-world environments that require combinatorial search and hierarchical credit assignment.

\paragraph{Sample Efficiency: World Models as Data Multipliers.}
 
The most direct motivation for world models in RL is the theoretical and empirical sample complexity gap between model-based and model-free methods. A learned dynamics model transforms each real transition $(s_t, a_t, r_t, s_{t+1})$ into a generator of arbitrarily many synthetic transitions, enabling the agent to amortize the cost of environment interaction across a much larger effective training dataset. \textbf{SimPLe}~\cite{kaiser2019model} was the first to demonstrate this at scale in pixel-based settings, iterating between collecting 6,400 real Atari frames, training a stochastic convolutional world model, and optimizing a PPO policy on 16,000 imagined rollouts per cycle—achieving a 2.1$\times$ improvement over Rainbow DQN at 100K steps and establishing the \textbf{Atari 100K benchmark} as the canonical evaluation regime for sample-efficient model-based RL. This iterative real-imagined loop, which Sutton originally termed the Dyna architecture~\cite{sutton1991dyna}, was subsequently refined by \textbf{IRIS}~\cite{micheli2023transformers} and \textbf{DIAMOND}~\cite{alonso2024diffusion}: IRIS discretizes frames via VQ-VAE into token sequences and models dynamics with a GPT-style transformer, using imagined rollouts for actor-critic training and achieving 1.046 human-normalized score (HNS) on Atari 100K; DIAMOND replaces autoregressive token prediction with denoising diffusion for holistic next-frame generation, producing sharper and more temporally consistent predictions that yield 1.463 HNS. The progression from SimPLe through IRIS and DIAMOND to \textbf{DreamerV3}~\cite{hafner2025mastering} over five years — all using exactly the same 100K real environment steps — quantifies the compounding gains from better world model representations, more stable training objectives, and tighter integration of model learning with policy optimization.
 
\paragraph{Credit Assignment: Imagination-Based Policy Gradient.}
 
Sparse reward environments pose the credit assignment problem in its sharpest form: the agent must identify which actions, among a long sequence, were causally responsible for an eventual reward—a problem whose difficulty grows with horizon length and reward sparsity. Model-free TD methods address this through bootstrapped value targets and eligibility traces, but remain limited by the variance of bootstrapped estimates and the difficulty of propagating sparse signals backward over long horizons. World models offer a fundamentally different solution: the agent can simulate the future consequences of actions forward through the learned dynamics, directly evaluating long-horizon returns without relying on sparse reward signals to propagate credit backward through time. The Dreamer series~\cite{hafner2019dream,hafner2020mastering,hafner2025mastering} operationalizes this through the \emph{RSSM} (Recurrent State-Space Model), which factorizes the latent state into a deterministic recurrent component $h_t = f_\theta(h_{t-1}, z_{t-1}, a_{t-1})$ and a stochastic component $z_t \sim q_\phi(z_t \mid h_t, x_t)$, trained via the ELBO objective combining reconstruction, reward prediction, and KL regularization (Eq.~\ref{eq:rssm_elbo}). The actor and critic are then optimized entirely within imagined $H$-step rollouts using the $\lambda$-return $G_t^\lambda = r_t + \gamma[(1-\lambda)\hat{v}_{t+1} + \lambda G_{t+1}^\lambda]$, which blends one-step TD targets (low variance) with Monte Carlo returns (low bias) to efficiently propagate credit over extended horizons.
\begin{equation}
\mathcal{L}_{\text{WM}} = \mathbb{E}_{q}\!\Bigl[\sum_t \ln p(x_t \mid h_t, z_t) + \ln p(r_t \mid h_t, z_t) - \beta\,\mathrm{KL}\bigl[q(z_t \mid h_t,x_t)\,\|\,p(z_t \mid h_t)\bigr]\Bigr].
\label{eq:rssm_elbo}
\end{equation}
The effectiveness of this approach in long-horizon credit assignment is best illustrated by the Minecraft diamond challenge: DreamerV3~\cite{hafner2025mastering} is the first algorithm to collect diamonds from scratch without human demonstrations, requiring sequential discovery of a 12-item technology tree (wood $\to$ planks $\to$ sticks $\to$ crafting table $\to$ \ldots $\to$ iron pickaxe $\to$ diamonds) under sparse binary rewards. Transformer-based world models further strengthen credit assignment by replacing GRU recurrence with attention over the full history of latent states. \textbf{TransDreamer}~\cite{chen2022transdreamer} introduces the Transformer State-Space Model (TSSM) and demonstrates substantial improvements over DreamerV2 on Memory Maze and POMDPSuite, where rewards depend on observations hundreds of steps in the past. Its successor \textbf{TransDreamerV3}~\cite{dongare2025transdreamerv3} integrates TSSM into the DreamerV3 framework, combining transformer-based long-range memory with DreamerV3's robustness techniques for stable multi-domain learning. \textbf{STORM}~\cite{zhang2023storm} takes a complementary approach by introducing stochastic uncertainty into transformer dynamics through prior-posterior Gaussian pairs, enabling explicit uncertainty quantification that both improves prediction calibration and provides a principled KL-based intrinsic reward for exploration; the result is a mean HNS of 1.311 on Atari 100K, with particularly large gains on partially observable games where uncertainty modeling is most valuable.
 
\paragraph{Planning and Look-Ahead Reasoning: From MCTS to Latent MPC.}
 
Beyond improving data efficiency and credit assignment, world models enable deliberate planning: the ability to mentally simulate action sequences and evaluate their long-term consequences before execution—a capability fundamentally unavailable to reactive model-free policies. Two planning paradigms have proven most effective. The first uses MCTS within the learned model at decision time, constructing a partial game tree whose nodes are latent states and whose edges are candidate actions, guided by value and policy network heuristics. \textbf{AlphaZero}~\cite{silver2018alphazero} established this paradigm using known game rules, achieving superhuman performance in Go, chess, and shogi through self-play. \textbf{MuZero}~\cite{schrittwieser2020mastering} generalized it to unknown environments by learning a value-equivalent model (three functions $h_\theta$ for representation, $g_\theta$ for dynamics, and $f_\theta$ for prediction of value, policy, and reward), trained end-to-end to minimize prediction error on planning-relevant targets without any reconstruction objective. MCTS then operates entirely within this learned latent space, achieving superhuman performance across 57 Atari games, Go, chess, and shogi simultaneously. \textbf{MuZero Reanalyze}~\cite{schrittwieser2021online} improved its sample efficiency by replaying stored trajectories through the current world model to relabel value and policy targets. \textbf{EfficientZero}~\cite{ye2021efficientzero} addressed MuZero's limitations in the low-data regime by augmenting it with three modifications that jointly solve the credit assignment and data efficiency problems simultaneously: a self-supervised consistency loss $\mathcal{L}_{\text{cons}} = \|g_\theta(s_t, a_t) - \mathrm{sg}[h_\theta(o_{t+1})]\|_2^2$ that provides dense dynamics supervision independent of reward sparsity, end-to-end value prefix learning to reduce compounding temporal error, and off-policy correction for replayed data. Together these achieve 194.3\% mean human performance on Atari 100K with two hours of real gameplay—500$\times$ fewer frames than DQN. \textbf{EfficientZero V2}~\cite{wang2024efficientzerov2} extends this framework to continuous action spaces via Gumbel-based sampling search, outperforming DreamerV3 in 50 of 66 tasks across Atari and DMControl without per-domain tuning. \textbf{Student of Games}~\cite{schmid2023student} unifies MCTS with counterfactual regret minimization for imperfect-information settings, achieving near-human performance in both chess and poker within a single model-based architecture.
 
The second planning paradigm uses MPC in continuous action spaces, sampling action sequences, evaluating them by rolling out through the world model, and executing the highest-value sequence in receding-horizon fashion. \textbf{TD-MPC2}~\cite{hansentd} combines a latent dynamics model with temporal difference learning and MPPI-based MPC, where the world model evaluates candidate trajectories and the value function bootstraps terminal returns. On HumanoidBench~\cite{sferrazza2024humanoidbench}—17 whole-body control tasks for a 56-DOF humanoid—TD-MPC2 outperforms DreamerV3 by over 40\% on average, demonstrating that tight integration of model-based look-ahead with value estimation is particularly powerful in high-dimensional continuous action spaces where exhaustive tree search is infeasible.
 
\paragraph{Partial Observability: Belief State Estimation Through World Models.}
 
A central challenge in many RL environments is partial observability: the agent's observation $o_t$ does not fully determine the underlying state $s_t$, so optimal decision-making requires integrating evidence across the full observation history $o_{1:t}$. Model-free methods address this heuristically through frame stacking or recurrent networks, but lack a principled mechanism for maintaining calibrated uncertainty over hidden states. World models provide a natural solution via Bayesian filtering: the RSSM posterior $q(z_t \mid h_t, o_t)$ represents the agent's belief about the current latent state given all past observations compressed into the deterministic state $h_t$, enabling principled uncertainty-aware planning. The advantage of this belief state formulation is most apparent in environments requiring long-range memory. On Memory Maze~\cite{pasukonis2022evaluating} (where the agent must navigate a 3D labyrinth and retrieve goal locations visited hundreds of steps earlier), \textbf{TransDreamer}~\cite{chen2022transdreamer} substantially outperforms DreamerV2 by replacing GRU recurrence with full-history attention in the TSSM, allowing the agent to selectively retrieve episodically relevant observations. \textbf{STORM}~\cite{zhang2023storm} similarly exploits transformer attention for partial observability, with its stochastic formulation additionally capturing residual state uncertainty through explicit prior-posterior pairs conditioned on hidden states. On the NetHack Learning Environment~\cite{kuttler2020nethack} (a roguelike with thousands of item types, stochastic events, and partial map visibility requiring persistent long-term memory), recurrent world model approaches demonstrate the capacity to maintain coherent beliefs over episodes of thousands of steps, far exceeding the horizon where standard frame-stacking policies degrade. The practical implication is that the choice of dynamics model architecture (GRU vs.\ transformer vs.\ state-space model) has the largest impact in partially observable settings, and this dimension should guide architecture selection for real-world RL deployments where full observability is rarely available.
 
\paragraph{Exploration: World Models Under Sparse and Deceptive Rewards.}
 
Effective exploration is among the most practically important and theoretically difficult aspects of RL: the agent must visit sufficiently diverse states to discover high-reward trajectories without a guiding signal. Standard entropy-based exploration (maximizing action distribution entropy) proves insufficient for hard exploration benchmarks with long reward chains or deceptive local optima. World models enable two principled approaches to this problem. The first exploits epistemic uncertainty: states where the world model has high prediction error are states that have been visited infrequently, and thus contain the most information to be gained from exploration. This motivates using world model prediction error as an intrinsic reward that augments the extrinsic task reward—a direction pursued by several curiosity-driven exploration methods that integrate naturally with RSSM-based world models. The second approach, embodied by \textbf{Optimistic World Models (OWM)}~\cite{mete2026optimistic}, draws on the reward-biased maximum likelihood estimation (RBMLE) principle from adaptive control theory to implement the ``optimism in the face of uncertainty'' (OFU) principle directly within the dynamics model. OWM augments the standard world model training objective with an optimistic dynamics loss:
\begin{equation}
\mathcal{L}_{\text{OWM}} = \mathcal{L}_{\text{WM}} - \alpha \cdot \mathbb{E}_{(s,a) \sim \mathcal{B}}\bigl[\hat{R}(\hat{s}' \mid s, a)\bigr], \quad \hat{s}' \sim g_\theta(s, a),
\label{eq:owm}
\end{equation}
which gently steers the learned transition distribution toward higher-reward outcomes, training the world model to generate optimistic imaginations without requiring explicit uncertainty estimates or constrained optimization. This plug-and-play modification is architecturally agnostic—instantiated as Optimistic DreamerV3 and Optimistic STORM, it achieves significant gains on sparse-reward Atari games that defeat standard entropy-based exploration, while adding minimal computational overhead. A complementary perspective is offered by \textbf{EfficientZero}'s~\cite{ye2021efficientzero} self-supervised consistency loss, which densifies the dynamics training signal independent of reward sparsity: by supervising the world model with the encoder's next-observation representation rather than relying solely on the sparse reward signal, EfficientZero ensures the dynamics model remains accurate even in regions where rewards are absent—enabling MCTS to plan reliably through extended reward-free passages that confound approaches relying on reward-supervised dynamics alone.
 
\paragraph{Generative Game Engines: From Simulators to Policy Training Substrates.}
 
A distinct but increasingly prominent paradigm treats environment simulation as a generative modeling problem, training high-capacity generative models to produce photorealistic action-conditioned frame sequences that can serve as both evaluation environments and policy training substrates. This approach prioritizes visual fidelity over latent compactness, and planning interpretability over computational efficiency. \textbf{GameGAN}~\cite{kim2020gamegan} was an early demonstration, training a GAN to simulate DOOM and Pac-Man from gameplay trajectories; while visually coherent, GAN training instability limited its utility as an RL training environment. \textbf{DIAMOND}~\cite{alonso2024diffusion} resolved this by replacing the adversarial objective with denoising diffusion: given a window of $k$ past frames and an action, the DDPM iteratively denoises a random noise tensor into the predicted next frame through a sequence of reverse diffusion steps conditioned via cross-attention, producing holistic per-frame predictions with 1.463 HNS on Atari 100K. \textbf{GameNGen}~\cite{valevski2024diffusion} scaled this paradigm dramatically—adapting Stable Diffusion to simulate DOOM at 20 fps with visual quality indistinguishable from the real game by human evaluators (PSNR 29.4 dB), trained on 900 hours of automated gameplay. The key insight of GameNGen is that consistent enemy AI behavior, ammo bookkeeping, and physically plausible projectile dynamics all emerge from a single diffusion model trained purely on gameplay video, without any explicit game rule supervision—suggesting that photorealistic game simulation may be an emergent property of large-scale generative models at sufficient training scale. 

\subsection{Scientific and Domain-specific Modeling}
\label{sec:scientific_modeling}
 
Scientific disciplines have long relied on numerical simulation to model complex physical systems: from atmospheric dynamics to molecular interactions to cosmic structure formation. These simulations, grounded in first-principles physics, provide high-fidelity predictions but incur enormous computational costs. A single high-resolution climate projection can require millions of CPU-hours, and atomistic molecular dynamics simulations of biologically relevant timescales remain intractable even on modern supercomputers. World models offer a compelling alternative by learning surrogate dynamics models directly from simulation data or observations, enabling orders-of-magnitude speedups while preserving physically meaningful predictions. Unlike the physics-informed architectures discussed in Section~\ref{sec:Physicsworldmodel}, which embed conservation laws and symmetries as architectural priors, this section surveys how world-model concepts are applied as \emph{domain-level tools} across specific scientific fields: weather and climate prediction, molecular and materials simulation, and cosmological modeling, where learned dynamics surrogates are transforming the practice of computational science.

\paragraph{Weather and climate prediction.}
Numerical weather prediction (NWP) has been the operational backbone of meteorology for decades, evolving the atmospheric state forward through discretized partial differential equations at enormous computational expense. The emergence of data-driven weather models represents a paradigm shift in this field. \textbf{Pangu-Weather}~\cite{bi2023pangu} introduced a 3D Earth-Specific Transformer architecture trained on 39 years of global reanalysis data from ERA5, becoming the first AI-based system to outperform the European Centre for Medium-Range Weather Forecasts (ECMWF) Integrated Forecasting System across all variables and lead times up to one week, at 0.25$^\circ$ spatial resolution. \textbf{GraphCast}~\cite{lam2023graphcast}, developed by DeepMind, adopted graph neural networks on an icosahedral mesh to model atmospheric dynamics, achieving superior accuracy to ECMWF's deterministic forecasts on over 90\% of 1,380 verification targets while producing 10-day forecasts in under one minute on a single GPU. These models can be interpreted as learned world models of the atmosphere: given the current and previous atmospheric states $(X_t, X_{t-1})$, they predict the next state $\hat{X}_{t+1}$ through a learned transition function that implicitly captures the governing fluid dynamics.

A critical limitation of these early models was their deterministic nature; they produced single-point forecasts that could not represent the inherent uncertainty of chaotic atmospheric dynamics. \textbf{GenCast}~\cite{price2025gencast}, also from DeepMind, addressed this by formulating weather prediction as a conditional diffusion process: starting from noise, a denoiser network iteratively refines candidate atmospheric states conditioned on recent observations, producing an ensemble of plausible future trajectories. GenCast outperformed ECMWF's operational ensemble forecast (ENS) on 97.2\% of 1,320 evaluation targets and demonstrated superior skill in predicting extreme weather events, tropical cyclone tracks, and wind power generation, while requiring only 8 minutes on a single TPU to generate a full 15-day ensemble. \textbf{NeuralGCM}~\cite{kochkov2024neuralgcm} pursued a complementary hybrid approach, embedding learned parameterizations within a differentiable general circulation model to combine the physical guarantees of NWP with the flexibility of neural networks, achieving competitive accuracy from weather to climate timescales.

Beyond medium-range forecasting, foundation-model approaches are extending the temporal scope of learned atmospheric world models. \textbf{ClimaX}~\cite{nguyen2023climax} introduced a vision-transformer backbone pretrained on heterogeneous climate datasets including CMIP6 simulations, demonstrating that a single architecture can be fine-tuned for weather forecasting, seasonal prediction, regional downscaling, and climate projection. \textbf{Aurora}~\cite{bodnar2025aurora}, pretrained on over one million hours of diverse Earth system data, achieved state-of-the-art accuracy using a 1.3-billion-parameter 3D Swin Transformer and has been integrated into ECMWF's operational evaluation suite alongside physics-based models.

\paragraph{Molecular simulation and computational chemistry.}
At the atomic scale, world models address a different but structurally analogous challenge: predicting the temporal evolution of molecular systems governed by quantum-mechanical interactions. Classical molecular dynamics (MD) numerically integrates Newton's equations of motion using empirical force fields, but the accuracy-cost tradeoff limits both the system sizes and timescales that can be practically explored. Machine learning interatomic potentials (MLIPs) have emerged as learned transition models that approximate the potential energy surface $U(\mathbf{r})$ and its gradients from quantum-mechanical training data, enabling MD simulations with near ab initio accuracy at a fraction of the cost.

Architecturally, these models parallel the state-space world models discussed in Section~\ref{sec:Physicsworldmodel}: the molecular configuration $\mathbf{r}_t$ serves as the state, the force field provides the transition dynamics, and the trajectory constitutes a rollout. \textbf{SchNet}~\cite{schutt2017schnet} introduced continuous-filter convolutional networks for learning molecular representations that respect rotational invariance. Subsequent equivariant architectures, including \textbf{NequIP}~\cite{batzner2022nequip} and \textbf{MACE}~\cite{batatia2022mace}, incorporated E(3)-equivariant message passing to achieve superior data efficiency and accuracy. The recent \textbf{GEMS} framework~\cite{unke2024gems} demonstrated that pretrained neural network potentials can simulate biomolecular dynamics across chemically diverse fragments, representing a step toward general-purpose molecular world models.

Beyond force-field surrogates, generative approaches directly tackle the sampling bottleneck that limits conventional MD. \textbf{Boltzmann generators}~\cite{noe2019boltzmann} introduced normalizing flows that learn invertible transformations between a simple Gaussian prior and the target Boltzmann distribution of a molecular system, enabling one-shot generation of equilibrium configurations without sequential simulation. This approach has been extended to transferable architectures that generalize across chemical space~\cite{klein2024transferable}, and to diffusion-based methods for modeling protein conformational dynamics~\cite{arts2023two}. In the drug discovery context, \textbf{AlphaFold 3}~\cite{abramson2024alphafold3} extended structure prediction from proteins to general biomolecular complexes including DNA, RNA, and small-molecule ligands, using a diffusion-based architecture that assembles predictions from an initial cloud of atoms. While primarily a structure prediction model rather than a dynamics simulator, AlphaFold 3 demonstrates that learned models can capture the physical constraints governing molecular interactions at scale, laying groundwork for future dynamics-capable molecular world models.

\paragraph{Cosmological simulation and astrophysics.}
At the largest physical scales, cosmological N-body simulations model the gravitational evolution of dark matter and baryonic matter over billions of years to predict the large-scale structure of the universe. These simulations are indispensable for interpreting observational data from galaxy surveys but are computationally prohibitive at the resolutions required for precision cosmology. Learned surrogate models have emerged as efficient alternatives that approximate the mapping from initial conditions to evolved cosmic structure.

The \textbf{Deep Density Displacement Model} (D$^3$M)~\cite{he2019learning} demonstrated that a deep neural network can learn to transform first-order perturbation theory approximations into accurate nonlinear structure predictions, achieving results comparable to full N-body simulations at a small fraction of the computational cost. Super-resolution approaches~\cite{ni2021superresolution} have further shown that generative adversarial networks can enhance low-resolution dark matter simulations to high-resolution counterparts containing up to 512 times more particles while preserving statistical properties of the matter distribution. The \textbf{CAMELS} project~\cite{villaescusanavarro2021camels} created thousands of cosmological hydrodynamic simulations with systematic parameter variations, providing large-scale training data for machine learning models that can marginalize over uncertainties in baryonic physics while extracting cosmological information, an approach that parallels the use of diverse training environments in reinforcement-learning-based world models.

These cosmological surrogates face a domain-specific variant of the long-horizon consistency challenge: small errors in predicted density fields compound over cosmic time, leading to incorrect predictions of halo formation and galaxy clustering. Addressing this requires incorporating physical priors such as mass conservation, momentum conservation, and the correct power spectrum of initial fluctuations, constraints that connect directly to the physics-informed methods discussed in Section~\ref{sec:Physicsworldmodel}.

Across these domains, scientific world models share common frontiers that distinguish them from perception-oriented counterparts. \emph{Physical consistency} is non-negotiable: unlike video prediction, where minor artifacts may be tolerable, scientific surrogates must respect conservation laws, symmetries, and thermodynamic constraints over extended rollouts, whether that means energy conservation in molecular trajectories, mass conservation in atmospheric models, or the correct matter power spectrum in cosmological rollouts. \emph{Uncertainty quantification} is equally essential, as demonstrated by GenCast's ensemble diffusion approach; deterministic point predictions are insufficient when the goal is to characterize full probability distributions over possible outcomes. Perhaps most acutely, \emph{out-of-distribution generalization} remains an open problem, since scientific models are routinely used to extrapolate beyond training conditions: novel greenhouse gas concentrations, untested thermodynamic regimes, or alternative cosmological parameters, where data-driven models may fail silently. Addressing these demands points toward hybrid architectures that couple learned components with physics-based constraints, and the convergence of foundation-model pretraining with domain-specific fine-tuning suggests that the next generation of scientific world models will achieve broader generalization across conditions, scales, and physical regimes, potentially enabling a paradigm of ``simulation-free'' scientific discovery in which learned surrogates serve as the primary engine for hypothesis generation, experimental design, and physical prediction.

\subsection{Medical Images and Video Recordings}
\label{sec:medical}

The application of world models to healthcare and medical imaging is an emerging frontier with unique requirements: patient safety demands extreme reliability, data is scarce and privacy constrained, and the stakes of inaccurate predictions are high. Unlike game or driving environments where errors result in low scores or simulated collisions, prediction failures in clinical settings can directly impact patient well-being. Yet, the core world-model paradigm---learning environment dynamics, predicting future states conditioned on actions, and using those predictions for planning---maps naturally onto the clinical workflow: a patient's physiological state evolves over time, clinical interventions (treatments, surgeries, medications) constitute actions, and the goal is to select the intervention that optimizes long-term patient outcomes.

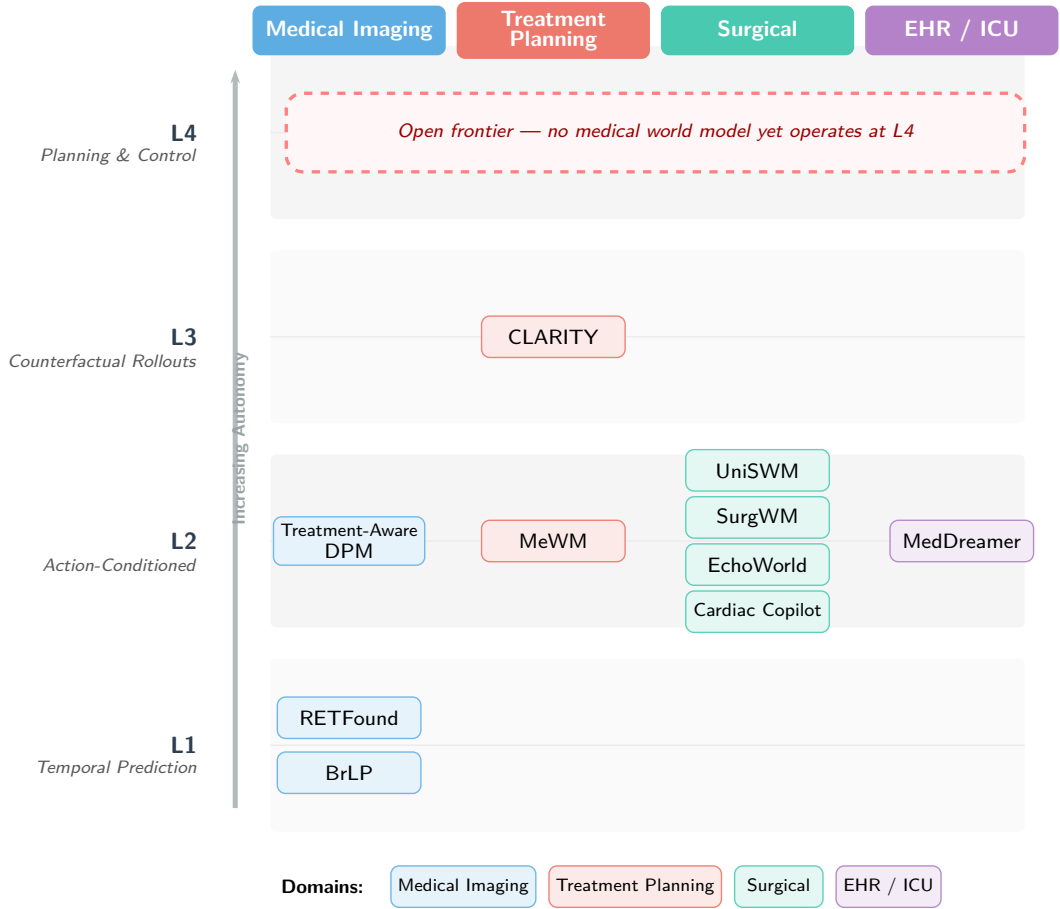
\begin{figure*}[!t]
\centering
\resizebox{0.85\textwidth}{!}{%
\begin{tikzpicture}[
    >=Stealth,
    modelnode/.style={
        rounded corners=3pt,
        font=\sffamily\fontsize{7.5}{9}\selectfont,
        inner sep=2.5pt,
        minimum height=15pt,
        minimum width=52pt,
        align=center,
        draw,
        line width=0.5pt
    },
    imgstyle/.style={modelnode, fill=colDriving!12, draw=colDriving!70},
    txstyle/.style={modelnode, fill=colRL!12, draw=colRL!70},
    surgstyle/.style={modelnode, fill=colArchitecture!12, draw=colArchitecture!70},
    ehrstyle/.style={modelnode, fill=colVideo!12, draw=colVideo!70},
    levellabel/.style={font=\sffamily\bfseries\fontsize{8.5}{10}\selectfont, text=yearcolor, anchor=east},
    leveldesc/.style={font=\sffamily\fontsize{6.5}{8}\selectfont, text=gray!70!black, anchor=east},
    colheader/.style={
        font=\sffamily\bfseries\fontsize{8}{9.5}\selectfont,
        text=white, rounded corners=3pt,
        minimum height=18pt, minimum width=70pt,
        align=center, inner sep=3pt
    },
    frontier/.style={draw=red!50, dashed, line width=1.2pt, rounded corners=6pt},
]

\def\colsep{2.6}
\def\rowsep{2.6}
\def\labeloffset{-1.8}

\def\colA{0}
\def\colB{\colsep}
\def\colC{2*\colsep}
\def\colD{3*\colsep}

\def\rowI{0}
\def\rowII{\rowsep}
\def\rowIII{2*\rowsep}
\def\rowIV{3*\rowsep}

\pgfmathsetmacro{\headerY}{3*\rowsep + 1.3}
\node[colheader, fill=colDriving!80]       at (\colA, \headerY) {Medical Imaging};
\node[colheader, fill=colRL!75]            at (\colB, \headerY) {Treatment\\[-2pt]Planning};
\node[colheader, fill=colArchitecture!80]  at (\colC, \headerY) {Surgical};
\node[colheader, fill=colVideo!75]         at (\colD, \headerY) {EHR / ICU};

\node[levellabel] at (\labeloffset, \rowIV)        {L4};
\node[leveldesc]  at (\labeloffset, \rowIV - 0.30) {\emph{Planning \& Control}};
\node[levellabel] at (\labeloffset, \rowIII)        {L3};
\node[leveldesc]  at (\labeloffset, \rowIII - 0.30) {\emph{Counterfactual Rollouts}};
\node[levellabel] at (\labeloffset, \rowII)          {L2};
\node[leveldesc]  at (\labeloffset, \rowII - 0.30)   {\emph{Action-Conditioned}};
\node[levellabel] at (\labeloffset, \rowI)            {L1};
\node[leveldesc]  at (\labeloffset, \rowI - 0.30)     {\emph{Temporal Prediction}};

\draw[-{Stealth[length=5pt,width=3.5pt]}, line width=1.8pt, colFoundation!60]
    (\labeloffset + 0.35, \rowI - 0.8) -- (\labeloffset + 0.35, \rowIV + 0.8);
\node[font=\sffamily\bfseries\fontsize{6.5}{8}\selectfont, text=colFoundation!80, rotate=90, anchor=south]
    at (\labeloffset + 0.65, 0.5*3*\rowsep) {Increasing Autonomy};

\pgfmathsetmacro{\bandL}{-1.0}
\pgfmathsetmacro{\bandR}{3*\colsep + 0.8}
\pgfmathsetmacro{\bandH}{1.1}
\begin{scope}[on background layer]
    \fill[gray!4,  rounded corners=2pt] (\bandL, \rowI   - \bandH) rectangle (\bandR, \rowI   + \bandH);
    \fill[gray!8,  rounded corners=2pt] (\bandL, \rowII  - \bandH) rectangle (\bandR, \rowII  + \bandH);
    \fill[gray!4,  rounded corners=2pt] (\bandL, \rowIII - \bandH) rectangle (\bandR, \rowIII + \bandH);
    \fill[gray!8,  rounded corners=2pt] (\bandL, \rowIV  - \bandH) rectangle (\bandR, \rowIV  + \bandH);
    \foreach \r in {\rowI, \rowII, \rowIII, \rowIV} {
        \draw[gray!15, line width=0.5pt] (\bandL, \r) -- (\bandR, \r);
    }
\end{scope}


\node[imgstyle] (brlp)     at (\colA, \rowI - 0.35) {BrLP};
\node[imgstyle] (retfound) at (\colA, \rowI + 0.35) {RETFound};
\node[imgstyle] (tadpm)    at (\colA, \rowII) {\fontsize{6.5}{8}\selectfont Treatment-Aware\\[-1pt]DPM};

\node[txstyle] (mewm)    at (\colB, \rowII)  {MeWM};
\node[txstyle] (clarity)  at (\colB, \rowIII) {CLARITY};

\node[surgstyle] (cardiac)   at (\colC, \rowII - 0.90) {\fontsize{6.8}{8}\selectfont Cardiac Copilot};
\node[surgstyle] (echoworld) at (\colC, \rowII - 0.30) {EchoWorld};
\node[surgstyle] (surgwm)    at (\colC, \rowII + 0.30) {SurgWM};
\node[surgstyle] (uniswm)    at (\colC, \rowII + 0.90) {UniSWM};

\node[ehrstyle] (meddreamer) at (\colD, \rowII) {MedDreamer};

\pgfmathsetmacro{\frontierW}{3*\colsep + 1.6}
\node[frontier, minimum width=\frontierW cm, minimum height=1.0cm, fill=red!3] (l4zone)
    at (0.5*3*\colsep, \rowIV) {};
\node[font=\sffamily\itshape\fontsize{7}{8.5}\selectfont, text=red!60!black]
    at (0.5*3*\colsep, \rowIV) {Open frontier --- no medical world model yet operates at L4};

\node[font=\sffamily\bfseries\fontsize{7}{8.5}\selectfont, anchor=west] (legtitle)
    at (\bandL, -1.8) {Domains:};
\node[imgstyle,  anchor=west, minimum width=48pt, font=\sffamily\fontsize{6.5}{8}\selectfont]
    (leg1) at ($(legtitle.east)+(0.2,0)$) {Medical Imaging};
\node[txstyle,   anchor=west, minimum width=52pt, font=\sffamily\fontsize{6.5}{8}\selectfont]
    (leg2) at ($(leg1.east)+(0.15,0)$) {Treatment Planning};
\node[surgstyle, anchor=west, minimum width=32pt, font=\sffamily\fontsize{6.5}{8}\selectfont]
    (leg3) at ($(leg2.east)+(0.15,0)$) {Surgical};
\node[ehrstyle,  anchor=west, minimum width=38pt, font=\sffamily\fontsize{6.5}{8}\selectfont]
    (leg4) at ($(leg3.east)+(0.15,0)$) {EHR / ICU};

\end{tikzpicture}%
}
\caption{Capability landscape of medical world models. Models are placed according to their clinical domain (columns) and the highest capability level they achieve on the L1--L4 rubric of Qazi et al.~\cite{chen2025medicalwm} (rows). L1 = temporal prediction; L2 = action-conditioned prediction; L3 = counterfactual rollouts; L4 = autonomous planning and control. The dashed red boundary at L4 indicates that no medical world model currently operates at the planning-and-control level, highlighting the principal open frontier.}
\label{fig:medical_landscape}
\end{figure*}

Qazi et al.~\cite{chen2025medicalwm} provided the first focused review of world models for clinical applications, proposing a four-level capability rubric that classifies medical world models by functional maturity: \textbf{L1} (temporal prediction)---forecasting the natural evolution of a patient's state without conditioning on interventions; \textbf{L2} (action-conditioned prediction)---predicting outcomes given specific treatment choices; \textbf{L3} (counterfactual rollouts)---simulating ``what-if'' scenarios for decision support, enabling comparison of multiple candidate interventions; and \textbf{L4} (planning and control)---closing the loop by using the learned dynamics model to autonomously recommend or optimize treatment strategies. This rubric provides a useful framework for assessing the maturity of medical world models across application domains. Figure~\ref{fig:medical_landscape} maps the principal medical world models onto this rubric, revealing that most systems cluster at L1--L2 with a clear gap at L4.



\subsubsection{Medical Imaging Analysis and Progressive Disease Prediction}

World models for medical imaging aim to capture the temporal dynamics of anatomical structures, enabling the prediction of how pathology evolves over time, such as tumor growth, neurodegenerative atrophy, or treatment-induced changes. In this context, these models typically operate across longitudinal time points or patient cohorts rather than in real-time interactive environments.

\textbf{Brain disease progression.} Latent diffusion models have recently emerged as a powerful paradigm for individual-level disease progression modeling using 3D brain MRI. Brain Latent Progression (BrLP)~\cite{puglisi2025brlp}, for example, operates in a compact latent space and incorporates subject-level metadata, including age, sex, diagnosis, and APOE genotype, to predict Alzheimer’s disease (AD) progression from 3D T1-weighted MRI. Trained on 11,730 scans from 2,805 subjects across three longitudinal AD datasets, BrLP achieved a 22\% improvement in volumetric accuracy across AD-relevant brain regions and a 43\% improvement in image similarity relative to prior approaches. An extended version~\cite{puglisi2025brlp} further scaled to more than 14,000 MRIs and demonstrated potential clinical utility in identifying fast progressors in a simulated retrospective study.

\textbf{Brain tumor growth prediction.} Treatment-aware diffusion probabilistic models~\cite{durrer2025tadpm} have been developed to predict future tumor morphology from multi-parametric MRI while conditioning on treatment plans, such as surgery type, chemotherapy regimen, and radiation dose. By incorporating treatment information as a conditioning signal to guide the diffusion process, these models can generate both future MRI volumes and corresponding tumor segmentation masks. Reportedly, they improve future tumor prediction accuracy by more than 16\% compared with treatment-agnostic baselines. This represents an L2-level capability, namely, action-conditioned prediction in which the “action” corresponds to the clinical treatment plan.

\textbf{Retinal disease progression.} In ophthalmology, AI systems have been developed to predict the progression of diabetic retinopathy (DR) and age-related macular degeneration (AMD) from fundus photographs and optical coherence tomography (OCT) scans. Although these approaches are not always explicitly framed as world models, they implement the core L1 capability of temporal prediction. The MICCAI 2024 DIAMOND and MARIO challenges~\cite{miccai2024diamond} benchmarked 21 deep learning methods for patient-specific prediction of DR and AMD progression. In addition, foundation models such as RETFound~\cite{zhou2023retfound}, which was pretrained using self-supervised masked autoencoding on 1.6 million unlabeled retinal images, have demonstrated strong transfer performance for disease detection and progression prediction, outperforming ImageNet-pretrained baselines across multiple ophthalmic and systemic disease tasks.

\textbf{General-purpose medical representations.} JEPA-style predictive representation learning~\cite{assran2023ijepa} offers a promising route toward general-purpose medical image representations without requiring large labeled datasets. By predicting masked image regions in representation space rather than reconstructing them in pixel space, these models can learn features that capture anatomically meaningful structure. Such representations may serve as effective encoders for downstream clinical world models.

\textbf{functional Magnetic Resonance Imaging.} An emerging direction within world models is the prediction of future brain activity from functional Magnetic Resonance Imaging (fMRI), where the goal is to model not only static representations but also the temporal evolution of brain states. Early work demonstrated that auto-regressive models can forecast resting-state dynamics from prior fMRI time series, showing that future brain states can be predicted while preserving functional connectivity structure~\cite{sun2024predicting}. This line of research has progressed to voxel-level prediction by Sun et al., which introduced a Swin Transformer–based framework for 4D fMRI forecasting~\cite{sun2025voxel}. By jointly modeling spatial and temporal dependencies, this approach predicts several seconds of future brain activity from prior scans while preserving BOLD contrast and dynamics, demonstrating that fine-grained whole-brain evolution is learnable at high spatial resolution.

More recently, the field has shifted toward foundation models for fMRI, moving beyond task-specific predictors to large-scale, generalizable representations. For example, NeuroSTORM was pretrained on tens of millions of fMRI frames across large cohorts, enabling transferable representations for prediction, diagnosis, and downstream analysis~\cite{wang2025towards}. Similarly, BrainLM and related models adopt self-supervised learning (e.g., masked modeling) to learn latent brain dynamics from thousands of hours of fMRI data, supporting both forecasting and cross-task generalization~\cite{caro2023brainlm}. These models mark a transition from narrow predictive models to scalable, reusable representations of brain activity. In parallel, multimodal foundation models are beginning to predict brain activity from external stimuli. Meta’s TRIBE and its recent extension TRIBE v2 demonstrate that large-scale models can predict whole-brain fMRI responses from video, audio, and language inputs in a zero-shot manner~\cite{d2025tribe}. Although primarily framed as encoding models, these approaches align with world-model perspectives by learning mappings between external environments and internal brain states.

Building on these advances, our recent work on fMRI foundation modelling~\cite{wang2025towards} further emphasizes learning unified, temporally consistent representations of brain dynamics that can support prediction, imputation, and downstream clinical tasks. This direction aligns closely with the concept of world models, where latent representations capture the evolving state of a complex system—in this case, the human brain.

Overall, the field is rapidly evolving from regional forecasting$\rightarrow$voxel$\rightarrow$level prediction $\rightarrow$ foundation models $\rightarrow$ multimodal brain simulators. This progression suggests that fMRI-based brain-state models are becoming increasingly capable of supporting in silico simulation of brain dynamics, with potential applications in scan acceleration, digital brain twins, disease monitoring, and counterfactual prediction of brain trajectories under different conditions.

\subsubsection{Tumor Evolution and Treatment Planning}

Treatment planning represents the transition from L1 (temporal prediction) to L2--L3 (action-conditioned prediction and counterfactual reasoning), where the world model explicitly simulates how a patient's disease state evolves \emph{in response to} specific therapeutic interventions.

\textbf{MeWM (Medical World Model).} Yang et al.~\cite{yang2025mewm} introduced MeWM, the first world model in medicine that visually predicts future disease states based on clinical decisions. MeWM comprises three components mirroring the canonical world-model architecture: (1) a \emph{policy model} implemented as a vision-language model that generates candidate treatment action combos (e.g., TACE protocol selection for hepatocellular carcinoma) from the patient's current CT scan and clinical context; (2) a \emph{dynamics model} implemented as an action-driven 3D diffusion model trained on longitudinal paired pre- and post-treatment CT volumes that simulates tumor progression or regression under the proposed treatment; and (3) an \emph{inverse dynamics model} that applies survival risk analysis to the simulated post-treatment tumor, quantitatively evaluating treatment efficacy and feeding back to refine the policy. In Turing tests evaluated by board-certified radiologists, MeWM's synthesized post-treatment tumors achieved state-of-the-art specificity. Its inverse dynamics model outperformed medical-specialized GPTs (including GPT-4o) in optimizing individualized TACE protocols, with F1-scores of 52.4\% (in-house) and 64.1\% (public dataset) versus GPT-4o's 42.0\% and 44.3\%. Critically, MeWM improved clinical decision-making for interventional physicians by 13\% F1-score in selecting optimal TACE protocols, demonstrating the practical value of L2--L3 world models in oncology.

\textbf{CLARITY.} Ding et al.~\cite{ding2025clarity} addressed limitations of MeWM---particularly its reliance on stochastic diffusion for visual reconstruction rather than causal physiological transitions, and its lack of patient-specific temporal context. CLARITY models the pre-to-post-treatment transformation as a smooth, interpretable trajectory within a structured latent space rather than reconstructing clinical images pixel-by-pixel. It integrates temporal context (time intervals between scans), clinical context (genomic alterations, demographics, therapeutic history), and an inverse survival evaluation that creates an adaptive feedback loop: predicted outcomes are fed back to the therapy policy agent, enabling iterative refinement of treatment recommendations. On the MU-Glioma-Post dataset for glioma treatment planning, CLARITY outperformed MeWM by 12\% and surpassed all medical-specialized large language models, representing an advance toward L3--L4 capability.

\subsubsection{Surgical Video and Robotic Surgery}

Surgical world models learn the dynamics of surgical scenes---tool-tissue interactions, anatomical deformations, bleeding, and instrument movements---to enable simulation, training, and autonomous assistance. These models operate at L2--L3, with L4 (autonomous surgical planning) as an aspirational frontier.

\textbf{SurgWM (Surgical Vision World Model).} Koju et al.~\cite{gao2025surgwm} presented the first action-controllable surgical visual world model. Inspired by Genie's~\cite{bruce2024genie} approach of learning latent actions from unlabeled video, SurgWM infers latent surgical actions from raw surgical video without requiring action annotations---a critical advantage given that obtaining action labels for surgical data is prohibitively expensive. The architecture comprises a video tokenizer, a surgical latent action model, and a surgical dynamics model, all using spatio-temporal transformers with causal temporal attention for autoregressive prediction. Trained on the SurgToolLoc-2022 dataset (24,695 clips from da Vinci robotic surgery), SurgWM generates action-controllable surgical video predictions, enabling applications in medical professional training and autonomous surgical agent development. Evaluation on FVD and SSIM confirmed both generation quality and controllability, with conditioning on ground-truth-derived actions yielding consistently better metrics.

\textbf{UniSWM (Unified Surgical World Model).} The UniSWM~\cite{unisurgwm2025} proposes a mixture-of-transformers (MoT) architecture that unifies three capabilities within a single model: structured understanding (hierarchical recognition of surgical phases, steps, atomic actions, and movements), long-horizon prediction (multi-step surgical trajectory forecasting), and fine-grained generation (action-conditioned video synthesis). UniSWM introduces UniSWM-DB, a diverse multimodal dataset containing 1.81 million samples for surgical training, and UniSWM-Bench, a comprehensive benchmark covering five understanding tasks, two prediction tasks, and three generation tasks across both in-body and operating room settings.

\textbf{Cardiac Copilot and EchoWorld.} Jiang et al.~\cite{jiang2024cardiaccopilot} introduced Cardiac Copilot, an AI system for real-time ultrasound probe guidance in echocardiography---addressing the severe global shortage of experienced cardiac sonographers. The core innovation is ``Cardiac Dreamer,'' a data-driven world model that represents cardiac spatial structure: given the current probe position, Cardiac Dreamer provides structure features of nearby cardiac planes in latent space, serving as a navigation map for autonomous plane localization. Trained on expert operation data from 125 clinical scans (188K sample pairs collected via a robotic arm with 6-DOF pose sensing), Cardiac Dreamer reduced navigation errors by up to 33\% across three standard echocardiographic planes (PLAX, PSAX-AV, PSAX-MV). Yue et al.~\cite{yue2025echoworld} (CVPR 2025) extended this with EchoWorld, a motion-aware world modeling framework pretrained on over one million ultrasound images from 200+ routine scans. EchoWorld employs masked anatomical prediction and motion-conditioned visual simulation during pretraining, then applies a motion-aware attention mechanism during fine-tuning to integrate historical visual-motion sequences, achieving further reductions in guidance error over Cardiac Copilot and existing visual backbones.

\subsubsection{Disease Progression Modeling from Electronic Health Records}

Applied to electronic health records (EHRs), world models learn to forecast clinical event sequences---lab results, medication changes, deterioration events---from structured temporal patient data, enabling proactive intervention planning. The irregular sampling, sparsity, and noise inherent in EHR data pose challenges absent in visual world modeling.

\textbf{MedDreamer.} Xu et al.~\cite{xu2025meddreamer} introduced MedDreamer, the first model-based RL framework that applies latent imagination to irregular healthcare data. MedDreamer uses a world model with an Adaptive Feature Integration (AFI) module that captures patient and treatment dynamics by modeling their interactions in higher-order latent spaces, addressing the challenges of irregular sampling, sparsity, and missing values in EHR data. Through a two-phase training pipeline, the model first grounds its policy in real clinical time progression, then refines it using horizon-based imagination mixed with real data. Evaluated on sepsis treatment and mechanical ventilation management using two large-scale EHR datasets, MedDreamer achieved relative mortality rate reductions of 20.9\% (sepsis) and 16.0\% (mechanical ventilation) compared to clinician baselines, outperforming both model-free (DDQN, EZ-Vent, DeepVent) and model-based (MBRL-BNN, DreamerV3) baselines. This represents L2--L3 capability on EHR data: action-conditioned prediction with imagination-based policy optimization.

\textbf{Broader EHR-based approaches.} The intersection of deep reinforcement learning and EHR data is an active area with multiple complementary approaches: Deep Q-Networks for disease prediction from EHR sequences, dynamic treatment regime optimization using offline RL, and personalized diagnostic decision pathways that formulate diagnosis as a sequential decision-making problem. However, most existing EHR-based systems operate at L1--L2 (temporal and action-conditioned prediction), with L3 (counterfactual validation of treatment alternatives) and L4 (autonomous closed-loop treatment optimization) remaining largely unrealized due to the absence of interventional ground truth and the ethical constraints of deploying autonomous clinical agents.

\subsubsection{Capability Assessment and Open Challenges}

As illustrated in Figure~\ref{fig:medical_landscape}, the majority of current medical world models operate at L1--L2; advancing to L3--L4 requires addressing several domain-specific challenges:

\begin{enumerate}
    \item \emph{Data scarcity and privacy:} Medical datasets are orders of magnitude smaller than those available for games or driving. A typical longitudinal oncology dataset contains hundreds to low thousands of patients, compared to millions of frames in Atari or thousands of hours of driving video. Strict privacy regulations (HIPAA, GDPR) further constrain data sharing and centralized model training. Federated learning and synthetic data generation via world models themselves offer partial solutions, but have not been validated at scale for clinical applications.
    \item \emph{Clinical validation and interventional ground truth:} Advancing from L2 to L3 requires validation of counterfactual predictions---``What would have happened under an alternative treatment?''---against outcomes that are inherently unobservable (the counterfactual was not administered). Qazi et al.~\cite{chen2025medicalwm} recommend multi-site holdout validation, matched cohort studies, and clinician adjudication, but prospective randomized evaluation of world-model recommendations remains ethically complex and logistically challenging.
    \item \emph{Safety, interpretability, and regulatory approval:} Deploying L4-level autonomous clinical world models requires not only accuracy but also calibrated uncertainty estimates (the model must know when it does not know), interpretability (clinicians must understand the basis for recommendations), and formal regulatory pathways (FDA 510(k) or De Novo for AI-as-medical-device). Overconfident predictions in oncology or surgery can be life-threatening; abstention policies and human-in-the-loop safeguards are essential.
    \item \emph{Heterogeneity across patient populations:} Patient populations vary across demographics, comorbidities, genetic backgrounds, and treatment histories. A world model trained on one institution's cohort may not generalize to another's. Domain adaptation, multi-site pretraining, and personalization through patient-specific conditioning (as in CLARITY~\cite{ding2025clarity} and BrLP~\cite{puglisi2025brlp}) are active areas of research.
    \item \emph{Multimodal state construction:} Clinical ``observations'' span imaging (CT, MRI, ultrasound, pathology slides), structured data (labs, vitals, medications), unstructured text (clinical notes, radiology reports), and genomic data. Constructing a complete patient state representation that integrates these heterogeneous modalities remains an open challenge, yet is necessary for world models that aspire to comprehensive clinical reasoning.
\end{enumerate}

\subsection{Educational Measurement}

\label{sec:edu_measurement}

Educational measurement studies how learners' knowledge, abilities, and reasoning processes can be inferred from observable educational data. 
The field spans a broad range of tasks, including ability estimation, student modeling, automated assessment, and learning analytics. 
Across these settings, the common objective is to recover latent cognitive constructs from partial observations and to characterize how these constructs evolve under instructional interactions.

From the perspective of world models, educational measurement can be reinterpreted as learning a dynamical model of a learner's cognitive environment. 
In this formulation, the learner is treated as a partially observable environment, instructional interventions (e.g., exercises, hints, feedback, or curriculum choices) play the role of actions, and student behaviors (e.g., answers, explanations, or scores) are treated as observations. 
This interpretation is consistent with the central goal of world models: learning latent state representations together with transition and observation mechanisms that support prediction, simulation, and downstream decision making.
For clarity, in this survey we use the term \emph{educational world model} to denote a world-model formulation of educational measurement, in which the learner is treated as a partially observable dynamical system whose latent cognitive state evolves under instructional actions.

Formally, let $s_t$ denote the learner's latent cognitive state at time step $t$, where $s_t$ may encode knowledge mastery, misconceptions, strategy use, or other unobserved cognitive variables. 
Let $a_t$ denote an instructional action applied at time $t$, such as assigning an exercise, providing feedback, or selecting the next learning activity. 
Let $o_t$ denote the observable outcome at time $t$, such as a correctness label, a score, a textual explanation, or another behavioral trace. 
Educational measurement under a world-model view can then be written as
\begin{equation}
s_{t+1} \sim p(s_{t+1}\mid s_t, a_t, o_t),
\label{eq:edu_transition}
\end{equation}
\begin{equation}
o_{t+1} \sim p(o_{t+1}\mid s_{t+1}, a_{t+1}).
\label{eq:edu_observation}
\end{equation}
In Eq.~\ref{eq:edu_transition}, the transition model describes how the latent cognitive state evolves after the learner interacts with an instructional action and produces an observation. 
In Eq.~\ref{eq:edu_observation}, the observation model maps the updated latent state to the next observable behavior. 
This decomposition closely parallels the structure of latent world models in model-based reinforcement learning, where hidden state, dynamics, and observation are modeled jointly.

\begin{figure}[t]
\centering
\begin{tikzpicture}[
    node distance=7mm and 9mm,
    box/.style={draw, rounded corners, align=center, minimum height=8mm, inner sep=4pt},
    smallbox/.style={draw, rounded corners, align=center, inner sep=3pt},
    arrow/.style={-{Latex[length=2.2mm,width=1.6mm]}, thick}
]
\node[box, minimum width=3.6cm] (obs) {Student interaction history\\{\footnotesize questions, responses, essays, feedback traces}};
\node[box, minimum width=3.0cm, below=of obs] (enc) {State inference / encoder};
\node[box, minimum width=3.0cm, below=of enc] (state) {Latent cognitive state $s_t$\\{\footnotesize mastery, ability, misconceptions, reasoning}};
\node[box, minimum width=3.4cm, right=18mm of state] (dyn) {Learning dynamics model\\{\footnotesize $p(s_{t+1}\mid s_t,a_t,o_t)$}};
\node[box, minimum width=3.4cm, below=of dyn] (obsmodel) {Observation model\\{\footnotesize $p(o_{t+1}\mid s_{t+1},a_{t+1})$}};
\node[smallbox, minimum width=2.8cm, above=of dyn] (action) {Instructional action $a_t$\\{\footnotesize exercise / hint / feedback / policy}};
\node[box, minimum width=3.0cm, left=18mm of obsmodel] (future) {Predicted / simulated outcomes\\{\footnotesize correctness, score, explanation, trajectory}};
\node[box, minimum width=3.3cm, below=of obsmodel] (plan) {Counterfactual simulation \&\\instructional planning};

\draw[arrow] (obs) -- (enc);
\draw[arrow] (enc) -- (state);
\draw[arrow] (state) -- (dyn);
\draw[arrow] (action) -- (dyn);
\draw[arrow] (dyn) -- node[right] {\footnotesize $s_{t+1}$} (obsmodel);
\draw[arrow] (obsmodel) -- (future);
\draw[arrow] (future) -- (plan);

\begin{scope}[on background layer]
\node[draw, dashed, rounded corners, fit=(enc)(state)(dyn)(obsmodel), inner sep=5pt, label=above:{\footnotesize Educational world model}] {};
\end{scope}
\end{tikzpicture}
\caption{Educational measurement viewed as a world model of learning dynamics. 
The figure combines the standard knowledge-tracing pipeline (interaction history $\rightarrow$ latent knowledge estimation $\rightarrow$ response prediction) with the latent-state formulation used in world models (state inference $\rightarrow$ dynamics $\rightarrow$ observation). 
This design is motivated by classical student modeling and knowledge tracing on the one hand, and by latent world-model architectures used in model-based reinforcement learning on the other.}
\label{fig:edu_world_model}
\end{figure}
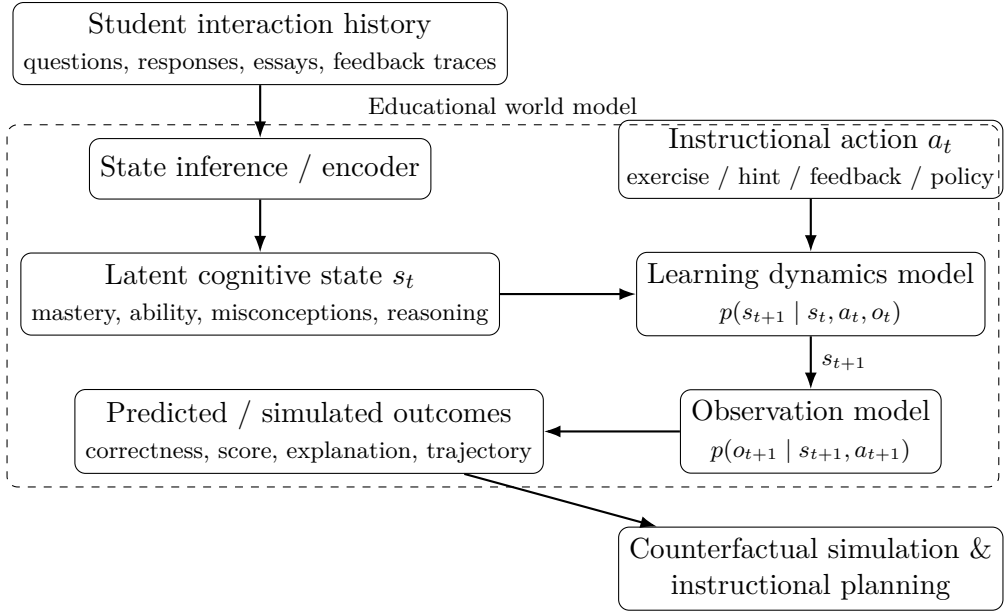

As illustrated in Fig.~\ref{fig:edu_world_model}, this perspective provides a unified way to connect traditional educational measurement with modern world-model architectures. 
Student interaction histories are first encoded into a latent knowledge state; a learning-dynamics module then models how this state changes under instructional actions; finally, an observation module predicts externally visible outcomes such as correctness, scores, or explanations. 
This view is useful because it not only covers response prediction, but also naturally suggests simulation of future learning trajectories and planning of instructional strategies.

\paragraph{Structural correspondence between education and world models.}
The mapping in Eqs.~\ref{eq:edu_transition}--\ref{eq:edu_observation} clarifies why educational measurement is a natural application domain for world-model thinking. 
The latent state $s_t$ plays the role of the learner's hidden knowledge space; the transition model captures learning, forgetting, and other state changes caused by instruction; and the observation model explains how latent cognition gives rise to visible behavior. 
Under this view, student modeling methods can be interpreted as domain-specific latent-dynamics models, even when they were not originally introduced using the vocabulary of world models.

\paragraph{Classical student models as early latent-dynamics models.}
Classical educational measurement already contains important ingredients of this formulation. 
Bayesian Knowledge Tracing (BKT) models each skill as a binary latent variable $L_t\in\{0,1\}$ indicating whether the skill is unmastered or mastered. 
The transition from one latent state to the next is governed by a learning probability
\begin{equation}
P(L_{t+1}=1\mid L_t=0)=p_{\mathrm{learn}},
\label{eq:bkt_transition}
\end{equation}
where $p_{\mathrm{learn}}$ denotes the probability that the learner acquires the skill between two interaction steps. 
The observation model then accounts for stochastic response noise through slip and guess parameters:
\begin{equation}
P(o_t=1\mid L_t=1)=1-p_{\mathrm{slip}}, \qquad
P(o_t=1\mid L_t=0)=p_{\mathrm{guess}},
\label{eq:bkt_obs}
\end{equation}
where $o_t=1$ means that the learner answers correctly, $p_{\mathrm{slip}}$ is the probability of making an error despite mastery, and $p_{\mathrm{guess}}$ is the probability of answering correctly without mastery. 
Although BKT is simple and skill-local, it is already an explicit latent-state transition model. \cite{corbett1994knowledge}

Item Response Theory (IRT) emphasizes the observation side of educational measurement. 
In the basic one-parameter logistic form, the probability that learner $i$ answers item $j$ correctly is
\begin{equation}
P(o_{ij}=1)=\frac{1}{1+\exp\big(-(\theta_i-\beta_j)\big)},
\label{eq:irt}
\end{equation}
where $\theta_i$ denotes the learner's latent ability and $\beta_j$ denotes the difficulty of the item. 
IRT is highly interpretable and foundational in psychometrics, but by itself it is primarily static: it measures latent ability from observations rather than modeling rich temporal state evolution. \cite{lord2012applications}

\paragraph{Neural latent-dynamics models for student state tracking.}
The shift from classical psychometrics to modern knowledge tracing reflects a shift from low-dimensional parametric state models to more expressive latent simulators. 
Deep Knowledge Tracing (DKT) uses recurrent neural networks to summarize interaction history into a continuous hidden state:
\begin{equation}
h_t = \mathrm{RNN}(h_{t-1}, x_t), \qquad
y_t = \sigma(W h_t + b),
\label{eq:dkt}
\end{equation}
where $x_t$ encodes the current interaction (typically the question-response pair), $h_t$ is the latent representation of the learner's state, $W$ and $b$ are learnable parameters, and $y_t$ is the predicted probability vector for future responses. 
Compared with BKT, DKT replaces hand-designed parametric transitions with a learned continuous-state dynamics model that can capture more complex dependencies in learning sequences. \cite{piech2015deep}

Subsequent work increasingly enriched this latent-dynamics view. 
DKVMN separates concept memory from mastery memory, allowing the model to represent which concepts exist and how well each concept is mastered over time. \cite{zhang2017dynamic}
SAKT uses self-attention to select relevant prior interactions for the current prediction, thereby improving modeling under sparse histories. \cite{pandey2019self}
SAINT introduces a deeper encoder-decoder Transformer formulation for knowledge tracing, while AKT integrates attentive sequence modeling with interpretable components inspired by cognitive and psychometric considerations. \cite{choi2020towards,ghosh2020context}
More recent models continue this trend: DTransformer emphasizes diagnostic and stable state estimation, HiTSKT models hierarchical/session-aware sequence structure, and LKT brings pretrained language-model representations into knowledge tracing when question and concept text are available. \cite{yin2023tracing,ke2024hitskt,lee2024language}
Together, these methods can be interpreted as increasingly expressive latent world models of learning processes.

\paragraph{Structured observation and generative assessment.}
A central challenge for educational world models is that the observation space is often far richer than a binary correct/incorrect label. 
Many authentic assessments involve open-ended text, explanations, short answers, essays, or multi-criteria rubrics. 
This makes the observation model in Eq.~\ref{eq:edu_observation} substantially harder: the system must map latent cognitive states to semantically rich outputs and, in many cases, also infer structured evidence from those outputs.

Recent LLM-based assessment studies show that large language models can achieve competitive performance in automated essay scoring and broader automated assessment settings, though validity, robustness, and rubric alignment remain central concerns. \cite{mansour2024can,pack2024large,emirtekin2025large}
Within this context, AutoSCORE provides a particularly natural example for this survey because it can be interpreted as a structured observation model. 
Instead of directly mapping a student response to a final score, AutoSCORE first extracts rubric-aligned scoring components from the response and then aggregates them into a final assessment:
\begin{equation}
o_t^{\mathrm{struct}} = \mathrm{Agent}_{\mathrm{extract}}(o_t^{\mathrm{text}}, \mathrm{Rubric}),
\label{eq:autoscore_extract}
\end{equation}
where $o_t^{\mathrm{text}}$ is the raw open-ended student response and $o_t^{\mathrm{struct}}$ is its structured rubric-aligned representation. 
A second scoring stage then maps this structured representation to the final score. 
From a world-model perspective, this is appealing because it turns unstructured observations into diagnostically meaningful components that can more directly support latent-state updating. \cite{wang2025autoscore}

\paragraph{Critical synthesis: from response prediction to imaginative instructional planning.}
A key insight from the literature is that educational measurement has become increasingly strong at \emph{predicting} immediate student performance, but it has not yet fully exploited the \emph{simulation and planning} capabilities that define modern world models. 
If the transition model $\hat{T}$ in Eq.~\ref{eq:edu_transition} were learned with sufficient fidelity, one could simulate future latent trajectories under counterfactual instructional choices:
\begin{equation}
\hat{s}_{t+1:t+H} = \mathrm{Simulate}\!\left(\hat{T}, s_t, \{a'_k\}_{k=t}^{t+H-1}\right),
\label{eq:counterfactual_sim}
\end{equation}
where $\hat{s}_{t+1:t+H}$ denotes the simulated future latent states over a horizon of length $H$, $\hat{T}$ is the learned transition model, $s_t$ is the current latent learner state, and $\{a'_k\}_{k=t}^{t+H-1}$ is a candidate sequence of future instructional actions. 
Such simulation would support counterfactual questions of direct pedagogical interest: which curriculum path is likely to maximize long-term retention, reduce cognitive overload, or accelerate mastery for a particular learner?

This research gap should be framed carefully. 
It does not mean that educational AI lacks student models; on the contrary, the field has developed a rich ecosystem of psychometric, sequential, and assessment models. 
Rather, the opportunity is that these methods have only partially been integrated into full simulation-based world-model pipelines. 
In this sense, educational measurement is already close to world modeling in spirit, but still early in its use of explicit imaginative rollouts, long-horizon planning, and adaptive counterfactual reasoning. 
That gap makes the domain especially promising for future work on intelligent tutoring systems, assessment-aware planning, and learner-specific adaptive world models.

\subsection{Business and Finance}
\label{sec:fin}

Business and financial systems represent a fundamentally different regime for world modeling, where the objective shifts from simulating deterministic physical dynamics to reasoning under severe uncertainty, hidden states, and strategic human interactions. Unlike classical environments governed by stable transition laws, corporate and financial ecosystems are shaped by latent socioeconomic factors and the adaptive behavior of competing agents. Consequently, world models in this domain must operate not as rigid physical engines, but as highly adaptable inferential and decision-support systems.

\begin{table}[htbp]
\centering
\small
\caption{World modeling in business and finance: core dimensions and methodologies.}
\label{tab:finance_wm_structure}
\renewcommand{\arraystretch}{1.3} 
\begin{tabularx}{0.95\linewidth}{@{} l >{\raggedright\arraybackslash}X >{\raggedright\arraybackslash}X >{\raggedright\arraybackslash}X @{}}
\toprule
\textbf{Dimension} & \textbf{Financial Context} & \textbf{Modeling Approach} & \textbf{Strategic Value} \\
\midrule
Latent State & Unobservable drivers & Belief POMDPs, latent variable models & Asset allocation under uncertainty \\
Market Dynamics & Non-stationary regime shifts & Adaptive state-space models & Long-horizon risk management \\
Financial Signals & Noisy, delayed, or alternative data & Bayesian filtering, representation learning & Real-time market inference \\
Interactions & Algorithmic \& competitor behaviors & Game-theoretic MARL & Anticipating strategic responses \\
Evaluation & Counterfactual strategy outcomes & Causal inference, offline RL & Risk-free policy backtesting \\
Reflexivity & Market self-impact and price impact & Policy-conditioned dynamics & Handling self-referential markets \\
\bottomrule
\end{tabularx}
\end{table}

\subsubsection{Beyond physical laws: the belief-modeling paradigm.}
Financial markets and corporate architectures are defined by severe partial observability, multi-agent strategic interactions, and structural non-stationarity. The underlying true state of the system—comprising institutional order flow, market sentiment, shifting consumer demand, and competitor strategies—is only indirectly observable through noisy or delayed signals.

Because of this, world models in business and finance cannot function as deterministic simulators; rather, they must operate strictly as belief models. The latent state does not correspond to a concrete spatial configuration, but instead represents a probability distribution over hidden economic drivers. This reframes the fundamental modeling objective: moving from predicting observable trajectories to continuously maintaining and updating coherent beliefs about latent factors, an approach that naturally aligns with the principles of partially observable Markov decision processes (POMDPs)~\cite{hambly2023recent}.

\subsubsection{Methodological trajectories.}
Existing applications of world models in this domain can be broadly categorized into three methodological trajectories. The first encompasses time-series predictive models, leveraging architectures such as recurrent neural networks and Temporal Fusion Transformers~\cite{lim2021temporal} to map historical observations directly to future asset prices or operational metrics. While highly effective for pure forecasting, these approaches typically lack an explicit, manipulable latent structure, thereby providing limited support for counterfactual intervention or causal decision analysis.

To address this limitation, a second class of methods employs latent dynamics models, predominantly inspired by state-space formulations. By introducing hidden variables that capture macroeconomic regimes (e.g., bull vs.\ bear markets), liquidity conditions, or latent risk factors, these models explicitly separate observable signals from their underlying generative drivers. This enables more robust reasoning over sudden regime shifts and long-horizon dependencies, bringing them significantly closer to the core principles of world modeling.

A third, increasingly critical direction formulates financial systems as multi-agent environments~\cite{liu2022finrl}. In this paradigm, market dynamics are viewed as emergent phenomena arising from the interactions of heterogeneous agents—such as algorithmic traders, institutions, and retail consumers—each pursuing distinct, often competing objectives. As a result, the evolution of the system depends not only on exogenous economic conditions but also on the strategic adaptations of other participants, intrinsically introducing a game-theoretic dimension into the world model.

\subsubsection{The imperative of counterfactual reasoning in corporate and financial strategy.}
In business and finance, strategic decision-making fundamentally revolves around evaluating alternatives: comparing different algorithmic trading strategies, portfolio allocations, dynamic pricing policies, or supply chain interventions. This operational reality necessitates rigorous counterfactual reasoning—estimating what would have transpired under a different decision while holding the underlying latent market conditions constant.

Formally, this is expressed by comparing expected outcomes under alternative interventions via the $do$-calculus framework~\cite{pearl2009causality}:
\begin{equation}
\Delta(a_t', a_t) = \mathbb{E}[R \mid do(a_t')] - \mathbb{E}[R \mid do(a_t)],
\end{equation}
which precisely quantifies the causal impact of a chosen action. In offline settings, causally grounded world models transform historical transaction logs and market data into a dynamic substrate for policy evaluation~\cite{levine2020offline}, enabling institutions to stress-test strategies without incurring real-world financial risk.

\subsubsection{Fundamental bottlenecks: non-stationarity, identifiability, and reflexivity.}
Despite these conceptual advances, the deployment of world models in finance faces formidable bottlenecks. First, non-stationarity is pervasive: macroeconomic conditions, regulatory frameworks, and collective agent behaviors evolve continuously, leading to severe distribution shifts that rapidly degrade learned dynamics. Second, non-identifiability poses a critical theoretical challenge; multiple latent configurations may explain the observed historical data equally well while yielding drastically divergent counterfactual predictions.

Third, and perhaps most uniquely, financial systems exhibit reflexivity~\cite{soros2013fallibility}—a phenomenon where the deployment of a model and its subsequent actions directly alter the environment it is attempting to model. This creates self-referential feedback loops where the environment is no longer independent of the agent's policy $\pi$:
\begin{equation}
p(z_{t+1} \mid z_t) \xrightarrow{\text{reflexivity}} p(z_{t+1} \mid z_t, \pi(z_t)).
\end{equation}
Such endogenous effects violate the standard reinforcement learning assumption of a fixed, exogenous environment, fundamentally complicating long-horizon prediction and robust control.

Ultimately, these characteristics dictate that world models for business and finance must transcend the role of passive environmental simulators. Instead, they must be rigorously engineered as belief-driven, strategically aware, and causally grounded systems, uniquely capable of operating amid deep uncertainty, adversarial interaction, and continuous structural evolution.

\section{Evaluation Protocols and Benchmarks}
\label{sec:evaluation}
Evaluating world models is fundamentally more complex than evaluating standard generative models or policies in isolation. A world model must simultaneously satisfy multiple requirement---perceptual fidelity, temporal consistency, physical plausibility, action-conditioned accuracy, and downstream task utility--yet no single metric or benchmark captures all of these dimensions. This section surveys the evaluation landscape along three axes: the metrics used to quantify model quality (Section~\ref{sec:eval_metrics}), the benchmark environments and datasets against which models are tested (Section~\ref{sec:eval_benchmarks}).

\subsection{Common Evaluation Metrics}
\label{sec:eval_metrics}

Evaluation metrics for world models can be organized into four families, each measuring a distinct facet of model quality.

\textbf{Perceptual quality metrics.}
These metrics assess the visual fidelity of generated observations. The \emph{Fr\'{e}chet Inception Distance} (FID)~\cite{heusel2017fid} computes the Fr\'{e}chet distance between the feature distributions (extracted from an Inception-v3 network trained on ImageNet) of generated and real image sets. Lower FID indicates higher distributional similarity. The \emph{Fr\'{e}chet Video Distance} (FVD)~\cite{unterthiner2019fvd} extends this to video by using features from an Inflated 3D ConvNet (I3D) trained on the Kinetics dataset, capturing both spatial quality and coarse temporal coherence. FVD has become the de facto standard for video generation evaluation, though it has known limitations: it prioritizes per-frame quality over temporal consistency (a phenomenon termed ``content bias''~\cite{ge2024contentdebiased}), and it is sensitive to sample size, yielding non-zero values even for identical distributions due to finite-sample estimation noise~\cite{unterthiner2019fvd}.

Frame-level metrics complement distributional measures. The Structural Similarity Index (SSIM)~\cite{wang2004ssim} evaluates luminance, contrast, and structural information between pairs of images, correlating better with human perception than pixel-wise MSE. The \emph{Peak Signal-to-Noise Ratio} (PSNR) measures pixel-level reconstruction accuracy. The \emph{Learned Perceptual Image Patch Similarity} (LPIPS)~\cite{zhang2018lpips} computes distances in deep feature space (typically VGGNet), achieving strong alignment with human perceptual judgments. However, all frame-level metrics are computed per-frame without access to temporal context, making them blind to inter-frame consistency---a critical property for world models that must generate coherent trajectories.

\textbf{Task-performance metrics.}
In model-based RL, the primary evaluation metric is \emph{cumulative episodic return}, often normalized against human performance. The \emph{Human-Normalized Score} (HNS) is defined as $\text{HNS} = (R_{\text{agent}} - R_{\text{random}}) / (R_{\text{human}} - R_{\text{random}})$, where $R$ denotes cumulative reward. Mean and median HNS across games are the standard aggregate statistics for the Atari 100k benchmark~\cite{kaiser2020simpl}. Agarwal et al.~\cite{agarwal2021precipice} advocated replacing mean/median with the \emph{interquartile mean} (IQM) and stratified bootstrap confidence intervals, demonstrating that mean HNS is dominated by outlier games and median HNS discards too much information. For continuous control (DeepMind Control Suite), raw episodic returns---averaged over seeds and normalized by an expert policy---are standard.

In autonomous driving, task metrics include \emph{driving score} (a composite of route completion and infraction penalties), \emph{success rate}, \emph{collision rate}, and planning-specific measures such as \emph{L2 trajectory error} and \emph{displacement error}. The CARLA Leaderboard~\cite{dosovitskiy2017carla} reports a driving score from 0 to 100. Bench2Drive~\cite{jia2024bench2drive} introduced multi-ability evaluation with fine-grained driving competency scores. In robotics, \emph{success rate} on manipulation or locomotion tasks is standard, evaluated over multiple trials with randomized initial conditions.

\textbf{Temporal consistency and dynamics metrics.}
VBench~\cite{huang2024vbench} decomposed video generation quality into 16 disentangled dimensions, including subject consistency, background consistency, temporal flickering, motion smoothness, and dynamic degree. VBench 2.0~\cite{huang2025vbench2} extended this to intrinsic faithfulness, adding physics plausibility, commonsense reasoning, and human anatomy consistency. WorldBench~\cite{upadhyay2026worldbench} proposed concept-specific physical evaluation using segmentation-based foreground mIoU against ground-truth physics simulations (generated via Kubric/PyBullet), revealing that state-of-the-art video world models achieve only 45\% mIoU on physical reasoning tasks. These emerging benchmarks represent a shift from distributional perceptual metrics toward diagnostic evaluation of specific physical and causal properties.

\textbf{Prediction accuracy metrics.}
For 3D world models, \emph{Chamfer distance} measures the geometric fidelity of predicted point clouds against ground truth. Copilot4D~\cite{zhang2024copilot4d} reported a 65\% reduction in Chamfer distance for 1-second LiDAR prediction. Occupancy-based models (OccWorld~\cite{wang2024occworld}) use \emph{Intersection-over-Union} (IoU) on predicted voxel grids. For offline RL world models, \emph{one-step prediction error} (MSE in latent or observation space) and \emph{multi-step rollout divergence} quantify dynamics model accuracy independently of downstream policy quality.

\subsection{Benchmark Environments and Datasets}
\label{sec:eval_benchmarks}

Benchmark environments for world models span four primary application domains, each with distinct evaluation protocols and data characteristics.

\textbf{Reinforcement learning benchmarks.}
The \emph{Atari 100k} benchmark~\cite{kaiser2020simpl} restricts agents to 100,000 environment steps (approximately 2 hours of real-time gameplay) across 26 games from the Arcade Learning Environment (ALE)~\cite{bellemare2013ale}. This data-efficiency constraint makes it the primary testbed for model-based RL methods. EfficientZero~\cite{ye2021efficientzero} achieved 194\% mean HNS; DIAMOND~\cite{alonso2024diamond} set the current record for agents trained entirely within a world model at 1.46 mean HNS. The \emph{DeepMind Control Suite} (DMC)~\cite{tassa2018dmc} provides 20+ continuous control tasks from pixels with a standard budget of $10^6$ environment steps, evaluating locomotion and manipulation under diverse physics. DreamerV3~\cite{hafner2023dreamerv3} demonstrated cross-domain generality on DMC with fixed hyperparameters. Additional RL benchmarks include \emph{Crafter}~\cite{hafner2022crafter} (a procedurally generated survival game testing 22 achievements), \emph{Minecraft} (open-ended exploration with sparse rewards), and \emph{Memory Maze}~\cite{samsami2024r2i} (testing long-term memory in world models).

\emph{D4RL}~\cite{fu2020d4rl} provides standardized offline RL datasets across locomotion, navigation, and manipulation tasks, enabling evaluation of world models that must learn dynamics from fixed datasets without online interaction. Recent work has shown that diffusion-based world models achieve 44\% improvement over one-step models on D4RL locomotion tasks. However, D4RL's tasks are increasingly saturated; D5RL~\cite{rafael2024d5rl} has been proposed with more realistic robotic tasks where current methods struggle to outperform behavioral cloning.

\textbf{Autonomous driving benchmarks.}
\emph{nuScenes}~\cite{caesar2020nuscenes} provides 1,000 driving scenes (5.5 hours) with 360\textdegree{} sensor coverage and 3D annotations, serving as the dominant evaluation dataset for driving world models. However, growing consensus holds that nuScenes was designed for perception rather than planning, and open-loop L2 trajectory error on nuScenes correlates poorly with closed-loop driving performance~\cite{jia2024bench2drive}. The \emph{CARLA simulator}~\cite{dosovitskiy2017carla} provides closed-loop evaluation via the CARLA Leaderboard, where agents drive through predefined routes and receive composite driving scores. \emph{Bench2Drive}~\cite{jia2024bench2drive} (NeurIPS 2024) introduced 10,000 clips with multi-ability closed-loop evaluation, explicitly arguing that L2 error is not a meaningful indicator and that the community should move beyond open-loop nuScenes planning results. \emph{NAVSIM}~\cite{dauner2024navsim} (NeurIPS 2024) provided the first direct comparison between model families previously benchmarked exclusively on either CARLA or nuScenes, revealing that simple methods can match large-scale architectures on challenging scenarios. \emph{nuPlan}~\cite{caesar2021nuplan} extends nuScenes with a closed-loop simulator for planning evaluation.

\textbf{Video generation and scene understanding benchmarks.} Measuring the gap between visual fidelity and physical understanding requires evaluation protocols that go beyond traditional perceptual metrics. The evaluation of video prediction and generation models has evolved substantially alongside model capabilities, progressing from simple pixel-level metrics to multidimensional assessment frameworks. Table~\ref{tab:video_eval_benchmarks} summarizes the principal evaluation protocols of video prediction and scene understanding.

Traditional metrics assess low-level visual quality: \textbf{Fr\'{e}chet Video Distance} (FVD)~\cite{unterthiner2019fvd} extends the image-level FID to video by comparing distributions of features extracted from a pretrained video classifier, and remains the most widely reported metric for unconditional and conditional video generation. \textbf{SSIM} and \textbf{LPIPS} measure structural similarity and perceptual distance at the frame level. However, Luo et al.~\cite{luo2025beyondfvd} demonstrated that FVD correlates poorly with human judgments for modern high-fidelity generators, motivating the development of more discriminative metrics.

Recent benchmarks take a multidimensional approach. \textbf{VBench}~\cite{huang2024vbench} introduced a comprehensive evaluation suite decomposing video generation quality into 16 fine-grained dimensions---including subject consistency, background consistency, temporal flickering, motion smoothness, aesthetic quality, and text-video alignment---enabling diagnosis of specific failure modes rather than relying on a single aggregate score. \textbf{VBench 2.0}~\cite{huang2025vbench2} extended this framework to assess \emph{intrinsic faithfulness}, evaluating whether generated videos obey real-world physical and commonsense constraints. \textbf{EvalCrafter}~\cite{liu2024evalcrafter} provided an evaluation framework spanning visual quality, text-video alignment, motion quality, and temporal coherence, benchmarking a broad range of video generation models. \textbf{WorldSimBench}~\cite{qin2025worldsimbench} focused specifically on evaluating video generation models as world simulators, introducing metrics for action controllability, physical consistency, and interactive fidelity that standard visual quality metrics fail to capture.

A persistent challenge across all evaluation approaches is the gap between perceptual quality and physical fidelity: a video can be visually stunning yet physically implausible, and no current metric reliably captures this distinction at scale. The development of evaluation protocols that assess physical understanding, causal consistency, and interactive reliability---beyond mere visual appeal---remains a critical open problem for the field.

\begin{table}[htbp]
\centering
\caption{Representative evaluation benchmarks and metrics for video prediction and scene understanding.}
\label{tab:video_eval_benchmarks}
\small
\begin{tabular}{llp{3.0cm}p{4.5cm}}
\toprule
\textbf{Benchmark / Metric} & \textbf{Year} & \textbf{Type} & \textbf{Key Focus} \\
\midrule
FVD~\cite{unterthiner2019fvd}          & 2019 & Distribution metric & Video-level distributional quality \\
VBench~\cite{huang2024vbench}          & 2024 & Multi-dim.\ benchmark & 16 fine-grained quality dimensions \\
VBench 2.0~\cite{huang2025vbench2}     & 2025 & Multi-dim.\ benchmark & Intrinsic faithfulness and physics \\
EvalCrafter~\cite{liu2024evalcrafter}  & 2024 & Multi-dim.\ benchmark & Visual, motion, and alignment quality \\
WorldSimBench~\cite{qin2025worldsimbench} & 2025 & Simulator benchmark & Action controllability and physical consistency \\
Beyond FVD~\cite{luo2025beyondfvd}     & 2025 & Metric analysis & Enhanced metrics beyond FVD limitations \\
\bottomrule
\end{tabular}
\end{table}

\textbf{Robotics benchmarks.}
Robot world models are evaluated through downstream task success rates on manipulation benchmarks such as \emph{RLBench}~\cite{james2020rlbench} (100 hand-designed tasks with proprioceptive and visual observations), \emph{CALVIN}~\cite{mees2022calvin} (language-conditioned long-horizon manipulation), and \emph{Meta-World}~\cite{yu2020metaworld} (50 robotic manipulation tasks for multi-task and meta-learning). DayDreamer~\cite{wu2023daydreamer} evaluated on real robot hardware, measuring locomotion distance and manipulation success rate from limited real-world interactions. DreMa~\cite{barcellona2025drema} reported one-shot policy learning accuracy on a Franka Emika Panda robot. Table~\ref{tab:eval_benchmarks} provides an overview of the principal benchmarks.

\begin{table*}[htbp]
\centering
\caption{Principal benchmark environments for world model evaluation.}
\label{tab:eval_benchmarks}
\renewcommand{\arraystretch}{1.25}
\setlength{\tabcolsep}{4pt}
\begin{tabular}{@{} >{\raggedright\arraybackslash}p{2.2cm}
                     >{\raggedright\arraybackslash}p{2.0cm}
                     >{\raggedright\arraybackslash}p{1.8cm}
                     >{\raggedright\arraybackslash}p{2.8cm}
                     >{\raggedright\arraybackslash}p{2.8cm}
                     >{\raggedright\arraybackslash}p{2.6cm} @{}}
\toprule
\textbf{Benchmark} & \textbf{Domain} & \textbf{Eval.\ Type} & \textbf{Budget / Scale} & \textbf{Primary Metrics} & \textbf{Top Methods} \\
\midrule
Atari 100k~\cite{kaiser2020simpl}  & RL / Games  & Online, closed  & 100k steps (${\sim}$2 hrs) & Mean/Median HNS, IQM  & DIAMOND, EfficientZero \\
DMC~\cite{tassa2018dmc}            & Cont.\ ctrl & Online, closed  & 1M steps                   & Episodic return        & DreamerV3, TD-MPC2 \\
D4RL~\cite{fu2020d4rl}             & Offline RL  & Offline         & Fixed datasets             & Normalized score       & DWM, LEQ \\
nuScenes~\cite{caesar2020nuscenes} & Driving     & Open-loop       & 5.5 hrs, 1k scenes        & L2 error, FID, FVD     & GAIA-1, Vista \\
CARLA~\cite{dosovitskiy2017carla}  & Driving     & Closed-loop     & Predefined routes          & Driving score, SR      & Bench2Drive entries \\
VBench~\cite{huang2024vbench}      & Video gen.  & Automated       & 16 dimensions              & Composite quality      & Sora, Kling \\
RLBench~\cite{james2020rlbench}    & Robotics    & Task success    & 100 tasks                  & Success rate           & PerAct, RoboMM \\
\bottomrule
\end{tabular}
\end{table*}

\section{Major Challenges and Limitations}
\label{sec:challenges}

Despite rapid progress, world models still face several fundamental challenges that span architectures, methodological families, and application domains. These are not simply engineering limitations, but often reflect deeper tensions between learned dynamics models and the complexity of real-world environments. This section examines seven major challenges: long-horizon consistency and error compounding, scalability and computational cost, the sim-to-real transfer gap, fragmented evaluation and benchmarking, data efficiency and representation learning, safety and robustness in high-stakes settings, and the still unresolved problem of multi-modal grounding.

\subsection*{Long-Horizon Consistency and Error Compounding}

Autoregressive world models generate future states by recursively feeding their own predictions as inputs, a process that causes small per-step inaccuracies to accumulate multiplicatively over the rollout horizon. Janner et al.~\cite{janner2019mbpo} formalized this risk by deriving monotonic improvement bounds showing that the return discrepancy between the learned model and the true environment grows linearly with both the rollout length~$k$ and the single-step model error~$\varepsilon_m$. In practice, this means that a world model with only 1\% per-step prediction error can diverge dramatically from reality after just tens of imagined steps, producing physically implausible or semantically incoherent trajectories. Ding et al.~\cite{ding2025dwm} demonstrated this empirically for offline RL, showing that predicted returns ``quickly collapse as rollout length increases'' when using standard one-step autoregressive dynamics.

\textbf{Controlled rollout horizons.}
The most widely adopted mitigation strategy is to limit the temporal extent of imagined rollouts. MBPO~\cite{janner2019mbpo} initiates short model-based rollouts (starting at $k{=}1$ and gradually extending to $k{=}25$) from real states stored in a replay buffer, then mixes these imagined transitions with real data for off-policy training. This \emph{branched rollout} design prevents long-horizon error accumulation while still gaining substantial sample efficiency---achieving the theoretical monotonic improvement guarantee within the trust region of the learned dynamics. DreamerV3~\cite{hafner2025mastering} employs a complementary approach: rather than limiting rollout length, it stabilizes long-horizon imagination through a suite of robustness techniques including symlog-transformed predictions that compress large value ranges, percentile-based return normalization, and KL balancing with free bits that prevents posterior collapse. Together, these enable stable actor-critic training across 150+ diverse tasks with rollout horizons of 15 steps using a single fixed hyperparameter set.

\textbf{Multi-step and parallel prediction.}
An alternative to sequential one-step rollouts is to predict multiple future steps simultaneously. The Diffusion World Model (DWM)~\cite{ding2025dwm} replaces the autoregressive paradigm with a diffusion process that generates multi-step future states and rewards concurrently in a single forward pass, bypassing the recursive error chain entirely and achieving a 44\% performance improvement over one-step models on D4RL locomotion benchmarks. Diffusion Forcing~\cite{chen2024diffusionforcing} offers a theoretically motivated hybrid: it assigns independent noise levels to each token in a sequence during training, unifying the variable-length generation of autoregressive models with the holistic coherence of full-sequence diffusion. This enables stable rollouts well past the training horizon where standard autoregressive baselines diverge, because the model learns to denoise partial trajectories at multiple corruption levels simultaneously.

\textbf{Latent overshooting and consistency objectives.}
PlaNet~\cite{hafner2019planet} introduced \emph{latent overshooting}, a multi-step variational inference objective that regularizes the world model's latent predictions over extended horizons by comparing multi-step prior predictions with posterior-inferred states. While subsequent work in the Dreamer family found that latent overshooting may be unnecessary when actor-critic optimization is performed in imagination~\cite{schiewer2024hierarchicallimits}, the underlying principle---providing dense supervision beyond one-step prediction---has proven influential. EfficientZero~\cite{ye2021efficientzero} implements a related idea through a self-supervised temporal consistency loss that aligns multi-step model predictions with encoder representations of actual future observations, providing rich dynamics supervision independent of reward sparsity.

\textbf{Long-context architectures.}
The limited temporal memory of recurrent world models (RSSM, GRU) exacerbates compounding error, since information from distant past observations is progressively lost. Transformer-based world models~\cite{micheli2023transformers, zhang2023storm} address this through full-history attention but incur quadratic computational cost. State-space models offer a scalable alternative: Po et al.~\cite{po2025longcontext} demonstrated that linear RNN architectures with structured state-space parameterizations can capture long-range temporal dependencies at constant per-frame cost, outperforming transformer baselines on spatial retrieval tasks in Memory Maze and Minecraft. StateSpaceDiffuser~\cite{savov2025statespacediffuser} integrates a Mamba-based state-space model into a diffusion world model, where the SSM branch provides persistent temporal memory while the diffusion branch generates high-fidelity frames, maintaining coherent visual context for an order of magnitude more steps than diffusion-only baselines. EDELINE~\cite{lee2025edeline} similarly couples Mamba SSMs with diffusion dynamics, introducing dynamic loss harmonization to balance multi-task learning of reward and termination prediction over extended horizons.

\textbf{Hierarchical temporal abstraction.}
Hierarchical world models reduce compounding error by operating at multiple temporal resolutions, so that high-level predictions span longer horizons with fewer recursive steps. Director~\cite{hafner2022director} decomposes planning into a high-level ``manager'' that selects latent subgoals and a low-level ``worker'' that executes short-horizon motor commands, enabling successful navigation in 3D mazes that flat world models cannot solve. THICK~\cite{gumbsch2024thick} learns a hierarchy of world models with adaptive temporal abstractions from discrete latent dynamics, where lower levels update sparsely in time and higher levels predict only at context-change boundaries. Schiewer et al.~\cite{schiewer2024hierarchicallimits} provide a systematic evaluation of hierarchical MBRL, documenting both the benefits (reduced effective rollout depth) and the challenges (difficulty of learning appropriate abstraction boundaries, risk of model exploitation at each level).

Despite these advances, long-horizon consistency remains the single most significant technical bottleneck for world models. The fundamental tension is that extending the imagination horizon increases the planning capacity of the agent but simultaneously amplifies the risk of exploiting model inaccuracies. No current approach fully resolves this trade-off: controlled rollout horizons sacrifice long-term planning, multi-step prediction methods struggle with complex visual observations, and hierarchical models introduce additional learning challenges at each level. Developing world models that maintain calibrated predictions over hundreds or thousands of steps---matching the temporal horizons required for real-world planning in domains such as autonomous driving and medical treatment---remains a critical open problem.

\subsection*{Scalability and Computational Cost}

State-of-the-art world models span a computational spectrum that ranges from lightweight latent dynamics models trainable on a single GPU to foundation-scale systems requiring thousands of accelerators and millions of GPU-hours. GAIA-1~\cite{hu2023gaia1}, a 9-billion-parameter generative world model for autonomous driving, pairs a 6.5B autoregressive transformer with a 2.6B video diffusion decoder trained on 4,700 hours of real-world driving data. NVIDIA's Cosmos~\cite{nvidia2025cosmos} released open-weight world foundation models at 4B, 7B, 12B, and 14B parameters trained on over 20 million hours of video. DeepMind's Genie~\cite{bruce2024genie} scales to 11 billion parameters with a spatiotemporal video tokenizer and autoregressive dynamics model trained on unlabeled internet video. These scales make training prohibitively expensive for most research groups: the Open-Sora project~\cite{opensora2024} estimated that reproducing a commercial-quality video generation model costs approximately \$200K in compute, and even this represents a substantial reduction from the estimated costs of proprietary systems like Sora~\cite{liu2024sora}.

\textbf{Inference-time bottlenecks.}
For interactive applications---robotics, game simulation, autonomous driving---inference latency is as critical as training cost. Diffusion-based world models are particularly affected because each predicted frame requires multiple denoising iterations. DIAMOND~\cite{alonso2024diamond} demonstrated that the EDM framework~\cite{karras2022edm} enables stable autoregressive generation with as few as three denoising steps, but this still limits throughput compared to single-pass autoregressive models. GameNGen~\cite{valevski2025gamengen} achieved real-time DOOM simulation at 20 FPS on a single TPU by fine-tuning Stable Diffusion with four denoising steps, and further distillation reduced this to a single step at 50 FPS. However, scaling such approaches to higher-resolution, more complex environments remains challenging. Oasis~\cite{oasis2024} demonstrated real-time Minecraft-like world generation at 360p and 20 FPS using a Diffusion Transformer architecture on NVIDIA H100 GPUs, but resolution and visual fidelity remain limited compared to offline generation systems.

\textbf{Consistency and distillation approaches.}
Consistency models~\cite{song2023consistency} offer a principled framework for reducing the sampling cost of diffusion models by training networks that map any noise level directly to the clean data manifold in a single step. Latent consistency models~\cite{luo2023lcm} extended this to latent diffusion, achieving high-quality image generation in 2--4 steps with only 32 A100 GPU-hours of training. For video, DOLLAR~\cite{ding2025dollar} combines variational score distillation with consistency distillation for few-step video generation, achieving up to 278$\times$ acceleration over the teacher model's sampling procedure. Applying these distillation techniques specifically to world model inference---where the quality of each predicted frame directly affects downstream planning---is an active research direction, though the interaction between distillation-induced approximation errors and compounding rollout errors remains poorly understood.

\textbf{Token reduction and efficient architectures.}
A complementary strategy reduces computational cost by minimizing the number of tokens processed per timestep. $\Delta$-IRIS~\cite{micheli2024deltairis} encodes stochastic frame-to-frame deltas rather than full observations, dramatically reducing per-timestep token count while achieving state-of-the-art performance on Crafter with an order-of-magnitude training speedup over the original IRIS. STORM~\cite{zhang2023storm} fuses each observation into a single continuous latent token (versus IRIS's 16 discrete tokens per frame), enabling transformer-based world modeling with substantially lower attention cost and achieving 126.7\% mean human performance on Atari 100k with only 4.3 hours of training on a single GPU. These results suggest that architectural efficiency innovations can deliver disproportionate cost reductions without sacrificing downstream task performance, potentially democratizing world model research beyond well-resourced industrial laboratories.

\textbf{Architecture-level trade-offs.}
The choice among RSSM, transformer, diffusion, and state-space architectures involves fundamental trade-offs between computational cost, memory, and predictive capacity. RSSM-based models (Dreamer family) are the most computationally efficient, with $O(1)$ per-step cost independent of history length, but compress all temporal information into a fixed-size hidden state that limits long-range memory. Transformer-based models (IRIS, STORM) offer superior long-range dependency modeling but incur $O(T^2)$ attention cost in context length $T$, which becomes prohibitive for long episodes with high-resolution observations. State-space models~\cite{po2025longcontext} provide $O(T)$ complexity with strong long-range modeling, but their application to world modeling is still nascent. Diffusion models (DIAMOND, GameNGen) achieve the highest visual fidelity but require multiple sequential denoising steps per frame, making them the slowest at inference unless distillation is applied. No single architecture currently dominates across all evaluation dimensions, and the optimal choice depends critically on the deployment constraints of the target application.

\subsection*{Sim-to-Real Transfer and Domain Gap}

World models trained in simulation benefit from unlimited data, perfect state access, and safe exploration, but the resulting models often fail when deployed in the real world due to systematic discrepancies between simulated and physical environments. This \emph{sim-to-real gap} manifests along two largely orthogonal axes: the \emph{visual domain gap}, arising from differences in rendering quality, lighting, textures, and sensor characteristics; and the \emph{dynamics domain gap}, arising from inaccurate contact models, simplified friction and deformation physics, and unmodeled latencies and noise in real actuators and sensors.

\textbf{Domain randomization.}
The most established approach to bridging the visual domain gap is domain randomization, which trains models on a distribution of simulated environments with randomly varied visual and physical parameters, encouraging representations that are invariant to superficial variation. Tobin et al.~\cite{tobin2017domainrand} demonstrated that object detectors trained on randomized simulated images transfer to real-world perception without any real training data. Chen et al.~\cite{chen2022understandingdr} subsequently provided a theoretical framework establishing sharp bounds on the sim-to-real gap under domain randomization, proving that transfer can succeed under mild distributional coverage conditions without requiring real-world samples. In the world model setting, domain randomization has been applied to train both visual encoders and dynamics models, with the expectation that randomization forces the model to capture invariant physical structure rather than simulation-specific artifacts. However, domain randomization alone is often insufficient for the dynamics gap, as it randomizes parameters within a fixed physics engine whose structural assumptions (e.g., rigid-body contact models) may not match reality.

\textbf{Direct real-world learning.}
DayDreamer~\cite{wu2023daydreamer} demonstrated an alternative paradigm: training Dreamer-style world models directly on real-world interaction data without any simulation. A quadruped robot learned to walk in approximately one hour, and a robotic arm learned pick-and-place in ten minutes, using only real sensor data for both world model training and imagination-based policy optimization. This approach completely avoids the sim-to-real gap but is constrained by the cost and risk of real-world data collection, limiting its applicability to domains where hardware is accessible and failure consequences are manageable. Data augmentation techniques such as DrQ~\cite{yarats2021drq}, which applies random image shifts to improve sample efficiency in pixel-based RL, can partially compensate for the limited visual diversity of real-world datasets.

\textbf{Modular sim-to-real adaptation.}
A promising middle ground exploits the modular structure of world models to selectively adapt only the components most affected by domain shift. SimDist~\cite{levy2026simdist} pretrains the full world model stack (encoder, dynamics, reward, and value models) in simulation using privileged state information, then freezes all components except the dynamics model, which is fine-tuned on a small amount of real-world data via supervised learning. This \emph{simulation distillation} approach leverages the insight that the encoder and value representations learned in simulation are largely transferable, while the dynamics model---which must capture real contact physics, friction, and latency---requires adaptation. Notably, SimDist documented that naive end-to-end RL fine-tuning frequently causes catastrophic forgetting of simulation-acquired knowledge, motivating the modular freezing strategy. ReDRAW~\cite{lanier2025redraw} takes a complementary approach by learning latent-state dynamics residuals: a base world model pretrained in simulation is augmented with a lightweight residual correction network that captures the sim-to-real dynamics gap without modifying the base model's representations, demonstrated on both MuJoCo domains and a physical Duckiebot lane-following task.

\textbf{Teacher-student distillation.}
TWIST~\cite{yamada2023twist} introduces a teacher-student framework in which a teacher world model is trained with full state information in simulation, then supervises a student world model that receives only domain-randomized image observations. The student inherits the teacher's dynamics understanding while learning visual representations robust to domain shift. Simulation-Guided Fine-Tuning (SGFT)~\cite{yin2025sgft} uses simulation-learned value functions for reward shaping during real-world exploration, achieving at least 2$\times$ improvement in sample efficiency on real robot manipulation tasks compared to baselines that do not exploit simulation priors.

\textbf{Conditional world generation for domain transfer.}
At the foundation model scale, Cosmos-Transfer1~\cite{nvidia2025cosmostransfer} reframes the domain gap as a conditional generation problem: given simulated trajectories with semantic segmentation, depth maps, and edge inputs, a diffusion-based world model generates photorealistic video that preserves the scene structure and motion of the simulation while adopting realistic visual appearance. This approach has been demonstrated for both robotics and autonomous driving sim-to-real transfer, producing training data that narrows the visual domain gap without requiring real-world data collection for each new task. GenSim2~\cite{hua2024gensim2} complements this by using large language models to automatically generate diverse simulated tasks, assets, and scenarios, expanding coverage over the task variation that most often breaks sim-trained policies at deployment.

\textbf{Remaining challenges.}
Despite these advances, the sim-to-real gap remains one of the most practically significant obstacles in deploying world models. Contact-rich manipulation---involving friction, deformation, and multi-point contact---is particularly challenging because the physics engines underlying most simulators make simplifying assumptions (rigid-body contacts, Coulomb friction) that diverge significantly from real-world mechanics. Multi-modal sensory integration compounds the problem: world models that fuse vision with proprioception and tactile sensing must bridge domain gaps in each modality simultaneously. Furthermore, the real world is non-stationary in ways that simulation is not: lighting changes, object wear, and environmental drift mean that even a well-adapted world model may degrade over time without continual adaptation mechanisms.

\subsection*{Evaluation and Benchmarking}

No consensus exists on how to evaluate world models comprehensively, and the fragmentation of evaluation protocols across application domains---RL, autonomous driving, video generation, robotics---impedes systematic comparison of architectural innovations and obscures fundamental limitations of current approaches.

\textbf{The perceptual quality trap.}
The most widely adopted metrics---Fr\'{e}chet Inception Distance (FID), Fr\'{e}chet Video Distance (FVD), SSIM, PSNR, and LPIPS---measure visual fidelity but not decision-relevant properties. Ge et al.~\cite{ge2024contentbias} demonstrated that FVD is strongly biased toward per-frame appearance quality over temporal realism: one can halve FVD scores simply by selecting static, motion-free videos, exposing a fundamental disconnect between the metric and the temporal dynamics that world models are designed to capture. Luo et al.~\cite{luo2025beyondfvd} identified three additional critical FVD limitations---non-Gaussianity of I3D feature distributions, insensitivity to temporal distortions, and impractical sample-size requirements---and proposed JEDi (JEPA Embedding Distance), which requires only 16\% of the samples needed for reliable FVD estimation while better correlating with human temporal quality judgments. Liu et al.~\cite{liu2024fvmd} introduced Fr\'{e}chet Video Motion Distance (FVMD), which evaluates motion consistency separately from appearance and reveals cases where FVD and VBench give inconsistent assessments compared to human judgment, particularly for videos with intensive motion. These critiques collectively suggest that the field's reliance on FVD as a primary evaluation metric may be systematically misdirecting research toward perceptual quality at the expense of temporal and physical consistency.

\textbf{Physics and causal reasoning benchmarks.}
A growing family of benchmarks specifically targets the physical understanding of world models. WorldBench~\cite{upadhyay2026worldbench} provides concept-specific, disentangled physics evaluation using physically simulated ground-truth videos, testing intuitive physics properties (object permanence, scale consistency) and physical parameter estimation (friction, viscosity) in isolation. Its central finding is sobering: all tested state-of-the-art world models lack robust physical consistency, with foreground mIoU degrading sharply after 5--9 frames of autoregressive prediction. Physion~\cite{bear2022physion} evaluates physical prediction using realistic 3D simulations across eight physical scenarios, comparing model predictions against human performance and finding that object-centric representations are necessary but insufficient for human-level physical reasoning. Physion++~\cite{tung2023physionpp} extends this by requiring online inference of latent physical properties (mass, friction, elasticity), revealing that all current models fail to reach human-level prediction when property inference is required. IntPhys~\cite{riochet2022intphys} adapts the violation-of-expectation paradigm from developmental psychology, testing whether models can distinguish physically possible from impossible events involving object permanence and spatio-temporal continuity---with current models performing near chance. PhysBench~\cite{chow2025physbench} evaluates 75 vision-language models across over 10,000 entries and finds that even models excelling at commonsense reasoning struggle with physical world understanding.

\textbf{World model-specific benchmarks.}
Several recent benchmarks target world models specifically rather than video generation in general. WorldModelBench~\cite{li2025worldmodelbench} evaluates world modeling capabilities across 7 application domains and 56 subdomains with 67K human labels, assessing both instruction-following and physics-adherence dimensions. WorldSimBench~\cite{qin2025worldsimbench} introduces a dual evaluation framework with explicit perceptual evaluation and implicit manipulative evaluation (measuring video-action consistency) across open-ended environments, autonomous driving, and robot manipulation. WorldScore~\cite{duan2025worldscore} provides the first unified benchmark for world generation, evaluating 19+ models on controllability, quality, and dynamics through a decomposition into next-scene generation with camera trajectories. VBench~\cite{huang2024vbench} decomposes video generation quality into 16 hierarchical dimensions (subject consistency, motion smoothness, temporal flickering, spatial relationships, etc.), and VBench 2.0~\cite{huang2025vbench2} extends this to intrinsic faithfulness, evaluating physics plausibility, commonsense reasoning, and human anatomy consistency---arguing that prior benchmarks focus on visual convincingness rather than real-world adherence.

\textbf{The open-loop vs.\ closed-loop evaluation gap.}
In autonomous driving, the disconnect between open-loop evaluation (predicting future trajectories given recorded past) and closed-loop performance (executing in an interactive simulator) has been documented as a critical limitation. Li et al.~\cite{li2024egostatus} exposed that models achieve strong open-loop results on nuScenes by primarily relying on ego-vehicle status (velocity, heading) rather than perception, fundamentally undermining the validity of open-loop planning metrics. Bench2Drive~\cite{jia2024bench2drive} explicitly argues that open-loop L2 trajectory error does not reliably predict closed-loop driving ability, providing 2 million annotated frames across 44 interactive scenarios for proper closed-loop evaluation. NAVSIM~\cite{dauner2024navsim} bridges this gap using non-reactive simulation on real-world nuPlan data, hosting a CVPR 2024 competition that attracted 143 teams and 463 submissions. These findings have broader implications: any world model evaluated solely in open-loop mode may appear far more capable than it is in deployment, where each prediction influences future observations through the agent's actions.

\textbf{Statistical rigor.}
Agarwal et al.~\cite{agarwal2021precipice} demonstrated that standard RL evaluation practices---reporting mean or median scores across a small number of games and seeds---produce unreliable conclusions. Mean human-normalized scores are dominated by outlier games, while median scores discard too much information. They proposed performance profiles, stratified bootstrap confidence intervals, and the interquartile mean as robust aggregate metrics, releasing the \texttt{rliable} library for standardized statistical analysis. Despite receiving the NeurIPS 2021 Outstanding Paper Award and widespread citation, many world model papers continue to report only mean/median scores without confidence intervals, making published comparisons difficult to interpret and potentially misleading.

\textbf{Toward unified evaluation.}
The field would benefit from a unified evaluation protocol that measures a world model's core capabilities---dynamics prediction accuracy, temporal consistency, physical plausibility, action-conditioned controllability, and downstream task utility---in a domain-agnostic manner. Currently, an RL world model measured by HNS on Atari cannot be meaningfully compared with a driving world model measured by FVD on nuScenes or a robotics world model measured by manipulation success rate. Developing such a unified framework, potentially combining diagnostic physics probes~\cite{upadhyay2026worldbench}, perceptual quality decomposition~\cite{huang2024vbench}, and task-performance normalization~\cite{agarwal2021precipice}, represents one of the most impactful methodological contributions the field could pursue.

\subsection*{Data Efficiency and Representation Learning}

While world models improve data efficiency relative to model-free methods, they still require substantial data---especially in high-dimensional observation spaces. Self-supervised and foundation-model approaches (JEPA~\cite{lecun2022path}, V-JEPA~\cite{bardes2024vjepa}) address this by pretraining on large unlabeled datasets, but effectively transferring these representations to downstream tasks with limited fine-tuning data remains challenging.

\subsection*{Safety, Robustness, and Interpretability}

Deploying world models in safety-critical applications (autonomous driving, medical decision-making, robotics) requires that the model's predictions be reliable, calibrated, and interpretable. Overconfident predictions can lead to dangerous actions; underconfident predictions can lead to overly conservative behavior. Formal verification of learned world models and methods for quantifying and communicating predictive uncertainty to downstream decision-makers are essential for real-world deployment.

\subsection*{Multimodality and Grounding}

Real-world environments are inherently multimodal: visual, auditory, tactile, proprioceptive, and linguistic information are all relevant for building accurate world models. Current approaches predominantly focus on visual inputs, with limited integration of other modalities. Grounding world models in physical reality---ensuring that predictions respect physical laws, object permanence, and causal relationships---remains an open challenge.

\section{Discussion and Future Directions}
\label{sec:future}

The preceding sections have traced the evolution of world models from early symbolic frame-based representations to the current generation of billion-parameter neural simulators, highlighting advances in architectures, methodological paradigms, reasoning strategies, and an increasingly diverse range of application domains. In this final section, we step back from the technical details to consider a broader set of questions that, in our view, are likely to shape the future trajectory of the field. Rather than revisiting the specific challenges summarized in Section~\ref{sec:challenges}, we focus instead on cross-cutting themes, unresolved tensions, and emerging research directions that transcend individual application areas.

\subsection*{Toward a Unified Definition of World Models}

One of the most striking observations emerging from this survey is that there is still no universally accepted definition of what constitutes a world model. In model-based reinforcement learning, the term retains a relatively precise meaning, typically referring to a learned transition model coupled with a reward predictor that enables planning through imagination. Yet, the same label is now applied to video generation systems that produce photorealistic footage without any explicit notion of action or reward, to language models whose internal representations may or may not encode causal structure, and to domain-specific simulators in areas such as medicine and climate science, where design constraints differ substantially from those of game-playing agents.

This ambiguity is not merely terminological; it has important practical consequences for how research is organized, how models are evaluated, and how progress is assessed. A video generation model that performs well on FVD, for example, may be described as a “world model” despite lacking action-conditioned prediction, whereas a lightweight RSSM that supports effective planning in a low-dimensional latent space may be regarded as insufficiently general. In our view, the field would benefit from a clearer distinction between \emph{predictive world models}, which forecast future states conditioned on actions and support downstream decision-making, and \emph{generative world simulators}, which generate realistic sensory data without necessarily enabling interactive control. Both classes of systems are valuable, but conflating them obscures the distinct technical challenges they entail and makes meaningful cross-domain comparison more difficult.

\subsection*{The Convergence of World Models and Foundation Models}

Perhaps the most consequential trend identified in this survey is the ongoing convergence of world models and foundation models. Systems such as Cosmos~\cite{nvidia2025cosmos}, Genie~\cite{bruce2024genie}, and V-JEPA 2~\cite{meta2025vjepa2} are trained on internet-scale video corpora and are intended to function as general-purpose simulators that can be adapted to downstream tasks through fine-tuning or prompting. This trajectory closely parallels the evolution of language models from task-specific systems to general-purpose reasoning engines, and it raises a comparable set of conceptual and practical questions.

The potential advantages of this shift are substantial. In principle, a single pretrained world model could provide a shared representational substrate for robotics, autonomous driving, scientific modeling, and medical simulation, thereby reducing the need for domain-specific data collection and extensive task-specific engineering. At the same time, this promise must be interpreted with caution. Foundation world models are trained predominantly on internet video, which is heavily biased toward particular visual domains, such as indoor scenes, driving footage, and gaming content, while underrepresenting the physical interactions and contact dynamics that are especially critical for robotic manipulation. Whether broad visual pretraining can yield representations that transfer effectively to domains governed by fundamentally different physical principles, such as molecular dynamics, fluid simulation, or surgical planning, remains an open empirical question.

Moreover, the sheer scale of these models raises important concerns about accessibility. When training a competitive world model requires tens of thousands of GPU-hours and access to proprietary datasets, meaningful progress becomes concentrated within a small number of industrial laboratories. The open-source efforts reviewed in this paper, including Open-Sora, Oasis, and related projects, are encouraging developments, but the gap between open and closed systems remains substantial. Ensuring that world model research remains accessible to the broader academic community is therefore not only a matter of equity, but also a scientific imperative, since diversity of approaches, assumptions, and perspectives has historically been central to progress in machine learning.

\subsection*{Grounding, Causality, and the Limits of Prediction}

A recurring theme across the application domains reviewed in this survey is the tension between perceptual fidelity and physical grounding. Contemporary video-based world models can generate strikingly realistic imagery, yet they often violate basic physical principles: objects may interpenetrate, gravity may behave inconsistently, and conservation laws may be ignored. The evaluation literature reinforces this concern. For example, WorldBench~\cite{upadhyay2026worldbench} reports that state-of-the-art models achieve only 45

This limitation is consequential because the most impactful applications of world models, including robotics, autonomous driving, and medical decision-making, require predictions that are not only visually plausible but also physically reliable. A driving world model that occasionally generates phantom obstacles or fails to preserve vehicle momentum is of little value for safety-critical planning, regardless of how strong its FVD performance may be. Bridging this gap will likely require moving beyond purely data-driven approaches toward architectures that incorporate stronger physical inductive biases, whether through equivariant network designs, physics-informed loss functions, or hybrid neuro-symbolic systems that combine learned components with known governing equations.

A related and deeper question is whether world models can, or should, learn causal structure. The distinction between correlation and causation is foundational to scientific reasoning, because it determines whether a model can support genuine counterfactual inference, asking not only “what will happen next?” but also “what would have happened if a different action had been taken?” Current world models are typically trained on observational data and optimized for next-step prediction, which, in general, is insufficient for identifying causal relationships~\cite{pearl2009causality}. The extent to which causal structure can be recovered from sufficiently rich and diverse training data, and the architectural innovations required to support such reasoning, therefore remains an important open question with implications that extend far beyond any single application domain.

\subsection*{Multimodal Integration and Embodied Intelligence}

The world models surveyed in this paper operate primarily on visual inputs, occasionally augmented with language descriptions or proprioceptive signals. By contrast, biological organisms construct their internal models of the world from a far richer sensory repertoire, including touch, hearing, proprioception, vestibular feedback, and olfaction. The relative neglect of non-visual modalities in current research is partly a practical consequence of the fact that vision is the modality for which large-scale datasets and pretrained encoders are most readily available. However, it also points to a deeper conceptual limitation in how current world models are formulated.

For embodied agents acting in the physical world, tactile and proprioceptive feedback is often more informative than vision for contact-rich tasks such as grasping, insertion, and tool use. The limited number of systems that incorporate tactile sensing into world models~\cite{yang2023unisim} already illustrates the value of multimodal dynamics modeling. Nevertheless, systematic methods for integrating heterogeneous sensory streams—each characterized by different sampling rates, noise profiles, and information content—into a unified predictive framework remain at an early stage of development. Advancing this line of research will likely require not only architectural innovations, but also new datasets and evaluation benchmarks capable of capturing the richness of multimodal embodied interaction.

Language occupies a distinctive position within this landscape. As discussed in Section~\ref{sec:methodology}, language-augmented world models can ground abstract instructions in physical dynamics while also providing a natural interface for specifying goals, constraints, and task structure. The rapid development of vision-language-action models, such as RT-2, OpenVLA, and $\pi_0$, suggests that language may increasingly function as a universal task-specification layer for embodied world models. At the same time, current systems of this kind rely heavily on the implicit world knowledge encoded in pretrained language models, and it remains unclear whether such implicit knowledge can substitute for an explicit learned dynamics model when precise physical reasoning is required.

\subsection*{Safety, Reliability, and Deployment Considerations}

The deployment of world models in safety-critical domains—such as autonomous driving, surgical planning, and drug dosing—raises concerns that extend well beyond predictive accuracy. In these settings, the consequences of model failure can be severe and irreversible, rendering the standard machine learning workflow of training a model, evaluating it on a held-out test set, and then deploying it insufficient for real-world use.

Three issues warrant particular attention. First, \emph{calibration}: the uncertainty estimates produced by a world model must accurately reflect the true likelihood of error. Overconfident predictions may cause an autonomous vehicle to accelerate into a situation the model has mischaracterized, whereas underconfident predictions may render a surgical planning system so conservative that it becomes practically unusable. Achieving well-calibrated uncertainty quantification in high-dimensional, autoregressive prediction settings remains largely unsolved, and the interaction between calibration and compounding errors over long rollout horizons is still poorly understood.

Second, \emph{distributional robustness}: real-world environments are inherently non-stationary, and any world model trained on one distribution will inevitably encounter situations that differ from those seen during training. The long tail of rare events—such as unusual weather conditions, atypical patient presentations, or novel road geometries—is precisely where accurate prediction is most critical, yet it is also where data-driven models are most vulnerable to failure. Domain randomization, sim-to-real transfer, and continual learning offer partial remedies, but none yet provides the level of formal safety assurance required in high-stakes applications.

Third, \emph{interpretability}: when world model predictions inform consequential decisions, practitioners must be able to understand the basis of those predictions. The latent representations learned by current world models are often opaque, and although object-centric and structured representations offer some advantages for interpretability, they have not yet been demonstrated at the scale and complexity required for practical deployment. Developing methods that provide meaningful and trustworthy explanations of world model predictions—without substantially sacrificing predictive performance—therefore remains an important direction for future research.

\subsection*{Expanding the Application Frontier}

This survey has documented the application of world models to domains that, only a few years ago, would have seemed unlikely candidates, including educational measurement, clinical decision-making, cosmological simulation, and financial markets. These emerging applications are important not only because of their practical relevance, but also because of the theoretical questions they bring to the forefront.

Medical world models, for example, must operate under conditions of extreme data scarcity, strict privacy constraints, and demanding safety requirements. The four-level capability framework proposed by Qazi et al.~\cite{chen2025medicalwm}, which ranges from temporal prediction to counterfactual rollouts and ultimately autonomous planning, provides a useful lens for assessing progress. At present, however, most systems remain concentrated at the lower levels of this hierarchy. Reaching Level 4, in which a world model can autonomously recommend treatment strategies, will require advances not only in calibration, interpretability, and reliability, but also in regulatory acceptance and clinical trust, challenges that are not purely technical.

Educational world models present a different set of difficulties. In this context, the “environment” is a human learner whose internal state—including knowledge, misconceptions, and motivation—is only indirectly observable and whose dynamics are shaped by cognitive, social, and developmental processes that differ fundamentally from those of physical systems. Whether the architectural assumptions that work well in physical environments, such as Markovian state transitions, stationary dynamics, and dense reward signals, can be meaningfully transferred to educational settings remains an open question.

Financial world models face yet another class of challenges, including reflexivity, non-stationarity, and strategic interaction, all of which depart substantially from the physical settings in which world models were originally developed. The fact that deploying a model can alter the very environment it seeks to model~\cite{soros2013fallibility} introduces a fundamental complication for standard reinforcement learning formulations. This suggests that world models for strategic environments may require game-theoretic or multi-agent formulations that move beyond the traditional single-agent POMDP framework.

To summarize, these diverse applications point to a common lesson: the world model paradigm is sufficiently flexible to extend across a wide range of domains, but each domain imposes its own constraints that challenge the limits of existing methods. The most productive directions for future research will likely emerge from sustained interaction between these domain-specific demands and the general-purpose architectural innovations being developed across the broader world model community.

\subsection*{The Road Ahead}

We conclude with several speculative observations regarding the future direction of the field. The integration of latent reasoning with world models—as exemplified by Coconut~\cite{hao2024coconut}, LCDrive~\cite{tan2025lcdrive}, and FutureX~\cite{xiang2025futurex}—suggests that world models may evolve from passive simulators into active reasoning engines capable of supporting multi-step deliberation in latent space. If this trajectory continues, the distinction between “thinking” and “simulating” may become increasingly blurred, with world models serving as a shared substrate for both perception and cognition.

The development of hierarchical and compositional world models—systems that decompose complex environments into reusable components and reason across multiple levels of temporal and spatial abstraction—remains an important but still largely unrealized objective. Existing hierarchical approaches~\cite{hafner2022director, gumbsch2024thick} have demonstrated promising results in structured environments, but extending these ideas to open-ended real-world settings characterized by heterogeneous objects, uncertain interactions, and long-horizon task structure remains a major challenge.

Finally, the question of what it would mean for a world model to “understand” the world, rather than merely predict it, remains both philosophically significant and practically consequential. The gap between current systems and the robust, flexible, and causal world understanding exhibited even by relatively simple biological organisms remains substantial. Narrowing this gap will require not only larger models and more data, but also conceptual advances in how intelligence represents, learns, and reasons about the structure of physical and social reality. This survey has sought to map the current state of the field; its most important chapters are still to be explored.

\newpage
\bibliographystyle{unsrtnat}
\bibliography{ref}

@book{johnson1983mental,
  title={Mental Models: Towards a Cognitive Science of Language, Inference, and Consciousness},
  author={Johnson-Laird, Philip N.},
  year={1983},
  publisher={Harvard University Press}
}

@techreport{minsky1974framework,
  title={A Framework for Representing Knowledge},
  author={Minsky, Marvin},
  year={1974},
  institution={MIT AI Laboratory},
  number={Memo 306}
}

@book{moravec1988mind,
  title={Mind Children: The Future of Robot and Human Intelligence},
  author={Moravec, Hans},
  year={1988},
  publisher={Harvard University Press}
}

@book{sutton2018reinforcement,
  title={Reinforcement Learning: An Introduction},
  author={Sutton, Richard S. and Barto, Andrew G.},
  edition={2nd},
  year={2018},
  publisher={MIT Press}
}

@article{ha2018world,
  title={World Models},
  author={Ha, David and Schmidhuber, J{\"u}rgen},
  journal={arXiv preprint arXiv:1803.10122},
  year={2018},
  note={Also published as: Recurrent World Models Facilitate Policy Evolution, NeurIPS 2018}
}

@article{kaiser2020simpl,
  title={Model-based reinforcement learning for {Atari}},
  author={Kaiser, Lukasz and Babaeizadeh, Mohammad and Milos, Piotr and Osinski, Blazej and Campbell, Roy H. and Czechowski, Konrad and Erhan, Dumitru and Finn, Chelsea and Kozakowski, Piotr and Levine, Sergey and Mohiuddin, Afroz and Sepassi, Ryan and Tucker, George and Zoph, Henryk},
  journal={International Conference on Learning Representations},
  year={2020}
}

@article{hafner2019dream,
  title={Dream to Control: Learning Behaviors by Latent Imagination},
  author={Hafner, Danijar and Lillicrap, Timothy and Ba, Jimmy and Norouzi, Mohammad},
  journal={arXiv preprint arXiv:1912.01603},
  year={2019}
}

@inproceedings{hafner2020mastering,
  title={Mastering Atari with Discrete World Models},
  author={Hafner, Danijar and Lillicrap, Timothy and Norouzi, Mohammad and Ba, Jimmy},
  booktitle={International Conference on Learning Representations},
  year={2021}
}

@article{hafner2025mastering,
  title={Mastering Diverse Domains through World Models},
  author={Hafner, Danijar and Pasukonis, Jurgis and Ba, Jimmy and Lillicrap, Timothy},
  journal={Nature},
  year={2025}
}

@article{schrittwieser2020mastering,
  title={Mastering Atari, Go, Chess and Shogi by Planning with a Learned Model},
  author={Schrittwieser, Julian and Antonoglou, Ioannis and Hubert, Thomas and Simonyan, Karen and Sifre, Laurent and Schmitt, Simon and Guez, Arthur and Lockhart, Edward and Hassabis, Demis and Graepel, Thore and Lillicrap, Timothy and Silver, David},
  journal={Nature},
  volume={588},
  number={7839},
  pages={604--609},
  year={2020}
}

@article{lecun2022path,
  title={A Path Towards Autonomous Machine Intelligence},
  author={LeCun, Yann},
  journal={OpenReview preprint},
  year={2022},
  note={Version 0.9.2, 2022-06-27}
}

@misc{lecun_ami_2025,
  title={Advanced Machine Intelligence ({AMI}): Building {AI} Systems that Understand the Physical World},
  author={LeCun, Yann},
  year={2025},
  note={Announced November 2025. \url{https://www.advancedmachineintelligence.com}}
}

@article{openai2023gpt4,
  title={{GPT-4} Technical Report},
  author={{OpenAI}},
  journal={arXiv preprint arXiv:2303.08774},
  year={2023}
}

@article{zhong2025evaluation,
  title={Evaluation of {OpenAI} o1: Opportunities and Challenges of {AGI}},
  author={Zhong, Tianyang and Liu, Zhengliang and Pan, Yi and Zhang, Yutong and Zhang, Zeyu and Zhou, Yifan and Liang, Shizhe and Wu, Zihao and Lyu, Yanjun and Shu, Peng and others},
  journal={arXiv preprint arXiv:2409.18486},
  year={2025}
}

@article{hamrick2019analogues,
  title={Analogues of mental simulation and imagination in deep learning},
  author={Hamrick, Jessica B},
  journal={Current Opinion in Behavioral Sciences},
  volume={29},
  pages={8--16},
  year={2019},
  publisher={Elsevier}
}

@article{moerland2023model,
  title={Model-based reinforcement learning: A survey},
  author={Moerland, Thomas M and Broekens, Joost and Plaat, Aske and Jonker, Catholijn M},
  journal={Foundations and Trends{\textregistered} in Machine Learning},
  volume={16},
  number={1},
  pages={1--118},
  year={2023},
  publisher={Now Publishers, Inc.}
}

@inproceedings{sutton1990dyna,
  title={Integrated architectures for learning, planning, and reacting based on approximating dynamic programming},
  author={Sutton, Richard S},
  booktitle={Proceedings of the seventh international conference on machine learning},
  pages={216--224},
  year={1990}
}

@inproceedings{hansen2022tdmpc,
  title={Temporal difference learning for model predictive control},
  author={Hansen, Nicklas and Wang, Xiaolong and Su, Hao},
  booktitle={International Conference on Machine Learning},
  pages={8487--8506},
  year={2022},
  organization={PMLR}
}

@inproceedings{grimm2020value,
  title={The value equivalence principle for model-based reinforcement learning},
  author={Grimm, Christopher and Barreto, Andr{\'e} and Singh, Satinder and Silver, David},
  booktitle={Advances in Neural Information Processing Systems},
  volume={33},
  pages={5541--5552},
  year={2020}
}

@inproceedings{ye2021efficientzero,
  title={Mastering atari games with limited data},
  author={Ye, Weirui and Liu, Shaohuai and Kurutach, Thanard and Abbeel, Pieter and Gao, Yang},
  booktitle={Advances in Neural Information Processing Systems},
  volume={34},
  pages={25476--25488},
  year={2021}
}

@inproceedings{janner2019mbpo,
  title={When to trust your model: Model-based policy optimization},
  author={Janner, Michael and Fu, Justin and Zhang, Marvin and Levine, Sergey},
  booktitle={Advances in Neural Information Processing Systems},
  volume={32},
  year={2019}
}

@inproceedings{lambert2020objective,
  title={Objective mismatch in model-based reinforcement learning},
  author={Lambert, Nathan and Amos, Brandon and Yadan, Omry and Calandra, Roberto},
  booktitle={Learning for Dynamics and Control},
  pages={761--770},
  year={2020},
  organization={PMLR}
}

@inproceedings{schwarzer2020spr,
  title={Data-Efficient Reinforcement Learning with Self-Predictive Representations},
  author={Schwarzer, Max and Anand, Ankesh and Goel, Rishab and Hjelm, R Devon and Courville, Aaron and Bachman, Philip},
  booktitle={International Conference on Learning Representations},
  year={2021}
}

@book{pearl2009causality,
  title={Causality: Models, Reasoning, and Inference},
  author={Pearl, Judea},
  edition={2nd},
  year={2009},
  publisher={Cambridge University Press}
}

@book{pearl2018book,
  title={The Book of Why: The New Science of Cause and Effect},
  author={Pearl, Judea and Mackenzie, Dana},
  year={2018},
  publisher={Basic Books}
}

@inproceedings{buesing2019woulda,
  title={Woulda, coulda, shoulda: Counterfactually-guided policy search},
  author={Buesing, Lars and Weber, Theophane and Zwols, Yori and Racaniere, Sebastien and Guez, Arthur and Lespiau, Jean-Baptiste and Heess, Nicolas},
  booktitle={International Conference on Learning Representations},
  year={2019}
}

@article{schoelkopf2021toward,
  title={Toward causal representation learning},
  author={Sch{\"o}lkopf, Bernhard and Locatello, Francesco and Bauer, Stefan and Ke, Nan Rosemary and Kalchbrenner, Nal and Goyal, Anirudh and Bengio, Yoshua},
  journal={Proceedings of the IEEE},
  volume={109},
  number={5},
  pages={812--834},
  year={2021},
  publisher={IEEE}
}

@book{peters2017elements,
  title={Elements of Causal Inference: Foundations and Learning Algorithms},
  author={Peters, Jonas and Janzing, Dominik and Sch{\"o}lkopf, Bernhard},
  year={2017},
  publisher={The MIT Press}
}

@inproceedings{locatello2019challenging,
  title={Challenging common assumptions in the unsupervised learning of disentangled representations},
  author={Locatello, Francesco and Bauer, Stefan and Lucic, Mario and Raetsch, Gunnar and Gelly, Sylvain and Sch{\"o}lkopf, Bernhard},
  booktitle={International Conference on Machine Learning},
  pages={4114--4124},
  year={2019},
  organization={PMLR}
}

@inproceedings{dehaan2019causal,
  title={Causal confusion in imitation learning},
  author={de Haan, Pim and Dinesh, Dinesh and Levine, Sergey},
  booktitle={Advances in Neural Information Processing Systems},
  volume={32},
  year={2019}
}

@article{arjovsky2019irm,
  title={Invariant risk minimization},
  author={Arjovsky, Martin and Bottou, L{\'e}on and Gulrajani, Ishaan and Lopez-Paz, David},
  journal={arXiv preprint arXiv:1907.02893},
  year={2019}
}

@inproceedings{seitzer2021causal,
  title={Causal Influence Detection for Improving Efficiency in Reinforcement Learning},
  author={Seitzer, Maximilian and Sch{\"o}lkopf, Bernhard and Martius, Georg},
  booktitle={Advances in Neural Information Processing Systems},
  volume={34},
  pages={22905--22918},
  year={2021}
}

@inproceedings{thomas2017independently,
  title={Independently controllable factors},
  author={Thomas, Valentin and Bengio, Emmanuel and Fedus, William and Pondard, Jules and Beaudoin, Philippe and Larochelle, Hugo and Pineau, Joelle and Precup, Doina and Bengio, Yoshua},
  booktitle={Advances in Neural Information Processing Systems},
  volume={30},
  year={2017}
}

@article{lim2021temporal,
  title={Temporal Fusion Transformers for interpretable multi-horizon time series forecasting},
  author={Lim, Bryan and Ar{\i}k, Sercan {\"O} and Loeff, Nicolas and Pfister, Tomas},
  journal={International Journal of Forecasting},
  volume={37},
  number={4},
  pages={1748--1764},
  year={2021},
  publisher={Elsevier}
}

@article{hambly2023recent,
  title={Recent advances in reinforcement learning in finance},
  author={Hambly, Ben and Xu, Ruihai and Yang, Huining},
  journal={Mathematical Finance},
  volume={33},
  number={3},
  pages={437--503},
  year={2023},
  publisher={Wiley Online Library}
}

@inproceedings{liu2022finrl,
  title={{FinRL-Meta}: Market Environments and Benchmarks for Data-Driven Financial Reinforcement Learning},
  author={Liu, Xiao-Yang and Xia, Jingyang and Wang, Hongyang and Yang, Ziyi and others},
  booktitle={Thirty-sixth Conference on Neural Information Processing Systems Datasets and Benchmarks Track},
  year={2022}
}

@article{levine2020offline,
  title={Offline reinforcement learning: Tutorial, review, and perspectives on open problems},
  author={Levine, Sergey and Kumar, Aviral and Tucker, George and Fu, Justin},
  journal={arXiv preprint arXiv:2005.01643},
  year={2020}
}

@article{soros2013fallibility,
  title={Fallibility, reflexivity, and the human uncertainty principle},
  author={Soros, George},
  journal={Journal of Economic Methodology},
  volume={20},
  number={4},
  pages={309--329},
  year={2013},
  publisher={Taylor \& Francis}
}

@article{liu2024sora,
  title={Sora: A Review on Background, Technology, Limitations, and Opportunities of Large Vision Models},
  author={Liu, Yixin and Zhang, Kai and Li, Yuan and Yan, Zhiling and Gao, Chujie and Chen, Ruoxi and Yuan, Zhengqing and Huang, Yue and Sun, Hanchi and Gao, Jianfeng and He, Lifang and Sun, Lichao},
  journal={arXiv preprint arXiv:2402.17177},
  year={2024}
}

@article{assran2025vjepa2,
  title={{V-JEPA} 2: Self-Supervised Video Models Enable Understanding, Prediction and Planning},
  author={Assran, Mahmoud and others},
  journal={arXiv preprint},
  year={2025}
}

@article{bruce2024genie,
  title={Genie: Generative Interactive Environments},
  author={Bruce, Jake and Dennis, Michael and Edwards, Ashley and Parker-Holder, Jack and Shi, Yuge and Hughes, Edward and Lai, Matthew and Mavalankar, Aditi and Steiber, Richie and Apps, Chris and others},
  journal={arXiv preprint arXiv:2402.15391},
  year={2024}
}

@article{nvidia2025cosmos,
  title={Cosmos: World Foundation Model Platform for Physical {AI}},
  author={{NVIDIA}},
  journal={arXiv preprint},
  year={2025}
}

@article{ding2025understanding,
  title={Understanding World or Predicting Future? A Comprehensive Survey of World Models},
  author={Ding, Jingtao and Zhang, Yunke and Shang, Yu and Zhang, Yuheng and Zong, Zefang and Feng, Jie and Yuan, Yuan and Su, Hongyuan and Li, Nian and Piao, Jinghua and Deng, Yucheng and Sukiennik, Nicholas and Gao, Chen and Xu, Fengli and Li, Yong},
  journal={ACM Computing Surveys},
  volume={58},
  number={3},
  pages={1--38},
  year={2025},
  publisher={ACM},
  doi={10.1145/3746449}
}

@article{li2025embodied,
  title={A Comprehensive Survey on World Models for Embodied {AI}},
  author={Li, Xinqing and others},
  journal={arXiv preprint arXiv:2510.16732},
  year={2025}
}

@article{guan2024world,
  title={World Models for Autonomous Driving: An Initial Survey},
  author={Guan, Yanchen and Cui, Haicheng and others},
  journal={arXiv preprint arXiv:2403.02622},
  year={2024}
}

@article{li2025steprobot,
  title={A Step Toward World Models: A Survey on Robotic Manipulation},
  author={Li, Xuan and others},
  journal={arXiv preprint arXiv:2511.02097},
  year={2025}
}

@misc{worldbench2025survey,
  title={3D and 4D World Modeling: A Survey},
  author={Chen, Steven C. H. and others},
  year={2025},
  howpublished={\url{https://worldbench.github.io/survey}}
}

@misc{vaswani2023attentionneed,
      title={Attention Is All You Need}, 
      author={Ashish Vaswani and Noam Shazeer and Niki Parmar and Jakob Uszkoreit and Llion Jones and Aidan N. Gomez and Lukasz Kaiser and Illia Polosukhin},
      year={2023},
      eprint={1706.03762},
      archivePrefix={arXiv},
      primaryClass={cs.CL},
      url={https://arxiv.org/abs/1706.03762}, 
}

@inproceedings{wei2022chain,
  title={Chain-of-Thought Prompting Elicits Reasoning in Large Language Models},
  author={Wei, Jason and Wang, Xuezhi and Schuurmans, Dale and Bosma, Maarten
          and Ichter, Brian and Xia, Fei and Chi, Ed H. and Le, Quoc V.
          and Zhou, Denny},
  booktitle={Advances in Neural Information Processing Systems},
  volume={35},
  year={2022}
}

@article{hao2024coconut,
  title={Training Large Language Models to Reason in a Continuous Latent Space},
  author={Hao, Shibo and others},
  journal={arXiv preprint arXiv:2412.06769},
  year={2024}
}

@article{tan2025lcdrive,
  title={Latent Chain-of-Thought World Modeling for End-to-End Driving},
  author={Tan, Shuhan and Chitta, Kashyap and Chen, Yuxiao and Tian, Ran
          and You, Yurong and Wang, Yan and Luo, Wenjie and Cao, Yulong
          and Kr{\"a}henb{\"u}hl, Philipp and Pavone, Marco and Ivanovic, Boris},
  journal={arXiv preprint arXiv:2512.10226},
  year={2025}
}

@article{xiang2025futurex,
  title={FutureX: Enhance End-to-End Autonomous Driving with
         Chain-of-Thought Reasoning in Latent World Model},
  author={Xiang, Zhiyu and others},
  journal={arXiv preprint arXiv:2512.11226},
  year={2025}
}

@article{chen2025latentcotsurvey,
  title={Reasoning Beyond Language: A Comprehensive Survey on Latent
         Chain-of-Thought Reasoning},
  author={Chen, Xinghao and others},
  journal={arXiv preprint arXiv:2505.16782},
  year={2025}
}

@article{lin2022survey,
  title={A Survey of Transformers},
  author={Lin, Tianyang and Wang, Yuxin and Liu, Xiangyang and Qiu, Xipeng},
  journal={AI Open},
  volume={3},
  pages={111--132},
  year={2022}
}

@inproceedings{gu2022s4,
  title={Efficiently modeling long sequences with structured state spaces},
  author={Gu, Albert and Goel, Karan and R{\'e}, Christopher},
  booktitle={International Conference on Learning Representations},
  year={2022}
}

@article{gu2023mamba,
  title={Mamba: Linear-time sequence modeling with selective state spaces},
  author={Gu, Albert and Dao, Tri},
  journal={arXiv preprint arXiv:2312.00752},
  year={2023}
}

@inproceedings{hafner2019planet,
  title={Learning latent dynamics for planning from pixels},
  author={Hafner, Danijar and Lillicrap, Timothy and Fischer, Ian and Villegas, Ruben and Ha, David and Lee, Honglak and Davidson, James},
  booktitle={Proceedings of the 36th International Conference on Machine Learning},
  pages={2555--2565},
  year={2019},
  organization={PMLR}
}

@inproceedings{hafner2020dreamer,
  title={Dream to control: Learning behaviors by latent imagination},
  author={Hafner, Danijar and Lillicrap, Timothy and Ba, Jimmy and Norouzi, Mohammad},
  booktitle={International Conference on Learning Representations},
  year={2020}
}

@article{hafner2023dreamerv3,
  title={Mastering diverse domains through world models},
  author={Hafner, Danijar and Pasukonis, Jurgis and Ba, Jimmy and Lillicrap, Timothy},
  journal={Nature},
  year={2025},
  publisher={Nature Publishing Group},
  note={arXiv preprint arXiv:2301.04104, 2023}
}

@article{hochreiter1997lstm,
  title={Long short-term memory},
  author={Hochreiter, Sepp and Schmidhuber, J{\"u}rgen},
  journal={Neural Computation},
  volume={9},
  number={8},
  pages={1735--1780},
  year={1997}
}

@article{cho2014gru,
  title={Learning phrase representations using {RNN} encoder-decoder for statistical machine translation},
  author={Cho, Kyunghyun and van Merri{\"e}nboer, Bart and Gulcehre, Caglar and Bahdanau, Dzmitry and Bougares, Fethi and Schwenk, Holger and Bengio, Yoshua},
  journal={Proceedings of the Conference on Empirical Methods in Natural Language Processing},
  pages={1724--1734},
  year={2014}
}

@article{kingma2014vae,
  title={Auto-encoding variational {Bayes}},
  author={Kingma, Diederik P. and Welling, Max},
  journal={International Conference on Learning Representations},
  year={2014}
}

@article{vaswani2017attention,
  title={Attention is all you need},
  author={Vaswani, Ashish and Shazeer, Noam and Parmar, Niki and Uszkoreit, Jakob and Jones, Llion and Gomez, Aidan N. and Kaiser, {\L}ukasz and Polosukhin, Illia},
  journal={Advances in Neural Information Processing Systems},
  volume={30},
  year={2017}
}

@article{babaeizadeh2018sv2p,
  title={Stochastic variational video prediction},
  author={Babaeizadeh, Mohammad and Finn, Chelsea and Erhan, Dumitru and Campbell, Roy H. and Levine, Sergey},
  journal={International Conference on Learning Representations},
  year={2018}
}

@article{denton2018svg,
  title={Stochastic video generation with a learned prior},
  author={Denton, Emily and Fergus, Rob},
  journal={International Conference on Machine Learning},
  pages={1174--1183},
  year={2018}
}

@inproceedings{kim2020gamegan,
  title={Learning to simulate dynamic environments with {GameGAN}},
  author={Kim, Seung Wook and Zhou, Yuhao and Philion, Jonah and Torralba, Antonio and Fidler, Sanja},
  booktitle={Proceedings of the IEEE/CVF Conference on Computer Vision and Pattern Recognition},
  pages={1231--1240},
  year={2020}
}

@inproceedings{alonso2024diamond,
  title={Diffusion for world modeling: Visual details matter in {Atari}},
  author={Alonso, Eloi and Jelley, Adam and Micheli, Vincent and Kanervisto, Anssi and Storber, Amos and Vinyals, Oriol and Fleuret, Fran{\c{c}}ois},
  booktitle={Advances in Neural Information Processing Systems},
  year={2024},
  note={NeurIPS 2024 Spotlight}
}

@inproceedings{oprea2020review,
  title={A review on deep learning techniques for video prediction},
  author={Oprea, Sergiu and Martinez-Gonzalez, Pablo and Garcia-Garcia, Alberto and Castro-Vargas, John Alejandro and Orts-Escolano, Sergio and Garcia-Rodriguez, Jose and Argyros, Antonis},
  booktitle={IEEE Transactions on Pattern Analysis and Machine Intelligence},
  volume={44},
  pages={2806--2826},
  year={2022}
}

@article{openai2024sora,
  title={Video generation models as world simulators},
  author={{OpenAI}},
  journal={OpenAI Technical Report},
  year={2024}
}

@article{parkerholder2024genie2,
  title={Genie 2: A large-scale foundation world model},
  author={Parker-Holder, Jack and others},
  journal={Google DeepMind Technical Report},
  year={2024}
}

@article{kong2025_3d4d,
  title={3{D} and 4{D} world modeling: A survey},
  author={Kong, Lingdong and others},
  journal={arXiv preprint arXiv:2509.07996},
  year={2025}
}

@article{chen2025medicalwm,
  title={Beyond generative {AI}: World models for clinical prediction, counterfactuals, and planning},
  author={Chen, Irene Y. and others},
  journal={arXiv preprint arXiv:2511.16333},
  year={2025}
}

@article{assran2023ijepa,
  title={Self-supervised learning from images with a joint-embedding predictive architecture},
  author={Assran, Mahmoud and Duval, Quentin and Misra, Ishan and Bojanowski, Piotr and Vincent, Pascal and Rabbat, Michael and LeCun, Yann and Ballas, Nicolas},
  journal={Proceedings of the IEEE/CVF Conference on Computer Vision and Pattern Recognition},
  pages={15619--15629},
  year={2023}
}

@article{yang2025mewm,
  title={Medical World Model: Generative simulation of tumor evolution for treatment planning},
  author={Yang, Zuoyu and others},
  journal={arXiv preprint arXiv:2506.02327},
  year={2025}
}

@article{gao2025surgwm,
  title={Surgical vision world model},
  author={Gao, Yuan and others},
  journal={arXiv preprint},
  year={2025}
}

@article{unisurgwm2025,
  title={Unified Surgical World Model for structured understanding, long-horizon prediction, and fine-grained generation},
  author={{UniSWM Authors}},
  journal={ICLR 2026 Submission},
  year={2025}
}

@article{bardes2024vjepa,
  title={Revisiting feature prediction for learning visual representations from video},
  author={Bardes, Adrien and Garrido, Quentin and Ponce, Jean and Chen, Xinlei and Rabbat, Michael and LeCun, Yann and Assran, Mahmoud and Ballas, Nicolas},
  journal={arXiv preprint arXiv:2404.08471},
  year={2024}
}

@article{corbett1994knowledge,
  title={Knowledge tracing: Modeling the acquisition of procedural knowledge},
  author={Corbett, Albert T and Anderson, John R},
  journal={User modeling and user-adapted interaction},
  volume={4},
  number={4},
  pages={253--278},
  year={1994},
  publisher={Springer}
}

@book{lord2012applications,
  title={Applications of item response theory to practical testing problems},
  author={Lord, Frederic M},
  year={2012},
  publisher={Routledge}
}

@article{piech2015deep,
  title={Deep knowledge tracing},
  author={Piech, Chris and Bassen, Jonathan and Huang, Jonathan and Ganguli, Surya and Sahami, Mehran and Guibas, Leonidas J and Sohl-Dickstein, Jascha},
  journal={Advances in neural information processing systems},
  volume={28},
  year={2015}
}

@inproceedings{zhang2017dynamic,
  title={Dynamic key-value memory networks for knowledge tracing},
  author={Zhang, Jiani and Shi, Xingjian and King, Irwin and Yeung, Dit-Yan},
  booktitle={Proceedings of the 26th international conference on World Wide Web},
  pages={765--774},
  year={2017}
}

@article{pandey2019self,
  title={A self-attentive model for knowledge tracing},
  author={Pandey, Shalini and Karypis, George},
  journal={arXiv preprint arXiv:1907.06837},
  year={2019}
}

@inproceedings{choi2020towards,
  title={Towards an appropriate query, key, and value computation for knowledge tracing},
  author={Choi, Youngduck and Lee, Youngnam and Cho, Junghyun and Baek, Jineon and Kim, Byungsoo and Cha, Yeongmin and Shin, Dongmin and Bae, Chan and Heo, Jaewe},
  booktitle={Proceedings of the seventh ACM conference on learning@ scale},
  pages={341--344},
  year={2020}
}

@inproceedings{ghosh2020context,
  title={Context-aware attentive knowledge tracing},
  author={Ghosh, Aritra and Heffernan, Neil and Lan, Andrew S},
  booktitle={Proceedings of the 26th ACM SIGKDD international conference on knowledge discovery \& data mining},
  pages={2330--2339},
  year={2020}
}

@inproceedings{yin2023tracing,
  title={Tracing knowledge instead of patterns: Stable knowledge tracing with diagnostic transformer},
  author={Yin, Yu and Dai, Le and Huang, Zhenya and Shen, Shuanghong and Wang, Fei and Liu, Qi and Chen, Enhong and Li, Xin},
  booktitle={Proceedings of the ACM web conference 2023},
  pages={855--864},
  year={2023}
}

@article{ke2024hitskt,
  title={HiTSKT: A hierarchical transformer model for session-aware knowledge tracing},
  author={Ke, Fucai and Wang, Weiqing and Tan, Weicong and Du, Lan and Jin, Yuan and Huang, Yujin and Yin, Hongzhi},
  journal={Knowledge-Based Systems},
  volume={284},
  pages={111300},
  year={2024},
  publisher={Elsevier}
}

@article{lee2024language,
  title={Language model can do knowledge tracing: Simple but effective method to integrate language model and knowledge tracing task},
  author={Lee, Unggi and Bae, Jiyeong and Kim, Dohee and Lee, Sookbun and Park, Jaekwon and Ahn, Taekyung and Lee, Gunho and Stratton, Damji and Kim, Hyeoncheol},
  journal={arXiv preprint arXiv:2406.02893},
  year={2024}
}

@inproceedings{mansour2024can,
  title={Can large language models automatically score proficiency of written essays?},
  author={Mansour, Watheq Ahmad and Albatarni, Salam and Eltanbouly, Sohaila and Elsayed, Tamer},
  booktitle={Proceedings of the 2024 Joint International Conference on Computational Linguistics, Language Resources and Evaluation (LREC-COLING 2024)},
  pages={2777--2786},
  year={2024}
}

@article{pack2024large,
  title={Large language models and automated essay scoring of English language learner writing: Insights into validity and reliability},
  author={Pack, Austin and Barrett, Alex and Escalante, Juan},
  journal={Computers and Education: Artificial Intelligence},
  volume={6},
  pages={100234},
  year={2024},
  publisher={Elsevier}
}

@article{emirtekin2025large,
  title={Large language model-powered automated assessment: a systematic review},
  author={Emirtekin, Emrah},
  journal={Applied Sciences},
  volume={15},
  number={10},
  pages={5683},
  year={2025},
  publisher={MDPI}
}

@article{wang2025autoscore,
  title={AutoSCORE: Enhancing Automated Scoring with Multi-Agent Large Language Models via Structured Component Recognition},
  author={Wang, Yun and Ding, Zhaojun and Wu, Xuansheng and Sun, Siyue and Liu, Ninghao and Zhai, Xiaoming},
  journal={arXiv preprint arXiv:2509.21910},
  year={2025}
}

@inproceedings{ho2020ddpm,
  author    = {Ho, Jonathan and Jain, Ajay and Abbeel, Pieter},
  title     = {Denoising Diffusion Probabilistic Models},
  booktitle = {Advances in Neural Information Processing Systems},
  volume    = {33},
  pages     = {6840--6851},
  year      = {2020},
}

@inproceedings{hafner2019dreamerv1,
  author    = {Hafner, Danijar and Lillicrap, Timothy and Ba, Jimmy and Norouzi, Mohammad},
  title     = {Dream to Control: Learning Behaviors by Latent Imagination},
  booktitle = {International Conference on Learning Representations (ICLR)},
  year      = {2020},
}

@inproceedings{hafner2021dreamerv2,
  author    = {Hafner, Danijar and Lillicrap, Timothy and Norouzi, Mohammad and Ba, Jimmy},
  title     = {Mastering Atari with Discrete World Models},
  booktitle = {International Conference on Learning Representations (ICLR)},
  year      = {2021},
}

@article{hafner2025dreamerv3,
  author    = {Hafner, Danijar and Pasukonis, Jurgis and Ba, Jimmy and Lillicrap, Timothy},
  title     = {Mastering Diverse Control Tasks through World Models},
  journal   = {Nature},
  year      = {2025},
}

@inproceedings{valevski2024gamengen,
  author    = {Valevski, Dani and Leviathan, Yaniv and Arar, Moab and Fruchter, Shlomi},
  title     = {Diffusion Models Are Real-Time Game Engines},
  booktitle = {The Thirteenth International Conference on Learning Representations (ICLR)},
  year      = {2025},
  note      = {arXiv:2408.14837},
}

@inproceedings{yang2023unisim,
  author    = {Yang, Mengjiao and Du, Yilun and Ghasemipour, Kamyar and Tompson, Jonathan and Schuurmans, Dale and Abbeel, Pieter},
  title     = {Learning Interactive Real-World Simulators},
  booktitle = {The Twelfth International Conference on Learning Representations (ICLR)},
  year      = {2024},
  note      = {Outstanding Paper Award},
}

@inproceedings{rombach2022ldm,
  author    = {Rombach, Robin and Blattmann, Andreas and Lorenz, Dominik and Esser, Patrick and Ommer, Bj{\"o}rn},
  title     = {High-Resolution Image Synthesis with Latent Diffusion Models},
  booktitle = {Proceedings of the IEEE/CVF Conference on Computer Vision and Pattern Recognition (CVPR)},
  pages     = {10684--10695},
  year      = {2022},
}

@inproceedings{karras2022edm,
  author    = {Karras, Tero and Aittala, Miika and Aila, Timo and Laine, Samuli},
  title     = {Elucidating the Design Space of Diffusion-Based Generative Models},
  booktitle = {Advances in Neural Information Processing Systems},
  volume    = {35},
  pages     = {26565--26577},
  year      = {2022},
}

@inproceedings{janner2022diffuser,
  author    = {Janner, Michael and Du, Yilun and Tenenbaum, Joshua B. and Levine, Sergey},
  title     = {Planning with Diffusion for Flexible Behavior Synthesis},
  booktitle = {Proceedings of the 39th International Conference on Machine Learning (ICML)},
  series    = {Proceedings of Machine Learning Research},
  volume    = {162},
  pages     = {9902--9915},
  publisher = {PMLR},
  year      = {2022},
}

@inproceedings{ajay2023dd,
  author    = {Ajay, Anurag and Du, Yilun and Gupta, Abhi and Tenenbaum, Joshua B. and Jaakkola, Tommi S. and Agrawal, Pulkit},
  title     = {Is Conditional Generative Modeling All You Need for Decision-Making?},
  booktitle = {The Eleventh International Conference on Learning Representations (ICLR)},
  year      = {2023},
}

@inproceedings{micheli2023iris,
  author    = {Micheli, Vincent and Alonso, Eloi and Fleuret, Fran{\c{c}}ois},
  title     = {Transformers Are Sample-Efficient World Learners},
  booktitle = {The Eleventh International Conference on Learning Representations (ICLR)},
  year      = {2023},
}

@inproceedings{zhang2023storm,
  title={{STORM}: Efficient stochastic transformer based world models for reinforcement learning},
  author={Zhang, Weipu and others},
  booktitle={Advances in Neural Information Processing Systems},
  year={2023}
}

@inproceedings{peebles2023dit,
  author    = {Peebles, William and Xie, Saining},
  title     = {Scalable Diffusion Models with Transformers},
  booktitle = {Proceedings of the IEEE/CVF International Conference on Computer Vision (ICCV)},
  pages     = {4195--4205},
  year      = {2023},
}

@article{schrittwieser2020muzero,
  title={Mastering {Atari}, {Go}, chess and shogi by planning with a learned model},
  author={Schrittwieser, Julian and Antonoglou, Ioannis and Hubert, Thomas and Simonyan, Karen and Sifre, Laurent and Schmitt, Simon and Guez, Arthur and Lockhart, Edward and Hassabis, Demis and Graepel, Thore and Lillicrap, Timothy and Silver, David},
  journal={Nature},
  volume={588},
  number={7839},
  pages={604--609},
  year={2020},
  publisher={Nature Publishing Group}
}

@article{Zidan2026,
  title = {A Theoretical and Experimental Exploration in Permutation Randomization on Nonsmooth Nonconvex Optimization},
  ISSN = {2705-1064},
  url = {http://dx.doi.org/10.37256/cm.7220268464},
  DOI = {10.37256/cm.7220268464},
  journal = {Contemporary Mathematics},
  publisher = {Universal Wiser Publisher Pte. Ltd},
  author = {Zidan,  Arif Hassan and Jahin,  Afrar and Bao,  Yu and Liu,  Tianming and Zhang,  Wei},
  year = {2026},
  month = apr,
  pages = {2382–2403}
}

@article{meta2025vjepa2,
  title={{V-JEPA} 2: Self-supervised video models enable understanding, prediction and planning},
  author={{Meta AI}},
  journal={arXiv preprint arXiv:2506.09985},
  year={2025}
}

@article{puglisi2025brlp,
  title={Brain Latent Progression: Individual-based spatiotemporal disease progression on {3D} brain {MRI}s via latent diffusion},
  author={Puglisi, Lemuel and Alexander, Daniel C. and Rav{\`i}, Daniele},
  journal={Medical Image Analysis},
  year={2025}
}

@article{hu2023gaia1,
  author    = {Hu, Anthony and Russell, Lloyd and Yeo, Hudson and Murez, Zak and Fedoseev, George and Kendall, Alex and Sherwin, Jamie and Corrado, Pedro},
  title     = {{GAIA-1}: A Generative World Model for Autonomous Driving},
  journal   = {arXiv preprint arXiv:2309.17080},
  year      = {2023},
}

@inproceedings{wang2023drivedreamer,
  author    = {Wang, Xiaofeng and Zhu, Zheng and Huang, Guan and Chen, Xinze and Zhu, Jiagang and Lu, Jiwen},
  title     = {{DriveDreamer}: Towards Real-World-Driven World Models for Autonomous Driving},
  booktitle = {European Conference on Computer Vision (ECCV)},
  year      = {2024},
}

@inproceedings{zhao2024drivedreamer2,
  author    = {Zhao, Guosheng and Wang, Xiaofeng and Zhu, Zheng and Chen, Xinze and Huang, Guan and Bao, Xiaoyi and Wang, Xingang},
  title     = {{DriveDreamer-2}: {LLM}-Enhanced World Models for Diverse Driving Video Generation},
  booktitle = {Proceedings of the AAAI Conference on Artificial Intelligence},
  year      = {2025},
}

@inproceedings{zheng2024genad,
  author    = {Zheng, Lanqing and Li, Xinqing and others},
  title     = {{GenAD}: Generalized Predictive Model for Autonomous Driving},
  booktitle = {Proceedings of the IEEE/CVF Conference on Computer Vision and Pattern Recognition (CVPR)},
  year      = {2024},
}

@inproceedings{gao2024vista,
  author    = {Gao, Shenyuan and others},
  title     = {Vista: A Generalizable Driving World Model with High Fidelity and Versatile Controllability},
  booktitle = {Advances in Neural Information Processing Systems},
  volume    = {37},
  year      = {2024},
}

@article{wayve2025gaia2,
  author    = {{Wayve}},
  title     = {{GAIA-2}: A Controllable Multi-View Generative World Model for Autonomous Driving},
  journal   = {arXiv preprint arXiv:2503.20523},
  year      = {2025},
}

@inproceedings{zhang2024copilot4d,
  author    = {Zhang, Yunpeng and others},
  title     = {{Copilot4D}: Learning Unsupervised World Models for Autonomous Driving via Discrete Diffusion},
  booktitle = {The Twelfth International Conference on Learning Representations (ICLR)},
  year      = {2024},
}

@inproceedings{chi2023diffusionpolicy,
  author    = {Chi, Cheng and Feng, Siyuan and Du, Yilun and Xu, Zhenjia and Cousineau, Eric and Burchfiel, Benjamin and Song, Shuran},
  title     = {Diffusion Policy: Visuomotor Policy Learning via Action Diffusion},
  booktitle = {Robotics: Science and Systems (RSS)},
  year      = {2023},
}

@inproceedings{robodreamer2024,
  author    = {Zhou, Siyuan and Du, Yilun and Chen, Jiaben and Li, Yandong and Yeung, Dit-Yan and Gan, Chuang},
  title     = {{RoboDreamer}: Learning Compositional World Models for Robot Imagination},
  booktitle = {Proceedings of the 41st International Conference on Machine Learning (ICML)},
  series    = {Proceedings of Machine Learning Research},
  volume    = {235},
  pages     = {61885--61896},
  publisher = {PMLR},
  year      = {2024},
}

@article{drema2024,
  author    = {Barcellona, Leonardo and Borselli, Matteo and Sclafani, Francesco and Kambhampati, Vedant and others},
  title     = {Dream to Manipulate: Compositional World Models Empowering Robot Imitation Learning with Imagination},
  journal   = {arXiv preprint arXiv:2412.14957},
  year      = {2024},
}

@article{nvidia2025cosmos25,
  author    = {NVIDIA and Ali, Arslan and others},
  title     = {World Simulation with Video Foundation Models for Physical {AI}},
  journal   = {arXiv preprint arXiv:2511.00062},
  year      = {2025},
}

@inproceedings{che2025gamegenx,
  author    = {Che, Haoxuan and He, Xuanhua and Liu, Quande and Jin, Cheng and Chen, Hao},
  title     = {{GameGen-X}: Interactive Open-world Game Video Generation},
  booktitle = {The Thirteenth International Conference on Learning Representations (ICLR)},
  year      = {2025},
}

@inproceedings{wang2025occsora,
  author    = {Wang, Lening and Zheng, Wenzhao and others},
  title     = {{OccSora}: 4D Occupancy Generation Models as World Simulators for Autonomous Driving},
  booktitle = {The Thirteenth International Conference on Learning Representations (ICLR)},
  year      = {2025},
}

@inproceedings{yang2025cogvideox,
  author    = {Yang, Zhuoyi and Teng, Jiayan and Zheng, Wendi and Ding, Ming and Huang, Shiyu and Xu, Jiazheng and Yang, Yuanming and Hong, Wenyi and Zhang, Xiaohan and Feng, Guanyu and others},
  title     = {{CogVideoX}: Text-to-Video Diffusion Models with an Expert Transformer},
  booktitle = {The Thirteenth International Conference on Learning Representations (ICLR)},
  year      = {2025},
}

@inproceedings{kang2025howfar,
  author    = {Kang, Bingyi and Yue, Yang and Lu, Rui and Lin, Zhijie and Zhao, Yang and Wang, Kaixin and Huang, Gao and Feng, Jiashi},
  title     = {How Far is Video Generation from World Model: A Physical Law Perspective},
  booktitle = {Proceedings of the 42nd International Conference on Machine Learning (ICML)},
  year      = {2025},
}

@inproceedings{tesseract2025,
  author    = {Zhen, Haoyu and Sun, Qiao and Zhang, Hongxin and Li, Junyan and Zhou, Siyuan and Du, Yilun and Gan, Chuang},
  title     = {{TesserAct}: Learning 4D Embodied World Models},
  booktitle = {Proceedings of the IEEE/CVF International Conference on Computer Vision (ICCV)},
  year      = {2025},
  note      = {arXiv:2504.20995},
}

@inproceedings{he2025uwm,
  author    = {Zhu, Chuning and Yu, Raymond and Feng, Siyuan and Burchfiel, Benjamin and Shah, Paarth and Gupta, Abhishek},
  title     = {Unified World Models: Coupling Video and Action Diffusion for Pretraining on Large Robotic Datasets},
  booktitle = {Robotics: Science and Systems (RSS)},
  year      = {2025},
  note      = {arXiv:2504.02792},
}

@inproceedings{zhao2025drivedreamer4d,
  author    = {Zhao, Guosheng and Ni, Chaojun and Wang, Xiaofeng and Zhu, Zheng and Zhang, Xingle and Wang, Yida and Huang, Guan and Chen, Xinze and Wang, Boyuan and Zhang, Youyi and others},
  title     = {{DriveDreamer4D}: World Models Are Effective Data Machines for 4D Driving Scene Representation},
  booktitle = {Proceedings of the IEEE/CVF Conference on Computer Vision and Pattern Recognition (CVPR)},
  year      = {2025},
}

@inproceedings{irasim2025,
  author    = {Zhu, Fangqi and others},
  title     = {{IRASim}: Learning Interactive Real-Robot Action Simulators},
  booktitle = {Proceedings of the IEEE/CVF International Conference on Computer Vision (ICCV)},
  year      = {2025},
  note      = {arXiv:2406.14540},
}

@inproceedings{statespacediffuser2025,
  author    = {Savov, Nedko and Kazemi, Naser and Zhang, Deheng and Paudel, Danda Pani and Wang, Xi and Van Gool, Luc},
  title     = {{StateSpaceDiffuser}: Bringing Long Context to Diffusion World Models},
  booktitle = {Advances in Neural Information Processing Systems (NeurIPS)},
  year      = {2025},
  note      = {arXiv:2505.22246},
}

@article{chandra2025diwa,
  author    = {Chandra, Akshay L and Nematollahi, Iman and Huang, Chenguang and Welschehold, Tim and Burgard, Wolfram and Valada, Abhinav},
  title     = {{DiWA}: Diffusion Policy Adaptation with World Models},
  journal   = {arXiv preprint arXiv:2508.03645},
  year      = {2025},
}

@article{jiang2025world4rl,
  author    = {Jiang, Zhennan and Liu, Kai and Qin, Yuxin and Tian, Shuai and Zheng, Yupeng and Zhou, Mingcai and Yu, Chao and Li, Haoran and Zhao, Dongbin},
  title     = {{World4RL}: Diffusion World Models for Policy Refinement with Reinforcement Learning for Robotic Manipulation},
  journal   = {arXiv preprint arXiv:2509.19080},
  year      = {2025},
}

@misc{ding2025dwm,
      title={Diffusion World Model: Future Modeling Beyond Step-by-Step Rollout for Offline Reinforcement Learning}, 
      author={Zihan Ding and Amy Zhang and Yuandong Tian and Qinqing Zheng},
      year={2024},
      eprint={2402.03570},
      archivePrefix={arXiv},
      primaryClass={cs.LG},
      url={https://arxiv.org/abs/2402.03570}, 
}

@inproceedings{sun2024predicting,
  title={Predicting human brain states with transformer},
  author={Sun, Yifei and Cabezas, Mariano and Lee, Jiah and Wang, Chenyu and Zhang, Wei and Calamante, Fernando and Lv, Jinglei},
  booktitle={International Conference on Medical Image Computing and Computer-Assisted Intervention},
  pages={136--146},
  year={2024},
  organization={Springer}
}

@article{sun2025voxel,
  title={Voxel-level brain states prediction using swin transformer},
  author={Sun, Yifei and Chahine, Daniel and Wen, Qinghao and Liu, Tianming and Li, Xiang and Yuan, Yixuan and Calamante, Fernando and Lv, Jinglei},
  journal={IEEE Journal of Biomedical and Health Informatics},
  volume={29},
  number={12},
  pages={8719--8726},
  year={2025},
  publisher={IEEE}
}

@article{wang2025towards,
  title={Towards a general-purpose foundation model for fMRI analysis},
  author={Wang, Cheng and Jiang, Yu and Peng, Zhihao and Li, Chenxin and Bang, Changbae and Zhao, Lin and Lv, Jinglei and Sepulcre, Jorge and Yang, Carl and He, Lifang and others},
  journal={arXiv preprint arXiv:2506.11167},
  year={2025}
}

@article{zhou2023retfound,
  title={{RETFound}: A foundation model for generalizable disease detection from retinal images},
  author={Zhou, Yukun and Chia, Mark A. and Wagner, Siegfried K. and others},
  journal={Nature},
  volume={622},
  pages={156--163},
  year={2023}
}

@article{ding2025clarity,
  title={{CLARITY}: Medical world model for guiding treatment decisions by modeling context-aware disease trajectories in latent space},
  author={Ding, Tianxingjian and others},
  journal={arXiv preprint arXiv:2512.08029},
  year={2025}
}

@article{caro2023brainlm,
  title={BrainLM: A foundation model for brain activity recordings},
  author={Caro, Josue Ortega and Fonseca, Antonio H de O and Averill, Christopher and Rizvi, Syed A and Rosati, Matteo and Cross, James L and Mittal, Prateek and Zappala, Emanuele and Levine, Daniel and Dhodapkar, Rahul M and others},
  journal={BioRxiv},
  pages={2023--09},
  year={2023},
  publisher={Cold Spring Harbor Laboratory}
}

@article{d2025tribe,
  title={TRIBE: TRImodal Brain Encoder for whole-brain fMRI response prediction},
  author={d'Ascoli, St{\'e}phane and Rapin, J{\'e}r{\'e}my and Benchetrit, Yohann and Banville, Hubert and King, Jean-R{\'e}mi},
  journal={arXiv preprint arXiv:2507.22229},
  year={2025}
}

@inproceedings{jiang2024cardiaccopilot,
  title={Cardiac Copilot: Automatic probe guidance for echocardiography with world model},
  author={Jiang, Haojun and Sun, Zhenguo and Jia, Ning and Li, Meng and Sun, Yu and Luo, Shaqi and Song, Shiji and Huang, Gao},
  booktitle={Medical Image Computing and Computer Assisted Intervention -- MICCAI 2024},
  pages={190--199},
  year={2024},
  publisher={Springer}
}

@inproceedings{yue2025echoworld,
  title={{EchoWorld}: Learning motion-aware world models for echocardiography probe guidance},
  author={Yue, Yang and Wang, Yulin and Jiang, Haojun and Liu, Pan and Song, Shiji and Huang, Gao},
  booktitle={Proceedings of the IEEE/CVF Conference on Computer Vision and Pattern Recognition},
  pages={25993--26003},
  year={2025}
}

@article{xu2025meddreamer,
  title={med{D}reamer: Model-based reinforcement learning with latent imagination on complex {EHR}s for clinical decision support},
  author={Xu, Qianyi and others},
  journal={arXiv preprint arXiv:2505.19785},
  year={2025}
}

@inproceedings{greydanus2019hnn,
 author = {Greydanus, Samuel and Dzamba, Misko and Yosinski, Jason},
 booktitle = {Advances in Neural Information Processing Systems},
 editor = {H. Wallach and H. Larochelle and A. Beygelzimer and F. d\textquotesingle Alch\'{e}-Buc and E. Fox and R. Garnett},
 pages = {},
 publisher = {Curran Associates, Inc.},
 title = {Hamiltonian Neural Networks},
 url = {https://proceedings.neurips.cc/paper_files/paper/2019/file/26cd8ecadce0d4efd6cc8a8725cbd1f8-Paper.pdf},
 volume = {32},
 year = {2019}
}

@misc{cranmer2020lnn,
      title={Lagrangian Neural Networks}, 
      author={Miles Cranmer and Sam Greydanus and Stephan Hoyer and Peter Battaglia and David Spergel and Shirley Ho},
      year={2020},
      eprint={2003.04630},
      archivePrefix={arXiv},
      primaryClass={cs.LG},
      url={https://arxiv.org/abs/2003.04630}, 
}

@misc{lutter2019delan,
      title={Deep Lagrangian Networks: Using Physics as Model Prior for Deep Learning}, 
      author={Michael Lutter and Christian Ritter and Jan Peters},
      year={2019},
      eprint={1907.04490},
      archivePrefix={arXiv},
      primaryClass={cs.LG},
      url={https://arxiv.org/abs/1907.04490}, 
}

@inproceedings{
zhong2020symoden,
title={Symplectic ODE-Net: Learning Hamiltonian Dynamics with Control},
author={Yaofeng Desmond Zhong and Biswadip Dey and Amit Chakraborty},
booktitle={International Conference on Learning Representations},
year={2020},
url={https://openreview.net/forum?id=ryxmb1rKDS}
}

@inproceedings{finzi2020chnn,
 author = {Finzi, Marc and Wang, Ke Alexander and Wilson, Andrew G},
 booktitle = {Advances in Neural Information Processing Systems},
 editor = {H. Larochelle and M. Ranzato and R. Hadsell and M.F. Balcan and H. Lin},
 pages = {13880--13889},
 publisher = {Curran Associates, Inc.},
 title = {Simplifying Hamiltonian and Lagrangian Neural Networks via Explicit Constraints},
 url = {https://proceedings.neurips.cc/paper_files/paper/2020/file/9f655cc8884fda7ad6d8a6fb15cc001e-Paper.pdf},
 volume = {33},
 year = {2020}
}

@inproceedings{
toth2020hgn,
title={Hamiltonian Generative Networks},
author={Peter Toth and Danilo J. Rezende and Andrew Jaegle and Sébastien Racanière and Aleksandar Botev and Irina Higgins},
booktitle={International Conference on Learning Representations},
year={2020},
url={https://openreview.net/forum?id=HJenn6VFvB}
}

@inproceedings{chen2018neuralode,
 author = {Chen, Ricky T. Q. and Rubanova, Yulia and Bettencourt, Jesse and Duvenaud, David K},
 booktitle = {Advances in Neural Information Processing Systems},
 editor = {S. Bengio and H. Wallach and H. Larochelle and K. Grauman and N. Cesa-Bianchi and R. Garnett},
 pages = {},
 publisher = {Curran Associates, Inc.},
 title = {Neural Ordinary Differential Equations},
 url = {https://proceedings.neurips.cc/paper_files/paper/2018/file/69386f6bb1dfed68692a24c8686939b9-Paper.pdf},
 volume = {31},
 year = {2018}
}

@inproceedings{
li2020koopman,
title={Learning Compositional Koopman Operators for Model-Based Control},
author={Yunzhu Li and Hao He and Jiajun Wu and Dina Katabi and Antonio Torralba},
booktitle={International Conference on Learning Representations},
year={2020},
url={https://openreview.net/forum?id=H1ldzA4tPr}
}

@inproceedings{battaglia2016in,
 author = {Battaglia, Peter and Pascanu, Razvan and Lai, Matthew and Jimenez Rezende, Danilo and kavukcuoglu, koray},
 booktitle = {Advances in Neural Information Processing Systems},
 editor = {D. Lee and M. Sugiyama and U. Luxburg and I. Guyon and R. Garnett},
 pages = {},
 publisher = {Curran Associates, Inc.},
 title = {Interaction Networks for Learning about Objects, Relations and Physics},
 url = {https://proceedings.neurips.cc/paper_files/paper/2016/file/3147da8ab4a0437c15ef51a5cc7f2dc4-Paper.pdf},
 volume = {29},
 year = {2016}
}

@misc{battaglia2018relational,
      title={Relational inductive biases, deep learning, and graph networks}, 
      author={Peter W. Battaglia and Jessica B. Hamrick and Victor Bapst and Alvaro Sanchez-Gonzalez and Vinicius Zambaldi and Mateusz Malinowski and Andrea Tacchetti and David Raposo and Adam Santoro and Ryan Faulkner and Caglar Gulcehre and Francis Song and Andrew Ballard and Justin Gilmer and George Dahl and Ashish Vaswani and Kelsey Allen and Charles Nash and Victoria Langston and Chris Dyer and Nicolas Heess and Daan Wierstra and Pushmeet Kohli and Matt Botvinick and Oriol Vinyals and Yujia Li and Razvan Pascanu},
      year={2018},
      eprint={1806.01261},
      archivePrefix={arXiv},
      primaryClass={cs.LG},
      url={https://arxiv.org/abs/1806.01261}, 
}

@InProceedings{sanchez2020gns,
  title = 	 {Learning to Simulate Complex Physics with Graph Networks},
  author =       {Sanchez-Gonzalez, Alvaro and Godwin, Jonathan and Pfaff, Tobias and Ying, Rex and Leskovec, Jure and Battaglia, Peter},
  booktitle = 	 {Proceedings of the 37th International Conference on Machine Learning},
  pages = 	 {8459--8468},
  year = 	 {2020},
  editor = 	 {III, Hal Daumé and Singh, Aarti},
  volume = 	 {119},
  series = 	 {Proceedings of Machine Learning Research},
  month = 	 {13--18 Jul},
  publisher =    {PMLR},
  url = 	 {https://proceedings.mlr.press/v119/sanchez-gonzalez20a.html}
}

@inproceedings{
pfaff2021meshgraphnets,
title={Learning Mesh-Based Simulation with Graph Networks},
author={Tobias Pfaff and Meire Fortunato and Alvaro Sanchez-Gonzalez and Peter Battaglia},
booktitle={International Conference on Learning Representations},
year={2021},
url={https://openreview.net/forum?id=roNqYL0_XP}
}

@InProceedings{kipf2018nri,
  title = 	 {Neural Relational Inference for Interacting Systems},
  author =       {Kipf, Thomas and Fetaya, Ethan and Wang, Kuan-Chieh and Welling, Max and Zemel, Richard},
  booktitle = 	 {Proceedings of the 35th International Conference on Machine Learning},
  pages = 	 {2688--2697},
  year = 	 {2018},
  editor = 	 {Dy, Jennifer and Krause, Andreas},
  volume = 	 {80},
  series = 	 {Proceedings of Machine Learning Research},
  month = 	 {10--15 Jul},
  publisher =    {PMLR},
  url = 	 {https://proceedings.mlr.press/v80/kipf18a.html}
}

@misc{li2019dpinet,
      title={Learning Particle Dynamics for Manipulating Rigid Bodies, Deformable Objects, and Fluids}, 
      author={Yunzhu Li and Jiajun Wu and Russ Tedrake and Joshua B. Tenenbaum and Antonio Torralba},
      year={2019},
      eprint={1810.01566},
      archivePrefix={arXiv},
      primaryClass={cs.LG},
      url={https://arxiv.org/abs/1810.01566}, 
}

@InProceedings{li2020visualgrounding,
  title = 	 {Visual Grounding of Learned Physical Models},
  author =       {Li, Yunzhu and Lin, Toru and Yi, Kexin and Bear, Daniel and Yamins, Daniel and Wu, Jiajun and Tenenbaum, Joshua and Torralba, Antonio},
  booktitle = 	 {Proceedings of the 37th International Conference on Machine Learning},
  pages = 	 {5927--5936},
  year = 	 {2020},
  editor = 	 {III, Hal Daumé and Singh, Aarti},
  volume = 	 {119},
  series = 	 {Proceedings of Machine Learning Research},
  month = 	 {13--18 Jul},
  publisher =    {PMLR},
  url = 	 {https://proceedings.mlr.press/v119/li20j.html}
}

@inproceedings{locatello2020slotattention,
 author = {Locatello, Francesco and Weissenborn, Dirk and Unterthiner, Thomas and Mahendran, Aravindh and Heigold, Georg and Uszkoreit, Jakob and Dosovitskiy, Alexey and Kipf, Thomas},
 booktitle = {Advances in Neural Information Processing Systems},
 editor = {H. Larochelle and M. Ranzato and R. Hadsell and M.F. Balcan and H. Lin},
 pages = {11525--11538},
 publisher = {Curran Associates, Inc.},
 title = {Object-Centric Learning with Slot Attention},
 url = {https://proceedings.neurips.cc/paper_files/paper/2020/file/8511df98c02ab60aea1b2356c013bc0f-Paper.pdf},
 volume = {33},
 year = {2020}
}

@inproceedings{
kipf2022savi,
title={Conditional Object-Centric Learning from Video},
author={Thomas Kipf and Gamaleldin Fathy Elsayed and Aravindh Mahendran and Austin Stone and Sara Sabour and Georg Heigold and Rico Jonschkowski and Alexey Dosovitskiy and Klaus Greff},
booktitle={International Conference on Learning Representations},
year={2022},
url={https://openreview.net/forum?id=aD7uesX1GF_}
}

@inproceedings{
kipf2020cswm,
title={Contrastive Learning of Structured World Models},
author={Thomas Kipf and Elise van der Pol and Max Welling},
booktitle={International Conference on Learning Representations},
year={2020},
url={https://openreview.net/forum?id=H1gax6VtDB}
}

@inproceedings{
wu2023slotformer,
title={SlotFormer: Unsupervised Visual Dynamics Simulation with Object-Centric Models},
author={Ziyi Wu and Nikita Dvornik and Klaus Greff and Thomas Kipf and Animesh Garg},
booktitle={The Eleventh International Conference on Learning Representations },
year={2023},
url={https://openreview.net/forum?id=TFbwV6I0VLg}
}

@inproceedings{
ocrssm2023,
title={Learning to Compose: Improving Object Centric Learning by Injecting Compositionality},
author={Whie Jung and Jaehoon Yoo and Sungjin Ahn and Seunghoon Hong},
booktitle={The Twelfth International Conference on Learning Representations},
year={2024},
url={https://openreview.net/forum?id=HT2dAhh4uV}
}

@InProceedings{mosbach2025sold,
  title = 	 {{SOLD}: Slot Object-Centric Latent Dynamics Models for Relational Manipulation Learning from Pixels},
  author =       {Mosbach, Malte and Ewertz, Jan Niklas and Villar-Corrales, Angel and Behnke, Sven},
  booktitle = 	 {Proceedings of the 42nd International Conference on Machine Learning},
  pages = 	 {44911--44935},
  year = 	 {2025},
  editor = 	 {Singh, Aarti and Fazel, Maryam and Hsu, Daniel and Lacoste-Julien, Simon and Berkenkamp, Felix and Maharaj, Tegan and Wagstaff, Kiri and Zhu, Jerry},
  volume = 	 {267},
  series = 	 {Proceedings of Machine Learning Research},
  month = 	 {13--19 Jul},
  publisher =    {PMLR},
  url = 	 {https://proceedings.mlr.press/v267/mosbach25a.html}
}

@inproceedings{
engelcke2020genesis,
title={GENESIS: Generative Scene Inference and Sampling with Object-Centric Latent Representations},
author={Martin Engelcke and Adam R. Kosiorek and Oiwi Parker Jones and Ingmar Posner},
booktitle={International Conference on Learning Representations},
year={2020},
url={https://openreview.net/forum?id=BkxfaTVFwH}
}

@InProceedings{veerapaneni2020op3,
  title = 	 {Entity Abstraction in Visual Model-Based Reinforcement Learning},
  author =       {Veerapaneni, Rishi and Co-Reyes, John D. and Chang, Michael and Janner, Michael and Finn, Chelsea and Wu, Jiajun and Tenenbaum, Joshua and Levine, Sergey},
  booktitle = 	 {Proceedings of the Conference on Robot Learning},
  pages = 	 {1439--1456},
  year = 	 {2020},
  editor = 	 {Kaelbling, Leslie Pack and Kragic, Danica and Sugiura, Komei},
  volume = 	 {100},
  series = 	 {Proceedings of Machine Learning Research},
  month = 	 {30 Oct--01 Nov},
  publisher =    {PMLR},
  url = 	 {https://proceedings.mlr.press/v100/veerapaneni20a.html}
}

@inproceedings{
dreamweaver2025,
title={Dreamweaver: Learning Compositional World Models from Pixels},
author={Junyeob Baek and Yi-Fu Wu and Gautam Singh and Sungjin Ahn},
booktitle={The Thirteenth International Conference on Learning Representations},
year={2025},
url={https://openreview.net/forum?id=e5mTvjXG9u}
}

@article{stica2025, 
  title={Object-Centric World Models for Causality-Aware Reinforcement Learning}, 
  volume={40}, 
  url={https://ojs.aaai.org/index.php/AAAI/article/view/39642}, 
  DOI={10.1609/aaai.v40i29.39642}, 
  abstractNote={World models have been developed to support sample-efficient deep reinforcement learning agents. However, it remains challenging for world models to accurately replicate environments that are high-dimensional, non-stationary, and composed of multiple objects with rich interactions since most world models learn holistic representations of all environmental components. By contrast, humans perceive the environment by decomposing it into discrete objects, facilitating efficient decision-making. Motivated by this insight, we propose Slot Transformer Imagination with CAusality-aware reinforcement learning (STICA), a unified framework in which object-centric Transformers serve as the world model and causality-aware policy and value networks. STICA represents each observation as a set of object-centric tokens, together with tokens for the agent action and the resulting reward, enabling the world model to predict token-level dynamics and interactions. The policy and value networks then estimate token-level cause--effect relations and use them in the attention layers, yielding causality-guided decision-making. Experiments on object-rich benchmarks demonstrate that STICA consistently outperforms state-of-the-art agents in both sample efficiency and final performance.}, 
  number={29}, 
  journal={Proceedings of the AAAI Conference on Artificial Intelligence}, 
  author={Nishimoto, Yosuke and Matsubara, Takashi}, 
  year={2026}, 
  month={Mar.}, 
  pages={24585-24593} 
}

@inproceedings{
fiocwm2025,
title={Learning Interactive World Model for Object-Centric Reinforcement Learning},
author={Fan Feng and Phillip Lippe and Sara Magliacane},
booktitle={The Thirty-ninth Annual Conference on Neural Information Processing Systems},
year={2025},
url={https://openreview.net/forum?id=E0cjqfM55C}
}

@InProceedings{satorras2021egnn,
  title = 	 {E(n) Equivariant Graph Neural Networks},
  author =       {Satorras, V\'{\i}ctor Garcia and Hoogeboom, Emiel and Welling, Max},
  booktitle = 	 {Proceedings of the 38th International Conference on Machine Learning},
  pages = 	 {9323--9332},
  year = 	 {2021},
  editor = 	 {Meila, Marina and Zhang, Tong},
  volume = 	 {139},
  series = 	 {Proceedings of Machine Learning Research},
  month = 	 {18--24 Jul},
  publisher =    {PMLR},
  url = 	 {https://proceedings.mlr.press/v139/satorras21a.html}
}

@inproceedings{han2022sgnn,
 author = {Han, Jiaqi and Huang, Wenbing and Ma, Hengbo and Li, Jiachen and Tenenbaum, Josh and Gan, Chuang},
 booktitle = {Advances in Neural Information Processing Systems},
 editor = {S. Koyejo and S. Mohamed and A. Agarwal and D. Belgrave and K. Cho and A. Oh},
 pages = {26256--26268},
 publisher = {Curran Associates, Inc.},
 title = {Learning Physical Dynamics with Subequivariant Graph Neural Networks},
 url = {https://proceedings.neurips.cc/paper_files/paper/2022/file/a845fdc3f87751710218718adb634fe7-Paper-Conference.pdf},
 volume = {35},
 year = {2022}
}

@inproceedings{
segno2024,
title={{SEGNO}: Generalizing Equivariant Graph Neural Networks with Physical Inductive Biases},
author={Yang Liu and Jiashun Cheng and Haihong Zhao and Tingyang Xu and Peilin Zhao and Fugee Tsung and Jia Li and Yu Rong},
booktitle={The Twelfth International Conference on Learning Representations},
year={2024},
url={https://openreview.net/forum?id=3oTPsORaDH}
}

@inproceedings{
flowequivariant2025,
title={Flow Equivariant World Models: Structured Dynamics Outside the Field of View},
author={Hansen Lillemark and Benhao Huang and Fangneng Zhan and Yilun Du and T. Anderson Keller},
booktitle={NeurIPS 2025 Workshop on Space in Vision, Language, and Embodied AI},
year={2025},
url={https://openreview.net/forum?id=hq2nQ1K7E5}
}

@inproceedings{cranmer2020symbolic,
 author = {Cranmer, Miles and Sanchez Gonzalez, Alvaro and Battaglia, Peter and Xu, Rui and Cranmer, Kyle and Spergel, David and Ho, Shirley},
 booktitle = {Advances in Neural Information Processing Systems},
 editor = {H. Larochelle and M. Ranzato and R. Hadsell and M.F. Balcan and H. Lin},
 pages = {17429--17442},
 publisher = {Curran Associates, Inc.},
 title = {Discovering Symbolic Models from Deep Learning with Inductive Biases},
 url = {https://proceedings.neurips.cc/paper_files/paper/2020/file/c9f2f917078bd2db12f23c3b413d9cba-Paper.pdf},
 volume = {33},
 year = {2020}
}

@inproceedings{
agarwala2024cosmos,
title={Neurosymbolic Grounding for Compositional World Models},
author={Atharva Sehgal and Arya Grayeli and Jennifer J. Sun and Swarat Chaudhuri},
booktitle={The Twelfth International Conference on Learning Representations},
year={2024},
url={https://openreview.net/forum?id=4KZpDGD4Nh}
}

@InProceedings{ramesh2023pimbrl,
  title = 	 {Physics-Informed Model-Based Reinforcement Learning},
  author =       {Ramesh, Adithya and Ravindran, Balaraman},
  booktitle = 	 {Proceedings of The 5th Annual Learning for Dynamics and Control Conference},
  pages = 	 {26--37},
  year = 	 {2023},
  editor = 	 {Matni, Nikolai and Morari, Manfred and Pappas, George J.},
  volume = 	 {211},
  series = 	 {Proceedings of Machine Learning Research},
  month = 	 {15--16 Jun},
  publisher =    {PMLR},
  url = 	 {https://proceedings.mlr.press/v211/ramesh23a.html}
}

@inproceedings{
bear2022physion,
title={Physion: Evaluating Physical Prediction from Vision in Humans and Machines},
author={Daniel Bear and Elias Wang and Damian Mrowca and Felix Jedidja Binder and Hsiao-Yu Tung and RT Pramod and Cameron Holdaway and Sirui Tao and Kevin A. Smith and Fan-Yun Sun and Li Fei-Fei and Nancy Kanwisher and Joshua B. Tenenbaum and Daniel LK Yamins and Judith E Fan},
booktitle={Thirty-fifth Conference on Neural Information Processing Systems Datasets and Benchmarks Track (Round 1)},
year={2021},
url={https://openreview.net/forum?id=CXyZrKPz4CU}
}

@InProceedings{lin2024dynalang,
  title = 	 {Learning to Model the World With Language},
  author =       {Lin, Jessy and Du, Yuqing and Watkins, Olivia and Hafner, Danijar and Abbeel, Pieter and Klein, Dan and Dragan, Anca},
  booktitle = 	 {Proceedings of the 41st International Conference on Machine Learning},
  pages = 	 {29992--30017},
  year = 	 {2024},
  editor = 	 {Salakhutdinov, Ruslan and Kolter, Zico and Heller, Katherine and Weller, Adrian and Oliver, Nuria and Scarlett, Jonathan and Berkenkamp, Felix},
  volume = 	 {235},
  series = 	 {Proceedings of Machine Learning Research},
  month = 	 {21--27 Jul},
  publisher =    {PMLR},
  url = 	 {https://proceedings.mlr.press/v235/lin24g.html}
}

@INPROCEEDINGS{aljalbout2025limt,
  author={Aljalbout, Elie and Sotirakis, Nikolaos and van der Smagt, Patrick and Karl, Maximilian and Chen, Nutan},
  booktitle={2025 IEEE International Conference on Robotics and Automation (ICRA)}, 
  title={LIMT: Language-Informed Multi-Task Visual World Models}, 
  year={2025},
  volume={},
  number={},
  pages={8226-8233},
  doi={10.1109/ICRA55743.2025.11128817}}

@INPROCEEDINGS{nematollahi2025lumos,
  author={Nematollahi, Iman and DeMoss, Branton and Chandra, Akshay L and Hawes, Nick and Burgard, Wolfram and Posner, Ingmar},
  booktitle={2025 IEEE International Conference on Robotics and Automation (ICRA)}, 
  title={LUMOS: Language-Conditioned Imitation Learning with World Models}, 
  year={2025},
  volume={},
  number={},
  pages={8219-8225},
  doi={10.1109/ICRA55743.2025.11127988}}

@inproceedings{
ledwm2025,
title={Language-conditioned world model improves policy generalization by reading environmental descriptions},
author={Joe Nguyen and Stefan Lee},
booktitle={NeurIPS 2025 Workshop on Bridging Language, Agent, and World Models for Reasoning and Planning},
year={2025},
url={https://openreview.net/forum?id=ClrjcLTKBt}
}

@inproceedings{hao2023rap,
    title = "Reasoning with Language Model is Planning with World Model",
    author = "Hao, Shibo  and
      Gu, Yi  and
      Ma, Haodi  and
      Hong, Joshua  and
      Wang, Zhen  and
      Wang, Daisy  and
      Hu, Zhiting",
    editor = "Bouamor, Houda  and
      Pino, Juan  and
      Bali, Kalika",
    booktitle = "Proceedings of the 2023 Conference on Empirical Methods in Natural Language Processing",
    month = dec,
    year = "2023",
    address = "Singapore",
    publisher = "Association for Computational Linguistics",
    url = "https://aclanthology.org/2023.emnlp-main.507/",
    doi = "10.18653/v1/2023.emnlp-main.507",
    pages = "8154--8173"
}

@inproceedings{xie2025precondition,
    title = "Making Large Language Models into World Models with Precondition and Effect Knowledge",
    author = "Xie, Kaige  and
      Yang, Ian  and
      Gunerli, John  and
      Riedl, Mark",
    editor = "Rambow, Owen  and
      Wanner, Leo  and
      Apidianaki, Marianna  and
      Al-Khalifa, Hend  and
      Eugenio, Barbara Di  and
      Schockaert, Steven",
    booktitle = "Proceedings of the 31st International Conference on Computational Linguistics",
    month = jan,
    year = "2025",
    address = "Abu Dhabi, UAE",
    publisher = "Association for Computational Linguistics",
    url = "https://aclanthology.org/2025.coling-main.503/",
    pages = "7532--7545"
}

@inproceedings{wang2024bytesized32,
  title={Can language models serve as text-based world simulators?},
  author={Wang, Ruoyao and Todd, Graham and Xiao, Ziang and Yuan, Xingdi and C{\^o}t{\'e}, Marc-Alexandre and Clark, Peter and Jansen, Peter},
  booktitle={Proceedings of the 62nd Annual Meeting of the Association for Computational Linguistics (Volume 2: Short Papers)},
  pages={1--17},
  year={2024}
}

@InProceedings{carta2023glam,
  title = 	 {Grounding Large Language Models in Interactive Environments with Online Reinforcement Learning},
  author =       {Carta, Thomas and Romac, Cl\'{e}ment and Wolf, Thomas and Lamprier, Sylvain and Sigaud, Olivier and Oudeyer, Pierre-Yves},
  booktitle = 	 {Proceedings of the 40th International Conference on Machine Learning},
  pages = 	 {3676--3713},
  year = 	 {2023},
  editor = 	 {Krause, Andreas and Brunskill, Emma and Cho, Kyunghyun and Engelhardt, Barbara and Sabato, Sivan and Scarlett, Jonathan},
  volume = 	 {202},
  series = 	 {Proceedings of Machine Learning Research},
  month = 	 {23--29 Jul},
  publisher =    {PMLR},
  url = 	 {https://proceedings.mlr.press/v202/carta23a.html}
}

@inproceedings{
gkountouras2025causal,
title={Language Agents Meet Causality -- Bridging {LLM}s and Causal World Models},
author={John Gkountouras and Matthias Lindemann and Phillip Lippe and Efstratios Gavves and Ivan Titov},
booktitle={The Thirteenth International Conference on Learning Representations},
year={2025},
url={https://openreview.net/forum?id=y9A2TpaGsE}
}

@misc{liu2024lwm,
      title={World Model on Million-Length Video And Language With Blockwise RingAttention}, 
      author={Hao Liu and Wilson Yan and Matei Zaharia and Pieter Abbeel},
      year={2025},
      eprint={2402.08268},
      archivePrefix={arXiv},
      primaryClass={cs.LG},
      url={https://arxiv.org/abs/2402.08268}, 
}

@misc{fung2024vljepa,
      title={VL-JEPA: Joint Embedding Predictive Architecture for Vision-language}, 
      author={Delong Chen and Mustafa Shukor and Theo Moutakanni and Willy Chung and Jade Yu and Tejaswi Kasarla and Yejin Bang and Allen Bolourchi and Yann LeCun and Pascale Fung},
      year={2026},
      eprint={2512.10942},
      archivePrefix={arXiv},
      primaryClass={cs.CV},
      url={https://arxiv.org/abs/2512.10942}, 
}

@InProceedings{zhou2025dinowm,
  title = 	 {{DINO}-{WM}: World Models on Pre-trained Visual Features enable Zero-shot Planning},
  author =       {Zhou, Gaoyue and Pan, Hengkai and Lecun, Yann and Pinto, Lerrel},
  booktitle = 	 {Proceedings of the 42nd International Conference on Machine Learning},
  pages = 	 {79115--79135},
  year = 	 {2025},
  editor = 	 {Singh, Aarti and Fazel, Maryam and Hsu, Daniel and Lacoste-Julien, Simon and Berkenkamp, Felix and Maharaj, Tegan and Wagstaff, Kiri and Zhu, Jerry},
  volume = 	 {267},
  series = 	 {Proceedings of Machine Learning Research},
  month = 	 {13--19 Jul},
  publisher =    {PMLR},
  url = 	 {https://proceedings.mlr.press/v267/zhou25t.html}
}

@article{berg2025swm,
  title={Semantic world models},
  author={Berg, Jacob and Zhu, Chuning and Bao, Yanda and Durugkar, Ishan and Gupta, Abhishek},
  journal={arXiv preprint arXiv:2510.19818},
  year={2025}
}

@article{chen2025vlwm,
  title={Planning with reasoning using vision language world model},
  author={Chen, Delong and Moutakanni, Theo and Chung, Willy and Bang, Yejin and Ji, Ziwei and Bolourchi, Allen and Fung, Pascale},
  journal={arXiv preprint arXiv:2509.02722},
  year={2025}
}

@inproceedings{ge2024worldgpt,
author = {Ge, Zhiqi and Huang, Hongzhe and Zhou, Mingze and Li, Juncheng and Wang, Guoming and Tang, Siliang and Zhuang, Yueting},
title = {WorldGPT: Empowering LLM as Multimodal World Model},
year = {2024},
isbn = {9798400706868},
publisher = {Association for Computing Machinery},
address = {New York, NY, USA},
url = {https://doi.org/10.1145/3664647.3681488},
doi = {10.1145/3664647.3681488},
booktitle = {Proceedings of the 32nd ACM International Conference on Multimedia},
pages = {7346–7355},
numpages = {10},
location = {Melbourne VIC, Australia},
series = {MM '24}
}

@article{motus2024,
  title={Motus: A unified latent action world model},
  author={Bi, Hongzhe and Tan, Hengkai and Xie, Shenghao and Wang, Zeyuan and Huang, Shuhe and Liu, Haitian and Zhao, Ruowen and Feng, Yao and Xiang, Chendong and Rong, Yinze and others},
  journal={arXiv preprint arXiv:2512.13030},
  year={2025}
}

@misc{black2024susie,
      title={Zero-Shot Robotic Manipulation with Pretrained Image-Editing Diffusion Models}, 
      author={Kevin Black and Mitsuhiko Nakamoto and Pranav Atreya and Homer Walke and Chelsea Finn and Aviral Kumar and Sergey Levine},
      year={2023},
      eprint={2310.10639},
      archivePrefix={arXiv},
      primaryClass={cs.RO},
      url={https://arxiv.org/abs/2310.10639}, 
}

@article{ahn2022saycan,
  title={Do as i can, not as i say: Grounding language in robotic affordances},
  author={Ahn, Michael and Brohan, Anthony and Brown, Noah and Chebotar, Yevgen and Cortes, Omar and David, Byron and Finn, Chelsea and Fu, Chuyuan and Gopalakrishnan, Keerthana and Hausman, Karol and others},
  journal={arXiv preprint arXiv:2204.01691},
  year={2022}
}

@InProceedings{huang2023innermonologue,
  title = 	 {Inner Monologue: Embodied Reasoning through Planning with Language Models},
  author =       {Huang, Wenlong and Xia, Fei and Xiao, Ted and Chan, Harris and Liang, Jacky and Florence, Pete and Zeng, Andy and Tompson, Jonathan and Mordatch, Igor and Chebotar, Yevgen and Sermanet, Pierre and Jackson, Tomas and Brown, Noah and Luu, Linda and Levine, Sergey and Hausman, Karol and ichter, brian},
  booktitle = 	 {Proceedings of The 6th Conference on Robot Learning},
  pages = 	 {1769--1782},
  year = 	 {2023},
  editor = 	 {Liu, Karen and Kulic, Dana and Ichnowski, Jeff},
  volume = 	 {205},
  series = 	 {Proceedings of Machine Learning Research},
  month = 	 {14--18 Dec},
  publisher =    {PMLR},
  url = 	 {https://proceedings.mlr.press/v205/huang23c.html}
}

@inproceedings{
zeng2022socratic,
title={Socratic Models: Composing Zero-Shot Multimodal Reasoning with Language},
author={Andy Zeng and Maria Attarian and brian ichter and Krzysztof Marcin Choromanski and Adrian Wong and Stefan Welker and Federico Tombari and Aveek Purohit and Michael S Ryoo and Vikas Sindhwani and Johnny Lee and Vincent Vanhoucke and Pete Florence},
booktitle={The Eleventh International Conference on Learning Representations },
year={2023},
url={https://openreview.net/forum?id=G2Q2Mh3avow}
}

@INPROCEEDINGS{liang2023codepolicies,
  author={Liang, Jacky and Huang, Wenlong and Xia, Fei and Xu, Peng and Hausman, Karol and Ichter, Brian and Florence, Pete and Zeng, Andy},
  booktitle={2023 IEEE International Conference on Robotics and Automation (ICRA)}, 
  title={Code as Policies: Language Model Programs for Embodied Control}, 
  year={2023},
  volume={},
  number={},
  pages={9493-9500},
  doi={10.1109/ICRA48891.2023.10160591}}

@InProceedings{huang2023voxposer,
  title = 	 {VoxPoser: Composable 3D Value Maps for Robotic Manipulation with Language Models},
  author =       {Huang, Wenlong and Wang, Chen and Zhang, Ruohan and Li, Yunzhu and Wu, Jiajun and Fei-Fei, Li},
  booktitle = 	 {Proceedings of The 7th Conference on Robot Learning},
  pages = 	 {540--562},
  year = 	 {2023},
  editor = 	 {Tan, Jie and Toussaint, Marc and Darvish, Kourosh},
  volume = 	 {229},
  series = 	 {Proceedings of Machine Learning Research},
  month = 	 {06--09 Nov},
  publisher =    {PMLR},
  url = 	 {https://proceedings.mlr.press/v229/huang23b.html}
}

@inproceedings{
ma2024eureka,
title={Eureka: Human-Level Reward Design via Coding Large Language Models},
author={Yecheng Jason Ma and William Liang and Guanzhi Wang and De-An Huang and Osbert Bastani and Dinesh Jayaraman and Yuke Zhu and Linxi Fan and Anima Anandkumar},
booktitle={The Twelfth International Conference on Learning Representations},
year={2024},
url={https://openreview.net/forum?id=IEduRUO55F}
}

@inproceedings{fan2022minedojo,
 author = {Fan, Linxi and Wang, Guanzhi and Jiang, Yunfan and Mandlekar, Ajay and Yang, Yuncong and Zhu, Haoyi and Tang, Andrew and Huang, De-An and Zhu, Yuke and Anandkumar, Anima},
 booktitle = {Advances in Neural Information Processing Systems},
 editor = {S. Koyejo and S. Mohamed and A. Agarwal and D. Belgrave and K. Cho and A. Oh},
 pages = {18343--18362},
 publisher = {Curran Associates, Inc.},
 title = {MineDojo: Building Open-Ended Embodied Agents with Internet-Scale Knowledge},
 url = {https://proceedings.neurips.cc/paper_files/paper/2022/file/74a67268c5cc5910f64938cac4526a90-Paper-Datasets_and_Benchmarks.pdf},
 volume = {35},
 year = {2022}
}

@article{
wang2023voyager,
title={Voyager: An Open-Ended Embodied Agent with Large Language Models},
author={Guanzhi Wang and Yuqi Xie and Yunfan Jiang and Ajay Mandlekar and Chaowei Xiao and Yuke Zhu and Linxi Fan and Anima Anandkumar},
journal={Transactions on Machine Learning Research},
issn={2835-8856},
year={2024},
url={https://openreview.net/forum?id=ehfRiF0R3a},
note={}
}

@inproceedings{wang2023deps,
author = {Wang, Zihao and Cai, Shaofei and Chen, Guanzhou and Liu, Anji and Ma, Xiaojian and Liang, Yitao and CraftJarvis, Team},
title = {Describe, explain, plan and select: interactive planning with large language models enables open-world multi-task agents},
year = {2023},
publisher = {Curran Associates Inc.},
address = {Red Hook, NY, USA},
booktitle = {Proceedings of the 37th International Conference on Neural Information Processing Systems},
articleno = {1480},
numpages = {37},
location = {New Orleans, LA, USA},
series = {NIPS '23}
}

@ARTICLE{wang2023jarvis1,
  author={Wang, Zihao and Cai, Shaofei and Liu, Anji and Jin, Yonggang and Hou, Jinbing and Zhang, Bowei and Lin, Haowei and He, Zhaofeng and Zheng, Zilong and Yang, Yaodong and Ma, Xiaojian and Liang, Yitao},
  journal={IEEE Transactions on Pattern Analysis and Machine Intelligence}, 
  title={JARVIS-1: Open-World Multi-Task Agents With Memory-Augmented Multimodal Language Models}, 
  year={2025},
  volume={47},
  number={3},
  pages={1894-1907},
  doi={10.1109/TPAMI.2024.3511593}}

@inproceedings{lifshitz2023steve1,
 author = {Lifshitz, Shalev and Paster, Keiran and Chan, Harris and Ba, Jimmy and McIlraith, Sheila},
 booktitle = {Advances in Neural Information Processing Systems},
 editor = {A. Oh and T. Naumann and A. Globerson and K. Saenko and M. Hardt and S. Levine},
 pages = {69900--69929},
 publisher = {Curran Associates, Inc.},
 title = {STEVE-1: A Generative Model for Text-to-Behavior in Minecraft},
 url = {https://proceedings.neurips.cc/paper_files/paper/2023/file/dd03f856fc7f2efeec8b1c796284561d-Paper-Conference.pdf},
 volume = {36},
 year = {2023}
}

@InProceedings{driess2023palme,
  title = 	 {{P}a{LM}-E: An Embodied Multimodal Language Model},
  author =       {Driess, Danny and Xia, Fei and Sajjadi, Mehdi S. M. and Lynch, Corey and Chowdhery, Aakanksha and Ichter, Brian and Wahid, Ayzaan and Tompson, Jonathan and Vuong, Quan and Yu, Tianhe and Huang, Wenlong and Chebotar, Yevgen and Sermanet, Pierre and Duckworth, Daniel and Levine, Sergey and Vanhoucke, Vincent and Hausman, Karol and Toussaint, Marc and Greff, Klaus and Zeng, Andy and Mordatch, Igor and Florence, Pete},
  booktitle = 	 {Proceedings of the 40th International Conference on Machine Learning},
  pages = 	 {8469--8488},
  year = 	 {2023},
  editor = 	 {Krause, Andreas and Brunskill, Emma and Cho, Kyunghyun and Engelhardt, Barbara and Sabato, Sivan and Scarlett, Jonathan},
  volume = 	 {202},
  series = 	 {Proceedings of Machine Learning Research},
  month = 	 {23--29 Jul},
  publisher =    {PMLR},
  url = 	 {https://proceedings.mlr.press/v202/driess23a.html}
}

@article{brohan2023rt1,
  title={Rt-1: Robotics transformer for real-world control at scale},
  author={Brohan, Anthony and Brown, Noah and Carbajal, Justice and Chebotar, Yevgen and Dabis, Joseph and Finn, Chelsea and Gopalakrishnan, Keerthana and Hausman, Karol and Herzog, Alex and Hsu, Jasmine and others},
  journal={arXiv preprint arXiv:2212.06817},
  year={2022}
}

@InProceedings{zitkovich2023rt2,
  title = 	 {RT-2: Vision-Language-Action Models Transfer Web Knowledge to Robotic Control},
  author =       {Zitkovich, Brianna and Yu, Tianhe and Xu, Sichun and Xu, Peng and Xiao, Ted and Xia, Fei and Wu, Jialin and Wohlhart, Paul and Welker, Stefan and Wahid, Ayzaan and Vuong, Quan and Vanhoucke, Vincent and Tran, Huong and Soricut, Radu and Singh, Anikait and Singh, Jaspiar and Sermanet, Pierre and Sanketi, Pannag R. and Salazar, Grecia and Ryoo, Michael S. and Reymann, Krista and Rao, Kanishka and Pertsch, Karl and Mordatch, Igor and Michalewski, Henryk and Lu, Yao and Levine, Sergey and Lee, Lisa and Lee, Tsang-Wei Edward and Leal, Isabel and Kuang, Yuheng and Kalashnikov, Dmitry and Julian, Ryan and Joshi, Nikhil J. and Irpan, Alex and Ichter, Brian and Hsu, Jasmine and Herzog, Alexander and Hausman, Karol and Gopalakrishnan, Keerthana and Fu, Chuyuan and Florence, Pete and Finn, Chelsea and Dubey, Kumar Avinava and Driess, Danny and Ding, Tianli and Choromanski, Krzysztof Marcin and Chen, Xi and Chebotar, Yevgen and Carbajal, Justice and Brown, Noah and Brohan, Anthony and Arenas, Montserrat Gonzalez and Han, Kehang},
  booktitle = 	 {Proceedings of The 7th Conference on Robot Learning},
  pages = 	 {2165--2183},
  year = 	 {2023},
  editor = 	 {Tan, Jie and Toussaint, Marc and Darvish, Kourosh},
  volume = 	 {229},
  series = 	 {Proceedings of Machine Learning Research},
  month = 	 {06--09 Nov},
  publisher =    {PMLR},
  url = 	 {https://proceedings.mlr.press/v229/zitkovich23a.html}
}

@INPROCEEDINGS{openxembodiment2024,
  author={O’Neill, Abby and Rehman, Abdul and Maddukuri, Abhiram and Gupta, Abhishek and Padalkar, Abhishek and Lee, Abraham and Pooley, Acorn and Gupta, Agrim and Mandlekar, Ajay and Jain, Ajinkya and Tung, Albert and Bewley, Alex and Herzog, Alex and Irpan, Alex and Khazatsky, Alexander and Rai, Anant and Gupta, Anchit and Wang, Andrew and Singh, Anikait and Garg, Animesh and Kembhavi, Aniruddha and Xie, Annie and Brohan, Anthony and Raffin, Antonin and Sharma, Archit and Yavary, Arefeh and Jain, Arhan and Balakrishna, Ashwin and Wahid, Ayzaan and Burgess-Limerick, Ben and Kim, Beomjoon and Schölkopf, Bernhard and Wulfe, Blake and Ichter, Brian and Lu, Cewu and Xu, Charles and Le, Charlotte and Finn, Chelsea and Wang, Chen and Xu, Chenfeng and Chi, Cheng and Huang, Chenguang and Chan, Christine and Agia, Christopher and Pan, Chuer and Fu, Chuyuan and Devin, Coline and Xu, Danfei and Morton, Daniel and Driess, Danny and Chen, Daphne and Pathak, Deepak and Shah, Dhruv and Büchler, Dieter and Jayaraman, Dinesh and Kalashnikov, Dmitry and Sadigh, Dorsa and Johns, Edward and Foster, Ethan and Liu, Fangchen and Ceola, Federico and Xia, Fei and Zhao, Feiyu and Stulp, Freek and Zhou, Gaoyue and Sukhatme, Gaurav S. and Salhotra, Gautam and Yan, Ge and Feng, Gilbert and Schiavi, Giulio and Berseth, Glen and Kahn, Gregory and Wang, Guanzhi and Su, Hao and Fang, Hao-Shu and Shi, Haochen and Bao, Henghui and Ben Amor, Heni and Christensen, Henrik I and Furuta, Hiroki and Walke, Homer and Fang, Hongjie and Ha, Huy and Mordatch, Igor and Radosavovic, Ilija and Leal, Isabel and Liang, Jacky and Abou-Chakra, Jad and Kim, Jaehyung and Drake, Jaimyn and Peters, Jan and Schneider, Jan and Hsu, Jasmine and Bohg, Jeannette and Bingham, Jeffrey and Wu, Jeffrey and Gao, Jensen and Hu, Jiaheng and Wu, Jiajun and Wu, Jialin and Sun, Jiankai and Luo, Jianlan and Gu, Jiayuan and Tan, Jie and Oh, Jihoon and Wu, Jimmy and Lu, Jingpei and Yang, Jingyun and Malik, Jitendra and Silvério, João and Hejna, Joey and Booher, Jonathan and Tompson, Jonathan and Yang, Jonathan and Salvador, Jordi and Lim, Joseph J. and Han, Junhyek and Wang, Kaiyuan and Rao, Kanishka and Pertsch, Karl and Hausman, Karol and Go, Keegan and Gopalakrishnan, Keerthana and Goldberg, Ken and Byrne, Kendra and Oslund, Kenneth and Kawaharazuka, Kento and Black, Kevin and Lin, Kevin and Zhang, Kevin and Ehsani, Kiana and Lekkala, Kiran and Ellis, Kirsty and Rana, Krishan and Srinivasan, Krishnan and Fang, Kuan and Singh, Kunal Pratap and Zeng, Kuo-Hao and Hatch, Kyle and Hsu, Kyle and Itti, Laurent and Chen, Lawrence Yunliang and Pinto, Lerrel and Fei-Fei, Li and Tan, Liam and Fan, Linxi Jim and Ott, Lionel and Lee, Lisa and Weihs, Luca and Chen, Magnum and Lepert, Marion and Memmel, Marius and Tomizuka, Masayoshi and Itkina, Masha and Castro, Mateo Guaman and Spero, Max and Du, Maximilian and Ahn, Michael and Yip, Michael C. and Zhang, Mingtong and Ding, Mingyu and Heo, Minho and Srirama, Mohan Kumar and Sharma, Mohit and Kim, Moo Jin and Kanazawa, Naoaki and Hansen, Nicklas and Heess, Nicolas and Joshi, Nikhil J and Suenderhauf, Niko and Liu, Ning and Di Palo, Norman and Shafiullah, Nur Muhammad Mahi and Mees, Oier and Kroemer, Oliver and Bastani, Osbert and Sanketi, Pannag R and Miller, Patrick Tree and Yin, Patrick and Wohlhart, Paul and Xu, Peng and Fagan, Peter David and Mitrano, Peter and Sermanet, Pierre and Abbeel, Pieter and Sundaresan, Priya and Chen, Qiuyu and Vuong, Quan and Rafailov, Rafael and Tian, Ran and Doshi, Ria and Martín-Martín, Roberto and Baijal, Rohan and Scalise, Rosario and Hendrix, Rose and Lin, Roy and Qian, Runjia and Zhang, Ruohan and Mendonca, Russell and Shah, Rutav and Hoque, Ryan and Julian, Ryan and Bustamante, Samuel and Kirmani, Sean and Levine, Sergey and Lin, Shan and Moore, Sherry and Bahl, Shikhar and Dass, Shivin and Sonawani, Shubham and Song, Shuran and Xu, Sichun and Haldar, Siddhant and Karamcheti, Siddharth and Adebola, Simeon and Guist, Simon and Nasiriany, Soroush and Schaal, Stefan and Welker, Stefan and Tian, Stephen and Ramamoorthy, Subramanian and Dasari, Sudeep and Belkhale, Suneel and Park, Sungjae and Nair, Suraj and Mirchandani, Suvir and Osa, Takayuki and Gupta, Tanmay and Harada, Tatsuya and Matsushima, Tatsuya and Xiao, Ted and Kollar, Thomas and Yu, Tianhe and Ding, Tianli and Davchev, Todor and Zhao, Tony Z. and Armstrong, Travis and Darrell, Trevor and Chung, Trinity and Jain, Vidhi and Vanhoucke, Vincent and Zhan, Wei and Zhou, Wenxuan and Burgard, Wolfram and Chen, Xi and Wang, Xiaolong and Zhu, Xinghao and Geng, Xinyang and Liu, Xiyuan and Liangwei, Xu and Li, Xuanlin and Lu, Yao and Ma, Yecheng Jason and Kim, Yejin and Chebotar, Yevgen and Zhou, Yifan and Zhu, Yifeng and Wu, Yilin and Xu, Ying and Wang, Yixuan and Bisk, Yonatan and Cho, Yoonyoung and Lee, Youngwoon and Cui, Yuchen and Cao, Yue and Wu, Yueh-Hua and Tang, Yujin and Zhu, Yuke and Zhang, Yunchu and Jiang, Yunfan and Li, Yunshuang and Li, Yunzhu and Iwasawa, Yusuke and Matsuo, Yutaka and Ma, Zehan and Xu, Zhuo and Cui, Zichen Jeff and Zhang, Zichen and Lin, Zipeng},
  booktitle={2024 IEEE International Conference on Robotics and Automation (ICRA)}, 
  title={Open X-Embodiment: Robotic Learning Datasets and RT-X Models : Open X-Embodiment Collaboration0}, 
  year={2024},
  volume={},
  number={},
  pages={6892-6903},
  doi={10.1109/ICRA57147.2024.10611477}}

@article{ghosh2024octo,
  title={Octo: An open-source generalist robot policy},
  author={Team, Octo Model and Ghosh, Dibya and Walke, Homer and Pertsch, Karl and Black, Kevin and Mees, Oier and Dasari, Sudeep and Hejna, Joey and Kreiman, Tobias and Xu, Charles and others},
  journal={arXiv preprint arXiv:2405.12213},
  year={2024}
}

@InProceedings{kim2024openvla,
  title = 	 {OpenVLA: An Open-Source Vision-Language-Action Model},
  author =       {Kim, Moo Jin and Pertsch, Karl and Karamcheti, Siddharth and Xiao, Ted and Balakrishna, Ashwin and Nair, Suraj and Rafailov, Rafael and Foster, Ethan P and Sanketi, Pannag R and Vuong, Quan and Kollar, Thomas and Burchfiel, Benjamin and Tedrake, Russ and Sadigh, Dorsa and Levine, Sergey and Liang, Percy and Finn, Chelsea},
  booktitle = 	 {Proceedings of The 8th Conference on Robot Learning},
  pages = 	 {2679--2713},
  year = 	 {2025},
  editor = 	 {Agrawal, Pulkit and Kroemer, Oliver and Burgard, Wolfram},
  volume = 	 {270},
  series = 	 {Proceedings of Machine Learning Research},
  month = 	 {06--09 Nov},
  publisher =    {PMLR},
  url = 	 {https://proceedings.mlr.press/v270/kim25c.html}
}

@article{black2024pi0,
  title={$\pi_0$: A Vision-Language-Action Flow Model for General Robot Control},
  author={Black, Kevin and Brown, Noah and Driess, Danny and Esmail, Adnan and Equi, Michael and Finn, Chelsea and Fusai, Niccolo and Groom, Lachy and Hausman, Karol and Ichter, Brian and others},
  journal={arXiv preprint arXiv:2410.24164},
  year={2024}
}

@misc{
jiang2023vima,
title={{VIMA}: General Robot Manipulation with Multimodal Prompts},
author={Yunfan Jiang and Agrim Gupta and Zichen Zhang and Guanzhi Wang and Yongqiang Dou and Yanjun Chen and Li Fei-Fei and Anima Anandkumar and Yuke Zhu and Linxi Fan},
year={2023},
url={https://openreview.net/forum?id=hzjQWjPC04A}
}

@InProceedings{hong2024cogagent,
    author    = {Hong, Wenyi and Wang, Weihan and Lv, Qingsong and Xu, Jiazheng and Yu, Wenmeng and Ji, Junhui and Wang, Yan and Wang, Zihan and Dong, Yuxiao and Ding, Ming and Tang, Jie},
    title     = {CogAgent: A Visual Language Model for GUI Agents},
    booktitle = {Proceedings of the IEEE/CVF Conference on Computer Vision and Pattern Recognition (CVPR)},
    month     = {June},
    year      = {2024},
    pages     = {14281-14290}
}

@InProceedings{tan2024cradle,
  title = 	 {Cradle: Empowering Foundation Agents towards General Computer Control},
  author =       {Tan, Weihao and Zhang, Wentao and Xu, Xinrun and Xia, Haochong and Ding, Ziluo and Li, Boyu and Zhou, Bohan and Yue, Junpeng and Jiang, Jiechuan and Li, Yewen and An, Ruyi and Qin, Molei and Zong, Chuqiao and Zheng, Longtao and Wu, Yujie and Chai, Xiaoqiang and Bi, Yifei and Xie, Tianbao and Gu, Pengjie and Li, Xiyun and Zhang, Ceyao and Tian, Long and Wang, Chaojie and Wang, Xinrun and Karlsson, B\"{o}rje F. and An, Bo and Yan, Shuicheng and Lu, Zongqing},
  booktitle = 	 {Proceedings of the 42nd International Conference on Machine Learning},
  pages = 	 {58658--58725},
  year = 	 {2025},
  editor = 	 {Singh, Aarti and Fazel, Maryam and Hsu, Daniel and Lacoste-Julien, Simon and Berkenkamp, Felix and Maharaj, Tegan and Wagstaff, Kiri and Zhu, Jerry},
  volume = 	 {267},
  series = 	 {Proceedings of Machine Learning Research},
  month = 	 {13--19 Jul},
  publisher =    {PMLR},
  url = 	 {https://proceedings.mlr.press/v267/tan25h.html}
}

@inproceedings{
zhou2024webarena,
title={WebArena: A Realistic Web Environment for Building Autonomous Agents},
author={Shuyan Zhou and Frank F. Xu and Hao Zhu and Xuhui Zhou and Robert Lo and Abishek Sridhar and Xianyi Cheng and Tianyue Ou and Yonatan Bisk and Daniel Fried and Uri Alon and Graham Neubig},
booktitle={The Twelfth International Conference on Learning Representations},
year={2024},
url={https://openreview.net/forum?id=oKn9c6ytLx}
}

@inproceedings{koh2024visualwebarena,
    title = "{V}isual{W}eb{A}rena: Evaluating Multimodal Agents on Realistic Visual Web Tasks",
    author = "Koh, Jing Yu  and
      Lo, Robert  and
      Jang, Lawrence  and
      Duvvur, Vikram  and
      Lim, Ming  and
      Huang, Po-Yu  and
      Neubig, Graham  and
      Zhou, Shuyan  and
      Salakhutdinov, Russ  and
      Fried, Daniel",
    editor = "Ku, Lun-Wei  and
      Martins, Andre  and
      Srikumar, Vivek",
    booktitle = "Proceedings of the 62nd Annual Meeting of the Association for Computational Linguistics (Volume 1: Long Papers)",
    month = aug,
    year = "2024",
    address = "Bangkok, Thailand",
    publisher = "Association for Computational Linguistics",
    url = "https://aclanthology.org/2024.acl-long.50/",
    doi = "10.18653/v1/2024.acl-long.50",
    pages = "881--905"
}

@inproceedings{
liu2024agentbench,
title={AgentBench: Evaluating {LLM}s as Agents},
author={Xiao Liu and Hao Yu and Hanchen Zhang and Yifan Xu and Xuanyu Lei and Hanyu Lai and Yu Gu and Hangliang Ding and Kaiwen Men and Kejuan Yang and Shudan Zhang and Xiang Deng and Aohan Zeng and Zhengxiao Du and Chenhui Zhang and Sheng Shen and Tianjun Zhang and Yu Su and Huan Sun and Minlie Huang and Yuxiao Dong and Jie Tang},
booktitle={The Twelfth International Conference on Learning Representations},
year={2024},
url={https://openreview.net/forum?id=zAdUB0aCTQ}
}

@misc{deepmind2026genie3, 
  title={Genie 3 — Google DeepMind}, 
  url={https://deepmind.google/models/genie/}, 
  journal={Google DeepMind}, 
  author={Google DeepMind}, 
  year={2025},
  language={en} }

@InProceedings{shridhar2020alfred,
author = {Shridhar, Mohit and Thomason, Jesse and Gordon, Daniel and Bisk, Yonatan and Han, Winson and Mottaghi, Roozbeh and Zettlemoyer, Luke and Fox, Dieter},
title = {ALFRED: A Benchmark for Interpreting Grounded Instructions for Everyday Tasks},
booktitle = {Proceedings of the IEEE/CVF Conference on Computer Vision and Pattern Recognition (CVPR)},
month = {June},
year = {2020}
}

@ARTICLE{mees2022calvin,
  author={Mees, Oier and Hermann, Lukas and Rosete-Beas, Erick and Burgard, Wolfram},
  journal={IEEE Robotics and Automation Letters}, 
  title={CALVIN: A Benchmark for Language-Conditioned Policy Learning for Long-Horizon Robot Manipulation Tasks}, 
  year={2022},
  volume={7},
  number={3},
  pages={7327-7334},
  doi={10.1109/LRA.2022.3180108}}

@ARTICLE{lynch2023languagetable,
  author={Lynch, Corey and Wahid, Ayzaan and Tompson, Jonathan and Ding, Tianli and Betker, James and Baruch, Robert and Armstrong, Travis and Florence, Pete},
  journal={IEEE Robotics and Automation Letters}, 
  title={Interactive Language: Talking to Robots in Real Time}, 
  year={2023},
  volume={},
  number={},
  pages={1-8},
  doi={10.1109/LRA.2023.3295255}}

@article{gao2023magicdrive,
  title={Magicdrive: Street view generation with diverse 3d geometry control},
  author={Gao, Ruiyuan and Chen, Kai and Xie, Enze and Hong, Lanqing and Li, Zhenguo and Yeung, Dit-Yan and Xu, Qiang},
  journal={arXiv preprint arXiv:2310.02601},
  year={2023}
}

@article{min2023uniworld,
  title={Uniworld: Autonomous driving pre-training via world models},
  author={Min, Chen and Zhao, Dawei and Xiao, Liang and Nie, Yiming and Dai, Bin},
  journal={arXiv preprint arXiv:2308.07234},
  year={2023}
}

@inproceedings{jia2023think,
  title={Think twice before driving: Towards scalable decoders for end-to-end autonomous driving},
  author={Jia, Xiaosong and Wu, Penghao and Chen, Li and Xie, Jiangwei and He, Conghui and Yan, Junchi and Li, Hongyang},
  booktitle={Proceedings of the IEEE/CVF Conference on Computer Vision and Pattern Recognition},
  pages={21983--21994},
  year={2023}
}

@inproceedings{caesar2020nuscenes,
  title={nuscenes: A multimodal dataset for autonomous driving},
  author={Caesar, Holger and Bankiti, Varun and Lang, Alex H and Vora, Sourabh and Liong, Venice Erin and Xu, Qiang and Krishnan, Anush and Pan, Yu and Baldan, Giancarlo and Beijbom, Oscar},
  booktitle={Proceedings of the IEEE/CVF conference on computer vision and pattern recognition},
  pages={11621--11631},
  year={2020}
}

@inproceedings{sun2020scalability,
  title={Scalability in perception for autonomous driving: Waymo open dataset},
  author={Sun, Pei and Kretzschmar, Henrik and Dotiwalla, Xerxes and Chouard, Aurelien and Patnaik, Vijaysai and Tsui, Paul and Guo, James and Zhou, Yin and Chai, Yuning and Caine, Benjamin and others},
  booktitle={Proceedings of the IEEE/CVF conference on computer vision and pattern recognition},
  pages={2446--2454},
  year={2020}
}

@inproceedings{karnchanachari2024towards,
  title={Towards learning-based planning: The nuplan benchmark for real-world autonomous driving},
  author={Karnchanachari, Napat and Geromichalos, Dimitris and Tan, Kok Seang and Li, Nanxiang and Eriksen, Christopher and Yaghoubi, Shakiba and Mehdipour, Noushin and Bernasconi, Gianmarco and Fong, Whye Kit and Guo, Yiluan and others},
  booktitle={2024 IEEE International Conference on Robotics and Automation (ICRA)},
  pages={629--636},
  year={2024},
  organization={IEEE}
}

@article{gu2024review,
  title={A review of safe reinforcement learning: Methods, theories, and applications},
  author={Gu, Shangding and Yang, Long and Du, Yali and Chen, Guang and Walter, Florian and Wang, Jun and Knoll, Alois},
  journal={IEEE Transactions on Pattern Analysis and Machine Intelligence},
  volume={46},
  number={12},
  pages={11216--11235},
  year={2024},
  publisher={IEEE}
}

@inproceedings{zheng2025deepresearcher,
  title={Deepresearcher: Scaling deep research via reinforcement learning in real-world environments},
  author={Zheng, Yuxiang and Fu, Dayuan and Hu, Xiangkun and Cai, Xiaojie and Ye, Lyumanshan and Lu, Pengrui and Liu, Pengfei},
  booktitle={Proceedings of the 2025 Conference on Empirical Methods in Natural Language Processing},
  pages={414--431},
  year={2025}
}

@article{dong2026learning,
author = {Qi Lyu  and Jiahua Dong  and Baichen Liu  and Xudong Wang  and Wenqi Liang  and Duzhen Zhang  and Jiahang Tu  and Hongliu Li  and Hanbin Zhao  and Henghui Ding  and Yulun Zhang  and Zhi Han  and Nicu Sebe  and Fahad Shahbaz Khan  and Salman Khan  and Mubarak Shah  and Philip Torr  and Ming-Hsuan Yang  and Dacheng Tao },
title = {Learning to Model the World: A Survey of World Models in Artificial Intelligence},
journal = {TechRxiv},
volume = {2026},
number = {0305},
pages = {},
year = {2026},
doi = {10.36227/techrxiv.177274570.09578608/v1},
URL = {https://www.techrxiv.org/doi/abs/10.36227/techrxiv.177274570.09578608/v1},
eprint = {https://www.techrxiv.org/doi/pdf/10.36227/techrxiv.177274570.09578608/v1},}

@inproceedings{zuo2025gaussianworld,
  title={Gaussianworld: Gaussian world model for streaming 3d occupancy prediction},
  author={Zuo, Sicheng and Zheng, Wenzhao and Huang, Yuanhui and Zhou, Jie and Lu, Jiwen},
  booktitle={Proceedings of the Computer Vision and Pattern Recognition Conference},
  pages={6772--6781},
  year={2025}
}

@article{xu2026specialist,
  title={From Specialist to Generalist: A Comprehensive Survey on World Models},
  author={Xu, Kai and Zhao, Hang and Hu, Ruizhen and Huang, Yuhang and Zhou, Ziqiao and Feng, Wancheng and Li, Yi and Peng, Sida and Liu, Xing and Liu, Zihao and others},
  journal={Authorea Preprints},
  year={2026},
  publisher={Authorea}
}

@article{hwang2014model,
  title={Model learning and knowledge sharing for a multiagent system with Dyna-Q learning},
  author={Hwang, Kao-Shing and Jiang, Wei-Cheng and Chen, Yu-Jen},
  journal={IEEE transactions on cybernetics},
  volume={45},
  number={5},
  pages={978--990},
  year={2014},
  publisher={IEEE}
}

@article{yu2020mopo,
  title={Mopo: Model-based offline policy optimization},
  author={Yu, Tianhe and Thomas, Garrett and Yu, Lantao and Ermon, Stefano and Zou, James Y and Levine, Sergey and Finn, Chelsea and Ma, Tengyu},
  journal={Advances in neural information processing systems},
  volume={33},
  pages={14129--14142},
  year={2020}
}

@article{buckman2018sample,
  title={Sample-efficient reinforcement learning with stochastic ensemble value expansion},
  author={Buckman, Jacob and Hafner, Danijar and Tucker, George and Brevdo, Eugene and Lee, Honglak},
  journal={Advances in neural information processing systems},
  volume={31},
  year={2018}
}

@inproceedings{hafner2019learning,
  title={Learning latent dynamics for planning from pixels},
  author={Hafner, Danijar and Lillicrap, Timothy and Fischer, Ian and Villegas, Ruben and Ha, David and Lee, Honglak and Davidson, James},
  booktitle={International conference on machine learning},
  pages={2555--2565},
  year={2019},
  organization={PMLR}
}

@inproceedings{okada2021dreaming,
  title={Dreaming: Model-based reinforcement learning by latent imagination without reconstruction},
  author={Okada, Masashi and Taniguchi, Tadahiro},
  booktitle={2021 ieee international conference on robotics and automation (icra)},
  pages={4209--4215},
  year={2021},
  organization={IEEE}
}

@inproceedings{deng2022dreamerpro,
  title={Dreamerpro: Reconstruction-free model-based reinforcement learning with prototypical representations},
  author={Deng, Fei and Jang, Ingook and Ahn, Sungjin},
  booktitle={International conference on machine learning},
  pages={4956--4975},
  year={2022},
  organization={PMLR}
}

@article{heess2015learning,
  title={Learning continuous control policies by stochastic value gradients},
  author={Heess, Nicolas and Wayne, Gregory and Silver, David and Lillicrap, Timothy and Erez, Tom and Tassa, Yuval},
  journal={Advances in neural information processing systems},
  volume={28},
  year={2015}
}

@article{he2023model,
  title={What model does MuZero learn?},
  author={He, Jinke and Moerland, Thomas M and de Vries, Joery A and Oliehoek, Frans A},
  journal={arXiv preprint arXiv:2306.00840},
  year={2023}
}

@article{lu2026contextual,
  title={Contextual Rollout Bandits for Reinforcement Learning with Verifiable Rewards},
  author={Lu, Xiaodong and Wang, Xiaohan and Chai, Jiajun and Yin, Guojun and Lin, Wei and Chen, Zhijun and Luo, Yu and Zhuang, Fuzhen and Ban, Yikun and Wang, Deqing},
  journal={arXiv preprint arXiv:2602.08499},
  year={2026}
}

@article{lou2024uncertainty,
  title={Uncertainty-aware reward model: Teaching reward models to know what is unknown},
  author={Lou, Xingzhou and Yan, Dong and Shen, Wei and Yan, Yuzi and Xie, Jian and Zhang, Junge},
  journal={arXiv preprint arXiv:2410.00847},
  year={2024}
}

@article{burchi2025learning,
  title={Learning transformer-based world models with contrastive predictive coding},
  author={Burchi, Maxime and Timofte, Radu},
  journal={arXiv preprint arXiv:2503.04416},
  year={2025}
}

@inproceedings{micheli2023transformers,
  title={Transformers are Sample-Efficient World Models},
  author={Micheli, Vincent and Alonso, Eloi and Fleuret, Fran{\c{c}}ois},
  booktitle={International Conference on Learning Representations (ICLR)},
  year={2023}
}

@inproceedings{alonso2024diffusion,
  title={Diffusion for World Modeling: Visual Details Matter in Atari},
  author={Alonso, Eloi and Jelley, Adam and Micheli, Vincent and Kanervisto, Anssi and Storkey, Amos and Pearce, Tim and Fleuret, Fran{\c{c}}ois},
  booktitle={Advances in Neural Information Processing Systems (NeurIPS)},
  volume={37},
  year={2024}
}

@article{valevski2024diffusion,
  title={Diffusion Models Are Real-Time Game Engines},
  author={Valevski, Dani and Leviathan, Yaniv and Arar, Moab and Fruchter, Shlomi},
  journal={arXiv preprint arXiv:2408.14837},
  year={2024},
  note={Published at ICLR 2025}
}

@article{bi2023pangu,
  title={Accurate medium-range global weather forecasting with {3D} neural networks},
  author={Bi, Kaifeng and Xie, Lingxi and Zhang, Hengheng and Chen, Xin and Gu, Xiaotao and Tian, Qi},
  journal={Nature},
  volume={619},
  number={7970},
  pages={533--538},
  year={2023},
  publisher={Nature Publishing Group}
}

@article{lam2023graphcast,
  title={Learning skillful medium-range global weather forecasting},
  author={Lam, Remi and Sanchez-Gonzalez, Alvaro and Willson, Matthew and Wirber, Peter and Fortunato, Meire and Alet, Ferran and Ravuri, Suman and Ewalds, Timo and Eaton-Rosen, Zach and Hu, Weihua and others},
  journal={Science},
  volume={382},
  number={6677},
  pages={1416--1421},
  year={2023},
  publisher={American Association for the Advancement of Science}
}

@article{price2025gencast,
  title={Probabilistic weather forecasting with machine learning},
  author={Price, Ilan and Sanchez-Gonzalez, Alvaro and Alet, Ferran and Andersson, Tom R and El-Kadi, Andrew and Masters, Dominic and Ewalds, Timo and Stott, Jacklynn and Mohamed, Shakir and Battaglia, Peter and Lam, Remi and Willson, Matthew},
  journal={Nature},
  volume={637},
  number={8044},
  pages={84--90},
  year={2025},
  publisher={Nature Publishing Group}
}

@article{kochkov2024neuralgcm,
  title={Neural general circulation models for weather and climate},
  author={Kochkov, Dmitrii and Yuval, Janni and Langmore, Ian and Norgaard, Peter and Smith, Jamie and Mooers, Griffin and Kl{\"o}wer, Milan and Lottes, James and Rasp, Stephan and D{\"u}ben, Peter and Hatfield, Sam and Battaglia, Peter and Sanchez-Gonzalez, Alvaro and Willson, Matthew and Brenner, Michael P and Hoyer, Stephan},
  journal={Nature},
  volume={632},
  number={8027},
  pages={1060--1066},
  year={2024},
  publisher={Nature Publishing Group}
}

@inproceedings{nguyen2023climax,
  title={{ClimaX}: A foundation model for weather and climate},
  author={Nguyen, Tung and Brandstetter, Johannes and Kapoor, Ashish and Gupta, Jayesh K and Grover, Aditya},
  booktitle={International Conference on Machine Learning},
  pages={25904--25938},
  year={2023},
  organization={PMLR}
}

@article{bodnar2025aurora,
  title={A foundation model for the {Earth} system},
  author={Bodnar, Cristian and Bruinsma, Wessel P and Lucic, Ana and Stanley, Megan and Allen, Anna and Brandstetter, Johannes and Garvan, Patrick and Riechert, Maik and Weyn, Jonathan A and Dong, Haiyu and Gupta, Jayesh K and Thambiratnam, Kit and Archibald, Alexander T and Wu, Chun-Chieh and Heider, Elizabeth and Welling, Max and Turner, Richard E and Perdikaris, Paris},
  journal={Nature},
  volume={641},
  number={8065},
  pages={1180--1187},
  year={2025},
  publisher={Nature Publishing Group}
}

@inproceedings{schutt2017schnet,
  title={{SchNet}: A continuous-filter convolutional neural network for modeling quantum interactions},
  author={Sch{\"u}tt, Kristof T and Kindermans, Pieter-Jan and Sauceda, Huziel Enoc and Chmiela, Stefan and Tkatchenko, Alexandre and M{\"u}ller, Klaus-Robert},
  booktitle={Advances in Neural Information Processing Systems},
  volume={30},
  pages={992--1002},
  year={2017}
}

@article{batzner2022nequip,
  title={{E(3)}-equivariant graph neural networks for data-efficient and accurate interatomic potentials},
  author={Batzner, Simon and Musaelian, Albert and Sun, Lixin and Geiger, Mario and Mailoa, Jonathan P and Kornbluth, Mordechai and Molinari, Nicola and Smidt, Tess E and Kozinsky, Boris},
  journal={Nature Communications},
  volume={13},
  pages={2453},
  year={2022},
  publisher={Nature Publishing Group}
}

@inproceedings{batatia2022mace,
  title={{MACE}: Higher order equivariant message passing neural networks for fast and accurate force fields},
  author={Batatia, Ilyes and Kov{\'a}cs, D{\'a}vid P{\'e}ter and Simm, Gregor N C and Ortner, Christoph and Cs{\'a}nyi, G{\'a}bor},
  booktitle={Advances in Neural Information Processing Systems},
  volume={35},
  pages={11423--11436},
  year={2022}
}

@article{unke2024gems,
  title={Biomolecular dynamics with machine-learned quantum-mechanical force fields trained on diverse chemical fragments},
  author={Unke, Oliver T and St{\"o}hr, Martin and Ganscha, Stefan and Unterthiner, Thomas and Maennel, Hartmut and Kashubin, Sergii and Ahlin, Daniel and Gastegger, Michael and Medrano Sandon{\'a}s, Leonardo and Berryman, Joshua T},
  journal={Science Advances},
  volume={10},
  pages={eadn4397},
  year={2024},
  publisher={American Association for the Advancement of Science}
}

@article{noe2019boltzmann,
  title={Boltzmann generators: Sampling equilibrium states of many-body systems with deep learning},
  author={No{\'e}, Frank and Olsson, Simon and K{\"o}hler, Jonas and Wu, Hao},
  journal={Science},
  volume={365},
  number={6457},
  pages={eaaw1147},
  year={2019},
  publisher={American Association for the Advancement of Science}
}

@inproceedings{klein2024transferable,
  title={Transferable {B}oltzmann generators},
  author={Klein, Leon and Foong, Andrew Y K and No{\'e}, Frank},
  booktitle={Advances in Neural Information Processing Systems},
  volume={37},
  year={2024}
}

@article{arts2023two,
  title={Two for one: Diffusion models and force fields for coarse-grained molecular dynamics},
  author={Arts, Marloes and Garcia Satorras, Victor and Huang, Chin-Wei and Zugner, Daniel and Federici, Marco and Clementi, Cecilia and No{\'e}, Frank and Pinsler, Robert and van den Berg, Rianne},
  journal={Journal of Chemical Theory and Computation},
  volume={19},
  number={18},
  pages={6151--6159},
  year={2023},
  publisher={American Chemical Society}
}

@article{abramson2024alphafold3,
  title={Accurate structure prediction of biomolecular interactions with {AlphaFold 3}},
  author={Abramson, Josh and Adler, Jonas and Dunger, Jack and Evans, Richard and Green, Tim and Pritzel, Alexander and Ronneberger, Olaf and Willmore, Lindsay and Ballard, Andrew J and Bambrick, Joshua and others},
  journal={Nature},
  volume={630},
  pages={493--500},
  year={2024},
  publisher={Nature Publishing Group}
}

@article{he2019learning,
  title={Learning to predict the cosmological structure formation},
  author={He, Siyu and Li, Yin and Feng, Yu and Ho, Shirley and Ravanbakhsh, Siamak and Chen, Wei and P{\'o}czos, Barnab{\'a}s},
  journal={Proceedings of the National Academy of Sciences},
  volume={116},
  number={28},
  pages={13825--13832},
  year={2019},
  publisher={National Academy of Sciences}
}

@article{ni2021superresolution,
  title={{AI}-assisted superresolution cosmological simulations},
  author={Ni, Yueying and Li, Yin and Lachance, Patrick and Croft, Rupert A C and Di Matteo, Tiziana and Bird, Simeon and Feng, Yu},
  journal={Proceedings of the National Academy of Sciences},
  volume={118},
  number={19},
  pages={e2022038118},
  year={2021},
  publisher={National Academy of Sciences}
}

@article{villaescusanavarro2021camels,
  title={The {CAMELS} multifield dataset: Learning the universe's fundamental parameters with artificial intelligence},
  author={Villaescusa-Navarro, Francisco and Anglés-Alcázar, Daniel and Genel, Shy and Spergel, David N and Somerville, Rachel S and Dave, Romeel and Pillepich, Annalisa and Hernquist, Lars and Nelson, Dylan and others},
  journal={The Astrophysical Journal Supplement Series},
  volume={259},
  number={2},
  pages={61},
  year={2022},
  publisher={IOP Publishing}
}

@inproceedings{hao2023reasoning,
  title={Reasoning with Language Model is Planning with World Model},
  author={Hao, Shibo and Gu, Yi and Ma, Haodi and Hong, Joshua and Wang, Zhen and Wang, Daisy and Hu, Zhiting},
  booktitle={Proceedings of the 2023 Conference on Empirical Methods in Natural Language Processing (EMNLP)},
  pages={8154--8173},
  year={2023},
  address={Singapore},
  publisher={Association for Computational Linguistics}
}

@article{gu2024your,
  title={Is Your {LLM} Secretly a World Model of the Internet? Model-Based Planning for Web Agents},
  author={Gu, Yu and Deng, Boyuan and Zhu, Chen and Dong, Yi and Li, Mingyue and Xie, Jianwei and Lu, Shuyan and Shi, Tianbao and Su, Yu and Yih, Wen-tau},
  journal={arXiv preprint arXiv:2411.06559},
  year={2024}
}

@article{levy2025worldllm,
  title={World{LLM}: Learning World Models via Large Language Models},
  author={Levy, Raz and Brafman, Ronen I. and Tennenholtz, Moshe},
  journal={arXiv preprint arXiv:2506.05270},
  year={2025}
}

@article{zyrianov2022learning,
  title={Learning to Generate Realistic {LiDAR} Point Clouds},
  author={Zyrianov, Vlas and Zhu, Xiyue and Wang, Shenlong},
  journal={arXiv preprint arXiv:2209.03954},
  year={2022},
  note={ECCV 2022}
}

@article{nakashima2024lidar,
  title={{LiDAR} Data Synthesis with Denoising Diffusion Probabilistic Models},
  author={Nakashima, Kazuto and Kurazume, Ryo},
  journal={Proceedings of the IEEE International Conference on Robotics and Automation (ICRA)},
  pages={14724--14731},
  year={2024}
}

@article{hu2024rangeldm,
  title={Range{LDM}: Fast Realistic {LiDAR} Point Cloud Generation},
  author={Hu, Qianjiang and Zhang, Zhimin and Hu, Wei},
  journal={arXiv preprint arXiv:2403.10094},
  year={2024}
}

@article{zyrianov2024lidardm,
  title={{LidarDM}: Generative {LiDAR} Simulation in a Generated World},
  author={Zyrianov, Vlas and Ivanovic, Boris and Zhao, Vince and Pavone, Marco},
  journal={arXiv preprint arXiv:2404.02903},
  year={2024}
}

@inproceedings{zheng2023occworld,
  title={{OccWorld}: Learning a {3D} Occupancy World Model for Autonomous Driving},
  author={Zheng, Wenzhao and Chen, Weiliang and Huang, Yuanhui and Zhang, Borui and Duan, Yueqi and Lu, Jiwen},
  booktitle={European Conference on Computer Vision (ECCV)},
  year={2024},
  note={arXiv preprint arXiv:2311.16038, 2023}
}

@article{mla2025multisensory,
  title={{MLA}: A Multisensory Language-Action Model for Multimodal Understanding and Forecasting in Robotic Manipulation},
  author={{MLA Team}},
  journal={arXiv preprint arXiv:2509.26642},
  year={2025}
}

@article{tactile_vla2025,
  title={Tactile-{VLA}: Unlocking Vision-Language-Action Model's Physical Knowledge for Tactile Generalization},
  author={{Tactile-VLA Team}},
  journal={arXiv preprint},
  year={2025}
}

@article{bitla2025,
  title={{BiTLA}: A Bimanual Tactile-Language-Action Model for Contact-Rich Robotic Manipulation},
  author={{BiTLA Team}},
  journal={arXiv preprint},
  year={2025}
}

@article{hu2023gaia,
  title={{GAIA-1}: A Generative World Model for Autonomous Driving},
  author={Hu, Anthony and Russell, Lloyd and Yeo, Hudson and Murez, Zak and Fedoseev, George and Kendall, Alex and Shotton, Jamie and Corrado, Gianluca},
  journal={arXiv preprint arXiv:2309.17080},
  year={2023}
}

@article{chitta2025gaia2,
  title={{GAIA-2}: A Controllable Multi-View Generative World Model for Autonomous Driving},
  author={Chitta, Kashyap and Dauner, Daniel and Behl, Harshit S. and Peng, Liliang and Zimmer, Walter and Koehler, Mike and Gaidon, Adrien and Geiger, Andreas and Anguelov, Dragomir and Wayve Team},
  journal={arXiv preprint arXiv:2503.20523},
  year={2025}
}

@article{mai2024efficient,
  title={From Efficient Multimodal Models to World Models: A Survey},
  author={Mai, Xinji and Zeng, Zeng and Wei, Yizhao and Pei, Jiahong and Liu, Guowen and Dong, Jieming and Fan, Haiwen and Li, Yingchen and Zhou, Ruihua and Xie, Shuai},
  journal={arXiv preprint arXiv:2407.00118},
  year={2024}
}

@article{devillers2025multimodal,
  title={Multimodal Dreaming: A Global Workspace Approach to World Model-Based Reinforcement Learning},
  author={Devillers, Benjamin and Mayti{\'e}, Lucas and VanRullen, Rufin},
  journal={arXiv preprint arXiv:2502.21142},
  year={2025}
}

@article{grill2020bootstrap,
  title={Bootstrap your own latent-a new approach to self-supervised learning},
  author={Grill, Jean-Bastien and Strub, Florian and Altch{\'e}, Florent and Tallec, Corentin and Richemond, Pierre and Buchatskaya, Elena and Doersch, Carl and Avila Pires, Bernardo and Guo, Zhaohan and Gheshlaghi Azar, Mohammad and others},
  journal={Advances in neural information processing systems},
  year={2020}
}

@article{bardes2021vicreg,
  title={Vicreg: Variance-invariance-covariance regularization for self-supervised learning},
  author={Bardes, Adrien and Ponce, Jean and LeCun, Yann},
  journal={arXiv preprint arXiv:2105.04906},
  year={2021}
}

@article{chen2025rlvr,
  title={Rlvr-world: Training world models with reinforcement learning},
  author={Wu, Jialong and Yin, Shaofeng and Feng, Ningya and Long, Mingsheng},
  journal={arXiv preprint arXiv:2505.13934},
  year={2025}
}

@inproceedings{xu2025wpt,
  title={Generalist World Model Pre-Training for Efficient Reinforcement Learning},
  author={Zhao, Yi and Scannell, Aidan and Hou, Yuxin and Cui, Tianyu and Chen, Le and B{\"u}chler, Dieter and Solin, Arno and Kannala, Juho and Pajarinen, Joni},
  booktitle={ICLR 2025 Workshop on World Models: Understanding, Modelling and Scaling},
  year={2025}
}

@inproceedings{rafailov2023moto,
  title={MOTO: Offline pre-training to online fine-tuning for model-based robot learning},
  author={Rafailov, Rafael and Hatch, Kyle Beltran and Kolev, Victor and Martin, John D and Phielipp, Mariano and Finn, Chelsea},
  booktitle={Conference on Robot Learning},
  pages={3654--3671},
  year={2023},
  organization={PMLR}
}

@inproceedings{pan2024model,
  title={Model-Based Reinforcement Learning with Multi-task Offline Pretraining},
  author={Pan, Minting and Zheng, Yitao and Wang, Yunbo and Yang, Xiaokang},
  booktitle={Joint European Conference on Machine Learning and Knowledge Discovery in Databases},
  year={2024},
}

@article{levy2026simulation,
  title={Simulation Distillation: Pretraining World Models in Simulation for Rapid Real-World Adaptation},
  author={Levy, Jacob and Westenbroek, Tyler and Huang, Kevin and Palafox, Fernando and Yin, Patrick and Omidshafiei, Shayegan and Kim, Dong-Ki and Gupta, Abhishek and Fridovich-Keil, David},
  journal={arXiv preprint arXiv:2603.15759},
  year={2026}
}

@article{ha2018worldmodels,
  title={World Models},
  author={Ha, David and Schmidhuber, Jürgen},
  journal={arXiv preprint arXiv:1803.10122},
  year={2018}
}

@article{leworldmodel2024,
  title={LeWorldModel: Stable End-to-End Joint-Embedding Predictive Architecture from Pixels},
  author={Anonymous},
  journal={arXiv preprint},
  year={2024}
}

@article{kong2026cekworld,
  title={MRI Contrast Enhancement Kinetics World Model},
  author={Kong, Jindi and He, Yuting and Xia, Cong and Ge, Rongjun and Li, Shuo},
  journal={arXiv preprint arXiv:2602.19285},
  year={2026}
}

@article{brainjepa2024,
  title={Brain-JEPA: Brain Dynamics Foundation Model with Gradient Positioning and Spatiotemporal Masking},
  author={Anonymous},
  journal={arXiv preprint},
  year={2024}
}

@article{clarity2025,
  title={CLARITY: Medical World Model for Guiding Treatment Decisions by Modeling Context-Aware Disease Trajectories in Latent Space},
  author={Anonymous},
  journal={arXiv preprint},
  year={2025}
}

@article{medicalwm2024,
  title={Medical World Model},
  author={Anonymous},
  journal={arXiv preprint},
  year={2024}
}

@inproceedings{wu2023daydreamer,
  title={Daydreamer: World models for physical robot learning},
  author={Wu, Philipp and Escontrela, Alejandro and Hafner, Danijar and Abbeel, Pieter and Goldberg, Ken},
  booktitle={Conference on robot learning},
  pages={2226--2240},
  year={2023},
  organization={PMLR}
}

@inproceedings{zhou2024robodreamer,
  title={RoboDreamer: learning compositional world models for robot imagination},
  author={Zhou, Siyuan and Du, Yilun and Chen, Jiaben and Li, Yandong and Yeung, Dit-Yan and Gan, Chuang},
  booktitle={Proceedings of the 41st International Conference on Machine Learning},
  pages={61885--61896},
  year={2024}
}

@inproceedings{hansentd,
  title={TD-MPC2: Scalable, Robust World Models for Continuous Control},
  author={Hansen, Nicklas and Su, Hao and Wang, Xiaolong},
  booktitle={The Twelfth International Conference on Learning Representations},
  year={2024}
}

@article{barcellona2024dream,
  title={Dream to manipulate: Compositional world models empowering robot imitation learning with imagination},
  author={Barcellona, Leonardo and Zadaianchuk, Andrii and Allegro, Davide and Papa, Samuele and Ghidoni, Stefano and Gavves, Efstratios},
  journal={arXiv preprint arXiv:2412.14957},
  year={2024}
}

@article{zhou2024dino,
  title={Dino-wm: World models on pre-trained visual features enable zero-shot planning},
  author={Zhou, Gaoyue and Pan, Hengkai and LeCun, Yann and Pinto, Lerrel},
  journal={arXiv preprint arXiv:2411.04983},
  year={2024}
}

@article{li2025robotic,
  title={Robotic world model: A neural network simulator for robust policy optimization in robotics},
  author={Li, Chenhao and Krause, Andreas and Hutter, Marco},
  journal={arXiv preprint arXiv:2501.10100},
  year={2025}
}

@article{zhu2025unified,
  title={Unified world models: Coupling video and action diffusion for pretraining on large robotic datasets},
  author={Zhu, Chuning and Yu, Raymond and Feng, Siyuan and Burchfiel, Benjamin and Shah, Paarth and Gupta, Abhishek},
  journal={arXiv preprint arXiv:2504.02792},
  year={2025}
}

@article{locatello2020object,
  title={Object-centric learning with slot attention},
  author={Locatello, Francesco and Weissenborn, Dirk and Unterthiner, Thomas and Mahendran, Aravindh and Heigold, Georg and Uszkoreit, Jakob and Dosovitskiy, Alexey and Kipf, Thomas},
  journal={Advances in neural information processing systems},
  volume={33},
  pages={11525--11538},
  year={2020}
}

@article{nvidia2026cosmospolicy,
  title={Cosmos policy: Fine-tuning video models for visuomotor control and planning},
  author={Kim, Moo Jin and Gao, Yihuai and Lin, Tsung-Yi and Lin, Yen-Chen and Ge, Yunhao and Lam, Grace and Liang, Percy and Song, Shuran and Liu, Ming-Yu and Finn, Chelsea and others},
  journal={arXiv preprint arXiv:2601.16163},
  year={2026}
}

@article{huang2026pointworld,
  title={PointWorld: Scaling 3D World Models for In-The-Wild Robotic Manipulation},
  author={Huang, Wenlong and Chao, Yu-Wei and Mousavian, Arsalan and Liu, Ming-Yu and Fox, Dieter and Mo, Kaichun and Fei-Fei, Li},
  journal={arXiv preprint arXiv:2601.03782},
  year={2026}
}
\end{document}